# NATURAL LANGUAGE PROCESSING FOR AFRICAN LANGUAGES

DAVID IFEOLUWA ADELANI

A dissertation submitted towards the degree of

Doctor of Engineering (Dr.-Ing.)

of the Faculty of Mathematics and Computer Science of Saarland University

Saarbrücken, 2022

<>
Date of Defense: 27.06.2023

Dean of the Faculty: Prof. Dr. Jürgen Steimle

EXAMINATION COMMITTEE:
Chair: Prof. Dr. Vera Demberg
First Reviewer, Advisor: Prof. Dr. Dietrich Klakow
Second Reviewer: Prof. Dr. Alexander M. Fraser
Third Reviewer: Prof. Dr. Benoît Sagot
Committee Member: Dr. Volha Petukhova

Dedicated to the vibrant Masakhane NLP Community

# ABSTRACT

Recent advances in pre-training of word embeddings and language models leverage large amounts of unlabelled texts and self-supervised learning to learn distributed representations that have significantly improved the performance of deep learning models on a large variety of natural language processing tasks. Similarly, multilingual variants of these models have been developed from web-crawled multilingual resources like Wikipedia and Common crawl. However, there are some drawbacks to building these multilingual representation models. First, the models only include few low-resource languages in the training corpus, and additionally, the texts of these languages are often noisy or of low quality texts. Second, their performance on downstream NLP tasks is difficult to evaluate because of the absence of labelled datasets, therefore, they are typically only evaluated on English and other high-resource languages.

In this dissertation, we focus on languages spoken in Sub-Saharan Africa where all the indigenous languages in this region can be regarded as low-resourced in terms of the availability of labelled data for NLP tasks and unlabelled data found on the web. We analyse the noise in the publicly available corpora, and curate a high-quality corpus, demonstrating that the quality of semantic representations learned in word embeddings does not only depend on the amount of data but on the quality of pre-training data. We demonstrate empirically the limitations of word embeddings, and the opportunities the multilingual pre-trained language model (PLM) offers especially for languages unseen during pre-training and low-resource scenarios. We further study how to adapt and specialize multilingual PLMs to unseen African languages using a small amount of monolingual texts. To address the under-representation of the African languages in NLP research, we developed large scale human-annotated labelled datasets for 21 African languages in two impactful NLP tasks: named entity recognition and machine translation. We conduct an extensive empirical evaluation using state-of-the-art methods across supervised, weakly-supervised, and transfer learning settings.

In order to advance the progress of NLP for African languages, future work should focus on expanding benchmark datasets for African languages in other important NLP tasks like part of speech tagging, sentiment analysis, hate speech detection, and question answering. Another direction is to focus on development of Africa-centric PLMs. Lastly, research on speech that involves developing corpora and techniques that require zero or few paired speech-text data would be very essential for the survival of many under-resourced African languages.

# ZUSAMMENFASSUNG

Jüngste Fortschritte beim Pre-Training von Worteinbettungen und neuronalen Sprachmodellen nutzen große Mengen nicht gelabelter Texte und selbstüberwachtes Lernen zum Erlernen verteilter Repräsentationen, die die Leistung von Deep-Learning-Modellen bei einer Vielzahl von Aufgaben zur Verarbeitung natürlicher Sprache erheblich verbessert haben. In ähnlicher Weise wurden mehrsprachige Varianten dieser Modelle auf der Grundlage von mehrsprachigen Ressourcen aus dem Internet wie Wikipedia und Common Crawl entwickelt. Die Entwicklung dieser mehrsprachigen Repräsentationsmodelle birgt jedoch einige Nachteile. Erstens enthalten die Modelle nur wenige Sprachen mit geringen Ressourcen im Trainingskorpus, und außerdem sind die Texte dieser Sprachen oft von geringer Qualität. Zweitens ist ihre Leistung bei nachgelagerten NLP-Aufgaben schwer zu bewerten, da es keine gelabelten Datensätze gibt, weshalb sie nur für Englisch und andere Sprachen mit hohen Ressourcen bewertet werden.

In dieser Dissertation konzentrieren wir uns auf Sprachen, die in Afrika südlich der Sahara gesprochen werden. Alle einheimischen Sprachen in dieser Region können als ressourcenarm angesehen werden, was die Verfügbarkeit von gelabelten Daten für NLP-Aufgaben und von nicht gelabelten Daten aus dem Internet angeht. Wir analysieren das Rauschen in den öffentlich zugänglichen Korpora und kuratieren ein qualitativ hochwertiges Korpus, um zu zeigen, dass die Qualität der semantischen Repräsentationen, die mit Worteinbettungen gelernt werden, nicht nur von der Menge der Daten, sondern auch von der Qualität der Trainingsdaten abhängt.

Wir demonstrieren empirisch die Grenzen von Worteinbettungen und die Möglichkeiten, die mehrsprachige vortrainierte Sprachmodell (PLM) bietet. Wir konzentrieren uns hierbei insbesondere auf Sprachen, die kein Bestandteil der Trainingsdaten sind, sowie auf Szenarien mit geringen Mengen an gelabelten Daten.

Darüber hinaus untersuchen wir, wie man mehrsprachige vortrainierte Sprachmodelle an für sie unbekannte afrikanische Sprachen anpassen und spezialisieren kann, indem man eine kleine Menge von Texten in der jeweiligen Sprache verwendet. Um der Unterrepräsentation afrikanischer Sprachen in der NLP-Forschung entgegenzuwirken haben wir große, von Menschen gelabelte Datensätze für 21 afrikanische Sprachen in zwei wichtigen NLP-Aufgaben entwickelt: Eigennamenerkennung und maschinelle Übersetzung; und führen eine umfassende empirische Evaluierung von modernsten Methoden des Überwachten-, Schwach-Überwachten- und Transfer Lernens durch.

Um den Fortschritt von NLP für afrikanische Sprachen weiter voranzutreiben, sollte sich die zukünftige Arbeit auf die Erweiterung von Benchmark-Datensätzen für afrikanische Sprachen in anderen wichtigen NLP-Aufgaben wie der des Part-of-Speech-Tagging, der Sentiment-Analyse, der Erkennung von Hassreden und der Beantwortung von Fragen konzentrieren. Ein weiterer Bereich ist die Entwicklung von afrikazentrierten vortrainierten Sprachmodellen. Schließlich wäre die Erstellung von Korpora sowie die Erforschung und Entwicklung von Techniken, die keine oder nur wenige Sprach- oder Textdaten benötigen, sehr wichtig für das Überleben vieler afrikanischer Sprachen mit geringen Ressourcen.

## ACKNOWLEDGMENTS

First, I am very grateful to God for the successful completion of my PhD thesis, and for the wisdom and strength (Psalms 18:34, 144:1).

I am very grateful to my supervisor, Prof. Dr. Dietrich Klakow for the rare opportunity to pursue PhD at LSV despite only having little experience with NLP research. Thank you very much for your mentorship, guidance, and for encouraging me to pursue "NLP for African languages" for my PhD dissertation.

A big thank you to my PhD examination committee chaired by Prof. Dr. Vera Demberg, and my reviewers Prof. Dr. Alexander Fraser & Prof. Dr. Benoît Sagot, and Dr. Volha Petukhova.

Next, I would like to thank my close collaborators at LSV, DFKI, and LST including Thomas, Ali, Jesujoba, Michael, Marius, Dana, Cristina, Ernie, Dawei, Xiaoyu, Miaoran, Aditya and Aleena. Thank you to all the members of LSV (Anu, Olga, Fech Scen, Alexander, and Aravind) for the regular discussions and feedback on my research projects. I'm also grateful to our secretary Claudia, and Nico our server administrator for all the assistance.

A big thank you to all my virtual mentors and collaborators I met through the Masakhane community like Graham, Sebastian, Julia, Jade, Constantine, Chester, Shruti, Stephen, Angela, Machel, Peter, Bamba, Kathleen, Michael, and Iyanu. I say a big thank you to all the language coordinators I worked with at Masakhane who helped to supervise dataset collection, annotation and translation: Israel, Shamsuddeen, Chris, Happy, Jonathan, Perez, Aremu, Catherine, Derguene, Fatoumata, Godson, Allahsera, Victoire, Edwin, Valencia, Blessing, Andiswa, Rooweither, Bonaventure, Amelia and Tebogo. I have been privilege to work with a lot of authors, and I say thank you for your contributions: Joyce, Daniel, Seid, Tajuddeen, Ignatius, Andre, Verrah, Iroro, Davis, Samba, Tosin, Paul, Mofetoluwa, Gerald, Emmanuel, Chiamaka, Nkiruka, Eric, Samuel, Clemencia, Tobius, Temilola, Yvonne, Victor, Deborah, Maurice, Ayodele, Mouhamadane, Dibora, Henok, Kelechi, Degaga, Abdoulaye, Orevaoghene, Kelechi, Thierno, Abdoulaye, Adewale, Tendai, Salomey, Freshia, Guyo, Oreen, Gilles, Muhammad, Benjamin, Tunde, Mohamed, Millicent, Idris, Sam, Vukosi, Elvis, Neo, Odunayo, Tatiana, Damilola, and Adesina.

Also, I would like to acknowledge the support of my PhD through the EU-funded Horizon 2020 projects (COMPRISE and ROXANNE); Lacuna Funds that supported the machine translation and named entity recognition datasets; the Saarbrücken Graduate School of Computer Science for the support of my first 2 years in Grad school; DFKI for student jobs; MPI-SWS where I was a research assistant for al-most 2 years; and National Institute of Informatics (NII) in Japan for supporting my internship. From all these organizations, I have met great mentors and collaborators including COMPRISE collaborators (Emmanuel, Marc, Denis, Irina, Aurélien, Imran, Mehmet, Brij, Zaineb, Akira, Youssef, Gerrit, Raivis, and Álvaro), Graduate School adminstration (Dr. Michelle Carnell and Susanne Vohl), DFKI mentor (Alassane), MPI-SWS collaborators and mentors (Przemyslaw, Krishna, Isabel, and Ingmar), and NII collaborators (Isao, Junichi, Fuming, Huy, Haotian, and Ryota).

Futhermore, I thank the members of the Nigerian student community in Saarbrueken (Bright, Olamide, Eustace, Obaro, Teju, Bukola, Ema, Afolabi, Tari, Olachi, Onyi, Kunle), members of the FeG church (Anjara, Nadine, Joachim, Cyrile, Leslie, Cammy, Sven, Manasoa) , and the HausKreis group (Bettina, Teffi, JJ, Tolulope, Nathan, Adina, Damaris, Michele, and Brenda). Thank you for the wonderful time we spent together and the prayers.

Finally, I really appreciate my wife (Tolulope), my parents, in-laws, uncles, Adelani household in Abeokuta for their support, prayers and encouragements. I also thank my former boss at Federal Ministry of Communications, Mrs Moni Udoh, my former MSc Advisor at the AUST Abuja, Prof. Mamadou Kaba Traoré, and my former BSc Advisor at FUNAAB, Nigeria, Prof. Adesina Sodiya for their encouragement. A big thank you to my wife, Jens, Jesujoba, and Marius, for helping with proofreading this dissertation.

x

## contents

# Part I

INTRODUCTION AND BACKGROUND

INTRODUCTION

Africa has over 2,000 spoken languages (Eberhard, Simons, and Fennig, 2021), and many of these languages are spoken by millions or tens of millions of speakers. However, these languages are poorly represented in existing natural language processing (NLP) datasets, research, and tools (Martinus and Abbott, 2019). Most developments of NLP have been focused on the English language, other European languages, and a few Asian languages like Arabic, Mandarin Chinese and Japanese; these languages are regarded as Winners (class 5) in Joshi's classification of world languages (Joshi et al., 2020) based on the size of available labelled and unlabelled corpora on the web, with classes ranging from 0 to 5. While there are many low-resource languages in different regions of the world, the situation of African languages is grave, with all the indigenous African languages falling within Joshi's definition of low-resource languages (classes 0 to 3). Thus, limiting the opportunities of NLP to over 1.2 billion people living in Africa whose native languages are rarely supported by technology.

There are many factors responsible for the under-representation of African languages, some are data-related, and others are societal factors such as lack of government support for indigenous languages, weak language policies by many African countries, and the impact of colonialism. Effects of colonialism include suppression of African languages[1], post-colonial successors' maintenance of colonial linguistic hierarchies (Phillipson and Skutnabb-Kangas, 2010), and native speakers' perception that their language is inferior to the dominant colonial language. There are other diversity and data-related factors such as (1) the geographical and language diversity of NLP researchers ($\forall$ et al., 2020), resulting in a lower overall interest to work in this area of research, (2) a lack of labelled datasets for several NLP tasks, (3) the absence of large monolingual corpora on the web required to leverage self-supervised pre-training to boost the performance of NLP tasks on these languages, and (4) a lack of basic linguistic tools like dictionaries, morphological analyzers, spell-checkers and keyboards that support the correct orthography of the language. While there are efforts to address the keyboard issues with the launch of Gboard (Esch et al., 2019) (Google's Keyboard) on mobile phones, it will still take years for many low-resource languages to have a large amount of monolingual data on the web.

In this dissertation, we address two challenges of NLP for African languages: (1) the lack of labelled datasets and (2) the absence of large monolingual data needed for multilingual representation models like word embeddings and pre-trained language models – which serve as foundational models to build models for many NLP tasks.

---

[1] https://www.goethe.de/prj/zei/en/pos/22902448.html

Our approach combines **distant and weak supervision** (like leveraging expert rules (Ratner et al., 2019), external knowledge (Pan et al., 2017) or self-training (Liang et al., 2020; Paul et al., 2019)), **transfer learning** (Ruder et al., 2019) and **participatory research** (∀ et al., 2020) for the development of datasets and models for African languages[2]. We describe the three approaches below.

*Participatory research* design for low-resource NLP involves working together with language speakers, dataset curators, NLP practitioners, and evaluation experts for the development of NLP datasets and models. This was pioneered by Masakhane[3] for the development of machine translation models for African languages (∀ et al., 2020). We make use of the participatory research approach by collaborating with Masakhane for the creation of publicly available, high quality datasets for 21 languages in two impactful NLP tasks: named entity recognition (NER) (§7, §8) and machine translation (MT) (§10). We prioritize creating small labelled datasets like 2k sentences for NER and 2k-5k parallel sentences for MT due to the high cost of human annotation, and we leverage techniques such as distant and weak supervision, and transfer learning for improved performance.

We make use of *distant and weak supervision* for NER by leveraging expert rules (e.g. rules for identifying a DATE entity) and external knowledge (e.g. entity lists from Wikidata (Hedderich, Lange, and Klakow, 2021)) to create labeled data in a (semi-) automatic way. This can be combined with a few available labelled samples. Additionally, to alleviate some of the negative effects of the errors in automatic annotation, we integrate noise-handling methods (Hedderich and Klakow, 2018) to the NER models (§6). This approach is based on the assumption of availability of large unlabelled texts in the same language or domain of the labelled data. In some cases, this assumption does not hold for many low-resource languages; transfer learning provides an alternative to this approach.

*Transfer learning* has been shown to be very effective for both zero-shot (Artetxe, Ruder, and Yogatama, 2020; Conneau et al., 2020; Pfeiffer et al., 2020b) and few-shot scenarios (Hedderich et al., 2020; Lauscher et al., 2020) since the introduction of pre-trained language models (PLMs) for NLP (Devlin et al., 2019; Liu et al., 2019; Raffel et al., 2020), and their multilingual variants (Conneau et al., 2020; Xue et al., 2021). This can be used for knowledge transfer across different tasks (Aribandi et al., 2022; Poth et al., 2021), domains (Davody et al., 2022; Gururangan et al., 2020), and languages (Ansell et al., 2022; Pfeif-fer et al., 2021). We leverage transfer learning to develop a new state-of-the-art PLM known as AfroXLMR[4] by adapting an existing multilingual pre-trained language model (PLM) to 17 African languages including nine languages previously unseen during pre-training (§5). Similarly, for the NER task, we leverage transfer learning to obtain impressive zero-shot and few-shot performance (§8). Lastly, we leverage transfer learning for machine translation by demonstrating that the most effective strategy for transferring both to additional languages and to additional domains is to fine-tune large pre-trained models such as M2M-100 (Fan et al., 2021) on small quantities of high quality translation data (§10).

---

2 All the labelled datasets and models developed in this dissertation are available on Github at https://github.com/dadelani/africanlp-resources
3 https://www.masakhane.io/

## 1.1 STRUCTURE AND CONTRIBUTIONS

The structure of this dissertation is divided into five parts: (1) Introduction and Background that covers Chapters 1, 2, and 3. (2) Multilingual representation models that covers Chapters 4 and 5. (3) Named entity recognition for African languages that covers Chapters 6, 7, and 8. (4) Machine translation for African languages that covers Chapters 9 and 10. (5) Conclusion and Future work in Chapter 11.

The contributions of this dissertation are summarized by chapters below:

(a) In Chapter 2, we provide an overview of the language families, official status, geographical locations, online corpora size, and linguistic characteristics of 28 African languages. We highlight important linguistic characteristics of these languages like writing systems, word order, morphology and noun classes.

(b) In Chapter 3, we survey the NLP resources that are publicly available to develop NLP models for African languages such as unlabelled and labelled corpora, pre-trained word embeddings, and multilingual pre-trained language models (PLMs). We demonstrate empirically the limitations of word embeddings using NER as a case study, and the opportunities multilingual PLM offers especially for languages unseen during pre-training.

(c) In Chapter 4, we evaluate the quality of pre-trained FastText word embeddings for two African languages (Twi and Yorùbá) using a word similarity task. Our evaluations show that they are of poor quality because the pre-training corpora are either small or of poor quality. To remedy this, we trained FastText embeddings on high-quality curated corpora. Using the same curated corpus, we extended the analysis to BERT (Devlin et al., 2019).

---

4 https://huggingface.co/Davlan/afro-xlmr-large

(d) In Chapter 5, we develop a new multilingual PLM for African languages by *adapting* an existing multilingual PLM (like XLM-R (Conneau et al., 2020)) to 17 African languages, and three high resource languages widely used on the continent (English, French and Arabic). Adding the high resource languages during adaptation improves cross-lingual transfer performance from them to African languages. Our adaptation approach achieves the state-of-the-art compared to other multilingual PLMs.

(e) In Chapter 6, we develop NER models for two African languages (Hausa and Yorùbá) with only a few labelled sentences. We leverage techniques such as distant and weak supervision to create labelled data in a (semi-)automatic way and combine them with noise-handling methods to alleviate the errors introduced by automatic annotation.

(f) In Chapter 7, we employ the participatory research approach ($\forall$ et al., 2020) to create NER datasets (known as MasakhaNER) and models for 10 African languages by working together with native speakers, dataset curators, and evaluation experts in the Masakhane community. We conduct an extensive empirical evaluation using both supervised and transfer learning methods.

(g) In Chapter 8, we expand the MasakhaNER to 21 (typologically-diverse) African languages and annotate more sentences for existing languages (more than twice the initial dataset). We also study the behaviour of state-of-the-art cross-lingual transfer methods in an Africa-centric setting, demonstrating that the choice of source language significantly affects performance.

(h) In Chapter 9, we create MENYO-20k, the first multi-domain parallel corpus (with 20k parallel sentences) for the Yorùbá–English to address the challenge of lack of standardized evaluation datasets from diverse domains for the language. We provide several neural machine translation (MT) benchmarks and compare them to the performance of popular pre-trained (massively multilingual) MT models both for a heterogeneous test set and its subdomains.

(i) In Chapter 10, we investigate "how to optimally leverage existing pre-trained models to create low-resource translation systems for 21 African languages in a new domain". To answer the question, we create a new African news corpus covering 21 languages and demonstrate that the most effective strategy for transferring to a new domain is to fine-tune large pre-trained models on small quantities of high-quality translation data.

## 1.2 PUBLICATIONS

### 1.2.1 Publications related to this Dissertation

1. Alabi*, Amponsah-Kaakyire*, **Adelani** & España-Bonet (2020)
   *Massive vs. Curated Embeddings for Low-Resourced Languages: the Case of Yorùbá and Twi*
   In Proceedings of the 12th Language Resources and Evaluation Conference (LREC)
   https://aclanthology.org/2020.lrec-1.335/
   The details of this work will be discussed in Chapter 4

2. Alabi*, **Adelani***, Mosbach & Klakow (2022)
   *Adapting Pre-trained Language Models to African Languages via Multilingual Adaptive Fine-Tuning*
   In Proceedings of the 28th International Conference on Computational Linguistics (COLING)
   https://arxiv.org/abs/2204.06487
   The details of this work will be discussed in Chapter 5

3. **Adelani***, Hedderich*, Zhu*, van den Berg & Klakow (2020)
   *Distant Supervision and Noisy Label Learning for Low Resource Named Entity Recognition: A Study on Hausa and Yorùbá*
   Presented at the Practical Machine Learning for Developing Countries (PML4DC) & AfricaNLP @ICLR
   https://arxiv.org/abs/2003.08370
   The details of this work will be discussed in Chapter 6

4. **Adelani**, Abbott, Neubig, D'souza, Kreutzer, Lignos, Palen-Michel, Buzaaba, Rijhwani, Ruder, Mayhew & 50 more authors from Masakhane (2021)
   *MasakhaNER: Named Entity Recognition for African Languages*
   In Transactions of the Association for Computational Linguistics (TACL). Presented at EMNLP 2021
   https://aclanthology.org/2021.tacl-1.66/
   The details of this work will be discussed in Chapter 7

5. **Adelani**, Neubig, Ruder, Rijhwani, Beukman, Palen-Michel, Lignos, Alabi, 35 more authors, & Klakow (2022)
   *MasakhaNER 2.0: Africa-centric Transfer Learning for Named Entity Recognition*
   In Proceedings of the 2022 Conference on Empirical Methods in Natural Language Processing (EMNLP)
   The details of this work will be discussed in Chapter 8

6. **Adelani***, Ruiter*, Alabi*, Adebonojo, Ayeni, Adeyemi, Awokoya & España-Bonet (2021)
   *The Effect of Domain and Diacritics in Yorùbá –English Neural Machine Translation*

   In Proceedings of Machine Translation Summit (MT Summit) XVIII: Research Track
   https://aclanthology.org/2021.mtsummit-research.6/
   The details of this work will be discussed in Chapter 9

7. **Adelani**, Alabi, Fan, Kreutzer, Shen, Reid, Ruiter, Klakow, Nabende, Chang & 35 more authors (2022)
   *A Few Thousand Translations Go a Long Way! Leveraging Pre-trained Models for African News Translation*
   In Proceedings of the Conference of the North American Chapter of the ACL: Human Language Technologies (NAACL-NLT)
   https://aclanthology.org/2022.naacl-main.223/
   The details of this work will be discussed in Chapter 10

\* The first authors contributed equally

1.2.2 *Other Publications*

The publications listed below are not related to NLP for African languages, and therefore not discussed in this dissertation. However, they are research papers I worked on during my doctoral studies, they focus on topics in privacy in NLP, few-shot learning for NER and detection of online fake reviews generated by language models.

1. **Adelani**, Mai, Fang, Nguyen, Yamagishi & Echizen (2020)
   *Generating sentiment-preserving fake online reviews using neural language models and their human-and machine-based detection*
   In Proceedings of the 34th International Conference on Advanced Information Networking and Applications (AINA)
   https://arxiv.org/abs/1907.09177

2. **Adelani**, Davody, Kleinbauer & Klakow (2020)
   *Privacy guarantees for de-identifying text transformations*
   In Proceedings of Interspeech
   https://www.isca-speech.org/archive_v0/Interspeech_2020/abstracts/2208.html

3. Thomas, **Adelani**, Davody, Mogadala & Klakow (2020)
   *Investigating the Impact of Pre-trained Word Embeddings on Memorization in Neural Networks*
   In Proceedings of the 23rd International Conference on Text, Speech, and Dialogue (TSD)
   https://hal.inria.fr/hal-02880590

4. **Adelani**, Zhang, Shen, Davody, Kleinbauer & Klakow (2021)
   *Preventing Author Profiling through Zero-Shot Multilingual Back-Translation*
   In Proceedings of the 2021 Conference on Empirical Methods in

Natural Language Processing
https://aclanthology.org/2021.emnlp-main.684/

5. Davody, **Adelani**, Kleinbauer & Klakow (2022)
   *On the effect of normalization layers on Differentially Private training of deep Neural networks*
   Under Submission
   https://arxiv.org/abs/2006.10919

6. Davody, **Adelani**, Kleinbauer & Klakow (2022)
   *TOKEN is a MASK: Few-shot named entity recognition with pretrained language models*
   In Proceedings of the 25th International Conference on Text, Speech, and Dialogue (TSD)
   https://arxiv.org/abs/2206.07841

# GEOGRAPHICAL AND LINGUISTIC CHARACTERISTICS

This chapter provides an overview of the language families, geographical locations, and linguistic characteristics of African languages. We focus on the 31 languages covered in multilingual representation learning and NLP datasets developed during this thesis. 28 of the languages are indigenous to Africa, and the last three are English, French and Arabic—widely spoken on the continent. First, we provide distinguishing characteristics of the different language families in Africa. Second, we discuss the geographic locations of these families, including the population of native speakers and the official languages used in the different African countries. Lastly, we elaborate on their linguistic characteristics such as writing systems, tonality, diacritics, word order, inflectional morphology, and noun classes.

## 2.1 GEOGRAPHICAL LOCATIONS OF LANGUAGES

### 2.1.1 *Categorization by Language Family*

The widely spoken languages in Africa typically belong to six different language families: Afro-Asiatic, Niger-Congo, Nilo-Saharan, Khoisan, Austronesian, and Indo-European. Figure 2.1 shows the geographical locations of the language families in Africa. We provide a few distinguishing characteristics of the language families below:

1. **Niger-Congo**: is the largest language family in Africa by number of speakers and number of languages. Geographically, it stretches from West Africa to East and Southern Africa. According to Ethnologue (Eberhard, Simons, and Fennig, 2021), it comprises over 1,500 languages, of which over 500 are from the Bantu language sub-family category. The most spoken Niger-Congo language is Kiswahili, spoken by over 100M speakers in over 10 East and South-Eastern African countries. It is an official language in four East African languages (Kenya, Tanzania, Uganda, and Rwanda) and the only indigenous African language with official status in the African Union[1]. Other widely spoken Niger-Congo languages are Yorùbá, Fula, and Igbo, with over 35 million native speakers each. The most distinctive characteristic of the Niger-Congo languages is their use of a noun class system (see §2.2). Although there are few exceptions in

---

1 https://au.int/en/about/languages

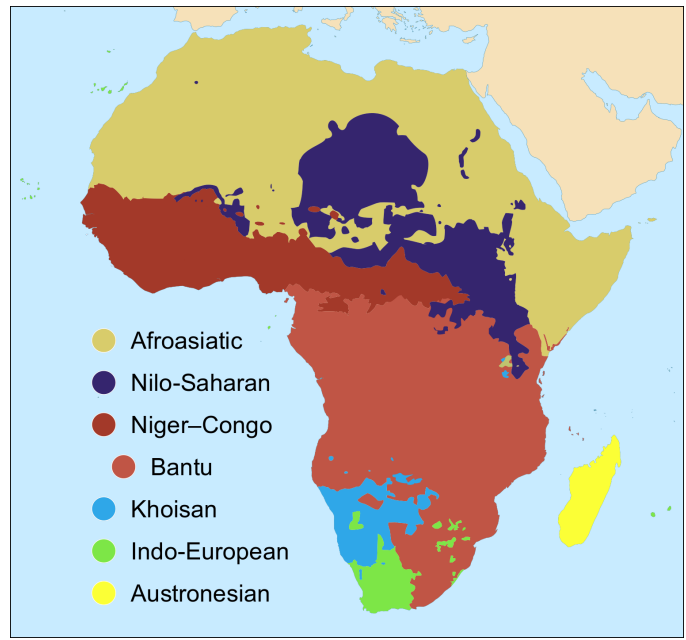

Figure 2.1: **Geographical locations of African language families**. Figure obtained from Wikipedia

West Africa. For example, the Mande and the Ijoid languages do not have noun classes despite being surrounded by languages with this attribute. Also, a large majority of the languages of this family are tonal. Another important characteristic is that many are morphologically-rich (Nichols and Bickel, 2013) or agglutinative, especially the Bantu languages. Although, there are also several isolating languages in the non-Bantu language sub-families like Kru, Gur, and Volta Niger.

2. **Afro-Asiatic** languages are spoken in Western Asia, North Africa, the Horn of Africa, and parts of West and Central Africa. It is geographically located in the Northern region of Africa, stretching from the West coast of Africa to the Red Sea and the Horn of Africa. It is the second biggest language family in Africa, spoken by over 300M people. The major sub-families of the Afro-Asiatic are Berber, Chadic, Cushitic, and Semitic. The languages with the most number of speakers are Arabic (Semitic), Hausa (Chadic), Oromo (Cushitic), Amharic (Semitic), Somali (Cushitic), and Tigrinya (Semitic). Oftentimes, many of these languages make use of different scripts, the popular scripts are: Arabic, Ge'ez, and Latin script. For example, the Berber languages make use of three different scripts: Latin, Arabic, and Libyco-Berber script. Due to the influence of Islam, most countries in the geographical location of the Afro-Asiatic family make use of Arabic as the official language except for Ethiopia. One of the distinct characteristics of Afro-Asiatic languages is prefixing verb conjugation (Voigt, 1987). The prefix may differ for singular or plural forms. They also show evidence of causative affixes. Furthermore, many of them make use of possessive suffixes. The Semitic languages often make use of non-concatenative morphology (Kastner and Tucker, 2019), and often make use of the Verb-Subject-Object (VSO) word order. Although there are exceptions, for example, Amharic, Oromo, and Somali make use of the Subject-Object-Verb (SOV) word order.

3. **Nilo-Saharan** languages are spoken in Central and East Africa and a few parts of West Africa. The largest is probably Dholuo in East Africa, Kanuri in North East Nigeria, and Songhay in West Africa. According to Dimmendaal (2016), some of the most stable characteristics are the causative prefix, number suffixes, reflexive markers, deictic[2] markers for singular or plural forms, and the use of negative verbs.

4. **Khoisan** languages are spoken in the South Western part of Africa in the Kalahari Desert, primarily in Namibia and Botswana. One major characteristic of this family is their extensive use of click sounds on consonants (Traill, 2015). It is probably the smallest language group in terms of population. Khoisan languages make use of the click consonants that are often absent in most African language families. Although, a few Southern-Bantu languages (like isiXhosa and isiZulu) that are geographically close to the Khoisan family have adopted some click consonants.

5. **Austronesian** languages are often found in Maritime Southeast Asia except for Malagasy, which is spoken in Africa by over 18M in Madagascar (Eberhard, Simons, and Fennig, 2021), a large island located in the Indian Ocean, close to Africa's mainland. The Malagasy language is closely related to the Barito languages found in Indonesia. Although, Malagasy has adopted many words from the surrounding Bantu languages and Arabic due to trade influence (The Editors of Encyclopedia Britannica, 2007).

6. **Indo-Europeans** language are widely spoken in Africa mostly due to colonization, which lasted from the 19th century to the late 1960s in most African countries. The Indo-European languages that are often spoken are: English, French, Portuguese, Spanish, and Afrikaans. These languages have official status in nearly all African countries till today. English, French, and Portuguese have official status in 23, 21, and 6 African countries, respectively. Spanish is an official language in only Equatorial Guinea, and Afrikaans is only an official language in South Africa. Indo-European languages are often the language for education, government services and business in Africa. African countries' reliance on Indo-European languages has negatively affected the development and use of indigenous African languages. For example, students are often punished for communicating with indigenous languages in schools (Alebiosu, 2016). Although for business, *creole languages* are often used by locals as an alternative, especially by people who lack formal education. Examples are Nigerian-Pidgin (also known as Naija), Cameroonian-Pidgin, Sheng (a mix of Kiswahili and English), Camfranglais (a mix of French, English and African languages).

---

2 A deictic expression is a word whose meaning varies depending on time or place.

In this thesis, we consider African languages spoken in all language families except for the Khoisan family. We cover 20 Niger-Congo languages, five Afro-Asiatic languages, four Indo-European languages, one Nilo-Saharan, and one Austronesian language in at least one of the following tasks: multilingual representation learning, named entity recognition, and machine translation.

### 2.1.2 *Categorization by Official Status*

Another approach to categorize African languages is by their **official status** in the countries where they are spoken. African languages that are official need to be prioritized in developing NLP applications since they have a large number of speakers. Here, we categorize the African countries based on their official languages, which are typically English, French, Arabic, Portuguese, Spanish, Kiswahili or other African languages. The African Union also recognize the first six languages as official languages. Although most indigenous African languages are not official, they are often regarded as *national* languages in their respective countries. Similarly, a few countries do not have Indo-European or Arabic as official languages for example Ethiopia, Eritrea and Mauritius. In such a case, they still use English, French or Arabic as the working language or language of education. Collecting information about each country's official language or working language is very important in building some NLP applications such as machine translation and question & answering (e.g. using a pivot language that is high-resourced). We provide the categorization below, which is also summarized in Table 2.1 and Figure 2.2.

ANGLOPHONE AFRICA   This refers to the English-speaking countries of Africa. About 21 countries in Africa make use of English as the official language. They are Nigeria, Tanzania, South Africa, Kenya, Uganda, Sudan, Ghana, Cameroon, Malawi, Zambia, Zimbabwe, Rwanda, South-Sudan, Sierra Leone, Liberia, Namibia, The Gambia, Botswana, Lesotho, Eswatini, and Seychelles.

FRANCOPHONE AFRICA   This refers to the French-speaking countries of Africa. About 21 countries in Africa make use of French as the official language. They are the Democratic Republic of Congo, Cameroon, Madagascar, Côte d'Ivoire, Niger, Burkina Faso, Mali, Senegal, Chad, Guinea, Rwanda, Benin, Burundi, Togo, Congo Republic, Central African Republic, Gabon, Equatorial Guinea, Djibouti, Comoros, and Seychelles.

ARABOPHONE AFRICA    This refers to the 11 Arabic-speaking countries of Africa. They are Egypt, Sudan, Algeria, Morocco, Chad, Somali, Tunisia, Libya, Mauritania, Djibouti, and Comoros.

LUSOPHONE AFRICA    This refers to the six Portuguese-speaking countries of Africa. They are Angola, Mozambique, Guinea-Bissau, Equatorial Guinea, Cape Verde and São Tomé and Príncipe.

HISPANOPHONE AFRICA    This refers to Spanish-speaking countries of Africa. Equatorial Guinea is the only African country that makes use of Spanish as the official language.

AFRICAN LANGUAGES WITH OFFICIAL STATUS    A few countries in Africa make use of an indigenous African language as the official language. Kiswahili is an official language in four countries i.e. Kenya, Tanzania, Uganda, and Rwanda. In South Africa, about 10 African languages are official. Similarly, Zimbabwe has about 14 African languages as official, while Rwanda has two African languages i.e. Kinyarwanda and Kiswahili. Also, Lesotho uses Sesotho as an official language. Ethiopia is the only country that does not have any Indo-European language or Arabic as an official language, probably because they were not colonized. They make use of Oromo, Amharic, Somali, Tigrinya, and Afar as official languages. Although, they have adopted English as the language of education after primary school.

NATIONAL LANGUAGES    While many African languages are not official. A subset of them is often categorized as **national** languages especially the most-spoken languages in the country. In some cases, only a few languages (like 1-4) are categorized as "national", for example, Nigeria has three: Hausa, Igbo, and Yorùbá. In some other cases, many languages are "national" e.g. Ghana has 10. Many countries do not have national languages, this concerns about 25 out of 54 African countries. In general, national languages are often prioritized by the government and taught in school, this helps many of them to have a presence on the web. However, non-national languages are often at the risk of being endangered since many natives do not learn how to write them.

In this thesis, we focus on African languages spoken in Anglophone, Francophone, and Arabophone Africa. Figure 2.2 shows the different regions of Africa where English, French and Arabic are official.

|   | Country | Pop (M) | Official Language | National / Regional Lang. |
|---|---|---|---|---|
| 1 | Nigeria | 211.4M | English | Hausa, Yorùbá, Igbo |
| 2 | Ethiopia | 117.8M | Oromo, Amharic, Somali, Tigrinya, Afar | Harari, Sidama |
| 3 | Egypt | 104.3M | Arabic | Egyptian Arabic |
| 4 | DR Congo | 92.4M | French | Kituba, Lingala, Kiswahili, Tshiluba |
| 5 | Tanzania | 61.5M | English, Kiswahili | |
| 6 | South Africa | 60.0M | English, isiZulu, isiXhosa, Afrikaans, Sepedi, Setswana, Sesotho, Xitsonga, siSwati, Tshivenda, isiNdebele | |
| 7 | Kenya | 55.0M | English, Kiswahili | |
| 8 | Uganda | 47.1M | English, Kiswahili | Luganda |
| 9 | Sudan | 44.9M | Arabic, English | |
| 10 | Algeria | 44.6M | Arabic, Berber | |
| 11 | Morocco | 37.3M | Arabic, Berber | |
| 12 | Angola | 33.9M | Portuguese | Umbundu, Kikongo, Kimbundu, Chokwe |
| 13 | Mozambique | 32.0M | Portuguese | |
| 14 | Ghana | 31.7M | English | Twi, Fante, Dagaara, Dagbani, Dangbe, Ewe, Frafra, Ga, Gonja, Nzema, |
| 15 | Cameroon | 27.2M | French, English | Cameroonian Pidgin, Fula, Ewondo, Igbo, Chadian Arabic, Camfranglais |
| 16 | Madagascar | 28.4M | Malagasy, French | |
| 17 | Côte d'Ivoire | 27.1M | French | |
| 18 | Niger | 25.1M | French | Buduma, Fulfulde, Gourmanchéma, Hausa, Kanuri, Zarma, Songhai, Tamasheq, Tassawaq, Tebu |
| 19 | Burkina Faso | 21.5M | French | |
| 20 | Mali | 20.8M | French | Bambara |
| 21 | Malawi | 19.6M | English, Chewa | Tumbuka, Yao, Lomwe, Sena, Tonga, Lambya, and Nyakyusa-Ngonde |
| 22 | Zambia | 18.9M | English | Many: (Most spoken: Bemba, Nyanja, Tonga, Tumbuka, Lozi) |
| 23 | Senegal | 17.2M | French | Wolof, Balanta, Jola-Fonyi, Mandinka, Mandjak, Mankanya, Noon, Pulaar, Serer, and Sonnike |
| 24 | Chad | 16.9M | Arabic, French | |
| 25 | Somalia | 16.4M | Somali, Arabic | |
| 26 | Zimbabwe | 15.1M | Chewa, Chibarwe, English, Kalanga, Tsoa, Nambya, Ndau, Ndebele, Shangani, Shona, sign language, Sotho, Tonga, Tswana, Venda, Xhosa | |
| 27 | Guinea | 13.5M | French | |
| 28 | Rwanda | 13.3M | Kinyawranda, French, English, Kiswahili | |
| 29 | Benin | 12.5M | French | All, most spoken: Fon, Yoruba, Bariba, Dendi, Mokole, Yom |
| 30 | Burundi | 12.3M | Kirundi, French | |
| 31 | Tunisia | 11.9M | Arabic | |
| 32 | South Sudan | 11.4M | English | Dinka, Nuer, Murle, Luo (e.g. Acholi), Ma'di, Otuho, Zande |
| 33 | Togo | 8.5M | French | Ewe, Kabiye |
| 34 | Sierra Leone | 8.1M | English, Krio | |
| 35 | Libya | 7.0M | Arabic | |
| 36 | Congo Republic | 5.7M | French | Kituba, Lingala |
| 37 | Liberia | 5.2M | English | |
| 38 | Central African Republic | 4.9M | French, Sango | |
| 39 | Mauritania | 4.8M | Arabic | Pulaar, Soninke, Wolof |
| 40 | Eritrea | 3.2M | None (Working languages: Tigrinya, Arabic, and English) | Tigrinya, Beja, Tigre, Kunama, Saho, Bilen, Nara, Afar |
| 41 | Namibia | 2.5M | English | Afrikaans, German, Otjiherero, Khoekhoegowab, Oshiwambo, RuKwangali, Setswana, siLozi, IKung, Gciriku, Thimbukushu |
| 42 | The Gambia | 2.5M | English | Mandinka, Pulaar, Wolof, Serer, Jola, Balanta, Hassaniya Arabic, Jola-Fonyi, Mandjak, Mankanya, Noon, Cangin, Dyula, Fula, Karon, Kassonke, Soninke |
| 43 | Botswana | 2.4M | English | Setswana |
| 44 | Gabon | 2.3M | French | Fang, Mbete, Myene, Nzebi, Punu, Teke, Vili |
| 45 | Lesotho | 2.2M | Sesotho, English | |
| 46 | Guinea-Bissau | 2.0M | Portuguese | Guinea-Bissau Creole, Balanta, Hassaniya Arabic, Jola-Fonyi, Mandinka, Mandjak, Mankanya, Noon, Pulaar, Serer, Soninke |
| 47 | Equitorial-Guinea | 1.5M | Spanish, French, Portuguese | Annobonese Creole, Igbo, Bube, Fang, Kombe |
| 48 | Mauritius | 1.3M | None (Working languages: English and French) | |
| 49 | Eswatini | 1.2M | Swazi, English | |
| 50 | Djibouti | 1.0M | Arabic, French | Somali, Afar |
| 51 | Comoros | 0.9M | Comorian, French, Arabic | |
| 52 | Cape Verde | 0.6M | Portuguese | Cape Verdean Creole |
| 53 | São Tomé and Príncipe | 0.2M | Portuguese | Forro, Angolar, Principense |
| 54 | Seychelles | 0.1M | English, French, Seychellois (French-based Creole) | |

Table 2.1: **African countries, their population (Pop (M) in millions), official and national languages (obtained from Wikipedia).** Population estimates were obtained from the World Bank.

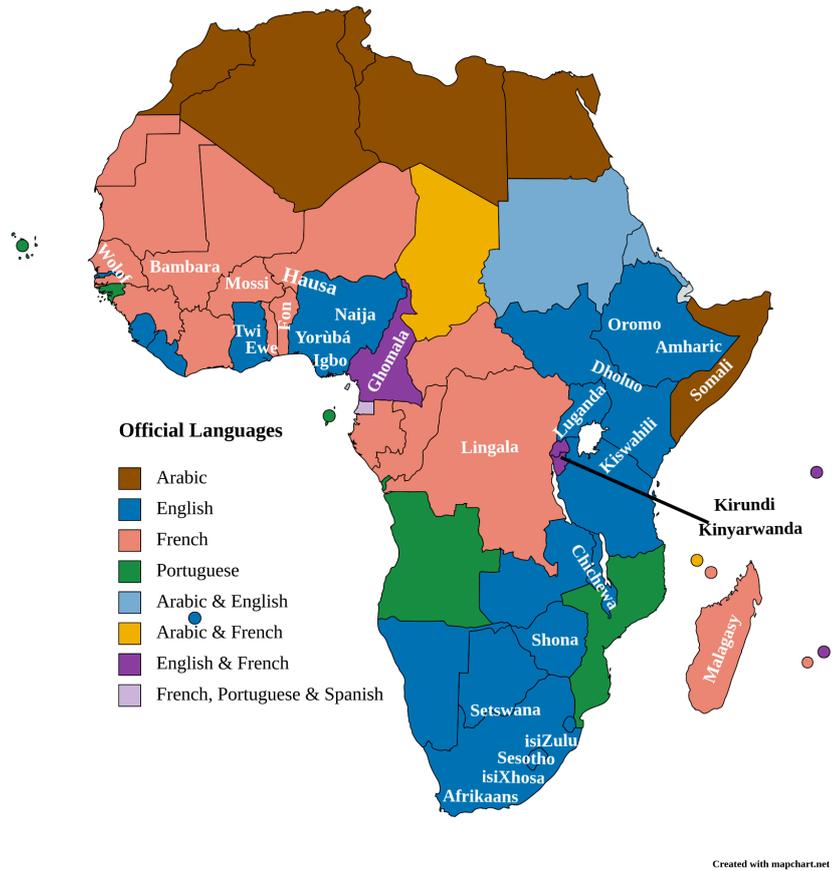

Figure 2.2: **Geographical locations of African languages and Official languages spoken in different countries**.

### 2.1.3 *Categorization by Region*

We can also categorize African languages based on the region of Africa they are native to. Africa is divided into five regions: Northern Africa, West Africa, East Africa, Central Africa, and Southern Africa. In some cases, a few languages are spoken across more than one region. For example, Kiswahili is spoken in East and Central Africa, while Hausa is spoken in West and Central Africa.

Here, we categorize 28 indigenous African languages covered in multilingual representation learning, named entity recognition, news topic classification, and machine translation into different regions. Figure 2.2 shows the regions of Africa where each language is native to. We cover ten languages from West Africa, two from Central Africa, nine languages from East Africa, and seven languages from Southern Africa. We further summarize all the languages, their region, language family and population estimates in Table 2.2.

| Language | Family | African Region | Native Countries | No. of Speakers L1 | L1+L2 |
|---|---|---|---|---|---|
| Afrikaans (afr) | Indo-European/ Germanic | South | South Africa, and Namibia | 7M | 17M |
| Amharic (amh) | Afro-Asiatic / Semitic | East | Ethiopia | 32M | 57M |
| Bambara (bam) | NC / Mande | West | Mali | 4M | 14M |
| Ghomálá' (bbj) | NC / Grassfields | Central | Cameroon | 1M | 1M+ |
| Éwé (ewe) | NC / Kwa | West | Ghana, Togo, and Benin | 5M | 6M |
| Fon (fon) | NC / Volta-Niger | West | Benin, Nigeria, and Togo | 2M | 2M+ |
| Hausa (hau) | Afro-Asiatic / Chadic | West | Nigeria, Niger, Cameroon Chad, Benin, and Ghana | 51M | 77M |
| Igbo (ibo) | NC / Volta-Niger | West | Nigeria | 31M | 31M+ |
| Kinyarwanda (kin) | NC / Bantu | East | Rwanda, Uganda, DR Congo, and Tanzania | 13M | 13M+ |
| Lingala (lin) | NC / Bantu | Central | DR Congo, Rep. of Congo, Central African Republic, Angola and South Sudan | 20M | 40M |
| Luganda (lug) | NC / Bantu | East | Uganda | 6M | 11M |
| Dholuo (luo) | Nilo-Saharan | East | Kenya, and Tanzania | 5M | 5M+ |
| Malagasy (mlg) | Austronesian / Barito | South | Madagascar | 18M | 18M+ |
| Mossi (mos) | NC / Gur | West | Burkina Faso, Côte d'Ivoire, Benin Ghana, Mali, Togo, and Niger | 8M | 8M+ |
| Naija (pcm) | English-Creole | West | Nigeria | 8M | 121M |
| Chichewa (nya) | NC / Bantu | East & South | Malawi, Zambia, Mozambique, and Zimbabwe | 14M | 14M+ |
| Oromo (orm) | Afro-Asiatic / Cushitic | East | Ethiopia, and Kenya | 37M | 37M+ |
| Kirundi (run) | NC / Bantu | East | Burundi | 11M | 11M+ |
| Sesotho (sot) | NC / Bantu | South | Lesotho, South African, and Zimbabwe | 6M | 14M |
| Somali (som) | Afro-Asiatic / Cushitic | East | Ethiopia, Eritrea, Somalia, and Djibouti | 22M | 22M |
| chiShona (sna) | NC / Bantu | South | Zimbabwe, and Mozambique | 7M | 11M |
| Kiswahili (swa) | NC / Bantu | East & Central | Tanzania, Kenya, Comoros, Uganda, Rwanda, Burundi, DR Congo, Somali, Mozambique, Zambia, Malawi, and Madagascar. | 16M | 71-106M |
| Setswana (tsn) | NC / Bantu | South | Botswana, and South Africa | 6M | 14M |
| Akan/Twi (twi) | NC / Kwa | West | Ghana | 8M | 9M |
| Wolof (wol) | NC / Senegambia | West | Senegal, Mauritania, Gambia | 6M | 12M |
| isiXhosa (xho) | NC / Bantu | South | South Africa | 8M | 19M |
| Yorùbá (yor) | NC / Volta-Niger | West | Benin, Nigeria, Togo | 44M | 46M |
| isiZulu (zul) | NC / Bantu | South | South Africa, Lesotho, and Eswatini | 12M | 28M |

Table 2.2: **Languages, Language, family (NC: Niger-Congo), region in Africa, native countries in Africa, number of speakers obtained from Ethnologue**

### 2.1.4 *Categorization by Online Corpora size*

Lastly, we can categorize African languages based on their online corpora size in terms of unlabelled resources and labelled resources, this indicates the "low-resourced"-ness of languages. We follow Joshi's (Joshi et al., 2020) taxonomy of categorizing World languages into six classes: 0-The left-Behinds, 1-The Scraping-Bys, 2-The Hopefuls, 3-The Rising Stars, 4-The Underdogs, and 5-The Winners. They make use of the number of Wikipedia pages as a measure for unlabeled data resources, and LDC catalog[3] and ELRA Map[4] as an indication for labelled datasets. The definitions of the six classes based on Joshi et al. (2020) are below:

1. **0-The Left-Behinds**: they have exceptionally limited resources, for example, Fon, Fulfulde, and Frisian.

2. **1-The Scraping-Bys**: they have some unlabeled data and could be in a better position in a matter of years with an organized movement of collecting labelled datasets. For example, Luganda, Kinyarwanda, and Romansh.

3. **2-The Hopefuls**: These languages have a small set of labelled datasets, and there are researchers and language support communities striving to keep them alive. For example, isiZulu, Yorùbá, and Irish.

4. **3-The Rising Stars**: They have a strong web presence and typically benefit from unsupervised pre-training. Although they have few labelled data. For example, Indonesian, Afrikaans, and Hebrew.

5. **4-The Underdogs**: They have a large amount of unlabeled data, and dedicated NLP communities conducting research on these languages. Although slightly less labelled datasets, compared to the Winners. For example, Russian, Dutch, and Korean.

6. **5-The Winners**: They have a dominant online presence, and there are massive government and industrial investments in the development of the languages. For example, English, French, and Arabic.

Joshi et al. (2020) classify most low-resourced languages in classes 0-3. Afrikaans is the only African language with a class 3. Others are in classes 0-2, including Kiswahili. This shows that most African languages lack labelled resources and large monolingual corpora on the web.

---

[3] https://catalog.ldc.upenn.edu/
[4] http://catalog.elra.info/en-us/

The use of Wikipedia size of a language is often not a true indication of the unlabelled resources of a language. mBERT (Devlin et al., 2019) was trained on the top 104 languages with the largest number of Wikipedia pages, which is about 100GB of data. However, XLM-R was trained on 2.4TB of data by sourcing data from the Common Crawl[5], which is collected from the Web Archive[6]. Common Crawl data with the right language identification gives a better indication of the "low-resourced"-ness of a language ($\forall$ et al., 2020). Therefore, we argue that Joshi's classification can be modified based on the size of unlabelled resources of Common Crawl data. For the Common Crawl size, we make use of CC100-XL (Lin et al., 2021) which is an extension of CC100 used to train XLM-R model. We assign a language $l$ with CC100-XL size $s$ to a class $c_l$ based on this:

$$c_l = \begin{cases} 5, & \text{if } s > 50GB \\ 4, & \text{if } 5GB < s \leq 50GB \\ 3, & \text{if } 500MB < s \leq 5GB \\ 2, & \text{if } 50MB < s \leq 500MB \\ 1, & \text{if } 5MB < s \leq 50MB \text{ or has Wikipedia} \\ 0, & \text{otherwise} \end{cases}$$

Table 2.3 shows the Joshi's classification of the language, Wikipedia size per language, and the common crawl size (CC100-XL).

## 2.2 LINGUISTIC CHARACTERISTICS

Here, we consider the different linguistic characteristics of African languages such as writing systems, tonality, diacritics, word order, inflectional morphology from WALS (Dryer and Haspelmath, 2013), and noun class system.

### 2.2.1 Writing Systems or Script

African languages mainly employ four major writing scripts: Latin, Arabic, N'ko and Ge'ez. Our focus languages mostly use either Latin, Ge'ez or Arabic script. Arabic is the only language in this thesis that make use of the Arabic script. During the pre-colonial period, the Ajami script, a variant of Arabic, was used for Hausa, Kiswahili and Yorùbá but it has now declined in popularity. While N'ko is still actively used by the Mande languages like Bambara, the most widely used writing script for the language is Latin. However, some languages use additional letters that go beyond the standard Latin script, e.g., "ɛ", "ɔ", "ŋ", "ẹ", and more than one character letters like "bv", "gb", "mpf", "ntsh". 22 of the languages are tonal except for Naija, Kiswahili and Wolof. 11 of the languages make use of diacritics (e.g., é, ë, ñ). Table 2.4 provides the summary of the languages, their writing systems[7], tonality and indication of the use of diacritics.

---

5 https://commoncrawl.org/
6 https://web.archive.org/

### 2.2.2 *Word Order*

The four common word orders are Subject-Object-Verb (SOV), Subject-Verb-Object (SVO), Verb-Subject-Object (VSO), and Verb-Object-Subject (VOS). Most of our focus languages make use of the Subject-Verb-Object word order. Bambara uses both SVO and SOV word order. Amharic, Oromo, and Somali make use of the SOV word order. Arabic uses the VSO word order while Malagasy makes use of the VOS word order. The details of the word order are in Table 2.5.

### 2.2.3 *Morphology and Noun classes*

Many African languages are morphologically rich. According to the World Atlas of Language Structures (WALS; Nichols and Bickel, 2013), 22 of our languages employ strong prefixing or suffixing inflections. Niger-Congo languages are known for their system of noun classification. 15 of the languages *actively* make use of between 6–20 noun classes, including all Bantu languages and Ghomálá', Mossi, Akan and Wolof (Babou and Loporcaro, 2016; Bodomo and Marfo, 2002; Nurse and Philippson, 2006; Payne, Pacchiarotti, and Bosire, 2017). While noun classes are often marked using affixes on the headword in Bantu languages, some non-Bantu languages, e.g., Wolof make use of a dependent such as a determiner that is not attached to the headword.

For the other Niger-Congo languages such as Fon, Ewe, Igbo and Yorùbá, the use of noun classes is merely *vestigial* (Konoshenko and Shavarina, 2019). For example, Yorùbá only distinguishes between human and non-human nouns. Bambara is the only Niger-Congo language without noun classes, and some have argued that the Mande family should be regarded as an independent language family. Three of our languages from the Southern Bantu family (chiShona, isiXhosa and isiZulu) capitalize proper names after the noun class prefix as in the language names themselves. This characteristic limits the transfer learning from languages without this feature for some NLP tasks e.g. NER, since NER models overfit on capitalization (Mayhew, Tsygankova, and Roth, 2019). Table 2.5 provides the summary of the word order, inflectional morphology (obtained from WALS) and noun class systems for our focus languages.

---

7 https://omniglot.com/writing

| African Language | Joshi's Class | Wikipedia Size (MB) | CC100-XL Size (MB) | New Suggested Class |
| --- | --- | --- | --- | --- |
| English (eng) | 5 | 19,000 | 3,324,450 | 5 |
| French (fra) | 5 | 5,100 | 303,760 | 5 |
| Arabic (ara) | 5 | 1,300 | 64,340 | 5 |
| Kiswahili (swa) | 2 | 43.8 | 3,190 | 3 |
| Afrikaans (afr) | 3 | 157.4 | 3,040 | 3 |
| Somali (som) | 1 | 10.7 | 1,560 | 3 |
| Amharic (amh) | 2 | 17.5 | 850 | 3 |
| Hausa (hau) | 2 | 36.7 | 420 | 2 |
| Malagasy (mlg) | 1 | 48.7 | 380 | 2 |
| Yorùbá (yor) | 2 | 7.7 | 140 | 2 |
| isiZulu (zul) | 2 | 4.6 | 140 | 2 |
| Oromo (orm) | 1 | 2.2 | 90 | 2 |
| isiXhosa (xho) | 2 | 1.8 | 90 | 2 |
| Igbo (ibo) | 1 | 3.8 | 60 | 2 |
| Luganda (lug) | 1 | 4.0 | 50 | 1 |
| Wolof (wol) | 2 | 2.9 | 30 | 1 |
| Setswana (tsn) | 2 | 1.3 | 30 | 1 |
| Lingala (lin) | 1 | 1.0 | 20 | 1 |
| Bambara (bam) | 1 | 0.3 | 20 | 1 |
| Kinyarwanda (kin) | 1 | 3.6 | – | 1 |
| chiShona (sna) | 1 | 3.4 | – | 1 |
| Chichewa (nya) | 1 | 1.2 | – | 1 |
| Kirundi (run) | 1 | 0.4 | – | 1 |
| Sesotho (sot) | 1 | 0.6 | – | 1 |
| Éwé (ewe) | 1 | 0.2 | – | 1 |
| Akan/Twi (twi) | 1 | 2.6 | – | 1 |
| Dholuo (luo) | 0 | – | – | 0 |
| Mossi (mos) | 0 | – | – | 0 |
| Fon (fon) | 0 | – | – | 0 |
| Naija (pcm) | 0 | – | – | 0 |
| Ghomálá' (bbj) | – | – | – | 0 |

Table 2.3: **Languages and Online Corpora Size**. Language, Joshi's class (Joshi et al., 2020), Wikipedia size (MB), CC-100-XL (MB) and a new taxonomy class modification of Joshi's class. 0-The left-Behinds, 1-The Scraping-Bys, 2-The Hopefuls, 3-The Rising Stars, 4-The Underdogs, and 5-The Winners.

| Language | No. of Letters | Latin Letters Omitted | Letters added | Tonality | diacritics |
|---|---|---|---|---|---|
| Amharic (amh) | 33 | – | Ge'ez alphabet | no | no |
| Afrikaans (afr) | 26 | – | – | no | yes |
| Bambara (bam) | 27 | q,v,x | ɛ, ɔ, ɲ, ŋ | yes, 2 tones | yes |
| Ghomálá' (bbj) | 40 | q, w, x, y | bv, dz, ə, aə, ɛ, gh, ny, nt, ŋ, ŋk, ɔ, pf, mpf, sh, ts, ʉ, zh, ' | yes, 5 tones | yes |
| Éwé (ewe) | 35 | c, j, q | ɖ, dz, ɛ, ƒ, gb, ɣ, kp, ny, ŋ, ɔ, ts, ʋ | yes, 3 tones | yes |
| Fon (fon) | 33 | q | ɖ, ɛ,gb, hw, kp, ny, ɔ, xw | yes, 3 tones | yes |
| Hausa (hau) | 44 | p,q,v,x | ɓ, ɗ, ƙ, yˈ, kw, ƙw, gw, ky, ƙy, gy, sh, ts | yes, 2 tones | no |
| Igbo (ibo) | 34 | c, q, x | ch, gb, gh, gw, kp, kw, nw, ny, ọ, ȯ, sh, ụ | yes, 2 tones | yes |
| Kinyarwanda (kin) | 30 | q, x | cy, jy, nk, nt, ny, sh | yes, 2 tones | no |
| Lingala (lin) | 40 | j, q, x | ɛ, gb, kp, mb, mf, mp, mv, nd, ng, ngb, nk, ns, nt, ny, nz, ɔ, ts | yes, 2 tones | no |
| Luganda (lug) | 25 | h, q, x | ŋ, ny | yes, 3 tones | no |
| Dholuo (luo) | 31 | c, q, x, v, z | ch, dh, mb, nd, ng', ng, ny, nj, th, sh | yes, 4 tones | no |
| Mossi (mos) | 26 | c, j, q, x | ', ɛ, ɩ, ʋ | yes, 2 tones | yes |
| Malagasy (mlg) | 21 | c, q, u, w, x | – | no | yes |
| Chichewa (nya) | 31 | q, x, y | ch, kh, ng, ŋ, ph, tch, th, ŵ | yes, 2 tones | no |
| Oromo (orm) | 32 | – | ch, dh, ny, ph, sh, th, ' | yes, 2 tones | no |
| Naija (pcm) | 26 | – | – | no | no |
| Kirundi (run) | 30 | q, x | cy, jy, nk, nt, ny, sh | yes, 2 tones | no |
| Sesotho (sot) | 29 | c, l, q, x | bh, ch, dh, nh, sh, vh, zh | yes, 2 tones | no |
| Shona (sna) | 29 | c, l, q, x | bh, ch, dh, nh, sh, vh, zh | yes, 2 tones | no |
| Somali (som) | 29 | c, l, q, x | bh, ch, dh, nh, sh, vh, zh | yes, 2 tones | no |
| Kiswahili (swa) | 33 | x, q | ch, dh, gh, kh, ng', ny, sh, th, ts | no | yes |
| Setswana (tsn) | 36 | c, q, v, x, z | ê, kg, kh, ng, ny, ô, ph, š, th, tl, tlh, ts, tsh, tš, tšh | yes, 2 tones | no |
| Akan/Twi (twi) | 22 | c,j,q,v,x,z | ɛ, ɔ | yes, 5 tones | no |
| Wolof (wol) | 29 | h,v,z | ŋ, à, é, ë, ó, ñ | no | yes |
| isiXhosa (xho) | 68 | – | bh, ch, dl, dy, dz, gc, gq, gr, gx, hh, hl, kh, kr, lh, mh, ng, ngc, ngh, ngq, ngx, nkq, nkx, nh, nkc, nx, ny, nyh, ph, qh, rh, sh, th, ths, thsh, ts, tsh, ty, tyh, wh, xh, yh, zh | yes, 2 tones | no |
| Yorùbá (yor) | 25 | c, q, v, x, z | ẹ, gb, ṣ , ọ | yes, 3 tones | yes |
| isiZulu (zul) | 55 | – | nx, ts, nq, ph, hh, ny, gq, hl, bh, nj, ch, ngc, ngq, th, ngx, kl, ntsh, sh, kh, tsh, ng, nk, gx, xh, gc, mb, dl, nc, qh | yes, 3 tones | no |

Table 2.4: **Linguistic Characteristics of the Languages.** Language, Letters, Tonality and Diacritics.

| Language | Word Order | Morphological typology | Inflectional Morphology (WALS) | Noun Classes |
| --- | --- | --- | --- | --- |
| Afrikaans (afr) | SVO | mostly analytic | strongly suffixing | absent |
| Amharic (amh) | SOV | agglutinative | weakly suffixing | absent |
| Bambara (bam) | SVO & SOV | isolating | strongly suffixing | absent |
| Ghomálá' (bbj) | SVO | agglutinative | strong prefixing | active, 6 |
| Éwé (ewe) | SVO | isolating | equal prefixing and suffixing | vestigial |
| Fon (fon) | SVO | isolating | little affixation | vestigial |
| Hausa (hau) | SVO | agglutinative | little affixation | absent |
| Igbo (ibo) | SVO | agglutinative | little affixation | vestigial |
| Kinyarwanda (kin) | SVO | agglutinative | strong prefixing | active, 16 |
| Lingala (lin) | SVO | agglutinative | strong prefixing | active, 15 |
| Luganda (lug) | SVO | agglutinative | strong prefixing | active, 20 |
| Dholuo (luo) | SVO | agglutinative | equal prefixing and suffixing | absent |
| Malagasy (mlg) | VOS | agglutinative | little affixation | absent |
| Mossi (mos) | SVO | isolating | strongly suffixing | active, 11 |
| Chichewa (nya) | SVO | agglutinative | strong prefixing | active, 17 |
| Naija (pcm) | SVO | mostly analytic | strongly suffixing | absent |
| Oromo (orm) | SOV | agglutinative | strongly suffixing | absent |
| Kirundi (kin) | SVO | agglutinative | strong prefixing | active, 16 |
| Shona (sna) | SVO | agglutinative | strong prefixing | active, 20 |
| Somali (som) | SOV | agglutinative | strongly suffixing | absent |
| Sesotho (sot) | SVO | agglutinative | strong prefixing | active, 15 |
| Kiswahili (swa) | SVO | agglutinative | strongly suffixing | active, 18 |
| Setswana (tsn) | SVO | agglutinative | strong prefixing | active, 18 |
| Akan/Twi (twi) | SVO | isolating | strong prefixing | active, 6 |
| Wolof (wol) | SVO | agglutinative | strongly suffixing | active, 10 |
| isiXhosa (xho) | SVO | agglutinative | strong prefixing | active, 17 |
| Yorùbá (yor) | SVO | isolating | little affixation | vestigial, 2 |
| isiZulu (zul) | SVO | agglutinative | strong prefixing | active, 17 |

Table 2.5: **Linguistic Characteristics of the Languages.** Word Order, Morphology, and Noun classes

# THE STATE OF NLP FOR AFRICAN LANGUAGES  3

This chapter describes the state of NLP for African languages based on the available NLP resources needed to develop NLP models. We focus on a survey of available datasets (unlabelled and labelled corpora), and pre-trained models (word embeddings and multilingual pre-trained language models) needed to build NLP models. Furthermore, we demonstrate empirically the limitations of word embeddings and the opportunities the multilingual pre-trained language model offers, especially for languages unseen during pre-training and low-resource training samples scenario. Lastly, we highlight current efforts by AI/NLP communities in Africa to address the under-representation of African languages in NLP.

## 3.1 NLP DATASETS FOR AFRICAN LANGUAGES

### 3.1.1 *Unlabelled Corpora*

Nowadays, many NLP models that are developed rely on pre-trained models to achieve impressive performance, especially in the absence of large training data. However, these pre-trained models are trained on large unlabelled texts which are often not available for many low-resource languages. The lack of a large monolingual corpus is one of the major challenges in developing NLP models for African languages. We describe the common unlabelled corpora that are available for African languages below:

BIBLE   The entire Bible books have been translated into over 700 languages, and portions of the Bible have been translated into over 2200 languages[1]. Bible can be regarded as the most available resource for all languages, including low-resourced and endangered languages (McCarthy et al., 2020; Resnik, Olsen, and Diab, 1999). In Africa, the Bible has been translated into over 750 languages [2]. The Bible provides a few thousand sentences (around 8,000 - 30,000 verses depending on the language) that can be used for several NLP tasks like training word embedding and machine translation models (due to the availability of parallel translation in several languages).

JEHOVAH WITNESS PUBLICATIONS AND JW300   Apart from the Bible, the second largest corpus for African languages can be obtained

---
[1] https://www.wycliffe.net/resources/statistics/
[2] http://missionscatalyst.net/?p=5326

from the Jehovah Witness website (`jw.org`). Agić and Vulić ([2019](#)) created **JW300** by automatically aligning the sentences of the `jw.org` to create a multilingual parallel translation dataset for over 300 languages. The vast majority of texts are from *Awake!* and *Watchtower* magazines that were mainly translated from the English language to other languages. The JW300 corpus covers over 100 African languages. Unlike the Bible with only a few thousand sentences, there are about 25 African languages (`afr, xho, swa, zul, mlg, tsn, nya, sna, tso, amh, ewe, nso, twi, lin, kin, ibo, yor, bem, run, efi, hau, lug, umb, mos, tiv`) with over 200K parallel sentences (with English as the source language) which makes it ideal for representation learning (Gowda et al., 2021). Although, parallel sentences from one African language to another are usually smaller. Unfortunately, JW300 is no longer publicly available due to copyright issues.

WIKIPEDIA    The Wikipedia is a multilingual platform with volunteer contributors from around the world. Wikipedia currently supports 314 languages[3] including 36 African languages[4]. However, most of the African languages have few articles in general. Afrikaans, Malagasy, and Kiswahili have the largest number of articles with 104K, 95K, and 74K respectively. Others are smaller and often not following the right orthography. In general, the numbers of articles from Wikipedia for different languages are much smaller than that of JW300. Although one advantage of Wikipedia is that it is in the general domain, unlike JW300 and the Bible, which are in the religious domain.

FILTERED COMMON CRAWL CORPUS    Recently, researchers and Big Tech companies like Google and Meta (formerly Facebook) began to filter the large collection of multilingual Web Archive to create large monolingual corpora for several languages. This involves building language identification tools to classify a large collection of web texts into different languages, followed by additional filtering to remove non-language or toxic contents from each language monolingual corpus. XLM-R (Conneau et al., 2020), MT5 (Xue et al., 2021) and XGLM (Lin et al., 2021) language models were pre-trained on **CC-100** (with 100 languages), **mC4** (with 101 languages), and **CC100-XL** (with 134 languages), respectively. The corpora are often released alongside the pre-trained language models. However, the CC100-XL is currently not publicly available. Despite the effort to filter and clean monolingual corpus for different languages, Kreutzer et al. (2021) analyzed the data and found that the extracted corpus for low-resource languages, especially African languages, is of poor quality. To address this, Ortiz Suárez, Sagot, and Romary (2019) and Abadji et al. (2022) released **OSCAR** – a cleaner version of the Common Crawl data, but it only includes a few African languages, and the ones that are represented have few sentences. For example, Afrikaans, Kiswahili, and Yorùbá have 47MB, 1.3MB, and 24.7KB sizes, respectively, while in CC100-XL, their sizes are 3GB, 3GB and 140MB.

---

3 https://en.wikipedia.org/wiki/List_of_Wikipedias
4 https://meta.wikimedia.org/wiki/African_languages

NEWS CORPUS    While filtered common crawl data are large, they are of poor quality for many African languages. A cleaner alternative which is also of a large size for a few African languages is the *news corpus*. A few popular sources are below:

1. **Voice of America (VOA[5])** is available in 11 African languages (`orm, amh, bam, hau, kin/run, lin, nde, sna, som, swa, tir`), and a large scrapped data of these languages are now publicly available (Palen-Michel, Kim, and Lignos, 2022).

2. **The British Broadcasting Corporation (BBC)** publishes in 10 African languages (`orm, amh, hau, ibo, kin/run, pcm, som, swa, tir, yor` ).

3. **Voice of Nigeria (VON)** publishes in four Nigerian languages and three foreign languages(`eng, fra, ara, hau, ibo, yor, fuv`)

4. **Global Voices** is a platform that hosts news articles submitted by volunteers contributors and journalists from around the world. Five African languages (amh, mlg, yor, swa, and ibo) have articles on their platform.

5. **Deutsche Welle (DW)** publishes in three African languages (amh, hau, and swa).

6. **Isolezwe** publishes in two South African languages: isiZulu[6] (zul) and isiXhosa[7] (xho).

7. **Other News Corpus**: There are many other news corpus sources which are often known by native speakers of the language. For example, Hausa (Legit[8] , Leadership[9], Premium Times[10], NNN[11], Daily Trust[12], and RFI[13], and others) and Yorùbá (Alaroye[14], Asejere[15], and Iroyin Awikonko[16]) newspapers.

---

5 https://www.voanews.com/
6 https://www.isolezwe.co.za/
7 https://www.isolezwelesixhosa.co.za/
8 https://hausa.legit.ng/
9 https://hausa.leadership.ng/
10 https://hausa.premiumtimesng.com/
11 https://nnn.ng/hausa/
12 https://aminiya.dailytrust.com/
13 https://www.rfi.fr/ha/duniya/
14 https://alaroye.org/
15 https://www.asejere.net/
16 https://www.awikonko.com.ng/

OTHER MONOLINGUAL CORPORA  There are a few other websites with some African language monolingual data like SADiLAR[17] (mostly resources for South African languages), Leipzig University corpus[18], LDC[19] (which requires payment), and HuggingFace Dataset (Lhoest et al., 2021) Hub[20].

3.1.2 *Labelled Corpora*

African languages often lack evaluation datasets in many NLP tasks. We consider NLP tasks (e.g. sentiment classification and natural language inference) that are often included in the popular benchmark datasets like XTREME-R (Ruder et al., 2021) – a benchmark for 50 typologically diverse languages, and language-specific benchmarks: English GLUE (Wang et al., 2018), French FLUE (Le et al., 2020), Chinese CLUE (Xu et al., 2020), Korean KLUE (Park et al., 2021), and Indonesian IndoNLU (Wilie et al., 2020) benchmark datasets. We survey African language labelled datasets in 10 popular NLP tasks below.

NER  NER is a classification task that identifies words in a text that refer to entities (such as dates, person, organization and location names). While the dataset exists in several languages, only a few African languages have NER datasets before this thesis. The largest dataset is WikiAnn corpus (Pan et al., 2017) covering 282 languages, the dataset consist of "silver-standard" labels created by transferring annotations from English to other languages through cross-lingual links in knowledge bases. However, only nine African languages are represented, most have fewer than 10k tokens (since African language articles are few on Wikipedia). Other NER datasets for African languages include SADiLAR (Eiselen, 2016) for ten South African languages based on government data, Amharic[21], Yorùbá (Alabi et al., 2020), Hausa (Hedderich et al., 2020), and Tigrinya (Yohannes and Amagasa, 2022). Additionally, the LORELEI language packs (Strassel and Tracey, 2016) include some African languages (Yorùbá, Hausa, Amharic, Somali, Twi, Kiswahili, Wolof, Kinyarwanda, and Zulu), but are not publicly available. In this thesis, we developed new NER datasets known as MasakhaNER (Chapter 7) for ten languages, and we also extended it to 21 languages (Chapter 8) to address the lack of NER evaluation datasets for African languages.

POS AND UD  Universal Dependencies (UD) (Nivre et al., 2016) is a collection of consistent annotation of grammar (parts of speech (POS), morphological features, and syntactic dependencies) across different human languages. UD dataset is only available for seven African languages [22]: Afrikaans (Augustinus et al., 2016), Amharic (Seyoum, Miyao, and Mekonnen, 2018), Bambara (Aplonova and Tyers, 2017), Beja (Kahane et al., 2021), Wolof (Dione, 2019), Yorùbá (Ishola and Zeman, 2020), and Naija (Caron et al., 2019). Although there are other POS datasets not part of the UD like Igbo POS (Onyenwe et al., 2019) and Setswana (Malema and Ishmael, 2022).

---

17 https://repo.sadilar.org/handle/20.500.12185/1
18 https://corpora.uni-leipzig.de/en
19 https://catalog.ldc.upenn.edu/
20 https://huggingface.co/datasets
21 https://github.com/uhh-lt/amharicmodels/tree/master/data/NER

NEWS TOPIC CLASSIFICATION   This involves the categorization of news titles or articles into several topics such as "health", "sports", "politics", and other pre-defined topics. A few African language datasets exist in this task [23], like Hausa and Yorùbá datasets from Hedderich et al. (2020), Amharic dataset (Azime and Mohammed, 2021), and Kinyarwanda & Kirundi datasets (Niyongabo et al., 2020). We developed new datasets for four African languages (Lingala, Naija, Somali, Malagasy, and isiZulu) in this thesis (Chapter 5).

SENTIMENT CLASSIFICATION   Sentiment classification is a popular text classification task to determine the sentiment or opinion expressed in a piece of text. The sentiments are often categorized into "positive", "negative", or "neutral". A few datasets in the Twitter domain exist for Amharic (Yimam et al., 2020) and Nigerian languages (Hausa, Igbo, Yorùbá and Naija) (Muhammad et al., 2022). [24] There is a recent one for movie reviews known as YOSM (Shode, Adelani, and Feldman, 2022). [25]

MACHINE TRANSLATION   Machine translation (MT) involves the automatic translation of sentences from a source language to a target language. One major challenge in MT development is the lack of parallel sentences for many language pairs in the world. The situation is more direr for low-resource languages, this has led to research on automatically aligned parallel sentences from web archive like Common Crawl but they are often of poor quality (Kreutzer et al., 2021). Examples of automatically aligned parallel sentences are WikiMatrix (Schwenk et al., 2021a), CCMatrix (Schwenk et al., 2021b), CC-Aligned (El-Kishky et al., 2020), but they often include few African languages. WikiMatrix, CCMatrix, and CC-Aligned have two, 14, and 20 African languages, respectively. Other alternatives that are cleaner are the Bible and JW300. Human translations have also been created for African languages by native speakers and African ML/NLP communities like Masakhane (∀ et al., 2020), GhanaNLP (Azunre et al., 2021b) etc. However, the curated sentences are often of smaller size because human translation is very costly. Apart from the training data, there is also an effort to create an evaluation dataset for diverse languages. Recently, Meta (formerly Facebook) released Flores-101 (Goyal et al., 2022), which is a many-to-many evaluation benchmark (with 3001 sentences per language) for 101 languages, and this has been expanded to 200 languages (NLLB-Team et al., 2022). This makes it possible to evaluate on any pair from the 200 languages covered in the dataset. We highlight some of the MT datasets in Table 3.1 i.e OPUS (Tiedemann, 2012), ParaCrawl (Bañón et al., 2020), WikiMatrix, CCMatrix, CCAligned, JW300, JHU Bible Corpus (McCarthy et al., 2020), IgboNLP (Ezeani et al., 2020), FFR v1.1 (Emezue and Dossou, 2020), BAM-FRA (Tapo et al., 2020), AI4D-MT (Siminyu et al., 2021), AfroNMT (Lakew, Negri, and Turchi, 2020), and Autshumato (McKellar, 2014).

---

[22] A few months after the submission of my thesis, we developed MasakhaPOS (Dione et al., 2023) for 20 African languages through collaboration with Masakhane.

[23] Similarly, for news topic classification, a few months after the submission of my thesis, we developed MasakhaNEWS (Adelani et al., 2023) for 14 African languages through participatory research with Masakhane

[24] With collaboration with Masakhane, NaijaSenti was extended to AfriSenti (Muhammad et al., 2023) to cover 14 African languages in 2023.

[25] Similarly, YOSM has been extended to NollySenti (Shode et al., 2023) for four Nigerian languages: Hausa, Igbo, Naija, and Yorùbá.

| Dataset name | Type | # Africa lang | African languages |
| --- | --- | --- | --- |
| OPUS | multi-way | 40+ | too many for the table |
| WikiMatrix | multi-way | 2 | mlg, swa |
| ParaCrawl | bilingual | 2 | som, swa |
| CCMatrix | multi-way | 2 | afr, amh, bam, hau, ibo, lug, mlg, orm, som, swa, wol, xho, yor, zul |
| CCAligned | multi-way | 2 | afr, amh, bam, ful, hau, ibo, lin, lug, mlg, orm, sna, som, sot, ssw, swa, tig, wol, xho, yor, zul |
| JW300 | multi-way | 100+ | too many for the table |
| JHU Bible Corpus | multi-way | 500+ | too many for the table |
| Flores-101 | multi-way | 20 | afr, amh, ful, hau, ibo, kam, lin, lug, luo, nso, nya, orm, sna, som, swa, umb, wol, xho, yor, zul |
| Flores-200 | multi-way | 55 | too many for the table |
| IgboNLP | bilingual | 1 | ibo |
| FFR v1.1 | bilingual | 1 | fon |
| BAM-FRA | bilingual | 1 | bam |
| AI4D-MT | bilingual | 5 | ewe, fon, lug, twi, yor |
| AfroNMT | bilingual | 5 | amh, orm, som, swa, tir |
| Autshumato | bilingual | 10 | afr, nbl, nso, sot, ssw, tsn, tso, ven, xho, zul |
| Created in this thesis | | | |
| MENYO-20k (Chapter 9) | bilingual | 1 | yor |
| MAFAND-MT(Chapter 10) | bilingual | 21 | amh, bam, bbj, ewe, fon, hau, ibo, kin, lug, luo, mos, nya, sna, swa, tsn, twi, wol, xho, yor, zul |

Table 3.1: **Machine Translation Benchmark Datasets**. Language, Type (either bilingual with one language pair or multi-way dataset with a language having multiple target languages, number of African languages, and list of African languages present in the dataset.

TEXT SUMMARIZATION   Summarization is a text generation task for generating a summary of an article. There are two approaches: *extractive* and *abstractive* summarization. Extractive summarization cuts out some segments of the article and concatenates them to produce a summary, while abstractive summarization generates summaries that may contain phrases that are absent in the article – similar to paraphrasing the article in a short form. There are two popular datasets for the abstractive summarization task that includes African languages, they are XL-Sum (Hasan et al., 2021) and MassiveSumm (Varab and Schluter, 2021). XL-Sum was created by crawling article-summary pairs from the BBC website covering 44 languages, including 10 African languages (amh, hau, ibo, orm, pcm, run, som, swa, tir, yor). MassiveSumm covers more languages, and 18 African languages (afr, amh, bam, ful, hau, ibo, nde, kin, lin, mlg orm, run, som, sna, swa, tir, xho, yor).

QUESTION & ANSWERING (QA)   QA involves providing an automatic answer to the questions of users. The only dataset that covers an African language is the TyDi-QA – a typologically diverse QA dataset covering 11 languages including Kiswahili (swa). [26]

NLI   The purpose of the natural language inference (NLI) task is to detect if two sentences entail, contradict or are neutral to each other. The original NLI dataset was created for English (MultiNLI) (Gururangan et al., 2018), and some portions of the dataset were later translated into 14 languages, known as the Cross-lingual NLI (XNLI) (Conneau et al., 2018) dataset. The only African language in XNLI is Kiswahili.

CAUSAL COMMONSENSE REASONING   The goal of the causal commonsense reasoning task is to decide, out of two sentences which one causally follows a premise sentence. The original dataset for this task was created for English, known as the COPA (Gordon, Kozareva, and Roemmele, 2012) dataset. (Ponti et al., 2020) later created the multilingual version known as (XCOPA) by translating the development and test sets of the English dataset into 11 languages. Similar to XNLI, Kiswahili is the only African language covered.

SLOT-FILLING AND INTENT DETECTION   The task is very important for the development of several Conversation AI applications like Amazon Alexa, and Apple Siri. Only a few African languages have slot-filling and intent detection datasets. The largest available dataset that covers many languages is MASSIVE-1M (FitzGerald et al., 2022), but it includes only three African languages: Afrikaans, Amharic, and Kiswahili.

---

26 Masakhane in 2023 created AfriQA (Ogundepo et al., 2023) using similar methodology to TyDi-QA

## 3.2 WORD EMBEDDING

A fundamental research problem in NLP is how to represent texts mathematically and use this representation to solve several NLP tasks. This representation can be in different granularities, it could be sentence representation, word representation, or character representation. Some approaches even consider sub-word units rather than words. To answer this, several methods have been developed to learn such a representation, and the most successful approach is based on **word embedding or representation**. A **word embedding** can be regarded as a fixed-sized *vector* (e.g 300-dimension), where each dimension's value corresponds to a feature that might have a semantic or grammatical interpretation (Turian, Ratinov, and Bengio, 2010). To learn a good embedding for a word that captures both semantic and syntactic relationships, previous works take inspiration from the quotes from *distributional hypothesis* which say: "*words that are used and occur in the same contexts tend to purport similar meanings*" (Harris, 1954), and ""*a word is characterized by the company it keeps* (Firth, 1957)". Bengio, Ducharme, and Vincent (2000) make use of a feed-forward neural network model to learn embedding for each word in a vocabulary of a large corpus by training the model to predict the next word given a fixed set of context words (or previous words). Other novel architectures that better capture distributional hypothesis are Word2Vec (Mikolov et al., 2013a,b) and GloVe (Pennington, Socher, and Manning, 2014). We discuss the architectures below.

### 3.2.1 *Word Embedding Architectures*

WORD2VEC   Word2Vec consist of two architectures (Mikolov et al., 2013a): Continuous Bag-of-Words (CBOW) and the Skip-gram models. In the CBOW, the left and right context words are trained to predict a center word using a feed-forward neural network, ignoring the order of context words – this explains the reason for the name "bag-of-words". The vector representation of surrounding words that are good enough to predict the center word are the learned word embeddings. In contrast, Skip-gram is trained in a reverse way, where a center word is trained to predict the surrounding left and right context words. The Skip-gram model was trained for the English language on 1 billion word corpus (Mikolov et al., 2013b), the resulting word embedding is a 300-dimension vector for each word in the vocabulary.

GLOVE   Word2Vec makes use of only local (or surrounding) contexts to learn word vectors but poorly utilize the statistics of the corpus, like word-to-word co-occurrence counts since they are trained on a separate local context windows (with predefined context size e.g 5 or 10). GloVe on the other hand, makes use of the global log-bilinear regression model[27] to learn word embeddings from the non-zero elements of a global co-occurrence counts rather than on the entire sparse matrix of co-occurrence counts or on individual context windows in a large corpus (like Skip-gram).

FASTTEXT    While Word2Vec and GloVe produce high-quality embedding for frequently occurring words in a large corpus, they often ignore rare words in the corpus and morphology of words, leading to a large Out-of-vocabulary (OOV) counts when trained for morphologically-rich languages (e.g. Turkish or isiZulu) or applied to new domains (e.g. medical or legal domains). To address this OOV problem, Bojanowski et al. (2017) trained a word embedding based on the Word2Vec architectures, but instead of training on words, they trained on character $n$-grams (e.g. $n \in [3, 6]$). With this approach, it is possible to have an embedding for a new word not found in the vocabulary, the embedding for a new word is computed from the embedding of the character $n$-grams. Apart from training FastText embedding for English, the authors also trained FastText for 294 languages in Wikipedia, and 157 languages identified from Common Crawl. This provides an important resource for many world languages. However, the quality is often lower for low-resource languages. Therefore, we performed an analysis of the quality of pre-trained FastText embedding on two African languages (see, Chapter 4).

COVE    Word2Vec, GloVe, and FastText embeddings are static word embeddings, i.e. a word has only one embedding. However, in practice, a word can mean different things in different contexts. For example, the word "bank" could mean "a financial organization" or "river bank". This calls for a focus on *contextualized word embedding*, where the embedding of a word differs depending on the context or sentence it occurs. CoVe (McCann et al., 2017) propose extracting contextualized embedding from a deep Long short-term memory (LSTM) (Hochreiter and Schmidhuber, 1997) encoder from sequence-to-sequence model trained for machine translation (MT). They showed that the context vectors (CoVe) improve performance over static word embedding (like GloVe) in several NLP tasks like sentiment analysis, question classification, entailment, and question answering tasks.

ELMO    ELMo builds on this idea of contextualized word embedding by learning word vectors from the internal states of a bidirectional LSTM language model (biLM) trained on a large monolingual corpus. They are exclusively trained on character $n$-grams. The trained biLM model provides three layers of representations for each input token, [28] also suitable for OOV words, due to the use of only character input. In contrast, static word embedding methods only provide one layer of representation for tokens in a fixed vocabulary.

---

27  This is formulated as a weighted least squares regression model
28  three layers of representation since it was trained using two biLSTM layers and a linear projection layer.

### 3.2.2 *Word embedding for African languages*

We create FastText embeddings for 21 African languages on curated monolingual corpus because of the poor quality of the pre-trained FastText embeddings[29] or their unavailability on the web. They are trained on curated monolingual corpus from clean sources like news articles (e.g. BBC, VOA or translated news from MAFAND-MT (Chapter 10) and religious texts (like Bible and JW300), MT560 (Gowda et al., 2021) on OPUS, and AI4D corpus (Siminyu et al., 2021). The trained FastText embeddings are available on Zenodo[30].

MONOLINGUAL CORPUS   Table 3.2 shows the curated monolingual corpus we used for training the FastText embeddings for the 21 languages.

| Language | Size (M) | Monolingual Corpus Source |
| --- | --- | --- |
| Amharic (amh) | 182.7 | MT560, Wikipedia, VOA, BBC |
| Bambara (bam) | 4.5 | Bible, MAFAND-MT |
| Ghomálá' (bbj) | 1.5 | Bible, MAFAND-MT |
| Éwé (ewe) | 64.1 | JW300, Wikipedia, MAFAND-MT |
| Fon (fon) | 4.8 | JW300, MAFAND-MT |
| Hausa (hau) | 115.1 | JW300, Wikipedia, VOA, BBC, and Voice of Nigeria |
| Igbo (ibo) | 70.7 | JW300, Wikipedia, BBC |
| Kinyarwanda (kin) | 161.3 | JW300, Wikipedia, VOA, KINNEWS, BBC |
| Luganda (lug) | 36.9 | JW300, Wikipedia, Bukkedde news |
| Dholuo (luo) | 14.2 | JW300, Ramoji news, MAFAND-MT |
| Mossi (mos) | 22.3 | JW300, MAFAND-MT |
| Chichewa (nya) | 89.1 | MT560, Wikipedia, AI4D corpus |
| Naija (pcm) | 56.9 | JW300, BBC |
| chiShona (sna) | 110.9 | MT560, Wikipedia, VOA |
| Kiswahili (swa) | 208.4 | JW300, Wikipedia, BBC, VOA, Global Voices |
| Setswana (tsn) | 102.0 | JW300, Wikipedia, Daily news |
| Akan/Twi (twi) | 69.2 | Twi Corpus in Chapter 4 |
| Wolof (wol) | 9.3 | Bible, Wikipedia, Defuwaxu, Saabal, MAFAND-MT |
| isiXhosa (xho) | 133.1 | JW300, Wikipedia, Isolezwe |
| Yorùbá (yor) | 78.2 | Yorùbá Corpus in Chapter 4 |
| isiZulu (zul) | 118.0 | JW300, Wikipedia, Isolezwe |

Table 3.2: **Languages, Monolingual corpus size, and Monolingual source for training FastText embeddings**

---

[29] https://fasttext.cc/docs/en/crawl-vectors.html
[30] Zenodo links are in https://github.com/dadelani/africanlp-resources

TRAINING OF WORD EMBEDDING  We train FastText embedding using a Skip-gram model with an embedding size of 300 dimensions, context window size of 5, 10 negative samples, minimum word count of 3, and *n*-grams ranging from 3 to 6 characters similar training setting of pre-trained models.

## 3.3 PRE-TRAINED LANGUAGE MODEL (PLM)

The task of the language model (LM) is to learn the probability of a sequence of tokens or predict the next token given previous tokens. This probability can be learned by several sequence modeling deep learning architectures like LSTM, or Transformer (Vaswani et al., 2017). While the most popular language models are autoregressive, there are non-autoregressive LMs like the *Masked Language Model*, e.g. BERT (Devlin et al., 2019) and RoBERTa (Liu et al., 2019). Rich contextualized word embedding can be extracted from these language models to initialize a task-specific NLP model, thus providing faster training and improved performance over static word embedding. Alternatively, they can be *fine-tuned* on a new task with impressive performance. **Fine-tuning on a new task** involves adding a linear layer to the LM, and training jointly the LM and the linear layer end-to-end. We explore a few popular language models and their multilingual variants below.

AUTOREGRESSIVE LM  The development of powerful and fluent LMs began with GPT (Radford and Narasimhan, 2018) using the decoder-only transformer model. GPT was trained on a large unlabelled corpus with 117M parameters. The GPT model achieves large improvement in performance on several tasks by simply *fine-tuning* on each specific task. This popularized the pre-training on a large corpus, followed by fine-tuning on the target task paradigm. Bigger models of GPT have been developed like GPT-2 (Radford et al., 2019) (1.5B parameters) and GPT-3 (Brown et al., 2020) (175B parameters). Due to the large model size, GPT-3 is difficult to fine-tune end-to-end for a new task. However, the authors of GPT-3 model showed that a frozen LM can be guided to perform different tasks through "in-context learning"—this involves providing to a LM an input which is a description of a new task with some examples demonstrating the task, followed by a final example with a text prompt, that the LM should complete.

MASKED LM (MLM)  MLM is a non-autoregressive model where a bidirectional LM is learned to predict [MASK] tokens using the left and right contexts. BERT was trained by *masking* 15% of sub-word units (at random) and a *next sentence prediction* task (a binary classification task to determine if two sentences follow each other). On the other hand, RoBERTa was only trained on only the MLM task but using a larger corpus. BERT and RoBERTa both showed large improvements over GPT on several natural language understanding (NLU) tasks, which shows that MLMs are better at capturing contextual representations (or better text encoders) than autoregressive LMs.

SEQUENCE-TO-SEQUENCE LM   While MLMs provide impressive performance on NLU tasks, they are not well suited for text generation tasks such as text summarization. Sequence-to-Sequence (or Seq-to-Seq) LMs such as T5 (Raffel et al., 2020) and BART (Lewis et al., 2020) are more suited for text generation tasks. Seq-to-Seq models often comprise two models, the "encoder" and the "decoder" which can be of different architectures. T5 was trained on the *text-infilling* task (similar to BERT) using the encoder-decoder of the standard transformer architecture. BART on the other hand was trained on the same text-infilling but using different architectures for the encoder and the decoder. They make use of a bidirectional transformer (like BERT) and an autoregressive transformer (similar to GPT-2) for the decoder. The multilingual variants of these models exist.

### 3.3.1 *Multilingual Pre-trained Language Models*

Many of the PLMs developed are based on the English language. There are multilingual variants of this model that are often pre-trained on several languages, including some low-resource languages.

AUTOREGRESSIVE LM   A few multilingual autoregressive LM exists that include African languages, for example, XGLM (Lin et al., 2021) was pre-trained on 134 languages, mGPT-2 (Shliazhko et al., 2022) was pre-trained on 60 languages, BLOOM[31] pre-trained on 46 languages, and PaLM (Chowdhery et al., 2022).

MASKED LM   mBERT and XLM-R (Conneau et al., 2020) are the most popular MLM models, they are trained on the BERT and RoBERTa architectures, respectively. mBERT was pre-trained on 104 languages with the largest Wikipedia articles and XLM-R was pre-trained on 100 languages with the largest contents on filtered Common Crawl corpus. There are also many other multilingual PLMs like InfoXLM (Chi et al., 2021), ERNIE-M (Ouyang et al., 2021), and RemBERT (Chung et al., 2021a). Unfortunately, only a few low-resource and African languages are represented, this has led to an active research on how to adapt them to unseen languages during pre-training, especially those with a different writing system (Chapter 5). To address this under-representation of African languages, Ogueji, Zhu, and Lin (2021) developed AfriBERTa, – a PLM pre-trained on 11 African languages, mostly from East and West Africa.

---

31 https://huggingface.co/bigscience/bloom

| PLM | PM Size | # Lang. | # African Lang. | African languages covered |
| --- | --- | --- | --- | --- |
| *Autoregressive LM* | | | | |
| XGLM 4.5B | 4.5B | 134 | 23 | afr, amh, bam, ful, hau, ibo, kon, lin, lug, mlg, nya, orm, sna, som, sot, ssw, swa, tir, tsn, wol, xho, yor, zul |
| mGPT-2 | 1.3B | 60 | 3 | afr, swa, yor |
| BLOOM | 300M – 13B | 46 | 21 | aka, bam, fon, ibo, kik, kin, lin, lug, nya, run, sna, sot, swa, tsn, tso, tum, twi, wol, xho, yor, zul |
| PaLM | 540B | 100 | 14 | afr, amh, hau, ibo, kin, mlg, nya, sna, som, sot, swa, xho, yor, zul |
| *Masked LM* | | | | |
| mBERT | 172M | 104 | 4 | afr, mlg, swa, yor |
| XLM-R | 276M-550M | 100 | 8 | afr, amh, hau, ibo, mlg, nya, sna, som, sot, swa, xho, yor, zul |
| InfoXLM | 276M-550M | 94 | 3 | afr, amh, swa |
| ERNIE-M | 300M – 13B | 101 | 13 | afr, amh, hau, ibo, mlg, nya, sna, som, sot, swa, xho, yor, zul |
| RemBERT | 575M | 110 | 12 | afr, amh, hau, ibo, mlg, nya, sna, som, swa, xho, yor, zul |
| AfriBERTa | 97M-126M | 11 | 11 | amh, hau, ibo, kin, run, orm, pcm, swa, som, tir, yor |
| *Seq-to-Seq LM* | | | | |
| mBART50 | 610M | 50 | 3 | afr, swa, xho |
| MT5/ByT5 | 300M – 13B | 101 | 13 | afr, amh, hau, ibo, mlg, nya, sna, som, sot, swa, xho, yor, zul |
| Charformer | 134M-206M | 101 | 13 | afr, amh, hau, ibo, mlg, nya, sna, som, sot, swa, xho, yor, zul |

Table 3.3: **Language coverage and parameter size (in millions) of multilingual pre-trained models**.

MULTILINGUAL SEQUENCE-TO-SEQUENCE LMS    There are a few popular Seq-to-Seq LMs like mBART50 (Tang et al., 2020), MT5 (Xue et al., 2021), ByT5 (a token-free T5) (Xue et al., 2022), and Charformer (Tay et al., 2022). mBART50 was trained on 50 languages including three African languages (Afrikaans, Kiswahili and isiXhosa). MT5, ByT5, and Charformer are pre-trained on the same corpus with 101 languages including 13 African languages.

We summarize different multilingual PLMs, their architecture, their parameter size, the number of languages supported, list of African languages covered in Table 3.3.

3.3.2 *Language-specific PLM*

Many PLMs are developed in English since it has numerous and challenging evaluation sets and is probably the language with the largest available texts on the web. For other languages, the options are to either use a multilingual PLM or pre-train a *language-specific PLM* from the scratch on a large monolingual data. In general, multilingual PLMs have been shown to have a slightly lower performance compared to language-specific PLMs for languages with large monolingual corpus (Antoun, Baly, and Hajj, 2020; Martin et al., 2020). These language-specific PLMs are available in many high-resourced languages like German (Scheible et al., 2020), French (Le et al., 2020), Chinese (Cui et al., 2021), and Arabic (Abdul-Mageed, Elmadany, and Nagoudi, 2021) but only a few are available for African languages like Amharic (Yimam et al., 2021), Afrikaans (Ralethe, 2020), Kiswahili (Martin et al., 2022), and Kinyarwanda (Nzeyimana and Niyongabo Rubungo, 2022). While developing language-specific PLMs is an interesting direction, there are a few limitations of this approach, like (1) lack of large monolingual data to pre-train from scratch, especially for languages with less than 1GB of monolingual data on the web. (2) ineffective cross-lingual transfer performance from high-resource languages—due to different vocabulary or script. Since many African languages lack labelled datasets, it would be beneficial to focus on developing multilingual PLMs for African languages that also include common transfer languages like English, French and Arabic — with numerous labelled datasets to transfer from. In Chapter 5 of this thesis, we develop multilingual PLMs in this direction to cover both common transfer languages and African languages jointly trained together.

3.4 COMPARISON OF WORD EMBEDDINGS AND MULTILINGUAL PLMS

Here, we compare the NER model performance of *fine-tuned* CNN-BiLSTM-CRF (Ma and Hovy, 2016) to XLM-R model for different training data sizes (e.g. 500, 1000, 2000 or 4000 sentences). For the CNN-BiLSTM-CRF model, we initialized with FastText embeddings from (Section 3.2.2). For Hausa, we make use of an existing Word2Vec model that was trained on a high quality large curated corpus (Abdulmumin and Galadanci, 2019). We evaluate the performance on 20 languages from the MasakhaNER 2.0 dataset (Chapter 8) and Amharic ("amh") dataset from MasakhaNER 1.0 (Chapter 7). The languages in the evaluation are: **amh**, **bam**, **bbj**, **ewe**, **fon**, **hau**, **ibo**, **kin**, **lug**, **luo**, **mos**, **nya**, **pcm**, **sna**, **swa**, **tsn**, **twi**, **wol**, **xho**, **yor**, and **zul**.

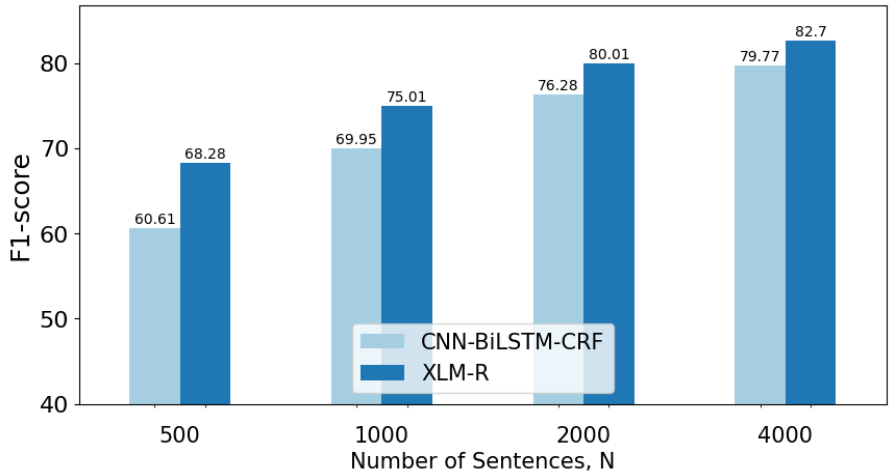

Figure 3.1: **Average F1-score of CNN-BiLSTM-CRF and XLM-R for different number of sentences**.

### 3.4.1 *Experimental Setup*

We experiment with different training data sizes of the MasakhaNER dataset, we fix the number of training sentences defined by variable $N$ to be either 500, 1000, 2000, and 4000, to analyze the gap in performance of the two models.

For the CNN-BiLSTM-CRF, we followed the same training configuration as (Hedderich et al., 2020). The hyper-parameters are: dropout of 0.5, batch-size of 10, SGD with a learning rate of 0.01, a decay of 0.05 and momentum of 0.9. Gradients are clipped with a value of 5 and the RNN dimension is 300. For the CNN, the character embedding dimension is 25 with 30 filters and a window-size of 3.

For the XLM-R fine-tuning, we make use of XLM-R-base model, and fine-tune on the NER dataset for 20 epochs using a batch size of 32, and a learning rate of 5e-5.

### 3.4.2 *Results*

Figure 3.1 shows the result for the average performance of CNN-BiLSTM-CRF and XLM-R-base model on NER. XLM-R-base outperforms the CNN-BiLSTM-CRF model for different numbers of sentences. However, we find the difference in performance to be wider for smaller number of sentences like $N = 500$ (+7.63 points) and $N = 1000$ (+5.1 points), while larger sentences see smaller improvements from using XLM-R-base, the performance with $N = 2000$ is (+3.73 points) and for $N = 4000$, the improvement is only (+2.93 points). This shows that making use of PLMs requires lesser data to give impressive results, also it saves the cost of annotation. For example, we could achieve on average 80 F1 using XLM-R-base with only 2,000 sentences, while CNN-BiLSTM-CRF requires twice the number of sentences to achieve similar results.

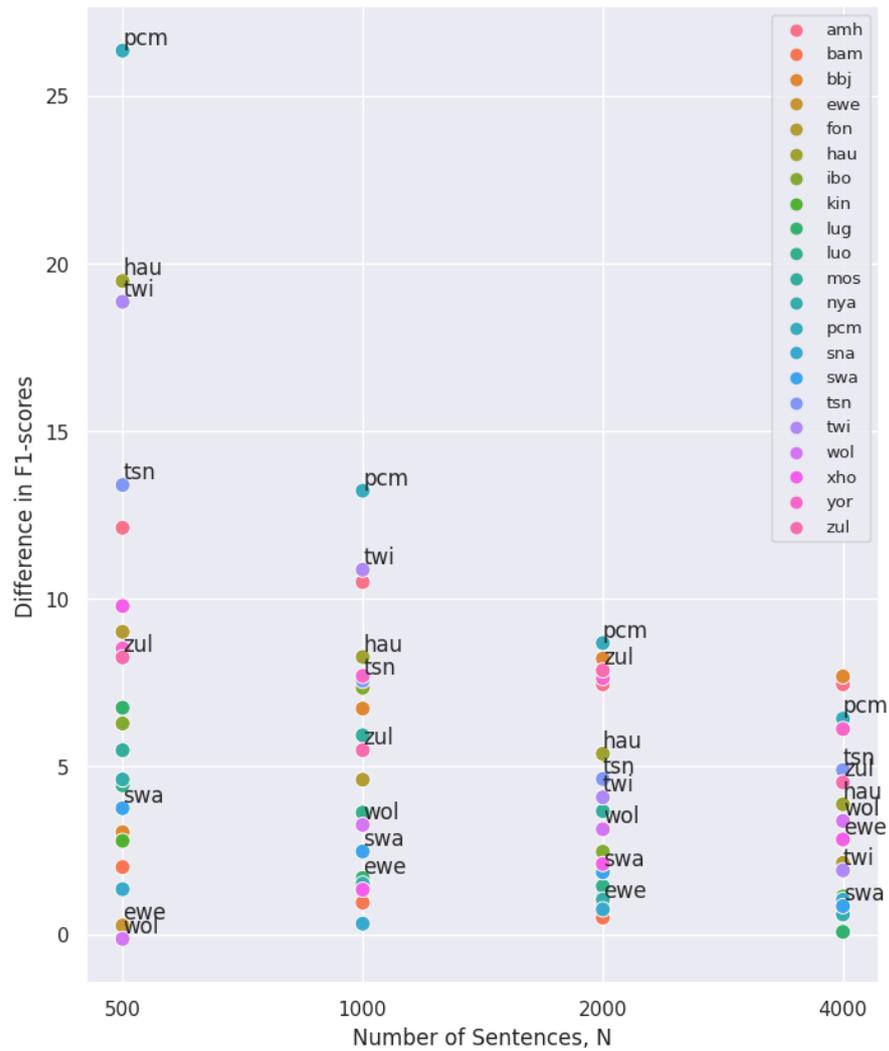

Figure 3.2: **Difference in F1 scores between BiLSTM and XLM-R**. Some languages are omitted in the figure to make the figure readable for a few languages.

Figure 3.2 shows the difference in F1-scores between XLM-R-base and CNN-BiLSTM-CRF for each language and the same number of sentences. The result shows a large difference for languages such as pcm, hau, twi, sna and xho with over 10 points when $N = 500$. The difference in F1-scores drops rapidly with more training sentences. Table 3.4 provides the results for all languages by their number of sentences. However, there are situations where the gap in performance is very low for $N = 500$, e.g. for ewe, and wol. But with more training sentences like $N = 1000$ or $N = 2000$, XLM-R-base gives a better result. This shows that having more training sentences is beneficial to both models, but PLM benefits the most, for example, when $N = 2000$,

| Language | 500 sentences CNN BiLSTM CRF | 500 sentences XLM-R base | 1000 sentences CNN BiLSTM CRF | 1000 sentences XLM-R base | 2000 sentences CNN BiLSTM CRF | 2000 sentences XLM-R base | 4000 sentences CNN BiLSTM CRF | 4000 sentences XLM-R large |
|---|---|---|---|---|---|---|---|---|
| amh | 53.8 | 65.9 | 60.6 | **71.1** | 63.6 | **71.1** | 63.6 | **71.1** |
| bam | 58.4 | 60.4 | 67.5 | 68.6 | 72.5 | 73.0 | 77.6 | **78.2** |
| bbj | 53.0 | 56.1 | 56.9 | 63.7 | 61.5 | 69.7 | 64.7 | **72.3** |
| ewe | 78.2 | 78.5 | 84.2 | 85.8 | 86.0 | 86.9 | 85.7 | **88.5** |
| fon | 55.3 | 64.3 | 69.4 | 74.1 | 76.7 | 78.7 | 78.9 | **81.0** |
| hau | 50.2 | 69.7 | 72.1 | 80.4 | 74.8 | 80.2 | 79.8 | **83.7** |
| ibo | 71.8 | 78.1 | 71.5 | 78.9 | 81.4 | 83.8 | 83.2 | **86.6** |
| kin | 63.9 | 66.6 | 70.8 | 72.2 | 75.7 | 76.8 | 79.6 | **80.7** |
| lug | 66.5 | 73.2 | 80.0 | 81.6 | 83.9 | 85.0 | 86.2 | **86.2** |
| luo | 49.2 | 53.6 | 60.1 | 63.7 | 73.0 | 74.5 | 77.4 | **78.4** |
| mos | 50.6 | 56.1 | 62.6 | 68.5 | 68.7 | 72.4 | 72.4 | **73.4** |
| nya | 79.3 | 83.9 | 83.5 | 85.0 | 85.7 | 86.8 | 88.2 | **88.8** |
| pcm | 52.4 | 78.7 | 71.1 | 84.3 | 77.0 | 85.7 | 80.9 | **87.3** |
| sna | 79.6 | 80.9 | 85.0 | 85.3 | 88.4 | 89.2 | 91.2 | **92.3** |
| swa | 82.4 | 86.1 | 86.7 | 89.2 | 89.4 | 91.3 | 90.8 | **91.6** |
| tsn | 53.7 | 67.0 | 64.0 | 71.6 | 77.2 | 81.8 | 81.5 | **86.4** |
| twi | 52.8 | 71.7 | 62.5 | 73.4 | 73.1 | 77.2 | 76.8 | **78.7** |
| wol | 63.0 | 62.8 | 69.4 | 72.7 | 73.8 | 77.0 | 77.7 | **81.1** |
| xho | 60.8 | 70.6 | 77.8 | 79.2 | 82.0 | 84.1 | 83.0 | **85.9** |
| yor | 52.8 | 61.3 | 62.1 | 69.8 | 71.6 | 79.3 | 78.1 | **84.2** |
| zul | 44.3 | 52.5 | 51.1 | 56.6 | 68.6 | 76.5 | 78.8 | **83.3** |
| AVG | 60.6 | 68.3 | 69.9 | 75.0 | 76.3 | 80.0 | 79.8 | **82.7** |

Table 3.4: **Comparison of CNN-BiLSTM-CRF model initialized with word embeddings with XLM-R-base**. Average is over 5 runs. The languages highlighted have similar performance for both models when they are trained on 500 sentences

the difference in F1-scores for wol exceeded 3 points, while for ewe, there is only a significant improvement for $N = 4000$. Regardless of the size of training sentences, PLM tends to give a better result than CNN-BiLSTM-CRF since they benefit from the rich contextual embeddings learned, unlike CNN-BiLSTM-CRF which relies on static word embeddings.

PLM provides an opportunity for many NLP practitioners working on low-resource languages to create labelled datasets with less number of samples depending on the task. For NER, we see that, with 2k sentences, we can already achieve very decent performance, and the additional benefit of more annotation is smaller. By investing more time in model development and transfer learning from related languages, we can further improve similar gains instead of more laborious annotations (see Chapter 8).

3.5 AFRICA ML/NLP COMMUNITIES

The recent rapid development of NLP for African languages that we have seen cannot happen without the support of several grass-root organizations across the continent like, Masakhane[32], Deep learning Indaba[33], Black-In-AI[34], ALTI[35], Data Science Network (formerly Data Science Nigeria)[36], Zindi[37], Knowledge for all[38], and Ghana NLP. A few research labs are also very active in NLP, like Makerere AI[39] and DSFSI[40] at the university of Pretoria. Adebara and Abdul-Mageed (2022) provide an overview of these AL/NLP communities in Africa.

It is important to highlight **Masakhane**, whose mission is to strengthen and spur NLP research in African languages, for Africans, by Africans. The organization has over 2,000 virtual members (who engage on slack[41]) mostly from Africa and from around the world. They pioneer participatory research ($\forall$ et al., 2020) to tackle the under-representation of African languages in NLP. In the participatory approach, speakers of the language, dataset curators, evaluation experts and language technologists are involved and work together for the development of datasets and NLP models. Masakhane has also been involved in the annual organization of AfricaNLP workshops at top AI and NLP conferences since 2020.[42]

A few organizations are also funding NLP dataset creation and model development for African languages like AI4D,[43] FairForward,[44] and Lacuna Fund.[45]

---

32 https://www.masakhane.io/
33 https://deeplearningindaba.com/2022/
34 https://blackinai.github.io/
35 http://www.alt-i.org/
36 https://www.datasciencenigeria.org/
37 https://zindi.africa/
38 https://www.k4all.org/
39 https://air.ug/
40 https://dsfsi.github.io/
41 https://slack.com/
42 https://africanlp.masakhane.io/
43 https://africa.ai4d.ai/
44 https://toolkit-digitalisierung.de/en/fair-forward/
45 https://lacunafund.org/language/

Part II

MULTILINGUAL REPRESENTATION LEARNING

# 4

# WORD EMBEDDINGS FOR AFRICAN LANGUAGES

This chapter[1] introduces the development of high-quality word embeddings for African languages which is essential for building NLP models for several tasks. First, we evaluate the quality of pre-trained FastText[2] embeddings of two African languages (Twi and Yorùbá) that have been trained on massive unlabelled texts like Wikipedia and Common crawl. Our evaluation on a word similarity task using wordsim-353 word pairs shows that pre-trained FastText embeddings have poor quality, this implies that the quality of word embeddings does not only depend on the quantity of pre-training corpus but also the quality. To remedy this, we trained FastText embeddings on high-quality curated corpora. We extend our analysis to pre-trained language models by evaluating the performance of multilingual BERT on a named entity recognition task.

## 4.1 INTRODUCTION

In recent years, word embeddings (Bojanowski et al., 2017; Mikolov et al., 2013b; Pennington, Socher, and Manning, 2014) have been proven to be very useful for training downstream natural language processing (NLP) tasks. Moreover, contextualized embeddings (Devlin et al., 2019; Peters et al., 2018) have been shown to further improve the performance of NLP tasks such as named entity recognition, question answering, and text classification when used as word features because they are able to resolve ambiguities of word representations when they appear in different contexts. Different deep learning architectures such as multilingual BERT (Devlin et al., 2019), LASER (Artetxe and Schwenk, 2019b) and XLM (Lample and Conneau, 2019) have proved successful in the multilingual setting. All these architectures learn the semantic representations from unannotated text, making them *cheap* given the availability of texts in online multilingual resources such as Wikipedia. However, the evaluation of such resources is usually done for the high-resourced languages, where one has a smorgasbord of tasks and test sets to evaluate on. This is the best-case scenario, i.e. languages with tonnes of data for training that generate high-quality models.

For low-resourced languages, the evaluation is more difficult and therefore normally ignored simply because of the lack of resources.

---

[1] This chapter is based on Alabi et al. (2020), my contributions include the development of BERT models and creation of NER evaluation dataset for Yorùbá language.
[2] https://fasttext.cc/docs/en/pretrained-vectors.html

In these cases, training data is scarce, and the assumption that the capability of deep learning architectures to learn (multilingual) representations in the high-resourced setting holds in the low-resourced one does not need to be true. In this work, we focus on two African languages, Yorùbá and Twi, and carry out several experiments to verify this claim. By a simple inspection of the word embeddings trained on Wikipedia by fastText[3], we observe that there are several non-Yorùbá or non-Twi words in the vocabularies of the word embeddings. For Twi, the vocabulary has only 935 words, and for Yorùbá we estimate that 135 k out of the 150 k words belong to other languages such as English, French and Arabic.

In order to improve the semantic representations for these languages, we collect online texts and study the influence of the quality and quantity of the data in the final models. We also examine the most appropriate architecture depending on the characteristics of each language. Finally, we translate test sets and annotate corpora to evaluate the performance of both our models together with fastText and BERT pre-trained embeddings which could not be evaluated otherwise for Yorùbá and Twi. The evaluation is carried out in a word similarity and relatedness task using the *wordsim-353* test set, and in a named entity recognition (NER) task where embeddings play a crucial role. Of course, the evaluation of the models in only two tasks is not exhaustive but it is an indication of the quality we can obtain for these two low-resourced languages as compared to others such as English where these evaluations are already available.

The rest of the paper is organized as follows. Related works are reviewed in Section 4.2 The two languages under study are described in the third section. We introduce the corpora and test sets in Section 4.4 The fifth section explores the different training architectures we consider, and the experiments that are carried out. Finally, discussion and concluding remarks are given in Section 4.6

## 4.2 related work

The large amount of freely available text in the internet for multiple languages is facilitating the massive and automatic creation of multilingual resources. The resource par excellence is Wikipedia[4], an online encyclopedia currently available in 314 languages[5]. Other initiatives such as Common Crawl[6] or the Jehovah's Witnesses site[7] are also repositories for multilingual data, usually assumed to be noisier than Wikipedia. Word and contextual embeddings have been pre-trained on these data, so that the resources are nowadays at hand for more than

---

[3] https://fasttext.cc/docs/en/pretrained-vectors.html
[4] https://www.wikipedia.org
[5] Number of languages in July 2022.
[6] https://commoncrawl.org
[7] https://www.jw.org

100 languages. Some examples include fastText word embeddings (Bojanowski et al., 2017; Grave et al., 2018), MUSE embeddings (Lample et al., 2018b), BERT multilingual embeddings (Devlin et al., 2019) and LASER sentence embeddings (Artetxe and Schwenk, 2019b). In all cases, embeddings are trained either simultaneously for multiple languages, joining high- and low-resource data, or following the same methodology.

On the other hand, different approaches try to specifically design architectures to learn embeddings in a low-resourced setting. Chaudhary et al. (2018) (Chaudhary et al., 2018) follow a transfer learning approach that uses phonemes, lemmas and morphological tags to transfer the knowledge from related high-resource language into the low-resource one. Jiang et al. (2018) (Jiang et al., 2018) apply Positive-Unlabeled Learning for word embedding calculations, assuming that unobserved pairs of words in a corpus also convey information, and this is specially important for small corpora.

In order to assess the quality of word embeddings, word similarity and relatedness tasks are usually used. *wordsim-353* (Finkelstein et al., 2001) is a collection of 353 pairs annotated with semantic similarity scores in a scale from 0 to 10. Even with the problems detected in this dataset (Camacho-Collados et al., 2017), it is widely used by the community. The test set was originally created for English, but the need for comparison with other languages has motivated several translations/adaptations. In Hassan and Mihalcea (2009), the test was translated manually into Spanish, Romanian and Arabic and the scores were adapted to reflect similarities in the new language. The reported correlation between the English scores and the Spanish ones is 0.86. Later, Joubarne and Inkpen (2011) show indications that the measures of similarity highly correlate across languages. Leviant and Reichart (2015) translated also wordsim-353 into German, Italian and Russian and used crowdsourcing to score the pairs. Finally, Jiang et al. (2018) translated with Google Cloud the test set from English into Czech, Danish and Dutch. In our work, native speakers translate wordsim-353 into Yorùbá and Twi, and similarity scores are kept unless the discrepancy with English is big (see Section 4.4.2 for details). A similar approach to our work is done for Gujarati in (Joshi, Koringa, and Mitra, 2019).

## 4.3 LANGUAGES UNDER STUDY

YORÙBÁ is a language in the West Africa with over 50 million speakers. It is spoken among other languages in Nigeria, republic of Togo, Benin Republic and Sierra Leone. It is also a language of Òrìsà in Cuba, Brazil, and some Caribbean countries. It is one of the three major languages in Nigeria and it is regarded as the third most spoken native African language. There are different dialects of Yorùbá in Nigeria (Adegbola, 2016; Asahiah, 2014; Fagbolu et al., 2015). However, in this paper our focus is the standard Yorùbá based upon a report from the 1974 Joint Consultative Committee on Education (Asahiah, Odejobi, and Adagunodo, 2017).

Standard Yorùbá has 25 letters without the Latin characters c, q, v, x and z. There are 18 consonants (b, d, f, g, gb, j[dz], k, l, m, n, p[kp], r, s, ṣ, t, w y[j]), 7 oral vowels (a, e, ẹ, i, o, ọ, u), five nasal vowels, (an, ẹn, in, ọn, un) and syllabic nasals (m̀, ḿ, ǹ, ń). Yorùbá is a tone language which makes heavy use of lexical tones which are indicated by the use of diacritics. There are three tones in Yorùbá namely low, mid and high which are represented as grave (\), macron (−) and acute (/) symbols respectively. These tones are applied on vowels and syllabic nasals. Mid tone is usually left unmarked on vowels and every initial or first vowel in a word cannot have a high tone. It is important to note that tone information is needed for correct pronunciation and to have the meaning of a word (Adegbola and Odilinye, 2012; Asahiah, Odejobi, and Adagunodo, 2017; Asahiah, 2014). For example, *owó* (money), *ọwọ̀* (broom), *òwò* (business), *ọ̀wọ̀* (honour), *ọwọ́* (hand), and *ọ̀wọ́* (group) are different words with different dots and diacritic combinations.

According to Asahiah (2014), Standard Yorùbá uses 4 diacritics, 3 are for marking tones while the fourth which is the dot below is used to indicate the open phonetic variants of letter "e" and "o" and the long variant of "s". Also, there are 19 single diacritic letters, 3 are marked with dots below (ẹ, ọ, ṣ) while the rest are either having the grave or acute accent. The four double diacritics are divided between the grave and the acute accent as well.

As noted in Asahiah (2014), most of the Yorùbá texts found in websites or public domain repositories (*i*) either use the correct Yorùbá orthography or (*ii*) replace diacritized characters with un-diacritized ones. This happens as a result of many factors, but most especially to the unavailability of appropriate input devices for the accurate application of the diacritical marks (Adegbola, 2016). This has led to research on restoration models for diacritics (Orife, 2018b), but the problem is not well solved and we find that most Yorùbá text in the public domain today is not well diacritized. Wikipedia is not an exception.

TWI is an Akan language of the Central Tano Branch of the Niger Congo family of languages. It is the most widely spoken of the about 80 indigenous languages in Ghana (Osam, 2003). It has about 9 million native speakers and about a total of 17–18 million Ghanaians have it as either first or second language. There are two mutually intelligible dialects, Asante and Akuapem, and sub-dialectical variants which are mostly unknown to and unnoticed by non-native speakers. It is also mutually intelligible with Fante and to a large extent Bono, another of the Akan languages. It is one of, if not the, easiest to learn to *speak* of the indigenous Ghanaian languages. The same is however not true when it comes to *reading* and especially *writing*. This is due to a number of easily overlooked complexities in the structure of the language. First of all, similarly to Yorùbá, Twi is a tonal language but written without diacritics or accents. As a result, words which are pronounced differently and unambiguous in speech tend to be ambiguous in writing. Besides, most of such words fit interchangeably in the same context and some of them can have more than two meanings. A simple example is:

> Me papa aba nti na me ne wo redi no yie no. Sɛ wo ara wo nim sɛ me papa ba a, me suban foforɔ adi.

| Description | Source URL | #tokens | Status | C1 | C2 | C3 |
| --- | --- | --- | --- | --- | --- | --- |
| *Yorùbá* | | | | | | |
| Lagos-NWU corpus | github.com/Niger-Volta-LTI | 24,868 | clean | ✓ | ✓ | ✓ |
| Alákòwé | alakoweyoruba.wordpress.com | 24,092 | clean | ✓ | ✓ | ✓ |
| Òrò Yorùbá | oroyoruba.blogspot.com | 16,232 | clean | ✓ | ✓ | ✓ |
| Èdè Yorùbá Rẹwà | deskgram.cc/edeyorubarewa | 4,464 | clean | ✓ | ✓ | ✓ |
| Doctrine $ Covenants | github.com/Niger-Volta-LTI | 20,447 | clean | ✓ | ✓ | ✓ |
| Yorùbá Bible | www.bible.com | 819,101 | clean | ✓ | ✓ | ✓ |
| GlobalVoices | yo.globalvoices.org | 24,617 | clean | ✓ | ✓ | ✓ |
| Jehova Witness | www.jw.org/yo | 170,203 | clean | ✓ | ✓ | ✓ |
| Ìrìnkèrindò nínú igbó elégbèje | manual | 56,434 | clean | ✓ | ✓ | ✓ |
| Igbó Olódùmarè | manual | 62,125 | clean | ✓ | ✓ | ✓ |
| JW300 Yorùbá corpus | opus.nlpl.eu/JW300.php | 10,558,055 | clean | ✗ | ✗ | ✓ |
| Yorùbá Tweets | twitter.com/yobamoodua | 153,716 | clean | ✓ | ✓ | ✓ |
| BBC Yorùbá | bbc.com/yoruba | 330,490 | noisy | ✗ | ✓ | ✓ |
| Voice of Nigeria Yorùbá news | von.gov.ng/yoruba | 380,252 | noisy | ✗ | ✗ | ✓ |
| Yorùbá Wikipedia | dumps.wikimedia.org/yowiki | 129,075 | noisy | ✗ | ✗ | ✓ |
| *Twi* | | | | | | |
| Bible | www.bible.com | 661,229 | clean | ✓ | ✓ | ✓ |
| Jehovah's Witness | www.jw.org/tw | 1,847,875 | noisy | ✗ | ✗ | ✓ |
| Wikipedia | dumps.wikimedia.org/twwiki | 5,820 | noisy | ✗ | ✓ | ✓ |
| JW300 Twi corpus | opus.nlpl.eu/JW300.php | 13,630,514 | noisy | ✗ | ✗ | ✓ |

Table 4.1: Summary of the corpora used in the analysis. The last 3 columns indicate in which dataset (C1, C2 or C3) are the different sources included (see text, Section 4.5.2).

This sentence could be translated as

> (*i*) I'm only treating you nicely because I'm in a good mood. You already know I'm a completely different person when I'm in a good mood.

> (*ii*) I'm only treating you nicely because my dad is around. You already know I'm a completely different person when my dad comes around.

Another characteristic of Twi is the fact that a good number of stop words have the same written form as content words. For instance,

"ɛna" or "na" could be the words "*and, then*", the phrase "*and then*" or the word "*mother*". This kind of ambiguity has consequences in several natural language applications where stop words are removed from text. Finally, we want to point out that words can also be written with or without prefixes. An example is this same *ɛna* and *na* which happen to be the same word with an omissible prefix across its multiple senses. For some words, the prefix characters are mostly used when the word begins a sentence and omitted in the middle. This however depends on the author/speaker. For the word embeddings calculation, this implies that one would have different embeddings for the same word found in different contexts.

## 4.4 DATA

We collect *clean* and *noisy* corpora for Yorùbá and Twi in order to quantify the effect of noise on the quality of the embeddings, where noisy has a different meaning depending on the language as it will be explained in the next subsections.

### 4.4.1 *Training Corpora*

For **Yorùbá**, we use several corpora collected by the Niger-Volta Language Technologies Institute[8] with texts from different sources, including the Lagos-NWU conversational speech corpus, fully-diacritized Yorùbá language websites and an online Bible. The largest source with clean data is the JW300 corpus. We also created our own small-sized corpus by web-crawling three Yorùbá language websites (Alàkọ̀wé, Ọ̀rọ̀ Yorùbá and Èdè Yorùbá Rẹwà in Table 4.1), some Yoruba Tweets with full diacritics and also news corpora (BBC Yorùbá and VON Yorùbá) with poor diacritics which we use to introduce noise. By noisy corpus, we refer to texts with incorrect diacritics (e.g in BBC Yorùbá), removal of tonal symbols (e.g in VON Yorùbá) and removal of all diacritics/under-dots (e.g some articles in Yorùbá Wikipedia). Furthermore, we got two manually typed fully-diacritized Yorùbá literature (Ìrìnkèrindò nínú igbó elégbèje and Igbó Olódùmarè) both written by Daniel Orowole Olorunfemi Fagunwa a popular Yorùbá author. The number of tokens available from each source, the link to the original source and the quality of the data is summarised in Table 4.1.

The gathering of clean data in **Twi** is more difficult. We use the Twi Bible as the base text as it has been shown that the Bible is the most available resource for low-resourced and endangered languages (Resnik, Olsen, and Diab, 1999). This is the cleanest of all the text we could obtain. In addition, we use the available (and small) Wikipedia dumps which are quite noisy, i.e. Wikipedia contains a good number of English words, spelling errors and Twi sentences formulated in a non-natural way (formulated as L2 speakers would speak Twi as compared to native speakers). Lastly, we added text crawled from JW.org (Jehovah's Witnesses, 2019) and the JW300 Twi corpus (Agić and Vulić, 2019). Notice that the Bible text, is mainly written in the Asante dialect whilst the last, Jehovah's Witnesses, was written mainly in the Akuapem dialect. The Wikipedia text is a mixture of the two dialects. This introduces a lot of noise into the embeddings as the spelling of most words differs especially at the end of the words due to the mixture of dialects. The JW300 Twi corpus also contains mixed dialects but is mainly Akuampem. In this case, the noise comes also from spelling errors and the uncommon addition of diacritics which are not standardised on certain vowels. Figures for Twi corpora are summarised in the bottom block of Table 4.1.

---

8 https://github.com/Niger-Volta-LTI/yoruba-text

| Entity type | Number of tokens | | | |
|---|---|---|---|---|
| | Total | Train | Val. | Test |
| ORG | 289 | 214 | 40 | 35 |
| LOC | 613 | 467 | 47 | 99 |
| DATE | 662 | 452 | 86 | 124 |
| PER | 688 | 469 | 109 | 110 |
| O | 23,988 | 17,819 | 2,413 | 4,867 |

Table 4.2: Number of tokens per named entity type in the Global Voices Yorùbá corpus.

### 4.4.2 *Evaluation Test Sets*

YORÙBÁ. One of the contribution of this work is the introduction of the wordsim-353 word pairs dataset for Yorùbá. All the 353 word pairs were translated from English to Yorùbá by 3 native speakers. The set is composed of 446 unique English words, 348 of which can be expressed as one-word translation in Yorùbá (e.g. *book* translates to *ìwé*). In 61 cases (most countries and locations but also other content words) translations are transliterations (e.g. *Doctor* is dókítà and *cucumber* is kùkúmbà.). 98 words were translated by short phrases instead of single words. This mostly affects words from science and technology (e.g. *keyboard* translates to pátákó ìtẹ̀wé —literally meaning typing board—, *laboratory* translates to ìyàrá ìṣèwádìí —research room—, and *ecology* translates to ìmọ̀ nípa àyíká while *psychology* translates to ìmọ̀ nípa èdá). Finally, 6 terms have the same form in English and Yorùbá therefore they are retained like that in the dataset (e.g. Jazz, Rock and acronyms such as FBI or OPEC).

We also annotate the Global Voices Yorùbá corpus to test the performance of our trained Yorùbá BERT embeddings on the named entity recognition task. The corpus consists of 26 k tokens which we annotate with four named entity types: DATE, location (LOC), organization (ORG) and personal names (PER). Any other token that does not belong to the four named entities is tagged with "O". The dataset is further split into training (70%), development (10%) and test (20%) partitions. Table 4.2 shows the number of named entities per type and partition.

TWI   Just like Yorùbá, the wordsim-353 word pairs dataset was translated for Twi. Out of the 353 word pairs, 274 were used in this case. The remaining 79 pairs contain words that translate into longer phrases.

The number of words that can be translated by a single token is higher than for Yorùbá. Within the 274 pairs, there are 351 unique English words which translated to 310 unique Twi words. 298 of the 310 Twi words are single word translations, 4 transliterations and 16 are used as is.

Even if Joubarne and Inkpen (2011) showed indications that semantic similarity has a high correlation across languages, different nuances between words are captured differently by languages. For instance, both *money* and *currency* in English translate into *sika* in Twi (and other 32 English words which translate to 14 Twi words belong to this category) and *drink* in English is translated as *Nsa* or *nom* depending on the part of speech (noun for the former, verb for the latter). 17 English words fall into this category. In translating these, we picked the translation that best suits the context (other word in the pair). In two cases, the correlation is not fulfilled at all: *soap–opera* and *star–movies* are not related in the Twi language and the score has been modified accordingly.

|  | Twi | | Yorùbá | |
|---|---|---|---|---|
| Model | Vocab Size | Spearman $\rho$ | Vocab Size | Spearman $\rho$ |
| F1: Pre-trained Model (Wiki) | 935 | 0.143 | 21,730 | 0.136 |
| F2: Pre-trained Model (Common Crawl & Wiki) | NA | NA | 151,125 | 0.073 |
| C1: Curated *Small* Dataset (Clean text) | 9,923 | 0.354 | 12,268 | 0.322 |
| C2: Curated *Small* Dataset (Clean + some noisy text) | 18,494 | **0.388** | 17,492 | 0.302 |
| C3: Curated *Large* Dataset (All Clean + Noisy texts) | 47,134 | 0.386 | 44,560 | **0.391** |

Table 4.3: FastText embeddings: Spearman $\rho$ correlation between human judgements and similarity scores on the wordSim-353 for the three datasets analysed (C1, C2 and C3). The comparison with massive fastText embeddings is shown in the top rows.

| | Twi | | Yorùbá | |
|---|---|---|---|---|
| **Model** | **Vocab Size** | **Spearman** $\rho$ | **Vocab Size** | **Spearman** $\rho$ |
| C1: Curated *Small* Dataset (Clean text) | 21,819 | 0.377 | 40,162 | 0.263 |
| C2: Curated *Small* Dataset (Clean + some noisy text) | 22,851 | **0.437** | 56,086 | 0.345 |
| C3: Curated *Large* Dataset (All Clean + Noisy texts) | 97,913 | 0.377 | 133,299 | **0.354** |

Table 4.4: CWE embeddings: Spearman $\rho$ correlation between human evaluation and embedding similarities for the three datasets analysed (C1, C2 and C3).

## 4.5 SEMANTIC REPRESENTATIONS

In this section, we describe the architectures used for learning word embeddings for the Twi and Yorùbá languages. Also, we discuss the quality of the embeddings as measured by the correlation with human judgements on the translated wordSim-353 test sets and by the F1 score in a NER task.

### 4.5.1 Word Embeddings Architectures

Modeling sub-word units has recently become a popular way to address out-of-vocabulary word problem in NLP especially in word representation learning (Bojanowski et al., 2017; Devlin et al., 2019; Sennrich, Haddow, and Birch, 2016b). A sub-word unit can be a character, character *n*-grams, or heuristically learned Byte Pair Encodings (BPE) which work very well in practice especially for morphologically rich languages. Here, we consider two word embedding models that make use of character-level information together with word information: Character Word Embedding (CWE) (Chen et al., 2015) and fastText (Bojanowski et al., 2017). Both of them are extensions of the Word2Vec architectures (Mikolov et al., 2013b) that model sub-word units, character embeddings in the case of CWE and character *n*-grams for fastText. CWE was introduced in 2015 to model the embeddings of characters jointly with words in order to address the issues of character ambiguities and non-compositional words especially in the Chinese language. A word or character embedding is learned in CWE using either CBOW or skipgram architectures, and then the final word embedding is computed by adding the character embeddings to the word itself:

$$x_j = \frac{1}{2}(w_j + \frac{1}{N_j}\sum_{k=1}^{N_j} c_k) \quad (4.1)$$

where $w_j$ is the word embedding of $x_j$, $N_j$ is the number of characters in $x_j$, and $c_k$ is the embedding of the *k*-th character $c_k$ in $x_j$.

Similarly, in 2017 fastText was introduced as an extension to skip-gram in order to take into account morphology and improve the representation of rare words. In this case the embedding of a word also includes the embeddings of its character *n*-grams:

$$x_j = \frac{1}{G_j + 1}(w_j + \sum_{k=1}^{G_j} g_k) \quad (4.2)$$

where $w_j$ is the word embedding of $x_j$, $G_j$ is the number of character *n*-grams in $x_j$ and $g_k$ is the embedding of the *k*-th *n*-gram.

Chen et al. (2015) also proposed three alternatives to learn multiple embeddings per character and resolve ambiguities: (*i*) position-based character embeddings where each character has different embeddings depending on the position it appears in a word, i.e., beginning, middle or end (*ii*) cluster-based character embeddings where a character can have *K* different cluster embeddings, and (*iii*) position-based cluster embeddings (CWE-LP) where for each position *K* different embeddings are learned. We use the latter in our experiments with CWE but no positional embeddings are used with fastText.

Finally, we consider a contextualized embedding architecture, Bidirectional Encoder Representations from Transformers (BERT) (Devlin et al., 2019). BERT is a masked language model based on the highly efficient and parallelizable Transformer architecture (Vaswani et al., 2017) known to produce very rich contextualized representations for downstream NLP tasks. The architecture is trained by jointly conditioning on both left and right contexts in all the transformer layers using two unsupervised objectives: Masked LM and Next-sentence prediction. The representation of a word is therefore learned according to the context it is found in. Training contextual embeddings needs of huge amounts of corpora which are not available for low-resourced languages such as Yorùbá and Twi. However, Google provided pre-trained multilingual embeddings for 102 languages[9] including Yorùbá (but not Twi).

### 4.5.2 *Experiments*

#### 4.5.2.1 *FastText Training and Evaluation*

As a first experiment, we compare the quality of fastText embeddings trained on (high-quality) curated data and (low-quality) massively extracted data for Twi and Yorùbá languages.

Facebook released pre-trained word embeddings using fastText for 294 languages trained on Wikipedia (Bojanowski et al., 2017) (identified as F1 in Table 4.3) and for 157 languages trained on Wikipedia and Common Crawl (Grave et al., 2018) (identified as F2 in Table 4.3). For Yorùbá, both versions are available but only embeddings trained on

---

9 https://github.com/google-research/bert/blob/master/multilingual.md

Wikipedia are available for Twi. We consider these embeddings the result of training on what we call *massively-extracted corpora*. Notice that training settings for both embeddings are not exactly the same, and differences in performance might come both from corpus size/quality but also from the background model. The 294-languages version is trained using skipgram, in dimension 300, with character *n*-grams of length 5, a window of size 5 and 5 negatives. The 157-languages version is trained using CBOW with position-weights, in dimension 300, with character *n*-grams of length 5, a window of size 5 and 10 negatives.

We want to compare the performance of these embeddings with the equivalent models that can be obtained by training on the different sources verified by native speakers of Twi and Yorùbá; what we call *curated corpora* and has been described in Section 4.4 For the comparison, we define 3 datasets according to the quality and quantity of textual data used for training: (*i*) *Curated Small Dataset (clean), C1*, about 1.6 million tokens for Yorùbá and over 735 k tokens for Twi. The clean text for Twi is the Bible and for Yoruba all texts marked under the C1 column in Table 4.1. (*ii*) In *Curated Small Dataset (clean + noisy), C2*, we add noise to the clean corpus (Wikipedia articles for Twi, and BBC Yorùbá news articles for Yorùbá). This increases the number of training tokens for Twi to 742 k tokens and Yorùbá to about 2 million tokens. (*iii*) *Curated Large Dataset, C3* consists of all available texts we are able to crawl and source out for, either clean or noisy. The addition of JW300 (Agić and Vulić, 2019) texts increases the vocabulary to more than 10 k tokens in both languages.

We train our fastText systems using a skipgram model with an embedding size of 300 dimensions, context window size of 5, 10 negatives and *n*-grams ranging from 3 to 6 characters similarly to the pre-trained models for both languages. Best results are obtained with minimum word count of 3. Table 4.3 shows the Spearman correlation between human judgements and cosine similarity scores on the wordSim-353 test set. Notice that pre-trained embeddings on Wikipedia show a very low correlation with humans on the similarity task for both languages ($\rho$=0.14) and their performance is even lower when Common Crawl is also considered ($\rho$=0.07 for Yorùbá). An important reason for the low performance is the limited vocabulary. The pre-trained Twi model has only 935 tokens. For Yorùbá, things are apparently better with more than 150 k tokens when both Wikipedia and Common Crawl are used but correlation is even lower. An inspection[10] of the pre-trained embeddings indicates that over 135 k words belong to other languages mostly English, French and Arabic. If we focus only on Wikipedia, we see that many texts are without diacritics in Yorùbá and often make use of mixed dialects and English sentences in Twi.

---

10 We used *langdetect* to have a rough estimation of the language of each word, assuming that words that are not detected are Yorùbá because the language is not supported by the tool.

The Spearman $\rho$ correlation for fastText models on the curated small dataset (clean), C1, improves the baselines by a large margin ($\rho = 0.354$ for Twi and 0.322 for Yorùbá) even with a small dataset. The improvement could be justified just by the larger vocabulary in Twi, but in the case of Yorùbá the enhancement is there with almost half of the vocabulary size. We found out that adding some noisy texts (C2 dataset) slightly improves the correlation for Twi language but not for the Yorùbá language. The Twi language benefits from Wikipedia articles because its inclusion doubles the vocabulary and reduces the bias of the model towards religious texts. However, for Yorùbá, noisy texts often ignore diacritics or tonal marks which increases the vocabulary size at the cost of an increment in the ambiguity too. As a result, the correlation is slightly hurt. One would expect that training with more data would improve the quality of the embeddings, but we found out with the results obtained with the C3 dataset, that only high-quality data helps. The addition of JW300 boosts the vocabulary in both cases, but whereas for Twi the corpus mixes dialects and is noisy, for Yorùbá it is very clean and with full diacritics. Consequently, the best embeddings for Yorùbá are obtained when training with the C3 dataset, whereas for Twi, C2 is the best option. In both cases, the curated embeddings improve the correlation with human judgements on the similarity task a $\Delta\rho = +0.25$ or, equivalently, by an increment on $\rho$ of 170% (Twi) and 180% (Yorùbá).

### 4.5.2.2 CWE Training and Evaluation

The huge ambiguity in the written Twi language motivates the exploration of different approaches to word embedding estimations. In this work, we compare the standard fastText methodology to include sub-word information with the character-enhanced approach with position-based clustered embeddings (CWE-LP as introduced in Section 4.5.1). With the latter, we expect to specifically address the ambiguity present in a language that does not translate the different oral tones on vowels into the written language.

The character-enhanced word embeddings are trained using a skip-gram architecture with cluster-based embeddings and an embedding size of 300 dimensions, context window-size of 5, and 5 negative samples. In this case, the best performance is obtained with a minimum word count of 1, and that increases the effective vocabulary that is used for training the embeddings with respect to the fastText experiments reported in Table 4.3.

We repeat the same experiments as with fastText and summarise them in Table 4.4. If we compare the relative numbers for the three datasets (C1, C2 and C3) we observe the same trends as before: the performance of the embeddings in the similarity task improves with the vocabulary size when the training data can be considered clean, but the performance diminishes when the data is noisy.

| Embedding Type | DATE | LOC | ORG | PER | F1-score |
|---|---|---|---|---|---|
| Pre-trained *uncased* Multilingual-bert (Multilingual vocab) | 44.6 | 33.9 | 12.1 | 5.7 | 27.1 ± 0.7 |
| Fine-tuned *uncased* Multilingual-bert (Multilingual vocab) | 64.0 | 65.3 | 38.8 | 47.4 | 56.4 ± 2.4 |
| Fine-tuned *uncased* Multilingual-bert (Yorùbá vocab) | 67.0 | 71.5 | 40.4 | 49.4 | 60.1 ± 0.8 |

Table 4.5: NER F1 score on Global Voices Yorùbá corpus after fine-tuning BERT for 10 epochs. Mean F1-score computed after 5 runs

According to the results, CWE is specially beneficial for Twi but not always for Yorùbá. Clean Yorùbá text, does not have the ambiguity issues at character-level, therefore the *n*-gram approximation works better when enough clean data is used ($\rho^{C3}_{CWE} = 0.354$ vs. $\rho^{C3}_{fastText} = 0.391$) but it does not when too much noisy data (no diacritics, therefore character-level information would be needed) is used ($\rho^{C2}_{CWE} = 0.345$ vs. $\rho^{C2}_{fastText} = 0.302$). For Twi, the character-level information reinforces the benefits of clean data and the best correlation with human judgements is reached with CWE embeddings ($\rho^{C2}_{CWE} = 0.437$ vs. $\rho^{C2}_{fastText} = 0.388$).

### 4.5.2.3 *BERT Evaluation on NER Task*

In order to go beyond the similarity task using static word vectors, we also investigate the quality of the multilingual BERT embeddings by fine-tuning a named entity recognition task on the Yorùbá Global Voices corpus.

One of the major advantages of pre-trained BERT embeddings is that fine-tuning of the model on downstream NLP tasks is typically computationally inexpensive, often with few number of epochs. However, the data the embeddings are trained on has the same limitations as that used in massive word embeddings. Fine-tuning involves replacing the last layer of BERT used optimizing the masked LM with a task-dependent linear classifier or any other deep learning architecture, and training all the model parameters end-to-end. For the NER task, we obtain the token-level representation from BERT and train a conditional random field classifier for sequence tagging.

Similar to our observations with non-contextualized embeddings, we find out that fine-tuning the pre-trained multilingual-uncased BERT for 10 epochs on the NER task gives an F1 score of 27. If we do the same experiment in English, F1 is 66.2 after 10 epochs. That shows how pre-trained embeddings by themselves do not perform well in downstream tasks on low-resource languages. To address this problem for Yorùbá, we fine-tune BERT masked language model on the Yorùbá corpus in two ways: (*i*) using the multilingual vocabulary, and (*ii*) using only Yorùbá vocabulary. In both cases diacritics are ignored to be consistent with the base model training. As expected, the fine-tuning of the pre-trained BERT on the Yorùbá corpus in the two configurations generates better representations than the base model. These models are able to achieve a better performance on the NER task with an average F1 score of over 56% (see Table 4.5 for the comparative). The fine-tuned BERT model with only Yorùbá vocabulary further increases by 4% in F1 score than the BERT model that uses the multilingual vocabulary. Although we do not have enough data to train BERT from scratch, we observe that fine-tuning BERT on a limited amount of monolingual data of a low-resource language helps to improve the quality of the embeddings. The same observation holds true for high-resource languages like German[11] and French (Martin et al., 2020).

## 4.6 SUMMARY AND DISCUSSION

In this paper, we present curated word and contextual embeddings for Yorùbá and Twi. For this purpose, we gather and select corpora and study the most appropriate techniques for the languages. We also create test sets for the evaluation of the word embeddings within a word similarity task (wordsim353) and the contextual embeddings within a NER task. Corpora, embeddings and test sets are available in github[12].

In our analysis, we show how massively generated embeddings perform poorly for low-resourced languages as compared to the performance for high-resourced ones. This is due both to the quantity but also the quality of the data used. While the Pearson $\rho$ correlation for English obtained with fastText embeddings trained on Wikipedia (WP) and Common Crawl (CC) are $\rho_{WP}$=0.67 and $\rho_{WP+CC}$=0.78, the equivalent ones for Yorùbá are $\rho_{WP}$=0.14 and $\rho_{WP+CC}$=0.07. For Twi, only embeddings with Wikipedia are available ($\rho_{WP}$=0.14). By carefully gathering high-quality data and optimising the models to the characteristics of each language, we deliver embeddings with correlations of $\rho$=0.39 (Yorùbá) and $\rho$=0.44 (Twi) on the same test set, still far from the high-resourced models, but representing an improvement over 170% on the task.

In a low-resourced setting, the data quality, processing and model selection is more critical than in a high-resourced scenario. We show how the characteristics of a language (such as diacritization in our case) should be taken into account in order to choose the relevant data and model to use. As an example, Twi word embeddings are significantly better when training on 742 k selected tokens than on 16 million noisy tokens, and when using a model that takes into account single character information (CWE-LP) instead of *n*-gram information (fastText).

---

11 https://deepset.ai/german-bert
12 https://github.com/ajesujoba/YorubaTwi-Embedding

Finally, we want to note that, even within a corpus, the quality of the data might depend on the language. Wikipedia is usually used as a high-quality freely available multilingual corpus as compared to noisier data such as Common Crawl. However, for the two languages under study, Wikipedia resulted to have too much noise: interference from other languages, text clearly written by non-native speakers, lack of diacritics and mixture of dialects. The JW300 corpus on the other hand, has been rated as high-quality by our native Yorùbá speakers, but as noisy by our native Twi speakers. In both cases, experiments confirm the conclusions.

# 5

## PRE-TRAINED LANGUAGE MODEL ADAPTATION FOR AFRICAN LANGUAGES

In Chapter 3 and Chapter 4, we discussed and compared word embeddings with pre-trained language models (PLMs) for African languages, and show that the latter, which is a single model, gives much better performance than using individual language word embeddings for building NLP models (see Figure 3.2). Our findings illustrate the opportunities of multilingual PLMs, especially for languages that do not have enough monolingual texts to train word embeddings.

This Chapter[1] focuses on the creation of a new multilingual PLM for African languages through *multilingual adaptive fine-tuning* (MAFT) on an existing multilingual PLM like XLM-R. MAFT involves fine-tuning a PLM on an aggregation of texts from multiple languages using the objective of the original PLM. The resulting model is known as AfroXLMR which is based on the XLM-R encoder, and it produces the best performance compared to all other multilingual PLMs on African language evaluation datasets from different tasks such as named entity recognition, news topic classification, and sentiment analysis. To further specialize XLM-R, we removed vocabulary tokens from the embedding layer that corresponds to non-African writing scripts before MAFT, thus reducing the model size by around 50%.

## 5.1 INTRODUCTION

Recent advances in the development of multilingual pre-trained language models (PLMs) like mBERT (Devlin et al., 2019), XLM-R (Conneau et al., 2020), and RemBERT (Chung et al., 2021a) have led to significant performance gains on a wide range of cross-lingual transfer tasks. Due to the *curse of multilinguality* (Conneau et al., 2020)—a trade-off between language coverage and model capacity—and the non-availability of pre-training corpora for many low-resource languages, multilingual PLMs are often trained on about 100 languages. Despite the limitations of language coverage, multilingual PLMs have been shown to transfer to several low-resource languages unseen during pre-training. Although, there is still a wide performance gap compared to languages seen during pre-training.

One of the most effective approaches to adapting to a new language is *language adaptive fine-tuning* (LAFT)—fine-tuning a multilingual PLM on monolingual texts in the target language using the same pre-training objective. This has been shown to lead to big gains in many cross-lingual transfer tasks (Pfeiffer et al., 2020b), and low-resource languages (Chau and Smith, 2021; Muller et al., 2021), including African languages (Adelani et al., 2021b; Alabi et al., 2020). Nevertheless, adapting a model to each of them takes large disk space, and limits their cross-lingual transfer abilities because they have been specialized for individual languages (Beukman, 2022).

---

[1] This chapter is based on Alabi et al. (2022) with Jesujoba Alabi and David Adelani contributing equally as first authors

An orthogonal approach to improve the coverage of low-resource languages is to include them in the pre-training data. An example of this approach is AfriBERTa (Ogueji, Zhu, and Lin, 2021), which trains on 11 African languages from the scratch. A downside of this approach is that it is resource intensive in terms of data and GPU compute.

Another alternative approach is parameter efficient fine-tuning like Adapters (Pfeiffer et al., 2020b) and sparse fine-tuning (Ansell et al., 2021), where the model is adapted to new languages by using a sparse network trained on a small monolingual text. Similar to LAFT, it requires language adaptation for every new target language. Although it takes little disk space, every target language-specific parameters needs to be stored.

In this paper, we propose *multilingual adaptive fine-tuning (MAFT)*—a language adaptation to multiple languages at once. We perform language adaptation on the 17 most-resourced African languages (Afrikaans, Amharic, Hausa, Igbo, Malagasy, Chichewa, Oromo, Naija, Kinyarwanda, Kirundi, Shona, Somali, Sesotho, Swahili, isiXhosa, Yorùbá, isiZulu) and three other high-resource languages widely spoken on the continent (English, French, and Arabic)—*simultaneously* to provide a single model for cross-lingual transfer learning for African languages. To further specialize the multilingual PLM, we follow the approach of Abdaoui, Pradel, and Sigel (2020) and remove vocabulary tokens from the embedding layer that corresponds to non-Latin and non-Ge'ez (used by Amharic) scripts before MAFT, thus effectively reducing the model size by 50%.

Our evaluation on two multilingual PLMs (AfriBERTa and XLM-R) and three NLP tasks (NER, news topic classification and sentiment classification) shows that our approach is competitive to performing LAFT on the individual languages, with the benefit of having a single model instead of a separate model for each of the target languages. Also, we show that our adapted PLM improves the zero-shot cross-lingual transfer abilities of parameter efficient fine-tuning methods like Adapters (Pfeiffer et al., 2020b) and sparse fine-tuning (Ansell et al., 2021).

As an additional contribution, and in order to cover more diverse African languages in our evaluation, we create a new evaluation corpus, ANTC—**A**frican **N**ews **T**opic **C**lassification – for Lingala, Somali, Naija, Malagasy, and isiZulu from pre-defined news categories of VOA, BBC, Global Voices, and Isolezwe newspapers. To further the research on NLP for African languages, we will make our code and data publicly available[2]. The PLMs are also available on HuggingFace[3].

## 5.2 RELATED WORK

MULTILINGUAL PLMS FOR AFRICAN LANGUAGES. The success of multilingual pre-trained language models (PLMs) such as mBERT (Devlin et al., 2019) and XLM-R (Conneau et al., 2020) for cross-lingual transfer in many natural language understanding tasks has encouraged the continuous development of multilingual models (Chi et al., 2021; Chung et al., 2021a; He, Gao, and Chen, 2021a; Luo et al., 2021; Ouyang et al., 2021). Most of these models cover 50 to 110 languages, and only a few African languages are represented due to lack of large monolingual corpora on the web. To address this under-representation, regional multilingual PLMs have been trained from scratch, such as AfriBERTa (Ogueji, Zhu, and Lin, 2021) or adapted from existing multilingual PLM through LAFT (Adelani et al., 2021b; Alabi et al., 2020; Muller et al., 2021; Pfeiffer et al., 2020b). AfriBERTa is a relatively small multilingual PLM (126M parameters) trained using the RoBERTa architecture and pre-training objective on 11 African languages. However, it lacks coverage of the languages from the southern region of the African continent, specifically the southern-Bantu languages. In our work, we extend to those languages since only a few of them have a large (>100MB size) monolingual corpus. We also do not specialize to a single language but apply MAFT which allows multilingual adaptation and preserves downstream performance on both high-resource and low-resource languages.

ADAPTATION OF MULTILINGUAL PLMS. It is not unusual for a new multilingual PLM to be initialized from an existing model. For example, Chi et al. (2021) trained InfoXLM by initializing the weights from XLM-R before training the model on a joint monolingual and translation corpus. Although they make use of a new training objective during adaptation. Similarly, Tang et al. (2020) extended the languages covered by mBART (Liu et al., 2020b) from 25 to 50 by first modifying the vocabulary and initializing the model weights of the original mBART before fine-tuning it on a combination of monolingual texts from the original 25 languages in addition to 25 new languages. Despite increasing the number of languages covered by their model, they did not observe a significant performance drop on downstream tasks. We take inspiration from these works for applying MAFT on African languages, but we do not modify the training objective during adaptation nor increase the vocabulary.

---

2 https://github.com/uds-lsv/afro-maft
3 https://huggingface.co/Davlan

COMPRESSING PLMS. One of the most effective methods for creating smaller PLMs is distillation, where a small student model is trained to reproduce the behaviour of a larger teacher model. This has been applied to many English PLMs (Jiao et al., 2020; Liu et al., 2020a; Sanh et al., 2019; Sun et al., 2020) and a few multilingual PLMs (Wang et al., 2021, 2020a). However, it often leads to a drop in performance compared to the teacher PLM. An alternative approach that does not lead to a drop in performance has been proposed by Abdaoui, Pradel, and Sigel (2020) for multilingual PLM. They removed unused vocabulary tokens from the embedding layer. This simple method significantly reduces the number of embedding parameters, thus reducing the overall model size since the embedding layer contributes the most to the total number of model parameters. In our paper, we combine MAFT with the method proposed by Abdaoui, Pradel, and Sigel (2020) to reduce the overall size of the resulting multilingual PLM for African languages. This is crucial because people from the under-represented communities in Africa may not have access to powerful GPUs to fine-tune large PLMs. Also, Google Colab[4] (free-version), which is widely used by individuals from under-represented communities without access to other compute resources, cannot run large models like XLM-R. Hence, it is important to provide smaller models that still achieve competitive downstream performance to these communities.

EVALUATION CORPORA IN AFRICAN LANGUAGES. One of the challenges of developing (multilingual) PLMs for African languages is the lack of evaluation corpora. There have been many efforts by communities like Masakhane to address this issue (Adelani et al., 2021b; ∀ et al., 2020). We only find two major evaluation benchmark datasets that cover a wide range of African languages: named entity recognition (NER) (Adelani et al., 2021b) and sentiment classification (Muhammad et al., 2022). In addition, there are also several news topic classification datasets (Azime and Mohammed, 2021; Hedderich et al., 2020; Niyongabo et al., 2020), but they are only available for a few African languages. Our work contributes novel news topic classification datasets (i.e. ANTC) for additional five African languages: lin, som, pcm, mlg, and zul.

## 5.3 DATA

### 5.3.1 *Adaptation corpora*

We perform MAFT on 17 African languages (Afrikaans, Amharic, Hausa, Igbo, Malagasy, Chichewa, Oromo, Naija, Kinyarwanda, Kirundi, Shona, Somali, Sesotho, Swahili, isiXhosa, Yorùbá, isiZulu)) covering the major African language families and three high resource languages (Arabic, French, and English) widely spoken in Africa. We selected the African languages based on the availability of a large amount of monolingual texts. We obtain the monolingual texts from three major sources: the mT5 pre-training corpus which is based on Common Crawl Corpus[5] (Xue et al., 2021), British Broadcasting Corporation (BBC) News, Voice of America News[6] (Palen-Michel, Kim, and Lignos, 2022), and some other news websites based in Africa. Table 5.1 provides a summary of the monolingual data, including their sizes and sources. We pre-processed the data by removing lines that consist of numbers or punctuation only, and lines with less than six tokens.

---

4 https://colab.research.google.com/

| Language | Source | Size (MB) | No. of sentences |
|---|---|---|---|
| Afrikaans (afr) | mC4 (subset) (Xue et al., 2021) | 752.2MB | 3,697,430 |
| Amharic (amh) | mC4 (subset), and VOA | 1,300MB | 2,913,801 |
| Arabic (ara) | mC4 (subset) | 1,300MB | 3,939,375 |
| English (eng) | mC4 (subset), and VOA | 2,200MB | 8,626,571 |
| French (fra) | mC4 (subset), and VOA | 960MB | 4,731,196 |
| Hausa (hau) | mC4 (all), and VOA | 594.1MB | 3,290,382 |
| Igbo (ibo) | mC4 (all), and AfriBERTa Corpus (Ogueji, Zhu, and Lin, 2021) | 287.5MB | 1,534,825 |
| Malagasy (mlg) | mC4 (all) | 639.6MB | 3,304,459 |
| Chichewa (nya) | mC4 (all), Chichewa News Corpus (Siminyu et al., 2021) | 373.8MB | 2,203,040 |
| Oromo (orm) | AfriBERTa Corpus, and VOA | 67.3MB | 490,399 |
| Naija (pcm) | AfriBERTa Corpus | 54.8MB | 166,842 |
| Rwanda-Rundi (kin/run) | AfriBERTa Corpus, KINNEWS & KIRNEWS (Niyongabo et al., 2020), and VOA | 84MB | 303,838 |
| chiShona (sna) | mC4 (all), and VOA | 545.2MB | 2,693,028 |
| Somali (som) | mC4 (all), and VOA | 1,000MB | 3,480,960 |
| Sesotho (sot) | mC4 (all) | 234MB | 1,107,565 |
| Kiswahili (swa) | mC4 (all) | 823.5MB | 4,220,346 |
| isiXhosa (xho) | mC4 (all), and Isolezwe Newspaper | 178.4MB | 832,954 |
| Yorùbá (yor) | mC4 (all), Alaroye News, Asejere News, Awikonko News, BBC, and Voice of Nigeria (VON) | 179.3MB | 897,299 |
| isiZulu (zul) | mC4 (all), and Isolezwe Newspaper | 700.7MB | 3,252,035 |

Table 5.1: Monolingual Corpora (after pre-processing – we followed AfriBERTa (Ogueji, Zhu, and Lin, 2021) approach) , their sources, size (MB), and number of sentences.

### 5.3.2 *Evaluation tasks*

We run our experiments on two sentence level classification tasks: news topic classification and sentiment classification, and one token level classification task: NER. We evaluate our models on English as well as diverse African languages with different linguistic characteristics.

---

[5] https://commoncrawl.org/
[6] https://www.voanews.com

#### 5.3.2.1 *Existing datasets*

NER. For the NER task, we evaluate on the MasakhaNER dataset (Adelani et al., 2021b), a manually annotated dataset covering 10 African languages (Amharic, Hausa, Igbo, Kinyarwanda, Luganda, Luo, Naija, Kiswahili, Wolof, and Yorùbá) with texts from the news domain. For English, we use data from the CoNLL 2003 NER task (Tjong Kim Sang and De Meulder, 2003), also containing texts from the news domain. For isiXhosa, we use the data from Eiselen (2016). Lastly, we evaluate on Arabic, we make use of the ANERCorp dataset (Benajiba, Rosso, and BenedíRuiz, 2007; Obeid et al., 2020).

NEWS TOPIC CLASSIFICATION. We use existing news topic datasets for Amharic (Azime and Mohammed, 2021), English – AG News corpus – (Zhang, Zhao, and LeCun, 2015), Kinyarwanda – KINNEWS – (Niyongabo et al., 2020), Kiswahili – new classification dataset– (David, 2020), and both Yorùbá and Hausa (Hedderich et al., 2020). For datasets without a development set, we randomly sample 5% of their training instances and use them as a development set.

SENTIMENT CLASSIFICATION. We use the NaijaSenti multilingual Twitter sentiment analysis corpus (Muhammad et al., 2022). This is a large code-mixed and monolingual sentiment analysis dataset, manually annotated for 4 Nigerian languages: Hausa, Igbo, Yorùbá and Pidgin. Additionally, we evaluate on the Amharic, and English Twitter sentiment datasets by Yimam et al. (2020) and Rosenthal, Farra, and Nakov (2017), respectively. For all the datasets above, we only make use of tweets with positive, negative and neutral sentiments.

#### 5.3.2.2 *Newly created dataset: ANTC Corpus*

We created a novel dataset, ANTC—**A**frican **N**ews **T**opic **C**lassification for five African languages. We obtained data from three different news sources: VOA, BBC[7] and isolezwe[8]. From the VOA data, we created datasets for Lingala and Somali. We obtained the topics from data released by Palen-Michel, Kim, and Lignos (2022) and used the provided URLs to get the news category from the websites. For Naija, Malagasy and isiZulu, we scrapped news topic from the respective news website (BBC Pidgin, Global Voices and isolezwe, respectively) directly based on their category. We noticed that some news topics are not mutually exclusive to their categories, therefore, we filtered such topics with multiple labels. Also, we ensured that each category has at least 200 samples. The categories include but are not limited to: Africa, Entertainment, Health, and Politics. The pre-processed datasets were divided into training, development, and test sets using stratified sampling with a ratio of 70:10:20. Table 5.2 provides details about the dataset size and news topic information.

---

7 https://www.bbc.com/pidgin
8 https://www.isolezwe.co.za

| Language | Train/ Dev/ Test | classes | # classes |
|---|---|---|---|
| *Newly created datasets* | | | |
| Lingala (lin) | 1536/ 220/ 440 | Rdc, Politiki/Politique, Bokengi/Securite, Justice, Bokolongono/Santé/Medecine | 5 |
| Naija (pcm) | 1165/ 167/ 333 | Entertainment, Africa, Sport, Nigeria, World | 5 |
| Somali (som) | 10K /1440/ 2879 | Soomaaliya (Somalia), Wararka (News), Caalamka (World), Maraykanka (United States), Afrika (Africa) | 6 |
| Malagasy (mlg) | 3905/ 559/ 1117 | Politika (Politics), Kolontsaina (Culture), Zon'olombelona (Human Rights), Siansa sy Teknolojia (Science and Technology), Tontolo iainana (Environment) | 5 |
| isiZulu (zul) | 2961/ 424/ 847 | Ezemidlalo (Sports), Ezokungcebeleka (Recreation), Imibono (Ideas), Ezezimoto (Automotive), Intandokazi (Favorites) | 5 |
| *Existing datasets* | | | |
| Amharic (amh) | 36K/ 5147 / 10K | Local News, Sport, Politics, International News, Business, Entertainment | 6 |
| English (eng) | 114K/ 6000/ 7,600 | World, Sports, Business, Sci/Tech | 4 |
| Hausa (hau) | 2045/ 290/ 582 | Africa, World, Health, Nigeria, Politics | 5 |
| Kinyarwanda (kin) | 16K/ 851/ 4254 | Politics, Sport, Economy, Health, Entertainment, History, Technology, Tourism, Culture, Fashion, Religion, Environment, Education, Relationship | 14 |
| Kiswahili (swa) | 21K/ 1111/ 7338 | Uchumi (Economic), Kitaifa (National), Michezo (Sports), Kimataifa (International), Burudani (Recreation), Afya (Health) | 6 |
| Yorùbá (yor) | 1340/ 189/ 379 | Nigeria, Africa, World, Entertainment, Health, Sport, Politics | 7 |

Table 5.2: Number of sentences in training, development and test splits. We provide automatic translation of some of the African language words to English (in Parenthesis) using Google Translate.

## 5.4 MULTILINGUAL PRE-TRAINED LANGUAGE MODELS

For our experiments, we use different multilingual PLMs that have been trained using a masked language model objective on large collections of monolingual texts from several languages. Table 5.3 shows the number of parameters as well as the African languages covered by each of the models we consider.

1. XLM-R (Conneau et al., 2020) has been pre-trained on 100 languages, including eight African languages. We make use of the XLM-R-base model with 270M parameters for MAFT because it was easier to adapt to more languages due to its smaller size compared to XLM-R-large. We additionally evaluated XLM-R-large when directly fine-tuned on the target language data to compare its performance to the MAFT-adapted models that are of smaller sizes.

| PLM | # Lang. | African languages covered |
| --- | --- | --- |
| XLM-R-base (270M) | 100 | afr, **amh**, **hau**, mlg, orm, **som**, **swa**, **xho** |
| AfriBERTa-large (126M) | 11 | **amh**, **hau**, **ibo**, **kin**, run, orm, **pcm**, **som**, **swa**, tir, **yor** |
| XLM-R-miniLM (117M) | 100 | afr, **amh**, **hau**, mlg, orm, **som**, **swa**, **xho** |
| XLM-R-large (550M) | 100 | afr, **amh**, **hau**, mlg, orm, **som**, **swa**, **xho** |
| AfroXLMR* (117M-270M) | 20 | afr, **amh**, **hau**, **ibo**, **kin**, run mlg, nya, orm, **pcm**, **sna**, **som**, sot, **swa**, **xho**, **yor**, **zul** |

Table 5.3: Language coverage and size for pre-trained language models. Languages in **bold** have evaluation datasets for either NER, news topic classification or sentiment analysis.

2. AfriBERTa (Ogueji, Zhu, and Lin, 2021) has been pre-trained only on African languages. Despite its smaller parameter size (110M), it has been shown to reach competitive performance to XLM-R-base on African language datasets (Adelani et al., 2021b; Hedderich et al., 2020).

3. XLM-R-miniLM (Wang et al., 2020a) is a distilled version of XLM-R-large with only 117M parameters.

HYPER-PARAMETERS FOR BASELINE MODELS  We fine-tune the baseline models for NER, news topic classification and sentiment classification for 50, 25, and 20 epochs, respectively. We use a learning rate of 5e-5 for all the tasks, except for sentiment classification where we use 2e-5 for XLM-R-base and XLM-R-large. The maximum sequence length is 164 for NER, 500 for news topic classification, and 128 for sentiment classification. The adapted models also make use of similar hyper-parameters.

## 5.5 MULTILINGUAL ADAPTIVE FINE-TUNING

We introduce MAFT as an approach to adapting a multi-lingual PLM to a new set of languages. Adapting PLMs has been shown to be effective when adapting to a new domain (Gururangan et al., 2020) or language (Adelani et al., 2021b; Alabi et al., 2020; Muller et al., 2021; Pfeiffer et al., 2020b). While previous work on multilingual adaptation has mostly focused on autoregressive sequence-to-sequence models such as mBART (Tang et al., 2020), in this work, we adapt non-autoregressive masked PLMs on monolingual corpora covering 20 languages. Crucially, during adaptation, we use the same objective that was also used during pre-training. The models resulting from MAFT were then fine-tuned on supervised NLP downstream tasks. As a first step, we applied MAFT to smaller models (XLM-R-base, AfriBERTa, and XLM-R-miniLM) before scaling it to a bigger PLM since one of our goals is to reduce model size. We name the model resulting after applying MAFT to XLM-R-base and XLM-R-miniLM as *AfroXLMR-base* and *AfroXLMR-mini*, respectively. For adaptation, we train on a combination of the monolingual corpora used for AfriMT5 adaptation by Adelani et al. (2022b). Details for each of the monolingual corpora and languages are provided in Table 5.1

HYPER-PARAMETERS FOR MAFT   The PLMs were trained on 3 epochs with a learning rate of 5e-5, using huggingface transformers (Wolf et al., 2020). We use a batch size of 32 for AfriBERTa and a batch size 10 for the other PLMs.

### 5.5.1 *Pre-trained LM vocabulary reduction*

Multilingual PLMs come with various parameter sizes, the larger ones having more than a hundred million parameters, which makes fine-tuning and deploying such models challenging due to resource constraints. One of the major factors that contribute to the parameter size of these models is the embedding matrix, whose size is a function of the vocabulary size of the model. While a large vocabulary size is essential for a multilingual PLM trained on hundreds of languages, some of the tokens in the vocabulary can be removed when they are irrelevant to the domain or language considered in the downstream task, thus reducing the vocabulary size of the model. Inspired by Abdaoui, Pradel, and Sigel (2020), we experiment with reducing the vocabulary size of the XLM-R-base model before adapting via MAFT. There are two possible vocabulary reductions in our setting: *(1) removal of tokens before MAFT or (2) removal of tokens after MAFT*. From our preliminary experiments, we find approach (1) to work better. We call the resulting model, *AfroXLMR-small*.

To remove non-African vocabulary sub-tokens from the pretrained XLM-base model, we concatenated the monolingual texts from 19 out of the 20 African languages except Amharic. Then, we apply `sentencepiece` to the Amharic monolingual texts, and concatenated texts separately using the original XLM-R-base tokenizer. The frequency of all the sub-tokens in the two separate monolingual corpora is computed, and we select the top-k most frequent tokens from the separate corpora. We used this separate sampling to ensure that a considerate number of Amharic sub-tokens are captured in the new vocabulary. We try to justify the choice of this approach in Section 5.5.3. We assume that the top-k most frequent tokens should be representative of the vocabulary of the whole 20 languages. We chose $k = 52.000$ from the Amharic sub-tokens which covers 99.8% of the Amharic monolingual texts, and $k = 60.000$ which covers 99.6% of the other 19 languages, and merged them. In addition, we include the top 1.000 tokens from the original XLM-R-base tokenizer in the new vocabulary to include frequent tokens that were not present in the new top-k tokens.[9] We note that our assumption above may not hold in the case of some very distant and low-resourced languages, as well as when there are domain differences between the corpora used during adaptation and fine-tuning. We leave the investigation of alternative approaches for vocabulary compression for future work.

| Method | Size | amh | ara | eng | hau | ibo | kin | lug | luo | pcm | swa | wol | xho | yor | avg |
|---|---|---|---|---|---|---|---|---|---|---|---|---|---|---|---|
| Finetune | | | | | | | | | | | | | | | |
| XLM-R-miniLM | 117M | 69.5 | 76.1 | 91.5 | 74.5 | 81.9 | 68.6 | 64.7 | 11.7 | 83.2 | 86.3 | 51.7 | 69.3 | 72.0 | 69.3 |
| AfriBERTa | 126M | 73.8 | 51.3 | 89.0 | 90.2 | 87.4 | 73.8 | 78.9 | 70.2 | 85.7 | 88.0 | 61.8 | 67.2 | 81.3 | 76.8 |
| XLM-R-base | 270M | 70.6 | 77.9 | 92.3 | 89.5 | 84.8 | 73.3 | 79.7 | 74.9 | 87.3 | 87.4 | 63.9 | 69.9 | 78.3 | 79.2 |
| XLM-R-large | 550M | 76.2 | 79.7 | 93.1 | 90.5 | 84.1 | 73.8 | 81.6 | 73.6 | 89.0 | 89.4 | 67.9 | 72.4 | 78.9 | 80.8 |
| MAFT + Finetune | | | | | | | | | | | | | | | |
| XLM-R-miniLM | 117M | 69.7 | 76.5 | 91.7 | 87.7 | 83.5 | 74.1 | 77.4 | 17.5 | 85.5 | 86.0 | 59.0 | 72.3 | 75.1 | 73.5 |
| AfriBERTa | 126M | 72.5 | 40.9 | 90.1 | 89.7 | 87.6 | 75.2 | 80.1 | 69.6 | 86.5 | 87.6 | 62.3 | 71.8 | 77.0 | 76.2 |
| XLM-R-base | 270M | 76.1 | 79.7 | **92.8** | 91.2 | 87.4 | 78.0 | 82.9 | 75.1 | 89.6 | 88.6 | 67.4 | 71.9 | 82.1 | 81.8 |
| XLM-R-base-v70k | 140M | 70.1 | 76.4 | 91.0 | 91.4 | 86.6 | 77.5 | 83.2 | **75.4** | 89.0 | 88.7 | 65.9 | 72.4 | 81.3 | 80.7 |
| XLM-R-base+LAFT | 270M x 13 | **78.0** | 79.1 | 91.3 | **91.5** | 87.7 | 77.8 | **84.7** | 75.3 | **90.0** | **89.5** | **68.3** | **73.2** | **83.7** | **82.3** |

Table 5.4: NER model comparison, showing F1-score on the test sets after 50 epochs averaged over 5 runs. Results are for all 4 tags in the dataset: PER, ORG, LOC, DATE/MISC. For LAFT, we multiplied the size of XLM-R-base by the number of languages as LAFT results in a single model per language.

| Method | Size | amh | eng | hau | kin | lin | mlg | pcm | som | swa | yor | zul | avg |
|---|---|---|---|---|---|---|---|---|---|---|---|---|---|
| Finetune | | | | | | | | | | | | | |
| XLM-R-miniLM | 117M | 70.4 | 94.1 | 77.6 | 64.2 | 41.2 | 42.9 | 67.6 | 74.2 | 86.7 | 68.8 | 56.9 | 67.7 |
| AfriBERTa | 126M | 70.7 | 93.6 | 90.1 | 75.8 | 55.4 | 56.4 | 81.5 | **79.9** | 87.7 | **82.6** | 71.4 | 76.8 |
| XLM-R-base | 270M | 71.1 | 94.1 | 85.9 | 73.3 | 56.8 | 54.2 | 77.3 | 78.8 | 87.1 | 71.1 | 70.0 | 74.6 |
| XLM-R-large | 550M | 72.7 | 94.5 | 86.2 | 75.1 | 52.2 | 63.6 | 79.4 | 79.2 | 87.5 | 74.8 | 78.7 | 76.7 |
| MAFT + Finetune | | | | | | | | | | | | | |
| XLM-R-miniLM | 117M | 69.5 | 94.1 | 86.7 | 72.0 | 51.7 | 55.3 | 78.1 | 77.7 | 87.2 | 74.0 | 60.3 | 73.3 |
| AfriBERTa | 126M | 68.8 | 93.7 | 89.5 | 76.5 | 54.9 | 59.7 | **82.2** | **79.9** | 87.7 | 80.8 | 76.4 | 77.3 |
| XLM-R-base | 270M | 71.9 | **94.6** | 88.3 | **76.8** | **58.6** | 64.7 | 78.9 | 79.1 | 87.8 | 80.2 | **79.6** | 78.2 |
| XLM-R-base-v70k | 140M | 70.4 | 94.2 | 87.7 | 76.1 | 56.8 | 64.4 | 76.1 | 79.4 | 87.4 | 76.9 | 77.4 | 76.9 |
| XLM-R-base+LAFT | 270M x 11 | **73.0** | 94.3 | **91.2** | 76.0 | 56.9 | **67.3** | 77.4 | 79.4 | **88.0** | 79.2 | 79.5 | **78.4** |

Table 5.5: News Topic Classification, showing F1-score on the test sets after 25 epochs averaged over 5 runs. For LAFT, we multiplied the size of XLM-R-base by the number of languages.

| Model | Size | amh | eng | hau | ibo | pcm | yor | Avg |
|---|---|---|---|---|---|---|---|---|
| Finetune | | | | | | | | |
| XLM-R-miniLM | 117M | 51.0 | 62.8 | 75.0 | 78.0 | 72.9 | 73.4 | 68.9 |
| AfriBERTa-large | 126M | 51.7 | 61.8 | 81.0 | **81.2** | 75.0 | 80.2 | 71.8 |
| XLM-R-base | 270M | 51.4 | 66.2 | 78.4 | 79.9 | 76.3 | 76.9 | 71.5 |
| XLM-R-large | 550M | 52.4 | **67.5** | 79.3 | 80.8 | **77.6** | 78.1 | 72.6 |
| MAFT+Finetune | | | | | | | | |
| XLM-R-miniLM | 117M | 51.3 | 63.3 | 77.7 | 78.0 | 73.6 | 74.3 | 69.7 |
| AfriBERTa | 126M | 53.6 | 63.2 | 81.0 | 80.6 | 74.7 | 80.4 | 72.3 |
| XLM-R-base | 270M | 53.0 | 65.6 | 80.7 | 80.5 | 77.5 | 79.4 | 72.8 |
| XLM-R-base-v70k | 140M | 52.2 | 65.3 | 80.6 | 81.0 | 77.4 | 78.6 | 72.5 |
| XLM-R-base+LAFT | 270M x 6 | **55.0** | 65.6 | **81.5** | 80.8 | 74.7 | **80.9** | **73.1** |

Table 5.6: Sentiment Classification, showing F1 evaluation on test sets after 20 epochs, averaged over 5 runs. We obtained the results for the baseline model results of "hau", "ibo", "pcm", and "yor" from Muhammad et al. (2022). For LAFT, we multiplied the size of XLM-R-base by the number of languages as LAFT results in a single model per language.

### 5.5.2 Results and discussion

#### 5.5.2.1 Baseline results

For the baseline models (top rows in Tables 5.4, 5.5, and 5.6), we directly fine-tune on each of the downstream tasks in the target language: NER, news topic classification and sentiment analysis.

PERFORMANCE OF LANGUAGES SEEN DURING PRE-TRAINING. For NER and sentiment analysis, we find XLM-R-large to give the best overall performance. We attribute this to the fact that it has a larger model capacity compared to the other PLMs. Similarly, we find AfriBERTa and XLM-R-base to give better results on languages they have been pre-trained on, and in most cases, AfriBERTa tends to perform better than XLM-R-base on languages they are both pre-trained on, for example, `amh`, `hau`, and `swa`. However, when the languages are unseen by AfriBERTa (e.g. `ara`, `eng`, `wol`, `lin`, `lug`, `luo`, `xho`, `zul`), it performs much worse than XLM-R-base and in some cases, even worse than the XLM-R-miniLM. This shows that it may be better to adapt to a new African language from a PLM that has seen numerous languages than one trained on a subset of African languages from scratch.

LAFT IS A STRONG BASELINE. Here, we applied LAFT to the XLM-R-base model (last row in Tables 5.4, 5.5, and 5.6). As our results show,

---

9 This introduced just a few new tokens which are mostly English tokens to the new vocabulary. For $k = 70.000$ we end up with 70.609 tokens.

applying LAFT on each language individually provides a significant improvement in performance across all languages and tasks we evaluated on. Sometimes, the improvement is very large, for example, (+7.4) F1 on Amharic NER and (+9.5) F1 for Zulu news-topic classification. The only exception is for English since XLM-R has already seen large amounts of English text during pre-training. Additionally, LAFT models tend to give slightly worse result when adaptation is performed on a smaller corpus[10].

5.5.2.2 *Multilingual adaptive fine-tuning results*

While LAFT provides an upper bound on downstream performance for most languages, our new approach is often competitive with LAFT. On average, the difference on NER, news topic, and sentiment classification is (−0.5), (−0.2), and (−0.3) F1, respectively. Crucially, compared to LAFT, MAFT results in a single adapted model which can be applied to many languages, while LAFT results in a new model for each language. Below, we discuss our results in more detail.

PLMS PRE-TRAINED ON MANY LANGUAGES BENEFIT THE MOST FROM MAFT. We found all the PLMs to improve after we applied MAFT. The improvement is the largest for the XLM-R-miniLM where, the performance improved by (+4.2) F1 for NER, and (+5.6) F1 for news topic classification. Although, the improvement was lower for sentiment classification (+0.8). Applying MAFT on XLM-R-base gave the overall best result. On average, there is an improvement of (+2.6), (+3.6), and (+1.5) on NER, news topic and sentiment classification, respectively. The main advantage of MAFT is that it allows us to use the same model for many African languages and cross-lingual transfer tasks instead of many models specialized to individual languages. This significantly reduces the required disk space to store the models, without sacrificing performance. Interestingly, there is no strong benefit of applying MAFT to AfriBERTa. In most cases the improvement is (< 0.6). We speculate that this is probably due to AfriBERTa's tokenizer having limited coverage. We leave a more detailed investigation of this for future work.

MORE EFFICIENT MODELS USING VOCABULARY REDUCTION. Applying vocabulary reduction helps to reduce the model size by more than 50% before applying MAFT. We find a slight reduction in performance as we remove more vocabulary tokens. The average performance of XLM-R-base-v70k reduces by (−1.6), (−1.5) and (−0.6) F1 for NER, news topic, and sentiment classification compared to the XLM-R-base+LAFT. Despite the reduction in performance compared to XLM-R-base+LAFT, they are still better than XLM-R-miniLM, which has a similar model size, with or without MAFT. We also find that their performance is better than that of the PLMs that have not undergone any adaptation. We find the largest reduction in performance on languages that make use of non-Latin scripts i.e. amh and ara—they make use of the Ge'ez script and Arabic script respectively. The vocabulary reduction impact the number of amh and ara subwords that are covered by our `tokenizer`.

---

10 We performed LAFT on English using VOA news corpus with about 906.6MB, much smaller than the CC-100 English corpus (300GB)

In summary, we recommend XLM-R-base+MAFT (i.e. AfroXLMR-base) for all languages on which we evaluated, including high-resource languages like English, French and Arabic. If there are GPU resource constraints, we recommend using XLM-R-base-v70k+MAFT (i.e. AfroXLMR-small).

### 5.5.3 *Investigating the drop in performance for models with reduced vocabulary*

Our results showed that applying vocabulary reduction reduced the model size, but we also observed a drop in performance for different languages across the downstream tasks, especially for Amharic, because it uses a non-Latin script. Hence, we compared different sampling strategies for selecting the top-k vocabulary sub-tokens. These includes: (i) concatenating the monolingual texts, and selecting the top-70k sub-tokens (ii) the exact approach described in Section 5.5.1. The resulting tokenizers from the two reduced methods are used to tokenize the sentences in the NER test sets for Amharic, Arabic, English, and Yorùbá. Table 5.7 shows the number of UNKs in the respective test set after tokenization and the F1 scores obtained on the NER task for the languages. The table shows that the original AfroXLMR tokenizer obtained the least number of UNKs for all languages, with the highest F1 scores. Although Yorùbá has 24 UNKs, the fact that Yorùbá was not seen during pretraining explains the reason behind this. Furthermore, using approach (i), gave 3704 UNKs for Amharic, but with approach (ii) there was a significant drop in the number of UNKs and an improvement in F1 score. We noticed a drop in the vocabulary coverage for the other languages as we increased the Amharic sub-tokens. Therefore, we concluded that there is no sweet spot in terms of the way to pick the vocabulary that covers all languages, and this calls for more research.

### 5.5.4 *Scaling MAFT to larger models*

Here, we applied MAFT to XLM-R-large using the same training setup as XLM-R-base. We refer to the new PLM as AfroXLMR-large. We also trained individual LAFT models using the monolingual data[11] from Adelani et al. (2021b). Table 5.8 shows the evaluation result on NER. On average over 13 languages, AfroXLMR-large improved over XLM-R-large by +2.8, which is very comparable to the improvement we obtained between AfroXLMR-base (81.8) and XLM-R-base (79.2). Surprisingly, the improvement is quite large (+3.5 to +6.3 points) for seven out of ten African languages: yor, luo, lug, kin, ibo, and amh. The most interesting observation is that AfroXLMR-large on average is either competitive or slightly better than individual language LAFT models, including languages not included in the MAFT training like lug, luo and wol. This implies that AfroXLMR-large (a single model) provides a better alternative to XLM-R-large+LAFT (for each language) in terms of performance on downstream tasks and disk space required to store individual LAFT models. AfroXLMR-large is the largest masked language model for African languages, and it achieves the state-of-the-art over all other multilingual PLM on the NER task. This shows that our MAFT approach is very effective and scales to larger PLMs.

---

11 For languages not in MasakhaNER, we use the same monolingual data in Table 5.1.

| Models | amh | | ara | | eng | | yor | |
|---|---|---|---|---|---|---|---|---|
| | #UNK | F1 | #UNK | F1 | #UNK | F1 | #UNK | F1 |
| AfroXLMR-base | 0 | 76.1 | 0 | 79.7 | 0 | 92.8 | 24 | 82.1 |
| Afro-XLM-R70k (i) | 3704 | 67.8 | 1403 | 76.2 | 44 | 90.6 | 5547 | 81.2 |
| Afro-XLM-R70k (ii) | 3395 | 70.1 | 669 | 76.4 | 54 | 91.0 | 6438 | 81.3 |

Table 5.7: Numbers of UNKs when the model tokenizers are applied on the NER test sets.

| Method | Size | amh | ara | eng | hau | ibo | kin | lug | luo | pcm | swa | wol | xho | yor | avg |
|---|---|---|---|---|---|---|---|---|---|---|---|---|---|---|---|
| XLM-R-large | 550M | 76.2 | 79.7 | 93.1 | 90.5 | 84.1 | 73.8 | 81.6 | 73.6 | 89.0 | 89.4 | 67.9 | 72.4 | 78.9 | 80.8 |
| XLM-R-large+LAFT | 550M x 13 | 79.9 | 81.3 | 92.2 | 91.7 | 87.7 | 78.4 | 86.2 | 78.2 | 91.1 | 90.3 | 68.8 | 72.7 | 82.9 | 83.2 |
| AfroXLMR-large | 550M | 79.7 | 80.9 | 92.2 | 91.2 | 87.7 | 79.1 | 86.7 | 78.1 | 91.0 | 90.4 | 69.6 | 72.9 | 85.2 | 83.4 |

Table 5.8: NER model comparison on XLM-R-large, showing F1-score on the test sets after 50 epochs averaged over 5 runs. Results are for all 4 tags in the dataset: PER, ORG, LOC, DATE/MISC.

## 5.6 CROSS-LINGUAL TRANSFER

Parameter-efficient fine-tuning methods like *adapters* (Houlsby et al., 2019) are appealing because of their modularity, portability, and composability across languages and tasks. Oftentimes, language adapters are trained on a general domain corpus like Wikipedia. However, when there is a mismatch between the target *domain* of the task and the *domain* of the language adapter, it could also impact the cross-lingual performance.

Here, we investigate how we can improve the cross-lingual abilities of our adapted PLM—AfroXLMR-base by training language adapters on the same domain as the target task. We make use of MasakhaNER dataset that is based on the news domain for experiments. We compare the performance of language adapters trained on *Wikipedia* and *news* domains. In addition to adapters, we experiment with another parameter-efficient method based on the Lottery-Ticket Hypothesis (Frankle and Carbin, 2019), i.e. LT-SFT (Ansell et al., 2021).

For the adapter approach, we make use of the MAD-X approach (Pfeiffer et al., 2020b)—an adapter-based framework that enables cross-lingual transfer to arbitrary languages by learning modular language and task representations. Specifically, we make use of MAD-X 2.0 (Pfeiffer et al., 2021) where the last adapter layers are dropped, which has been shown to improve performance. The setup is as follows: (1) We train language adapters via masked language modelling (MLM) individually on source and target languages, the corpora used are described in Table 5.9; (2) We train a task adapter by optimising a target task on labelled data in a source language. (3) At inference, task and language adapters are stacked together by substituting the source language adapter with a target language adapter.

We also make use of the Lottery Ticket Sparse Fine-tuning (LT-SFT) approach (Ansell et al., 2021), a parameter-efficient fine-tuning approach that has been shown to give competitive or better performance than the MAD-X 2.0 approach. The LT-SFT approach is based on the Lottery Ticket Hypothesis (LTH), which states that each neural model contains a sub-network (a "winning ticket") that, if trained again in isolation, can reach or even surpass the performance of the original model. The LTH is originally a compression approach, the authors of LT-SFT re-purposed the approach for cross-lingual adaptation by finding a sparse sub-networks for tasks and languages, that will later be composed together for zero-shot adaptation, similar to Adapters.

| **Language** | **Source** | **Size (MB)** | **No. of sentences** |
|---|---|---|---|
| Amharic (amh) | VOA (Palen-Michel, Kim, and Lignos, 2022) | 19.9MB | 72,125 |
| Hausa (hau) | VOA (Palen-Michel, Kim, and Lignos, 2022) | 46.1MB | 235,614 |
| Igbo (ibo) | BBC Igbo (Ogueji, Zhu, and Lin, 2021) | 16.6MB | 62,654 |
| Kinyarwanda (kin) | KINNEWS (Niyongabo et al., 2020) | 35.8MB | 61,910 |
| Luganda (lug) | Bukedde | 7.9MB | 67,716 |
| Luo (luo) | Ramogi FM news and MAFAND-MT (Adelani et al., 2022b) | 1.4MB | 8,684 |
| Naija (pcm) | BBC | 50.2MB | 161,843 |
| Kiswahilii (swa) | VOA (Palen-Michel, Kim, and Lignos, 2022) | 17.1MB | 88,314 |
| Wolof (wol) | Lu Defu Waxu, Saabal, Wolof Online, and MAFAND-MT (Adelani et al., 2022b) | 2.3MB | 13,868 |
| Yorùbá (yor) | BBC Yorùbá | 15.0MB | 117,124 |

Table 5.9: Monolingual News Corpora used for language adapter and SFT training, their sources and size (MB), and number of sentences.

5.6.1 *Experimental setup*

For our experiments, we followed the same setting as Ansell et al. (2021), that adapted English CoNLL03 (Tjong Kim Sang and De Meulder, 2003) as the source language to African languages (using MasakhaNER dataset) for the NER task using mBERT. Furthermore, we adapted the same CoNLL03 dataset to MasakhaNER[12] using other PLMs like XLMR-base and AfroXLMR-base. For the training of MAD-X 2.0 and sparse fine-tunings (SFT) for African languages, we make use of the monolingual texts from the news domain (see Table 5.9) since it matches the domain of the evaluation data. Unlike Ansell et al. (2021), that trained adapters and SFT on monolingual data from the Wikipedia domain except for luo and pcm, where the dataset is absent, we show that the domain used for training language SFT is also very important. For a fair comparison, we reproduced the result of Ansell et al. (2021) by training MAD-X 2.0 and LT-SFT on mBERT, XLM-R-base and AfroXLMR-base on target languages with the news domain corpus. But, we still make use of the pre-trained English language adapter[13] and SFT[14] for mBERT and XLM-R-base trained on the Wikipedia domain. For the AfroXLMR-base, we make use of the same English adapter and SFT as XLM-R-base because the PLM is already good for the English language. We make use of the same hyper-parameters reported in the LT-SFT paper.

HYPER-PARAMETERS FOR ADAPTERS  We train the task adapter using the following hyper-parameters: batch size of 8, 10 epochs, "pfeiffer" adapter config, adapter reduction factor of 8, and learning rate of 5e-5. For the language adapters, we make use of 100 epochs or maximum steps of 100K, minimum number of steps is 30K, batch size of 8, "pfeiffer+inv" adapter config, adapter reduction factor of 2, learning rate of 5e-5, and maximum sequence length of 256. For a fair comparison with adapter models trained on the Wikipedia domain, we used the same hyper-parameter settings (Ansell et al., 2021) for the news domain.

5.6.2 *Results and discussion*

Table 5.10 shows the results of MAD-X 2.0 and LT-SFT, we compare their performance to fully supervised setting, where we fine-tune XLM-R-base on the training dataset of each of the languages, and evaluate on the test-set. We find that both MAD-X 2.0 and LT-SFT using the news domain for African languages produce better performance (+4.2 on MAD-X and +3.7 on LT-SFT) than the ones trained largely on the Wikipedia domain. This shows that the domain of the data matters. Also, we find that training LT-SFT on XLM-R-base gives better performance than mBERT on all languages. For MAD-X, there are a few exceptions like hau, pcm, and yor. Overall, the best performance is obtained by training LT-SFT on AfroXLMR-base, and sometimes it gives better performance than the fully-supervised setting (e.g. as observed in kin and lug, wol yor languages). On both MAD-X and LT-SFT, AfroXLMR-base gives the best result since it has been firstly adapted on several African languages and secondly on the target domain of the target task. This shows that the MAFT approach is effective since the technique provides a better PLM that parameter-efficient methods can benefit from.

---

12 We excluded the MISC and DATE from CoNLL03 and MasakhaNER respectively to ensure same label configuration.
13 https://adapterhub.ml/
14 https://huggingface.co/cambridgeltl

| Method | amh | hau | ibo | kin | lug | luo | pcm | swa | wol | yor | avg |
|---|---|---|---|---|---|---|---|---|---|---|---|
| XLM-R-base (fully-supervised) | 69.7 | 91.0 | 86.2 | 73.8 | 80.5 | 75.8 | 86.9 | 88.7 | 69.6 | 78.1 | 81.2 |
| mBERT (MAD-X) (Ansell et al., 2021) | - | 83.4 | 71.7 | 65.3 | 67.0 | 52.2 | 72.1 | 77.6 | 65.6 | 74.0 | 69.9 |
| mBERT (MAD-X on **news domain**) | - | 86.0 | 77.6 | 69.9 | 73.3 | 56.9 | 78.5 | 80.2 | 68.8 | 75.6 | 74.1 |
| XLM-R-base (MAD-X on **news domain**) | 47.5 | 85.5 | 83.2 | 72.0 | 75.7 | 57.8 | 76.8 | 84.0 | 68.2 | 72.2 | 75.0 |
| AfroXLMR-base (MAD-X on **news domain**) | 47.7 | 88.1 | 80.9 | 73.0 | 80.1 | 59.2 | 79.9 | 86.9 | 69.1 | 75.6 | 77.0 |
| mBERT (LT-SFT) (Ansell et al., 2021) | - | 83.5 | 76.7 | 67.4 | 67.9 | 54.7 | 74.6 | 79.4 | 66.3 | 74.8 | 71.7 |
| mBERT (LT-SFT on **news domain**) | - | 86.4 | 80.6 | 69.2 | 76.8 | 55.1 | 80.4 | 82.3 | 71.6 | 76.7 | 75.4 |
| XLM-R-base (LT-SFT on **news domain**) | 54.1 | 87.6 | 81.4 | 72.7 | 79.5 | 60.7 | 81.2 | 85.5 | 73.6 | 73.7 | 77.3 |
| AfroXLMR-base (LT-SFT on **news domain**) | 54.0 | 88.6 | 83.5 | 73.8 | 81.0 | 60.7 | 81.7 | 86.4 | 74.5 | 78.7 | 78.8 |

Table 5.10: Cross-lingual Transfer using LT-SFT (Ansell et al., 2021) and evaluation on MasakhaNER. The full-supervised baselines are obtained from Adelani et al. (2021b) to measure performance gap when annotated datasets are available. Experiments are performed on 3 tags: PER, ORG, LOC. Average (avg) excludes amh. The best zero-shot transfer F1-scores are underlined.

## 5.7 CONCLUSION

In this work, we proposed and studied MAFT as an approach to adapt multilingual PLMs to many African languages with a single model. We evaluated our approach on three different NLP downstream tasks and additionally contribute a novel news topic classification dataset for 4 African languages. Our results show that MAFT is competitive to LAFT while providing a single model compared to many models specialized for individual languages. We went further to show that combining vocabulary reduction and MAFT leads to a 50% reduction in the parameter size of XLM-R while still being competitive to applying LAFT on individual languages. We hope that future work improves on vocabulary reduction to provide even smaller models with a strong performance on distant and low-resource languages. To further research on NLP for African languages and reproducibility, we released language adapters, language SFTs, AfroXLMR-large,

AfroXLMR-base, AfroXLMR-small, and AfroXLMR-mini models on the HuggingFace Model Hub[15].

---

[15] https://huggingface.co/Davlan

Part III

NAMED ENTITY RECOGNITION FOR AFRICAN LANGUAGES

# 6

## DISTANT SUPERVISION FOR AFRICAN NER

This Chapter[1] describes the development of named entity recognition for two African languages (Hausa and Yorùbá) that are widely spoken in Africa. However, since we only have few labelled examples, we leverage techniques such as distant and weak supervision to create labeled data in a (semi-) automatic way. Additionally, to alleviate some of the negative effects of the errors in automatic annotation, we integrate noise-handling methods to the NER models. We evaluate two different deep learning architectures (BiLSTM and multilingual BERT), and show that distant supervision can be successfully leveraged in a realistic low-resource scenario where it can more than double a classifier's performance.

### 6.1 INTRODUCTION

Named Entity Recognition (NER) is a classification task that identifies words in a text that refer to entities (such as dates, person, organization and location names). It is a core task of natural language processing and a component for many downstream applications like search engines, knowledge graphs and personal assistants. For high-resource languages like English, this is a well-studied problem with complex state-of-the-art systems reaching close to or above 90% F1-score on the standard datasets CoNLL03 (Baevski et al., 2019) and Ontonotes (Akbik, Blythe, and Vollgraf, 2018). In recent years, research has been extended to a larger pool of languages including those of developing countries (Cao et al., 2019; Christianson, Duncan, and Onyshkevych, 2018; Mayhew et al., 2019; Zhang et al., 2018). Often, for these languages (like Hausa and Yorùbá studied here), there exists a large population with access to digital devices and internet (and therefore digital text), but natural language processing (NLP) tools do not support them.

One key reason is the absence of labeled training data required to train these systems. While manually labeled, gold-standard data is often only available in small quantities, it tends to be much easier to obtain large amounts of unlabeled text. Distant and weak supervision methods can then be used to create labeled data in a (semi-) automatic way. Using context (Mahajan et al., 2018; Wang et al., 2019), external knowledge and resources (Li et al., 2017; Pan et al., 2017), expert rules (Ratner et al., 2016; Ratner et al., 2019) or self-training (Chen,

---

1 This chapter is based on Adelani et al. (2020) with David Adelani, Michael A. Hedderich and Dawei Zhu contributing equally as first authors

Zhang, and Gao, 2018; Paul et al., 2019), a corpus or dataset can be labeled quickly and cheaply. Additionally, a variety of noise-handling methods have been proposed to circumvent the negative effects that errors in this automatic annotation might have on the performance of a machine learning classifier.

In this work, we study two methods of distant supervision for NER: Automatic annotation rules and matching of lists of entities from an external knowledge source. While distant supervision has been successfully used for high resource languages, it is not straight forward that these also work in low-resource settings where the amount of available external information might be much lower. The knowledge graph of Wikidata e.g. contains 4 million person names in English while only 32 thousand such names are available in Yorùbá, many of which are Western names.

Orthogonally to distant supervision, the pre-training of word embeddings is a key component for training many neural NLP models. A vector representation for words is built in an unsupervised fashion, i.e. on unlabeled text. Standard embedding techniques include Word2Vec (Mikolov et al., 2013a), GloVe (Pennington, Socher, and Manning, 2014) and FastText (Bojanowski et al., 2017). In the last two years, contextualized word embeddings have been proposed (Devlin et al., 2019; Peters et al., 2018; Radford et al., 2019). At the cost of having a much larger model size, these vector representations take the context of words into account and have been shown to outperform other embeddings in many tasks. In this study, we evaluate both types of representations.

The key questions we are interested in this paper are: How do NER models perform for Hausa and Yorùbá, two languages from developing countries? Are distant-supervision techniques relying on external information also useful in low-resource settings? How do simple and contextual word embeddings trade-off in model size and performance?

## 6.2 BACKGROUND & METHODS

### 6.2.1 *Languages*

**Hausa language** is the second most spoken indigenous language in Africa with over 40 million native speakers (Eberhard, Simons, and Fennig, 2021), and one of the three major languages in Nigeria, along with Igbo and Yorùbá. The language is native to the Northern part of Nigeria and the southern part of Niger, and it is widely spoken in West and Central Africa as a trade language in eight other countries: Benin, Ghana, Cameroon, Togo, Côte d'Ivoire, Chad, Burkina Faso, and Sudan. Hausa has several dialects but the one regarded as standard Hausa is the *Kananci* spoken in the ancient city of Kano in Nigeria.

Kananci is the dialect popularly used in many local (e.g VON news[2]) and international news media such as BBC, VOA, DW and Radio France Internationale. Hausa is a tone language but the tones are often ignored in writings, the language is written in a modified Latin alphabet. Despite the popularity of Hausa as an important regional language in Africa and it's popularity in news media, it has very little or no labelled data for common NLP tasks such as text classification, named entity recognition and question answering.

**Yorùbá language** is the third most spoken indigenous language in Africa after Swahilli and Hausa with over 35 million native speakers (Eberhard, Simons, and Fennig, 2021). The language is native to the South-western part of Nigeria and the Southern part of Benin, and it is also spoken in other countries like Republic of Togo, Ghana, Côte d'Ivoire, Sierra Leone, Cuba and Brazil. Yorùbá has several dialects but the written language has been standardized by the 1974 Joint Consultative Committee on Education (Asahiah, Odejobi, and Adagunodo, 2017), it has 25 letters without the Latin characters (c, q, v, x and z) and with additional characters (ẹ, gb, ṣ , ọ). Yorùbá is a tone language and the tones are represented as diacritics in written text, there are three tones in Yorùbá namely low ( \), mid ("−") and high (/). The mid tone is usually ignored in writings. Often time articles written online including news articles[3] like BBC and VON ignore diacritics. Ignoring diacritics makes it difficult to identify or pronounce words except they are in a context. For example, *owó* (money), *ọwọ̀* (broom), *òwò* (business), *ọ̀wọ̀* (honour), *ọwọ́* (hand), and *ọ̀wọ́* (group) will be mapped to *owo* without diacritics. Similar to the Hausa language, there are few or no labelled datasets for NLP tasks.

### 6.2.2 *Datasets & Embeddings*

The Hausa data used in this paper is part of the LORELEI[4] language pack. It consists of Broad Operational Language Translation (BOLT) data gathered from news sites, forums, weblogs, Wikipedia articles and twitter messages. We use a split of 10k training and 1k test instances. Due to the Hausa data not being publicly available at the time of writing, we could only perform a limited set of experiments on it.

The Yorùbá NER data used in this work is the annotated corpus of Global Voices news articles[5] recently released by Alabi et al. (2020). The dataset consists of 1,101 sentences (26k tokens) divided into 709 training sentences, 113 validation sentences and 279 test sentences based on 65%/10%/25% split ratio. The named entities in the dataset are personal names (PER), organization (ORG), location (LOC) and date & time (DATE). All other tokens are assigned a tag of "O".

---

2 https://www.von.gov.ng/hausa/
3 https://www.von.gov.ng/yoruba/, and https://www.bbc.com/yoruba
4 https://www.darpa.mil/program/low-resource-languages-for-emergent-incidents
5 https://yo.globalvoices.org/

For the Yorùbá NER training, we make use of Yorùbá FastText embeddings (Alabi et al., 2020) and multilingual-BERT[6] that was trained on 104 languages including Yorùbá. Instead of the original FastText embeddings (Grave et al., 2018), we chose FastText embeddings trained on a multi-domain and high-quality dataset (Alabi et al., 2020) because it gave better word similarity scores.

6.2.3 *Distant and Weak Supervision*

In this work, we rely on two sources of distant supervision chosen for its ease of application:

**Rules** allow to apply the knowledge of domain experts without the manual effort of labeling each instance. They are especially suited for entities that follow specific patterns, like time phrases in text (see also (Strötgen et al., 2018)). We use them for the DATE entity. In Yoruba, date expressions are written with the keywords of "*ọjọ́*" (day), "*oṣù*" (month), and "*ọdún*" (year). Similarly, time expressions are written with keywords such as "*wákàtí*" (hour), "*ìṣéjú* (minute) and "*ìṣéjú-aaya* (seconds). Relative date and time expressions are also written with keywords "*lọdún*" (in the year), "*loṣù*" (in the month), "*lọsẹ̀*" (in the week), "*lọjọ́*" (in the day). An example of a date expression is:

"*8th of December, 2018*" in Yorùbá translates to "*ọjọ́ 8 oṣù Ọpẹ̀, ọdún 2018*"

**Lists of Entities** can be obtained from a variety of sources like gazetteers, dictionaries, phone books, census data and Wikipedia categories (Ratinov and Roth, 2009). In recent years, knowledge bases like Freebase and Wikidata have become another option to retrieve entity lists in a structured way. An entity list is created by extracting all names of that type from a knowledge source (e.g. all person names from Wikidata). If a word or token from the unlabeled text matches an entry in an entity list, it is assigned the corresponding label. Experts can add heuristics to this automatic labeling that improve the matching (Dembowski, Wiegand, and Klakow, 2017). These include e.g. normalizing the grammatical form of words or filtering common false positives.

Another popular method for low-resource NER is the use of cross-lingual information (Rahimi, Li, and Cohn, 2019). Alternatives to distant supervision are crowd-sourcing (Rehbein and Ruppenhofer, 2017) and non-expert annotations (Mayhew and Roth, 2018).

---

6 https://github.com/google-research/bert/blob/master/multilingual.md

6.2.4 *Learning With Noisy Labels*

The labels obtained through distant and weak supervision methods tend to contain a high amount of errors. In the Food101N dataset (Lee et al., 2018) around 20% of the automatically obtained labels are incorrect while for Clothing1M (Xiao et al., 2015) the noise rate is more than 60%. Learning with this additional, noisily labeled data can result in lower classification performance compared to just training on a small set of clean labels (cf. e.g. (Fang and Cohn, 2016)). A variety of techniques have been proposed to handle label noise like modelling the underlying noise process (Lange, Hedderich, and Klakow, 2019) and filtering noisy instances (Nguyen et al., 2020; Yang et al., 2018). (Frenay and Verleysen, 2014) gives an in-depth introduction into this field and (Algan and Ulusoy, 2019) survey more recent approaches, focusing on the vision domain.

In this work, we experiment with three noise handling techniques. The approach by Bekker and Goldberger (2016) estimates a noise channel using the EM algorithm. It treats all labels as possibly noisy and does not distinguish between a clean and a noisy part of the data. In contrast, the method by Hedderich and Klakow (2018) leverages the existence of a small set of gold standard labels, something that—in our experience—is often available even in low resource settings. Having such a small set of clean labels is beneficial both for the main model itself as well as for the noise handling technique. Both approaches model the relationship between clean and noisy labels using a confusion matrix. This allows adapting the noisy to the clean label distribution during training. For a setting with 5 labels, it only requires $5^2 = 25$ additional parameters to estimate which could be beneficial when only few training data is available. The technique by Veit et al. (2017) (adapted to NER by Hedderich and Klakow (2018)) learns a more complex neural network to clean the noisy labels before training with them. It also takes the features into account when cleaning the noise and it might, therefore, be able to model more complex noise processes. All three techniques can be easily added to the existing standard neural network architectures for NER.

## 6.3 MODELS & EXPERIMENTAL SETTINGS

**Hausa** Distant supervision on Hausa was performed using lists of person names extracted from Wikipedia data. Since we had limited access to the data, we tested a simplified binary NER-tagging setting (PERSON-tags only). As a base model, we used a Bi-LSTM model developed for Part-of-Speech tagging (Plank, Søgaard, and Goldberg, 2016). For noise handling, we apply the **Noise Channel** model by Bekker and Goldberger (2016).

**Yorùbá** For Yorùbá, the entity lists were created by extracting person, location and organization entities from Wikidata in English and Yorùbá. Additionally, a list of person names in Nigeria was obtained from a Yorùbá Name website[7] (8,365 names) and list of popular Hausa, Igbo, Fulani and Yorùbá people on Wikipedia (in total 9,241 names). As manual heuristic, a minimum name length of 2 was set for extraction of PER (except for Nigerian names), LOC and ORG. The Nigerian names were set to include names with a minimum length of 3. For the DATE label, a native Yorùbá speaker wrote some annotation rules using 11 "date keywords" ("*ojọ́*", "*ọsẹ̀*", "*osù*", "*ọdún*", "*wákàtí*", "*ḷodún*", "*ḷodún*-un", "*ọdún*-un" "*lọsẹ̀*", "*lọjọ́*", "*aago*") following these two criteria: (1) A token is a date keyword or follows a date keyword in a sequence. (2) A token is a digit. For Yorùbá, we evaluate four settings with different amounts of clean data, namely 1k, 2k, 4k and the full dataset. As distantly supervised data with noisy labels, the full dataset is used. Additionally, 19,559 words from 18 articles of the Global News Corpus (different from the articles in the training corpus) were automatically annotated.

The **Bi-LSTM** model consists of a Bi-LSTM layer followed by a linear layer to extract input features. The Bi-LSTM layer has a 300-dimensional hidden state for each direction. For the final classification, an additional linear layer is added to output predicted class distributions. For noise handling, we experiment with the **Confusion Matrix** model by Hedderich and Klakow (2018) and the **Cleaning** model by Veit et al. (2017). We repeat all the Bi-LSTM experiments 20 times and report the average F1-score (following the approach by Tjong Kim Sang and De Meulder (2003)) and the standard error.

The **BERT** model is obtained by *fine-tuning* the pre-trained BERT embeddings on NER data with an additional untrained CRF classifier. We fine-tuned all the parameters of BERT including that of the CRF end-to-end. This has been shown to give better performance than using word features extracted from BERT to train a classifier (Devlin et al., 2019). The evaluation result is obtained as an average of 5 runs, we report the F1-score and the standard error in the result section.

## 6.4 RESULTS

The results for Hausa are given in Table 6.1. Training with a mix of 50% clean and 50% distantly-supervised data performs 15 F1-score points below using the whole 100% clean data which is to be expected due to the lower quality of the distantly-supervised labels. Using the Noise Channel closes half of this gap. Due to the limited availability of the dataset, we could unfortunately not investigate this further, but it shows already the benefits that are possible through noise-handling.

---

7 http://www.yorubaname.com/

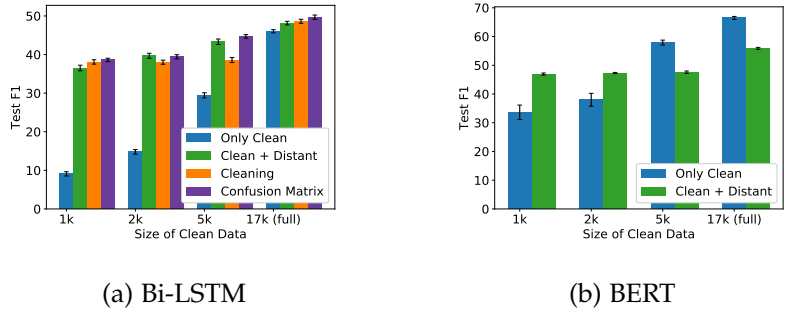

(a) Bi-LSTM          (b) BERT

Figure 6.1: F1-scores and standard error for Yorùbá.

| Data | Precision | Recall | F1 |
|---|---|---|---|
| 100% Clean | 79 | 45 | 57 |
| 50% Clean + 50% Distant | 59 | 38 | 42 |
| Noise Channel | 57 | 47 | 51 |

Table 6.1: Performance of the Bi-LSTM on Hausa (PER only)

An evaluation of the distant supervision for Yorùbá is given in Table 6.2. The quality of the automatically annotated labels differs between the classes. Locations perform better than person and organization names, probably due to locations being less diverse and better covered in Wikidata. With simple date rules, we obtain already a 48% F1-score. This shows the importance of leveraging the knowledge of native speakers in automatic annotations. Overall a decent annotation can be obtained by the distant supervision and it even outperforms some of the actual machine learning models in the low-resource setting. Table 6.3 compares using only Wikidata as data source versus adding additional, manually obtained lists of person names. While adding a list of Yorùbá names only improves recall slightly, the integration of Nigerian names helps to boost recall by 13 points.

The experimental results for Yorùbá are given in Figure 6.1. The setting differs from the experiments with Hausa in that there is a small clean training set and *additional*, distantly-supervised data. For the Bi-LSTM model, adding distantly-supervised labels always helps. In the low-resource settings with 1k and 2k labeled data, it more than doubles the performance. Handling the noise in the distant supervision can result in slight improvements. The noise-cleaning approach struggles somewhat while the confusion matrix architecture does give better results in the majority of the scenarios. Training on 5k labeled data with distantly supervised data and noise handling, one can obtain a performance close to using the full 17k manually labeled token.

The Bi-LSTM model has 1.50 million parameters (1.53 million for the cleaning model), while BERT has 110 million parameters. There is a clear trade-off between model size and performance. The BERT model is 70 times larger and obtains consistently better results due to its more complex, contextual embeddings pretrained on more data. Still, the F1-score also drops nearly half for the BERT model in the 1k setting compared to the full dataset. For 1k and 2k labeled data, the distant supervision helps to improve the model's performance. However, once the model trained only on clean data reaches a higher F1-score than the distant supervision technique, the model trained on clean and distantly-supervised data deteriorates. This suggests that the BERT model overfits too much on the noise in the distant supervision.

| Method | P | R | F1 |
|---|---|---|---|
| Overall | 41 | 41 | 41 |
| PER | 36 | 16 | 22 |
| LOC | 73 | 54 | 62 |
| ORG | 17 | 29 | 22 |
| DATE | 37 | 66 | 48 |

Table 6.2: Distant Supervision Quality (on test)

| Method | P | R | F1 |
|---|---|---|---|
| Wikidata | 50 | 25 | 33 |
| + Yorùbá Names | 47 | 25 | 33 |
| + Nigerian Names | 55 | 39 | 46 |

Table 6.3: Distant Supervision: Coverage of Wikidata for person names (on train)

## 6.5 CONCLUSION

In this study, we analysed distant supervision techniques and label-noise handling for NER in Hausa and Yorùbá, two languages from developing countries. We showed that they can be successfully leveraged in a realistic low-resource scenario to double a classifier's performance. If model size is not a constraint, the more complex BERT model clearly outperforms the smaller Bi-LSTM architecture. Nevertheless, there is still a large gap between the best performing model on Yorùbá with 66 F1-score and the state-of-the-art in English around 90.

We see several interesting follow-ups to these evaluations. In the future, we want to evaluate if noise handling methods can also allow the more complex BERT model to benefit from distant supervision. Regarding the model complexity, it would be interesting to experiment with more compact models like DistilBERT (Sanh et al., 2019) that reach a similar performance with a smaller model size for high-resource settings. We addressed the future works in Hedderich et al. (2020).

# 7

# MASAKHANER 1.0: INTRODUCING AFRICAN NER DATASET

This Chapter[1] describes the creation of the first large publicly available high-quality dataset for named entity recognition (NER) in ten African languages. This is an important step towards addressing the under-representation of the African continent in NLP research. We detail characteristics of the languages to help researchers understand the challenges that these languages pose for NER. We analyze our datasets and conduct an extensive empirical evaluation of state-of-the-art methods across both supervised and transfer learning settings. Transfer learning provides an alternative to distant supervision described in Chapter 6 by leveraging knowledge from an high-resource language (e.g. English) with large training data. We highlight some limitations of distant supervision in this section.

## 7.1 INTRODUCTION

Africa has over 2,000 spoken languages (Eberhard, Simons, and Fennig, 2021); however, these languages are scarcely represented in existing natural language processing (NLP) datasets, research, and tools (Martinus and Abbott, 2019). ∀ et al. (2020) investigate the reasons for these disparities by examining how NLP for low-resource languages is constrained by several societal factors. One of these factors is the geographical and language diversity of NLP researchers. For example, of the 2695 affiliations of authors whose works were published at the five major NLP conferences in 2019, only five were from African institutions (Caines, 2019). Conversely, many NLP tasks such as machine translation, text classification, part-of-speech tagging, and named entity recognition would benefit from the knowledge of native speakers who are involved in the development of datasets and models.

In this work, we focus on named entity recognition (NER)—one of the most impactful tasks in NLP (Lample et al., 2016; Sang and De Meulder, 2003). NER is an important information extraction task and an essential component of numerous products including spellcheckers, localization of voice and dialogue systems, and conversational agents. It also enables identifying African names, places and organizations for information retrieval. African languages are underrepresented in this crucial task due to lack of datasets, reproducible results, and researchers who understand the challenges that such languages present for NER.

---

1 This chapter is based on Adelani et al. (2021b)

In this paper, we take an initial step towards improving representation for African languages for the NER task, making the following contributions:

1. We bring together language speakers, dataset curators, NLP practitioners, and evaluation experts to address the challenges facing NER for African languages. Based on the availability of online news corpora and language annotators, we develop NER datasets, models, and evaluation covering ten widely spoken African languages.

2. We curate NER datasets from local sources to ensure relevance of future research for native speakers of the respective languages.

3. We train and evaluate multiple NER models for all ten languages. Our experiments provide insights into the transfer across languages, and highlight open challenges.

4. We release the datasets, code, and models to facilitate future research on the specific challenges raised by NER for African languages.

## 7.2 RELATED WORK

AFRICAN NER DATASETS   NER is a well-studied sequence labeling task (Yadav and Bethard, 2018) and has been the subject of many shared tasks in different languages (Benikova, Biemann, and Reznicek, 2014; Sangal, Sharma, and Singh, 2008; Shaalan, 2014; Tjong Kim Sang, 2002; Tjong Kim Sang and De Meulder, 2003). However, most of the available datasets are in high-resource languages. Although there have been efforts to create NER datasets for lower-resourced languages, such as the WikiAnn corpus (Pan et al., 2017) covering 282 languages, such datasets consist of "silver-standard" labels created by transferring annotations from English to other languages through cross-lingual links in knowledge bases. Because the WikiAnn corpus data comes from Wikipedia, it includes some African languages; though most have fewer than 10k tokens.

Other NER datasets for African languages include SADiLaR (Eiselen, 2016) for ten South African languages based on government data, and small corpora of fewer than 2K sentences for Yorùbá (Alabi et al., 2020) and Hausa (Hedderich et al., 2020). Additionally, the LORELEI language packs (Strassel and Tracey, 2016) include some African languages (Yorùbá, Hausa, Amharic, Somali, Twi, Kiswahili, Wolof, Kinyarwanda, and Zulu), but are not publicly available.

NER MODELS   Popular sequence labeling models for NER include the CRF (Lafferty, McCallum, and Pereira, 2001), CNN-BiLSTM (Chiu and Nichols, 2016), BiLSTM-CRF (Huang, Xu, and Yu, 2015), and CNN-BiLSTM-CRF (Ma and Hovy, 2016). The traditional CRF makes use of hand-crafted features like part-of-speech tags, context words and word capitalization. Neural NER models on the other hand are initialized with word embeddings like Word2Vec (Mikolov et al., 2013b), GloVe (Pennington, Socher, and Manning, 2014) and FastText (Bojanowski et al., 2017). More recently, pre-trained language models such as BERT (Devlin et al., 2019), RoBERTa (Liu et al., 2019), and LUKE (Yamada et al., 2020) have been applied to produce state-of-the-art results for the NER task. Multilingual variants of these models like mBERT and XLM-RoBERTa (Conneau et al., 2020) make it possible to train NER models for several languages using transfer learning. Language-specific parameters and adaptation to unlabeled data of the target language have yielded further gains (Pfeiffer et al., 2020b, 2021).

## 7.3 FOCUS LANGUAGES

| Language | Family | Speakers | Region |
| --- | --- | --- | --- |
| Amharic | Afro-Asiatic-Ethio-Semitic | 33M | East |
| Hausa | Afro-Asiatic-Chadic | 63M | West |
| Igbo | Niger-Congo-Volta-Niger | 27M | West |
| Kinyarwanda | Niger-Congo-Bantu | 12M | East |
| Luganda | Niger-Congo-Bantu | 7M | East |
| Dholuo | Nilo Saharan | 4M | East |
| Nigerian-Pidgin | English Creole | 75M | West |
| Kiswahili | Niger-Congo-Bantu | 98M | Central & East |
| Wolof | Niger-Congo-Senegambia | 5M | West & North West |
| Yorùbá | Niger-Congo-Volta-Niger | 42M | West |

Table 7.1: Language, family, number of speakers (Eberhard, Simons, and Fennig, 2021), and regions in Africa.

Table 7.1 provides an overview of the languages considered in this work, their language family, number of speakers and the regions in Africa where they are spoken. We chose to focus on these languages due to the availability of online news corpora, annotators, and most importantly because they are widely spoken native African languages. Both region and language family might indicate a notion of proximity for NER, either because of linguistic features shared within that family, or because data sources cover a common set of locally relevant entities. We highlight language specifics for each language to illustrate the diversity of this selection of languages in Section 7.3.1, and then showcase the differences in named entities across these languages in Section 7.3.2.

| Language | Sentence |
|---|---|
| English | The Emir of [Kano] turbaned [Zhang] who has spent [18 years] in [Nigeria] |
| Amharic | የካኖ ኢምር በናይጀርያ ፳፰ ዓመት ያሳለፈውን ዛንግን ዋና መሪ አደረጉት |
| Hausa | Sarkin [Kano] yayi wa [Zhang] wanda yayi [shekara 18] a [Najeriya] sarauta |
| Igbo | Onye Emir nke [Kano] kpubere [Zhang] okpu onye nke nọgoro [afọiri na asatọ] na [Naijiria] |
| Kinyarwanda | Emir w'i [Kano] yimitse [Zhang] wari umaze [imyaka 18] muri [Nijeriya] |
| Luganda | Emir w'e [Kano] yatikkidde [Zhang] amaze [emyaka 18] mu [Nigeria] |
| Dholuo | Emir mar [Kano] ne orwakone turban [Zhang] ma osedak [Nigeria] kwuom [higni 18] |
| Nigerian-Pidgin | Emir of [Kano] turban [Zhang] wey don spend [18 years] for [Nigeria] |
| Kiswahili | Emir wa [Kano] alimvisha kilemba [Zhang] ambaye alikaa [miaka 18] nchini [Nigeria] |
| Wolof | Emiiru [Kanó] dafa kaala kii di [Zhang] mii def [Nigeria] [fukki at ak juróom ñett] |
| Yorùbá | Ẹmíà ìlú [Kánò] wé láwàní lé orí [Zhang] ẹni tí ó ti lo [ọdún méjìdínlógún] ní orílè-èdè [Nàìjíríà] |

Table 7.2: Example of named entities in different languages. [PER], [LOC], and [DATE] are in colours purple, orange, and green respectively.

### 7.3.1 Language Characteristics

AMHARIC   (amh) uses the Fidel script consisting of 33 basic scripts, each of them with at least 7 vowel sequences. This results in more than 231 characters or Fidels. Numbers and punctuation marks are also represented uniquely with specific Fidels.

HAUSA   (hau) has 23-25 consonants, depending on the dialect and five short and five long vowels. Hausa has labialized phonemic consonants, as in /gw/ e.g. "agwagwa". As found in some African languages, implosive consonants also exist in Hausa, e.g. 'b, 'd, etc as in 'barna. Similarly, the Hausa approximant "r" is realized in two distinct manners: roll and trill, as in "rai" and "ra'ayi", respectively.

IGBO   (ibo) is an agglutinative language, with many frequent suffixes and prefixes (Emenanjo, 1978). A single stem can yield many word-forms by addition of affixes that extend its original meaning (Onyenwe and Hepple, 2016). Igbo is also tonal, with two distinctive tones (high and low) and a down-stepped high tone in some cases. The alphabet consists of 28 consonants and 8 vowels (A, E, I, Ị, O, Ọ, U, Ụ). In addition to the Latin letters (except *c*), Igbo contains the following digraphs: (ch, gb, gh, gw, kp, kw, nw, ny, sh).

KINYARWANDA   (kin) makes use of 24 Latin characters with 5 vowels similar to English and 19 consonants excluding q and x. Moreover, Kinyarwanda has 74 additional complex consonants (such as mb, mpw, and njyw). (Government, 2014) It is a tonal language with three tones: low (no diacritic), high (signaled by "/") and falling (signaled by "^"). The default word order is Subject-Verb-Object.

LUGANDA    (`lug`) is a tonal language with subject-verb-object word order. The Luganda alphabet is composed of 24 letters that include 17 consonants (p, v, f, m, d, t, l, r, n, z, s, j, c, g), 5 vowel sounds represented in the five alphabetical symbols (a, e, i, o, u) and 2 semi-vowels (w, y). It also has a special consonant *ng'*.

DHOLUO    (`luo`) is a tonal language with 4 tones (high, low, falling, rising) although the tonality is not marked in orthography. It has 26 Latin consonants without Latin letters (c, q, v, x and z) and additional consonants (ch, dh, mb, nd, ng', ng, ny, nj, th, sh). There are nine vowels (a, e, i, o, u, ɐ, ɛ, ɔ, ʊ) which are distinguished primarily by advanced tongue root (ATR) harmony (De Pauw, Wagacha, and Abade, 2007).

NIGERIAN-PIDGIN    (`pcm`) is a largely oral, national lingua franca with a distinct phonology from English, its lexifier language. Portuguese, French, and especially indigenous languages form the substrate of lexical, phonological, syntactic, and semantic influence on Nigerian-Pidgin (NP). English lexical items absorbed by NP are often phonologically closer to indigenous Nigerian languages, notably in the realization of vowels. As a rapidly evolving language, the NP orthography is undergoing codification and indigenization (Offiong Mensah, 2012; Ojarikre, 2013; Onovbiona, 2012).

KISWAHILI    (`swa`) is the most widely spoken language on the African continent. It has 30 letters including 24 Latin letters without characters (q and x) and six additional consonants (ch, dh, gh, ng', sh, th) unique to Kiswahili pronunciation.

WOLOF    (`wol`) has an alphabet similar to that of French. It consists of 29 characters, including all letters of the French alphabet except H, V and Z. It also includes the characters ŋ ("ng") and Ñ ("gn" as in Spanish). Accents are present, but limited in number (À, É, Ë, Ó). However, unlike many other Niger-Congo languages, Wolof is not a tonal language.

YORÙBÁ    (`yor`) has 25 Latin letters without the Latin characters (c, q, v, x and z) and with additional letters (ẹ, gb, ṣ , ọ). Yorùbá is a tonal language with three tones: low ("\"), middle ("−", optional) and high ("/"). The tonal marks and underdots are referred to as diacritics and they are needed for the correct pronunciation of a word. Yorùbá is a highly isolating language and the sentence structure follows Subject-Verb-Object.

7.3.2 *Named Entities*

Most of the work on NER is centered around English, and it is unclear how well existing models can generalize to other languages in terms of sentence structure or surface forms. In Hu et al. (2020)'s evaluation on cross-lingual generalization for NER, only two African languages were considered and it was seen that transformer-based models particularly struggled to generalize to named entities in Kiswahili. To highlight the differences across our focus languages, Table 7.2 shows an English[2] example sentence, with color-coded PER, LOC, and DATE entities, and the corresponding translations. The following characteristics of the languages in our dataset could pose challenges for NER systems developed for English:

- Amharic shares no lexical overlap with the English source sentence.
- While "Zhang" is identical across all Latin-script languages, "Kano" features accents in Wolof and Yorùbá due to its localization.
- The Fidel script has no capitalization, which could hinder transfer from other languages.
- Igbo, Wolof, and Yorùbá all use diacritics, which are not present in the English alphabet.
- The surface form of named entities (NE) is the same in English and Nigerian-Pidgin, but there exist lexical differences (e.g. in terms of how time is realized).
- Between the 10 African languages, "Nigeria" is spelled in 6 different ways.
- Numerical "18": Igbo, Wolof and Yorùbá write out their numbers, resulting in different numbers of tokens for the entity span.

7.4 DATA AND ANNOTATION METHODOLOGY

Our data was obtained from local news sources, in order to ensure relevance of the dataset for native speakers from those regions. The dataset was annotated using the ELISA tool (Lin et al., 2018) by native speakers who come from the same regions as the news sources and volunteered through the *Masakhane* community[3]. Annotators were not paid but are all part of the authors of this paper. The annotators were trained on how to perform NER annotation using the MUC-6 annotation guide[4]. We annotated four entity types: Personal name (PER), Location (LOC), Organization (ORG), and date & time (DATE). The annotated entities were inspired by the English CoNLL-2003 Corpus (Tjong Kim Sang, 2002). We replaced the MISC tag with the DATE tag following Alabi et al. (2020) as the MISC tag may be ill-defined and cause disagreement among non-expert annotators. We report the number of annotators as well as general statistics of the datasets in Table 7.3. For each language, we divided the annotated data into training, development, and test splits consisting of 70% training, 10%, and 20% of the data respectively.

---

[2] Although the original sentence is from BBC Pidgin https://www.bbc.com/pidgin/tori-51702073
[3] https://www.masakhane.io
[4] https://cs.nyu.edu/~grishman/muc6.html

| Language | Data Source | Train/ dev/ test | # Anno. | PER | ORG | LOC | DATE | % of Entities in Tokens | # Tokens |
|---|---|---|---|---|---|---|---|---|---|
| Amharic | DW & BBC | 1750/ 250/ 500 | 4 | 730 | 403 | 1,420 | 580 | 15.13 | 37,032 |
| Hausa | VOA Hausa | 1903/ 272/ 545 | 3 | 1,490 | 766 | 2,779 | 922 | 12.17 | 80,152 |
| Igbo | BBC Igbo | 2233/ 319/ 638 | 6 | 1,603 | 1,292 | 1,677 | 690 | 13.15 | 61,668 |
| Kinyarwanda | IGIHE news | 2110/ 301/ 604 | 2 | 1,366 | 1,038 | 2096 | 792 | 12.85 | 68,819 |
| Luganda | BUKEDDE news | 2003/ 200/ 401 | 3 | 1,868 | 838 | 943 | 574 | 14.81 | 46,615 |
| Dholuo | Ramogi FM news | 644/ 92/ 185 | 2 | 557 | 286 | 666 | 343 | 14.95 | 26,303 |
| Nigerian-Pidgin | BBC Pidgin | 2100/ 300/ 600 | 5 | 2,602 | 1,042 | 1,317 | 1,242 | 13.25 | 76,063 |
| Kiswahili | VOA Swahili | 2104/ 300/ 602 | 6 | 1,702 | 960 | 2,842 | 940 | 12.48 | 79,272 |
| Wolof | Lu Defu Waxu & Saabal | 1,871/ 267/ 536 | 2 | 731 | 245 | 836 | 206 | 6.02 | 52,872 |
| Yorùbá | GV & VON news | 2124/ 303/ 608 | 5 | 1,039 | 835 | 1,627 | 853 | 11.57 | 83,285 |

Table 7.3: Statistics of our datasets including their source, number of sentences in each split, number of annotators, number of entities of each label type, percentage of tokens that are named entities, and total number of tokens.

A key objective of our annotation procedure was to create high-quality datasets by ensuring a high annotator agreement. To achieve high agreement scores, we ran collaborative workshops for each language, which allowed annotators to discuss any disagreements. ELISA provides an entity-level F1-score and also an interface for annotators to correct their mistakes, making it easy to achieve inter-annotator agreement scores between 0.96 and 1.0 for all languages.

We report inter-annotator agreement scores in Table 7.4 using Fleiss' Kappa (Fleiss, 1971) at both the token and entity level. The latter considers each span an annotator proposed as an entity. As a result of our workshops, all our datasets have exceptionally high inter-annotator agreement. For Kinyarwanda, Dholuo, Kiswahili, and Wolof, we report perfect inter-annotator agreement scores ($\kappa = 1$). For each of these languages, two annotators annotated each token and were instructed to discuss and resolve conflicts among themselves. To shed more light on the few cases where annotators disagreed, we provide entity-level confusion matrices across all ten languages in Table 7.5. The most common disagreement is between organizations and locations.

| Dataset | Token Fleiss' κ | Entity Fleiss' κ | Disagreement from Type |
|---|---|---|---|
| **amh** | 0.987 | 0.959 | 0.044 |
| **hau** | 0.988 | 0.962 | 0.097 |
| **ibo** | 0.995 | 0.983 | 0.071 |
| **kin** | 1.000 | 1.000 | 0.000 |
| **lug** | 0.997 | 0.990 | 0.023 |
| **luo** | 1.000 | 1.000 | 0.000 |
| **pcm** | 0.989 | 0.966 | 0.048 |
| **swa** | 1.000 | 1.000 | 0.000 |
| **wol** | 1.000 | 1.000 | 0.000 |
| **yor** | 0.990 | 0.964 | 0.079 |

Table 7.4: Inter-annotator agreement for our datasets calculated using Fleiss' kappa ($\kappa$) at the token and entity level. Disagreement from type refers to the proportion of all entity-level disagreements, which are due only to type mismatch.

|  | DATE | LOC | ORG | PER |
|---|---|---|---|---|
| **DATE** | 32,978 | - | - | - |
| **LOC** | 10 | 70,610 | - | - |
| **ORG** | 0 | 52 | 35,336 | - |
| **PER** | 2 | 48 | 12 | 64,216 |

Table 7.5: Entity-level confusion matrix between annotators, calculated over all ten languages.

## 7.5 EXPERIMENTAL SETUP

### 7.5.1 *NER baseline models*

To evaluate baseline performance on our dataset, we experiment with three popular NER models: CNN-BiLSTM-CRF, multilingual BERT (mBERT), and XLM-RoBERTa (XLM-R). The latter two models are implemented using the HuggingFace transformers toolkit (Wolf et al., 2019). For each language, we train the models on the in-language training data and evaluate on its test data.

CNN-BILSTM-CRF  This architecture was proposed for NER by Ma and Hovy (2016). For each input sequence, we first compute the vector representation for each word by concatenating character-level encodings from a CNN and vector embeddings for each word. Following Rijhwani et al. (2020), we use randomly initialized word embeddings since we do not have high-quality pre-trained embeddings for all the languages in our dataset. Our model is implemented using the DyNet toolkit (Neubig et al., 2017).

MBERT    We fine-tune multilingual BERT (Devlin et al., 2019) on our NER corpus by adding a linear classification layer to the pre-trained transformer model, and train it end-to-end. mBERT was trained on 104 languages including only two African languages: Kiswahili and Yorùbá. We use the mBERT-base cased model with 12-layer Transformer blocks consisting of 768-hidden size and 110M parameters.

XLM-R    XLM-R (Conneau et al., 2020) was trained on 100 languages including Amharic, Hausa, and Kiswahili. The major differences between XLM-R and mBERT are (1) XLM-R was trained on Common Crawl while mBERT was trained on Wikipedia; (2) XLM-R is based on RoBERTa, which is trained with a masked language model (MLM) objective while mBERT was additionally trained with a next sentence prediction objective. We make use of the XLM-R base and large models for the baseline models. The XLM-R-base model consisting of 12 layers, with a hidden size of 768 and 270M parameters. On the other hand, the XLM-R-large has 24 layers, with a hidden size of 1024 and 550M parameters.

MEANE-BILSTM    This is a simple BiLSTM model with an additional linear classifier. For each input sequence, we first extract a sentence embedding from mBERT or XLM-R language model (LM) before passing it into the BiLSTM model. Following Reimers and Gurevych (2019), we make use of the mean of the 12-layer output embeddings of the LM (i.e *MeanE*). This has been shown to provide better sentence representations than the embedding of the `[CLS]` token used for fine-tuning mBERT and XLM-R.

LANGUAGE BERT    The mBERT and the XLM-R models only supports two and three languages under study respectively. One effective approach to adapt the pre-trained transformer models to new domains is "domain-adaptive fine-tuning" (Gururangan et al., 2020; Howard and Ruder, 2018)—fine-tuning on unlabeled data in the new domain, which also works very well when adapting to a new language (Alabi et al., 2020; Pfeiffer et al., 2020a). For each of the African languages, we performed *language-adaptive fine-tuning (LAFT)* on available unlabeled corpora mostly from JW300 (Agić and Vulić, 2019), indigenous news sources and XLM-R Common Crawl corpora (Conneau et al., 2020). This approach is quite useful for languages whose scripts are not supported by the multi-lingual transformer models like Amharic where we replace the vocabulary of mBERT by an Amharic vocabulary

| Language | Source | Size (MB) | No. sentences |
|---|---|---|---|
| amh | CC-100 (Conneau et al., 2020) | 889.7MB | 3,124,760 |
| hau | CC-100 | 318.4MB | 3,182,277 |
| ibo | JW300 (Agić and Vulić, 2019), CC-100, CC-Aligned (El-Kishky et al., 2020), and IgboNLP (Ezeani et al., 2020) | 118.3MB | 1,068,263 |
| kin | JW300, KIRNEWS (Niyongabo et al., 2020), and BBC Gahuza | 123.4MB | 726,801 |
| lug | JW300, CC-100, and BUKEDDE News | 54.0MB | 506,523 |
| luo | JW300 | 12.8MB | 160,904 |
| pcm | JW300, and BBC Pidgin | 56.9MB | 207,532 |
| swa | CC-100 | 1,800MB | 12,664,787 |
| wol | OPUS (Tiedemann, 2012) (excl. CC-Aligned), Wolof Bible (MBS, 2020), and news corpora (Lu Defu Waxu, Saabal, and Wolof Online) | 3.8MB | 42,621 |
| yor | JW300, Yoruba Embedding Corpus (Alabi et al., 2020), MENYO-20k (Adelani et al., 2021a), CC-100, CC-Aligned, and news corpora (BBC Yoruba, Asejere, and Alaroye). | 117.6MB | 910,628 |

Table 7.6: Monolingual Corpora, their sources, size, and number of sentences

before we perform language-adaptive fine-tuning, similar to Alabi et al. (2020).

Table 7.6 shows the monolingual corpus we used for the language adaptive fine-tuning. We provide the details of the source of the data, and their sizes. For most of the languages, we make use of JW300[5] and CC-100[6]. In some cases CC-Aligned (El-Kishky et al., 2020) was used, in such a case, we removed duplicated sentences from CC-100. For fine-tuning the language model, we make use of the HuggingFace (Wolf et al., 2019) code with learning rate 5e-5. However, for the Amharic BERT, we make use of a smaller learning rate of 5e-6 since the multilingual BERT vocabulary was replaced by Amharic vocabulary, so that we can slowly adapt the mBERT LM to understand Amharic texts. All language BERT models were pre-trained for 3 epochs ("ibo", "kin","lug","luo", "pcm","swa","yor") or 10 epochs ("amh", "hau","wol") depending on their convergence. The models can be found on HuggingFace Model Hub[7].

### 7.5.2 Improving the Baseline Models

In this section, we consider techniques to improve the baseline models such as utilizing gazetteers, transfer learning from other domains and languages, and aggregating NER datasets by regions. For these experiments, we focus on the PER, ORG, and LOC categories, because the gazetteers from Wikipedia do not contain DATE entities and some source domains and languages that we transfer from do not have the DATE annotation. We apply these modifications to the XLM-R model because it generally outperforms mBERT in our experiments (see Section 7.6).

---

[5] https://opus.nlpl.eu/
[6] http://data.statmt.org/cc-100/
[7] https://huggingface.co/Davlan

#### 7.5.2.1 *Gazetteers for NER*

Gazetteers are lists of named entities collected from manually crafted resources such as GeoNames or Wikipedia. Before the widespread adoption of neural networks, NER methods used gazetteers-based features to improve performance (Ratinov and Roth, 2009). These features are created for each *n*-gram in the dataset and are typically binary-valued, indicating whether that *n*-gram is present in the gazetteer.

Recently, Rijhwani et al. (2020) showed that augmenting the neural CNN-BiLSTM-CRF model with gazetteer features can improve NER performance for low-resource languages. We conduct similar experiments on the languages in our dataset, using entity lists from Wikipedia as gazetteers. For Dholuo and Nigerian-Pidgin, which do not have their own Wikipedia, we use entity lists from English Wikipedia.

### 7.5.3 *Distant Supervision for NER*

Here, we consider distant supervision introduced in Chapter 6 to improve the NER performance in the absence of labelled examples or few available examples, similar to Gazetteers in Section 7.5.2.1 but in low-resource scenario i.e. we assume availability of small labelled data. Specifically, we make use of list of entities from Wikidata, Geo-names, city names, Crunchbase companies, and other publicly available lists in English and African languages (Lison, Barnes, and Hubin, 2021; Saleva and Lignos, 2021). For African languages, list of entities are few, therefore, we only evaluate on four African languages with large Wikidata: Swahili, Hausa, Yorùbá and Igbo. We focus on personal name (PER), location (LOC), and organization entity types (ORG). We make use of 10k sentences from BBC (Igbo and Yorùbá) and VOA (Hausa and Swahili) articles for distant supervision. For generating automatic annotations, we make use of ANEA (Hedderich, Lange, and Klakow, 2021)—a tool to automatically annotate named entities in texts based on entity lists.

For the experiments, we compare the performance of training 100 sentences with XLM-R+LAFT (XLM-R-base adapted through LAFT, the best model) to *Confusion Matrix with smoothing* (Hedderich et al., 2020)—a noisy handling learning method. As an additional baseline, we evaluate on BOND (Liang et al., 2020)—an approach to improve distantly supervised labels with early stopping and self-training.

#### 7.5.3.1 Transfer Learning

Here, we focus on cross-domain transfer from Wikipedia to the news domain, and cross-lingual transfer from English and Kiswahili NER datasets to the other languages in our dataset.

DOMAIN ADAPTATION FROM WIKIANN  We make use of the WikiAnn corpus (Pan et al., 2017), which is available for five of the languages in our dataset: Amharic, Igbo, Kinyarwanda, Kiswahili and Yorùbá. For each language, the corpus contains 100 sentences in each of the training, development and test splits except for Kiswahili, which contains 1K sentences in each split. For each language, we train on the corresponding WikiAnn training set and either zero-shot transfer to our respective test set or additionally fine-tune on our training data.

CROSS-LINGUAL TRANSFER  For training the cross-lingual transfer models, we use the CoNLL-2003[8] NER dataset in English with over 14K training sentences and our annotated corpus. The reason for CoNLL-2003 is because it is in the same news domain as our annotated corpus. We also make use of the languages that are supported by the XLM-R model and are widely spoken in East and West Africa like Kiswahili and Hausa. The English corpus has been shown to transfer very well to low resource languages (Hedderich et al., 2020; Lauscher et al., 2020). We first train on either the English CoNLL-2003 data or our training data in Kiswahili, Hausa, or Nigerian-Pidgin before testing on the target African languages.

### 7.5.4 Aggregating Languages by Regions

As previously illustrated in Table 7.2, several entities have the same form in different languages while some entities may be more common in the region where the language is spoken. To study the performance of NER models across geographical areas, we combine languages based on the region of Africa that they are spoken in (see Table 7.1): (1) East region with Kinyarwanda, Luganda, Dholuo, and Kiswahili; (2) West Region with Hausa, Igbo, Nigerian-Pidgin, Wolof, and Yorùbá languages, (3) East and West regions—all languages except Amharic because of its distinct writing system.

### 7.5.5 Model Hyper-parameters for Reproducibility

For fine-tuning mBERT and XLM-R, we used the base and large models with maximum sequence length of 164 for mBERT and 200

---

8 We also tried OntoNotes 5.0 by combining FAC & ORG as "ORG" and GPE & LOC as "LOC" and others as "O" except "PER", but it gave lower performance in zero-shot transfer (19.38 F1) while CoNLL-2003 gave 37.15 F1.

| Lang. | In mBERT? | In XLM-R? | % OOV in Test Entities | CNN-BiLSTM CRF | mBERT-base MeanE / FTune | XLM-R-base MeanE / FTune | XLM-R Large FTune | lang. BERT FTune | lang. XLM-R FTune |
|---|---|---|---|---|---|---|---|---|---|
| amh | ✗ | ✓ | 72.94 | 52.08 | 0.0 / 0.0 | 63.57 / 70.62 | 76.18 | 60.89 | **77.97** |
| hau | ✗ | ✓ | 33.40 | 83.52 | 81.49 / 86.65 | 86.06 / 89.50 | 90.54 | 91.31 | **91.47** |
| ibo | ✗ | ✗ | 46.56 | 80.02 | 76.17 / 85.19 | 73.47 / 84.78 | 84.12 | 86.75 | **87.74** |
| kin | ✗ | ✗ | 57.85 | 62.97 | 65.85 / 72.20 | 63.66 / 73.32 | 73.75 | 77.57 | **77.76** |
| lug | ✗ | ✗ | 61.12 | 74.67 | 70.38 / 80.36 | 68.15 / 79.69 | 81.57 | 83.44 | **84.70** |
| luo | ✗ | ✗ | 65.18 | 65.98 | 56.56 / 74.22 | 52.57 / 74.86 | 73.58 | **75.59** | 75.27 |
| pcm | ✗ | ✗ | 61.26 | 67.67 | 81.87 / 87.23 | 81.93 / 87.26 | 89.02 | 89.95 | **90.00** |
| swa | ✓ | ✓ | 40.97 | 78.24 | 83.08 / 86.80 | 84.33 / 87.37 | 89.36 | 89.36 | **89.46** |
| wol | ✗ | ✗ | 69.73 | 59.70 | 57.21 / 64.52 | 54.97 / 63.86 | 67.90 | **69.43** | 68.31 |
| yor | ✓ | ✗ | 65.99 | 67.44 | 74.28 / 78.97 | 67.45 / 78.26 | 78.89 | 82.58 | **83.66** |
| avg | – | – | 57.50 | 69.23 | 64.69 / 71.61 | 69.62 / 78.96 | 80.49 | 80.69 | **82.63** |
| avg (excl. amh) | – | – | 55.78 | 71.13 | 71.87 / 79.88 | 70.29 / 79.88 | 80.97 | 82.89 | **83.15** |

Table 7.7: NER model comparison, showing F1-score on the test sets after 50 epochs averaged over 5 runs. This result is for all 4 tags in the dataset: PER, ORG, LOC, DATE. **Bold** marks the top score (tied if within the range of SE). mBERT and XLM-R are trained in two ways (1) MeanE: mean output embeddings of the 12 LM layers are used to initialize BiLSTM + Linear classifier, and (2) FTune: LM fine-tuned end-to-end with a linear classifier. Lang. BERT & Lang XLM-R (base) are models fine-tuned after language adaptive fine-tuning.

| Method | amh | hau | ibo | kin | lug | luo | pcm | swa | wol | yor | avg |
|---|---|---|---|---|---|---|---|---|---|---|---|
| CNN-BiLSTM-CRF | **50.31** | 84.64 | **81.25** | 60.32 | **75.66** | **68.93** | 62.60 | 77.83 | 61.84 | **66.48** | 68.99 |
| + Gazetteers | 49.51 | **85.02** | 80.40 | **64.54** | 73.85 | 65.44 | **66.54** | **80.16** | **62.44** | 65.49 | **69.34** |

Table 7.8: Improving NER models using Gazetteers. The result is only for 3 Tags: PER, ORG & LOC. Models trained for 50 epochs. Result is an average over 5 runs.

for XLM-R, batch size of 32, learning rate of 5e-5, and number of epochs 50. For the MeanE-BiLSTM model, the hyper-parameters are similar to fine-tuning the LM except for the learning rate that we set to be 5e-4, the BiLSTM hyper-parameters are: input dimension is 768 (since the embedding size from mBERT and XLM-R is 768) in each direction of LSTM, one hidden layer, hidden layer size of 64, and drop-out probability of 0.3 before the last linear layer. For the distant supervision experiments, we make use of hyper-parameters from Hedderich et al. (2020) and (Liang et al., 2020). All the experiments were performed on a single GPU (Nvidia V100).

## 7.6 RESULTS

### 7.6.1 Baseline Models

Table 7.7 gives the F1-score obtained by CNN-BiLSTM-CRF, mBERT and XLM-R models on the test sets of the ten African languages when training on our in-language data. We additionally indicate whether the language is supported by the pre-trained language models (✓). The percentage of entities that are of out-of-vocabulary (OOV; entities in the test set that are not present in the training set) is also reported alongside results of the baseline models. In general, the datasets with greater numbers of OOV entities have lower performance with the CNN-BiLSTM-CRF model, while those with lower OOV rates (Hausa, Igbo, Kiswahili) have higher performance. We find that the CNN-BiLSTM-CRF model performs worse than fine-tuning mBERT and XLM-R models end-to-end (FTune). We expect performance to be better (e.g., for Amharic and Nigerian-Pidgin with over 18 F1 point difference) when using pre-trained word embeddings for the initialization of the BiLSTM model rather than random initialization (we leave this for future work as discussed in Section 7.7).

Interestingly, the pre-trained language models (PLMs) have reasonable performance even on languages they were not trained on such as Igbo, Kinyarwanda, Luganda, Dholuo, and Wolof. However, languages supported by the PLM tend to have better performance overall. We observe that fine-tuned XLM-R-base models have significantly better performance on five languages; two of the languages (Amharic and Kiswahili) are supported by the pre-trained XLM-R. Similarly, fine-tuning mBERT has better performance for Yorùbá since the language is part of the PLM's training corpus. Although mBERT is trained on Kiswahili, XLM-R-base shows better performance. This observation is consistent with Hu et al. (2020) and could be because XLM-R is trained on more Kiswahili text (Common Crawl with 275M tokens) whereas mBERT is trained on a smaller corpus from Wikipedia (6M tokens[9]).

Another observation is that mBERT tends to have better performance for the non-Bantu Niger-Congo languages i.e., Igbo, Wolof, and Yorùbá, while XLM-R-base works better for Afro-Asiatic languages (i.e., Amharic and Hausa), Nilo-Saharan (i.e., Dholuo) and Bantu languages like Kinyarwanda and Kiswahili. We also note that the writing script is one of the primary factors influencing the transfer of knowledge in PLMs with regard to the languages they were not trained on. For example, mBERT achieves an F1-score of 0.0 on Amharic because it has not encountered the script during pre-training. In general, we find the fine-tuned XLM-R-large (with 550M parameters) to be better than XLM-R-base (with 270M parameters) and mBERT (with 110 parameters) in almost all languages. However, mBERT models perform slightly better for Igbo, Dholuo, and Yorùbá despite having fewer parameters.

We further analyze the transfer abilities of mBERT and XLM-R by extracting sentence embeddings from the LMs to train a BiLSTM model (*MeanE-BiLSTM*) instead of fine-tuning them end-to-end. Table 7.7 shows that languages that are not supported by mBERT or XLM-R generally perform worse than CNN-BiLSTM-CRF model (despite being randomly initialized) except for *kin*. Also, sentence embeddings extracted from mBERT often lead to better performance than XLM-R for languages they both do not support (like *ibo*, *kin*, *lug*, *luo*, and *wol*).

---

[9] https://github.com/mayhewsw/multilingual-data-stats

| Method | amh | hau | ibo | kin | lug | luo | pcm | swa | wol | yor | avg |
|---|---|---|---|---|---|---|---|---|---|---|---|
| XLM-R-base | 69.71 | 91.03 | 86.16 | 73.76 | 80.51 | 75.81 | 86.87 | 88.65 | 69.56 | 78.05 | 77.30 |
| WikiAnn zero-shot | 27.68 | – | 21.90 | 9.56 | – | – | – | 36.91 | – | 10.42 | – |
| eng-CoNLL zero-shot | – | 67.52 | 47.71 | 38.17 | 39.45 | 34.19 | 67.27 | 76.40 | 24.33 | 39.04 | 37.15 |
| pcm zero-shot | – | 63.71 | 42.69 | 40.99 | 43.50 | 33.12 | – | 72.84 | 25.37 | 35.16 | 36.81 |
| swa zero-shot | – | 85.35* | 55.37 | 58.44 | 57.65* | 42.88* | 72.87* | – | 41.70 | 57.87* | 52.32 |
| hau zero-shot | – | – | 58.41* | 59.10* | 59.78 | 42.81 | 70.74 | 83.19* | 42.81* | 55.97 | 53.14* |
| WikiAnn + finetune | 70.92 | – | 85.24 | 72.84 | – | – | – | 87.90 | – | 76.78 | – |
| eng-CoNLL + finetune | – | 89.73 | 85.10 | 71.55 | 77.34 | 73.92 | 84.05 | 87.59 | 68.11 | 75.77 | 75.30 |
| pcm + finetune | – | 90.78 | 86.42 | 71.69 | 79.72 | 75.56 | – | 87.62 | 67.21 | 78.29 | 76.48 |
| swa + finetune | – | 91.50 | 87.11 | 74.84 | 80.21 | 74.49 | 86.74 | – | 68.47 | 80.68 | 77.63 |
| hau + finetune | – | – | 86.84 | 74.22 | 80.56 | 75.55 | 88.03 | 87.92 | 70.20 | 79.44 | 77.80 |
| combined East Langs. | – | – | – | 75.65 | 81.10 | 77.56 | – | 88.15 | – | – | – |
| combined West Langs. | – | 90.88 | 87.06 | – | – | – | 87.21 | – | 69.70 | 80.68 | – |
| combined 9 Langs. | – | **91.64** | **87.94** | 75.46 | **81.29** | **78.12** | **88.12** | 88.10 | 69.84 | 80.59 | 78.87 |

Table 7.9: Transfer Learning Result (i.e. F1-score). 3 Tags: PER, ORG & LOC. WikiAnn, eng-CoNLL, and the annotated datasets are trained for 50 epochs. Fine-tuning is only for 10 epochs. Results are averaged over 5 runs and the total average (avg) is computed over ibo, kin, lug, luo, wol, and yor languages. The overall highest F1-score is in **bold**, and the best F1-score in zero-shot settings is indicated with an asterisk (*).

Lastly, we train NER models using *language BERT* models that have been adapted to each of the African languages via language-specific fine-tuning on unlabeled text. In all cases, fine-tuning language BERT and language XLM-R models achieves a $1-7\%$ improvement in F1-score over fine-tuning mBERT-base and XLM-R-base respectively. This approach is still effective for small sized pre-training corpora provided they are of good quality. For example, the Wolof monolingual corpus, which contains less than 50K sentences (see Table 7.6 in the Appendix) still improves performance by over 4% F1. Further, we obtain over 60% improvement in performance for Amharic BERT because mBERT does not recognize the Amharic script.

### 7.6.2 *Evaluation of Gazetteer Features*

Table 7.8 shows the performance of the CNN-BiLSTM-CRF model with the addition of gazetteer features as described in Section 7.5.2.1. On average, the model that uses gazetteer features performs better than the baseline. In general, languages with larger gazetteers, such as Kiswahili (16K entities in the gazetteer) and Nigerian-Pidgin (for which we use an English gazetteer with 2M entities), have more improvement in performance than those with fewer gazetteer entries, such as Amharic and Luganda (2K and 500 gazetteer entities respectively). This indicates that having high-coverage gazetteers is important for the model to take advantage of the gazetteer features.

| Source Language | PER | ORG | LOC |
| --- | --- | --- | --- |
| eng-CoNLL | 36.17 | 27.00 | 50.50 |
| pcm | 21.50 | 65.33 | 68.17 |
| swa | 55.00 | 69.67 | 46.00 |
| hau | 52.67 | 57.50 | 48.50 |

Table 7.10: Average per-named entity F1-score for the zero-shot NER using the XLM-R model. The average is computed over `ibo`, `kin`, `lug`, `luo`, `wol`, `yor` languages.

### 7.6.3 *Distant Supervision Results*

Table 7.11 shows the performance of automatically annotating 10k sentences by ANEA using only English *(eng) Wikidata*, *eng + language Wikidata* and *eng + language Wikidata & curated lists*. The latter gives the best result with 20 F1 points better than using only *eng Wikidata* and 12 points better than using *eng + language Wikidata*. The improvement with each language Wikidata is small because of their small size. We achieve a higher F1 for LOC reaching 68.7 on average, while ORG entity has a very low F1 (34.8).

In Table 7.12, we compare the performance of the automatically annotated labels by ANEA (*Distant*) with *BOND*. BOND improves over Distant by (+7 points). With additional clean 100 sentences that have been humanly annotated, we achieve an impressive performance of 76.2 which is (+12.6 points) better than BOND. However, we did not see any improvement in combining distantly annotated labels with 100 clean examples (i.e. Distant + Clean). Similarly, Confusion Matrix (CM) struggle to beat the performance of 100 clean examples baseline, especially for `ibo` and `swa`, although `hau` and `yor` result improve significantly by $(2.7 - 3.8)$ points. Interestingly, we find that by first initializing with the PLM parameters with the 100-samples baseline before applying Confusion Matrix gives the best result for all languages except for `swa`.

While we obtain very impressive result in general using distant supervision and noisy handling learning, generating automatic annotation is a bit time consuming especially aggregating a large list of entities (which requires native speakers input for low-resource languages) and running ANEA tool. We believe that investing some time by native speakers to annotate few sentences e.g 100 sentences would result in better improvement Hedderich et al., 2020; Lauscher et al.,

| Method | hau | ibo | swa | yor | avg |
|---|---|---|---|---|---|
| *eng Wikidata* | | | | | |
| All | 35.0 | 27.9 | 40.2 | 29.7 | 33.2 |
| LOC | 25.5 | 25.5 | 54.5 | 41.3 | 36.7 |
| ORG | 8.8 | 13.7 | 13.1 | 7.1 | 17.1 |
| PER | 50.0 | 39.2 | 41.1 | 29.9 | 33.7 |
| *eng + language Wikidata* | | | | | |
| All | 57.9 | 28.1 | 48.5 | 30.8 | 41.3 |
| LOC | 76.7 | 26.7 | 70.5 | 43.2 | 54.3 |
| ORG | 9.0 | 13.7 | 13.5 | 7.1 | 10.8 |
| PER | 52.7 | 38.9 | 42.4 | 30.3 | 41.1 |
| *eng + language Wikidata & curated list* | | | | | |
| All | 58.8 | 50.8 | 54.7 | 49.9 | 53.6 |
| LOC | 80.2 | 49.6 | 74.6 | 70.4 | 68.7 |
| ORG | 17.2 | 57.4 | 38.6 | 25.8 | 34.8 |
| PER | 48.9 | 47.1 | 41.1 | 31.5 | 42.1 |

Table 7.11: Distant supervision with ANEA. Evalutation performed on the MasakhaNER test set for hau, ibo, swa, and yor. We compare performance using different entity lists

2020. As shown in Table 7.12, the 100 sentences baseline is very strong and it was difficult to improve over it with noisy label learning. Lastly, with the recent advances in cross-lingual transfer learning approaches like parameter efficient fine-tuning methods (Pfeiffer et al., 2020a). We could achieve better improvement (see results in Table 5.10, in Chapter 5) with less human and time effort if we leverage cross-lingual knowledge from high-resource languages like English.

### 7.6.4 Transfer Learning Experiments

Table 7.9 shows the result for the different transfer learning approaches, which we discuss individually in the following sections. We make use of XLM-R-base model for all the experiments in this sub-section because the performance difference if we use XLM-R-large is small (<2%) as shown in Table 7.7 and because it is faster to train.

#### 7.6.4.1 Cross-domain Transfer

We evaluate cross-domain transfer from Wikipedia to the news domain for the five languages that are available in the WikiAnn (Pan et al., 2017) dataset. In the zero-shot setting, the NER F1-score is low: less than 40 F1-score for all languages, with Kinyarwanda and Yorùbá having less than 10 F1-score. This is likely due to the number of training sentences present in WikiAnn: there are only 100 sentences in the datasets of Amharic, Igbo, Kinyarwanda and Yorùbá. Although the Kiswahili corpus has 1,000 sentences, the 35 F1-score shows that transfer is not very effective. In general, cross-domain transfer is a challenging problem, and is even harder when the number of training examples from the source domain is small. Fine-tuning on the in-domain news NER data does not improve over the baseline (XLM-R-base).

| Method | hau | ibo | swa | yor | avg |
|---|---|---|---|---|---|
| *only noisy labels* | | | | | |
| **Distant** (10k noisy sentences) | 62.2 | 55.7 | 56.4 | 52.2 | 56.6 |
| **BOND** (10k noisy sentences) | 68.1 | 64.2 | 66.8 | 55.1 | 63.6 |
| *only clean labels* | | | | | |
| **Clean model** (100 clean sentences) | 77.7 | 79.3 | **84.8** | 63.1 | 76.2 |
| *clean + noisy labels* | | | | | |
| **Distant + Clean** (10K noisy + 100 clean) | 77.2 | 76.5 | 74.2 | 63.4 | 72.8 |
| **BOND + Clean** (10K noisy + 100 clean) | 77.5 | 77.1 | 78.9 | **68.4** | 75.5 |
| **CM [Distant]** | 80.4 | 77.7 | 82.5 | 66.9 | 76.9 |
| **Clean model → CM [Distant]** | **82.0** | **79.7** | 81.3 | 65.5 | **77.1** |

Table 7.12: Noisy label learning results on the MasakhaNER test set for hau, ibo, swa, and yor. We compare performance training on 10k distantly annotated sentences (Distant) and different approaches with or without additional clean 100 sentences. All models are based on XLM-R-base + LAFT, we took an average over 5 runs.

7.6.4.2 *Cross-Lingual Transfer*

ZERO-SHOT In the zero-shot setting we evaluated NER models trained on the English eng-*CoNLL03* dataset, and on the Nigerian-Pidgin (pcm), Kiswahili (swa), and Hausa (hau) annotated corpus. We excluded the MISC entity in the eng-*CoNLL03* corpus because it is absent in our target datasets. Table 7.9 shows the result for the (zero-shot) transfer performance. We observe that the closer the source and target languages are geographically, the better the performance. The pcm model (trained on only 2K sentences) obtains similar transfer performance as the eng-*CoNLL03* model (trained on 14K sentences). swa performs better than pcm and eng-*CoNLL03* with an improvement of over 14 F1 on average. We found that, on average, transferring from Hausa provided the best F1, with an improvement of over 16% and 1% compared to using the eng-*CoNLL* and swa data respectively. Per-entity analysis in Table 7.10 shows that the largest improvements are obtained for ORG. The pcm data was more effective in transferring to LOC and ORG, while swa and hau performed better when transferring to

7.6 RESULTS 109

PER. In general, zero-shot transfer is most effective when transferring from Hausa and Kiswahili.

FINE-TUNING We use the target language corpus to fine-tune the NER models previously trained on eng-*CoNLL*, `pcm`, and `swa`. On average, there is only a small improvement when compared to the XLM-R base model. In particular, we see significant improvement for Hausa, Igbo, Kinyarwanda, Nigerian-Pidgin, Wolof, and Yorùbá using either `swa` or `hau` as the source NER model.

### 7.6.5 *Regional Influence on NER*

We evaluate whether combining different language training datasets by region affects the performance for individual languages. Table 7.9 shows that all languages spoken in West Africa (`ibo`, `wol`, `pcm`, `yor`) except `hau` have slightly better performance (0.1–2.6 F1) when we train on their combined training data. However, for the East-African languages, the F1 score only improved (0.8–2.3 F1) for three languages (`kin`, `lug`, `luo`). Training the NER model on all nine languages leads to better performance on all languages except Kiswahili. On average over six languages (`ibo`, `kin`, `lug`, `luo`, `wol`, `yor`), the performance improves by 1.6 F1.

### 7.6.6 *Error analysis*

| Lang. | CNN-BiLSTM | | | | | mBERT-base | | | | | XLM-R-base | | | | |
|---|---|---|---|---|---|---|---|---|---|---|---|---|---|---|---|
| | all | 0-freq | 0-freq Δ | long | long Δ | all | 0-freq | 0-freq Δ | long | long Δ | all | 0-freq | 0-freq Δ | long | long Δ |
| amh | 52.89 | 40.98 | -11.91 | 45.16 | -7.73 | – | – | – | – | – | 70.96 | 68.91 | -2.05 | 64.86 | -6.10 |
| hau | 83.70 | 78.52 | -5.18 | 66.21 | -17.49 | 87.34 | 79.41 | -7.93 | 67.67 | -19.67 | 89.44 | 85.48 | -3.96 | 76.06 | -13.38 |
| ibo | 78.48 | 70.57 | -7.91 | 53.93 | -24.55 | 85.11 | 78.41 | -6.70 | 60.46 | -24.65 | 84.51 | 77.42 | -7.09 | 59.52 | -24.99 |
| kin | 64.61 | 55.89 | -8.72 | 40.00 | -24.61 | 70.98 | 65.57 | -5.41 | 55.39 | -15.59 | 73.93 | 66.54 | -7.39 | 54.96 | -18.97 |
| lug | 74.31 | 67.99 | -6.32 | 58.33 | -15.98 | 80.56 | 76.27 | -4.29 | 65.67 | -14.89 | 80.71 | 73.54 | -7.17 | 63.77 | -16.94 |
| luo | 66.42 | 58.93 | -7.49 | 54.17 | -12.25 | 72.65 | 72.85 | 0.20 | 66.67 | -5.98 | 75.14 | 72.34 | -2.80 | 69.39 | -5.75 |
| pcm | 66.43 | 59.73 | -6.70 | 47.80 | -18.63 | 87.78 | 82.40 | -5.38 | 77.12 | -10.66 | 87.39 | 83.65 | -3.74 | 74.67 | -12.72 |
| swa | 79.26 | 64.74 | -14.52 | 44.78 | -34.48 | 86.37 | 78.77 | -7.60 | 45.55 | -40.82 | 87.55 | 80.91 | -6.64 | 53.93 | -33.62 |
| wol | 60.43 | 49.03 | -11.40 | 26.92 | -33.51 | 66.10 | 59.54 | -6.56 | 19.05 | -47.05 | 64.38 | 57.21 | -7.17 | 38.89 | -25.49 |
| yor | 67.07 | 56.33 | -10.74 | 64.52 | -2.55 | 78.64 | 73.41 | -5.23 | 74.34 | -4.30 | 77.58 | 72.01 | -5.57 | 76.14 | -1.44 |
| avg (excl. amh) | 69.36 | 60.27 | -9.09 | 50.18 | -19.18 | 79.50 | 74.07 | -5.43 | 59.10 | -20.40 | 79.15 | 73.80 | -5.36 | 63.22 | -15.94 |

Table 7.13: F1 score for two varieties of hard-to-identify entities: zero-frequency entities that do not appear in the training corpus, and longer entities of four or more words.

Finally, to better understand the types of entities that were successfully identified and those that were missed, we performed fine-grained analysis of our baseline methods mBERT and XLM-R using the method of Fu, Liu, and Neubig (2020), with results shown in Table 7.13. Specifically, we found that across all languages, entities that were not contained in the training data (zero-frequency entities), and entities consisting of more than three words (long entities) were particularly difficult in all languages; compared to the F1 score over all entities, the scores dropped by around 5 points when evaluated on zero-frequency entities, and by around 20 points when evaluated on long entities. Future work on low-resource NER or cross-lingual representation learning may further improve on these hard cases.

## 7.7 CONCLUSION AND FUTURE WORK

We address the NER task for African languages by bringing together a variety of stakeholders to create a high-quality NER dataset for ten African languages. We evaluate multiple state-of-the-art NER models and establish strong baselines. We have released one of our best models that can recognize named entities in ten African languages on HuggingFace Model Hub[10]. We also investigate cross-domain transfer with experiments on five languages with the WikiAnn dataset, along with cross-lingual transfer for low-resource NER using the English CoNLL-2003 dataset and other languages supported by XLM-R.

---

10 https://huggingface.co/Davlan/xlm-roberta-large-masakhaner

# 8

## MASAKHANER 2.0: AFRICA-CENTRIC TRANSFER LEARNING FOR NER

In Chapter 7, we introduced the first large scale dataset for named entity recognition (NER) covering ten African languages. Despite the coverage, it excludes some languages from the Southern Africa region with distinct linguistic characteristics. We also show some impressive zero-shot transfer learning result from English language. We further show that transferring from an African language such as Kiswahili or Hausa gave better result (up to 15 points of F1-score) than transferring from English (see, Table 7.9). This shows that the choice of transfer language is very crucial.

This Chapter[1] describes the expansion of the MasakhaNER dataset to cover more typologically and geographically diverse languages. Specifically, we create the largest human-annotated NER dataset for 20 African languages, and we study the behavior of state-of-the-art cross-lingual transfer methods in an Africa-centric setting, demonstrating that the choice of source language significantly affects performance. We show that choosing the best transfer language improves zero-shot F1 scores by an average of 14 points across 20 languages compared to using English. Our results highlight the need for benchmark datasets and models that cover typologically-diverse African languages.

### 8.1 INTRODUCTION

Many African languages are spoken by millions or tens of millions of speakers. However, these languages are poorly represented in NLP research, and the development of NLP systems for African languages is often limited by the lack of datasets for training and evaluation (Adelani et al., 2021b).

Additionally, while there has been much recent work in using zero-shot cross-lingual transfer (Ebrahimi et al., 2022; Pfeiffer et al., 2020b; Ponti et al., 2020) to improve performance on tasks for low-resource languages with multilingual pretrained language models (PLMs) (Conneau et al., 2020; Devlin et al., 2019), the settings under which contemporary transfer learning methods work best are still unclear (Lauscher et al., 2020; Pruksachatkun et al., 2020; Xia et al., 2020). For example, several methods use English as the source language because of the availability of training data across many tasks (Hu et al., 2020; Ruder et al., 2021), but there is evidence that English is often not the best transfer language (Lin et al., 2019; Vries, Wieling, and Nissim, 2022),

---

1 This is based on Adelani et al., 2022a (under submission)

and the process of choosing the best source language to transfer from remains an open question.

There has been recent progress in creating benchmark datasets for training and evaluating models in African languages for several tasks such as machine translation (Adelani et al., 2022b; Reid et al., 2021; ∀ et al., 2020), sentiment analysis (Muhammad et al., 2022; Yimam et al., 2020), and other text classification tasks (Hedderich et al., 2020). In this paper, we focus on the standard NLP task of named entity recognition (NER) because of its utility in downstream applications such as question answering and information extraction. For NER, annotated datasets exist only in a few African languages (Adelani et al., 2021b; Alabi et al., 2020; Yohannes and Amagasa, 2022), the largest of which is the MasakhaNER dataset (Adelani et al., 2021b) (or MasakhaNER 1.0). While MasakhaNER 1.0 covers 10 African languages spoken mostly in West and East Africa, it does not include any languages spoken in Southern Africa, which have distinct syntactic and morphological characteristics and are spoken by 40 million people.

In this paper, we tackle two current challenges in developing NER models for African languages: (1) the lack of typologically- and geographically-diverse evaluation datasets for African languages; and (2) choosing the best transfer language for NER in an Africa-centric setting, which has not been previously explored in the literature.

To address the first challenge, we create the MasakhaNER 2.0 corpus, the largest human-annotated NER dataset for African languages. MasakhaNER 2.0 contains annotated text data from 20 languages widely spoken in Sub-Saharan Africa and is complementary to the languages present in previously existing datasets (e.g., Adelani et al., 2021b). We discuss our annotation methodology as well as perform benchmarking experiments on our dataset with state-of-the-art NER models based on multilingual PLMs.

In addition, to better understand the effect of source language on transfer learning, we extensively analyze different features that contribute to cross-lingual transfer, including linguistic characteristics of the languages (i.e., typological, geographical, and phylogenetic features) as well as data-dependent features such as entity overlap across source and target languages (Lin et al., 2019). We demonstrate that choosing the best transfer language(s) in both single-source and co-training setups leads to large improvements in NER performance in zero-shot settings; our experiments show an average of a 14 point increase in F1 score as compared to using English as source language across 20 target African languages.

## 8.2 RELATED WORK

AFRICAN NER DATASETS  NER is an important task in NLP involved in many benchmark datasets like XTREME (Hu et al., 2020)

and is necessary for numerous applications. Due to its importance, several efforts have been made to create both automatically annotated datasets like WikiAnn (Pan et al., 2017; Rahimi, Li, and Cohn, 2019) and human-annotated datasets in several languages (Benajiba, Rosso, and BenedíRuiz, 2007; Dumitrescu and Avram, 2020; Gruzitis et al., 2018; Johansen, 2019; Khairunnisa, Imankulova, and Komachi, 2020; Poostchi et al., 2016; Singh, Padia, and Joshi, 2019; Szarvas et al., 2006; Tjong Kim Sang, 2002; Tjong Kim Sang and De Meulder, 2003). WikiAnn (Rahimi, Li, and Cohn, 2019) covers 176 languages including nine African languages but their datasets are often small in size (100–1000 sentences) and of lower quality due to the automatic annotation process. There are some human-annotated NER datasets for African languages such as the SaDiLAR NER corpus (Eiselen, 2016) covering 10 South African languages, LORELEI (Strassel and Tracey, 2016), which covers nine African languages but is not open-sourced, and some individual language efforts for Amharic[2], Yorùbá (Alabi et al., 2020), Hausa (Hedderich et al., 2020), and Tigrinya (Yohannes and Amagasa, 2022). Closest to our work is the MasakhaNER 1.0 corpus (Adelani et al., 2021b), which covers 10 widely spoken languages in the news domain, but excludes languages from the southern region of Africa like isiZulu, isiXhosa, and chiShona with distinct syntactic features (e.g., noun prefixes and capitalization in between words) which limits transfer learning from other languages. We include five languages from Southern Africa in our new corpus.

NER MODELS  The state-of-the-art approach for developing NER models is by fine-tuning pre-trained language models (PLM) like BERT (Devlin et al., 2019). Prior to PLMs, CNN-BiLSTM-CRF (Ma and Hovy, 2016) was the prevailing approach though its performance depended on pre-trained word embeddings whose quality is often lower for low-resource languages due to the availability of only small and noisy monolingual corpora (Alabi et al., 2020; Kreutzer et al., 2021). Multilingual PLMs like mBERT, XLM-R (Conneau et al., 2020), RemBERT (Chung et al., 2021a), and mDeBERTa (He, Gao, and Chen, 2021b) have shown impressive performance on several NLP tasks even for low-resourced languages (Adelani et al., 2021b; Ebrahimi et al., 2022; Ruder et al., 2021) that are not covered by any multilingual PLM. Multilingual PLMs, to some extent, help alleviate the challenges posed by small monolingual data for training word by enabling positive transfer from related languages.

CROSS-LINGUAL TRANSFER  Leveraging cross-lingual transfer has the potential to drastically improve model performance without requiring large amounts of data in the target language (Conneau et al., 2020) but it is not always clear from which language we must transfer from (Lin et al., 2019; Vries, Wieling, and Nissim, 2022). To this end, recent work investigates methods for selecting good transfer languages and informative features. For instance, token overlap between the source and target language is a useful predictor of transfer performance for some tasks (Lin et al., 2019; Wu and Dredze, 2019). Linguistic distance (Lin et al., 2019; Vries, Wieling, and Nissim, 2022), word order (K et al., 2020; Pires, Schlinger, and Garrette, 2019) and script differences (Vries, Wieling, and Nissim, 2022), and syntactic similarity (Karamolegkou and Stymne, 2021) have also been shown to impact performance. Another research direction attempts to build models of transfer performance that predicts the best transfer language for a target language by using some linguistic and data-dependent features (Ahuja et al., 2022; Lin et al., 2019).

---

2 https://github.com/uhh-lt/amharicmodels/tree/master/data/NER

| Language | Family | African Region | No. of Speakers |
|---|---|---|---|
| Bambara (bam) | NC / Mande | West | 14M |
| Ghomálá' (bbj) | NC / Grassfields | Central | 1M |
| Éwé (ewe) | NC / Kwa | West | 7M |
| Fon (fon) | NC / Volta-Niger | West | 2M |
| Hausa (hau) | Afro-Asiatic / Chadic | West | 63M |
| Igbo (ibo) | NC / Volta-Niger | West | 27M |
| Kinyarwanda (kin) | NC / Bantu | East | 10M |
| Luganda (lug) | NC / Bantu | East | 7M |
| Dholuo (luo) | Nilo-Saharan | East | 4M |
| Mossi (mos) | NC / Gur | West | 8M |
| Naija (pcm) | English-Creole | West | 75M |
| Chichewa (nya) | NC / Bantu | South-East | 14M |
| chiShona (sna) | NC / Bantu | South | 12M |
| Kiswahili (swa) | NC / Bantu | East & Central | 98M |
| Setswana (tsn) | NC / Bantu | South | 14M |
| Akan/Twi (twi) | NC / Kwa | West | 9M |
| Wolof (wol) | NC / Senegambia | West | 5M |
| isiXhosa (xho) | NC / Bantu | South | 9M |
| Yorùbá (yor) | NC / Volta-Niger | West | 42M |
| isiZulu (zul) | NC / Bantu | South | 27M |

Table 8.1: **Languages, Language, family (NC: Niger-Congo), number of speakers.**

## 8.3 LANGUAGES AND THEIR CHARACTERISTICS

### 8.3.1 *Focus Languages*

Table 8.1 provides an overview of the languages in our MasakhaNER 2.0 corpus. We focus on 20 Sub-Saharan African languages with varying numbers of speakers (between 1M–100M) that are spoken by over 500M people in around 27 countries in the Western, Eastern, Central and Southern regions of Africa. The selected languages cover four language families. 17 languages belong to the Niger-Congo language family, and one language belongs to each of the Afro-Asiatic (Hausa), Nilo-Saharan (Dholuo), and English Creole (Naija) families. Although many languages belong to the Niger-Congo language family, they have different linguistic characteristics. For instance, Bantu languages (eight in our selection) make extensive use of affixes, unlike many languages of non-Bantu subgroups such as Gur, Kwa, and Volta-Niger.[3]

### 8.3.2 *Language Characteristics*

SCRIPT AND WORD ORDER   African languages mainly employ four major writing scripts: Latin, Arabic, N'ko and Ge'ez. Our focus languages mostly make use of the Latin script. During the pre-colonial period, the Ajami script, a variant of Arabic was used for Hausa, Kiswahili, and Yorùbá but has now declined in popularity. While N'ko is still actively used by the Mande languages like Bambara, the most widely used writing script for the language is Latin. However, some languages use additional letters that go beyond the standard Latin script, e.g., "ɛ", "ɔ", "ŋ", "ẹ", and more than one character letters like "bv", "gb", "mpf", "ntsh". 17 of the languages are tonal except for Naija, Kiswahili and Wolof. Nine of the languages make use of diacritics (e.g., é, ë, ñ). All languages use the SVO word order, while Bambara additionally uses the SOV word order.

MORPHOLOGY AND NOUN CLASSES   Many African languages are morphologically rich. According to the World Atlas of Language Structures WALS; Nichols and Bickel, 2013, 16 of our languages employ strong prefixing or suffixing inflections. Niger-Congo languages are known for their system of noun classification. 12 of the languages *actively* make use of between 6–20 noun classes, including all Bantu languages and Ghomálá', Mossi, Akan and Wolof (Babou and Loporcaro, 2016; Bodomo and Marfo, 2002; Nurse and Philippson, 2006; Payne, Pacchiarotti, and Bosire, 2017). While noun classes are often marked using affixes on the head word in Bantu languages, some non-Bantu languages, e.g., Wolof make use of a dependent such as a determiner that is not attached to the head word. For the other Niger-Congo languages such as Fon, Ewe, Igbo and Yorùbá, the use of noun classes is merely *vestigial* (Konoshenko and Shavarina, 2019). For example, Yorùbá only distinguishes between human and non-human nouns. Bambara is the only Niger-Congo language without noun classes, and some have argued that the Mande family should be regarded as an independent language family. Three of our languages from the Southern Bantu family (chiShona, isiXhosa and isiZulu) capitalize proper names after the noun class prefix as in the language names themselves. This characteristic may limit transfer from languages without this feature as NER models overfit on capitalization (Mayhew, Tsygankova, and Roth, 2019). Section 2.2 provides more details regarding the languages' linguistic characteristics.

---

[3] Our selection was constrained by the availability of volunteers that speak the languages in different NLP/AI communities in Africa.

| Language | Source | Train / dev / test | % Entities in Tokens | No. Tokens |
|---|---|---|---|---|
| Bambara (bam) | MAFAND-MT (Adelani et al., 2022b) | 4462/ 638/ 1274 | 6.5 | 155,552 |
| Ghomálá' (bbj) | MAFAND-MT (Adelani et al., 2022b) | 3384/ 483/ 966 | 11.3 | 69,474 |
| Éwé (ewe) | MAFAND-MT (Adelani et al., 2022b) | 3505/ 501/ 1001 | 15.3 | 90420 |
| Fon (fon) | MAFAND-MT (Adelani et al., 2022b) | 4343/ 621/ 1240 | 8.3 | 173,099 |
| Hausa (hau) | Kano Focus and Freedom Radio | 5716/ 816/ 1633 | 14.0 | 221,086 |
| Igbo (ibo) | IgboRadio and Ka ỌdỊ Taa | 7634/ 1090/ 2181 | 7.5 | 344,095 |
| Kinyarwanda (kin) | IGIHE, Rwanda | 7825/ 1118/ 2235 | 12.6 | 245,933 |
| Luganda (lug) | MAFAND-MT (Adelani et al., 2022b) | 4942/ 706/ 1412 | 15.6 | 120,119 |
| Dholuo (luo) | MAFAND-MT (Adelani et al., 2022b) | 5161/ 737/ 1474 | 11.7 | 229,927 |
| Mossi (mos) | MAFAND-MT (Adelani et al., 2022b) | 4532/ 648/ 1294 | 9.2 | 168,141 |
| Naija (pcm) | MAFAND-MT (Adelani et al., 2022b) | 5646/ 806/ 1613 | 9.4 | 206,404 |
| Chichewa (nya) | Nation Online Malawi | 6250/ 893/ 1785 | 9.3 | 263,622 |
| chiShona (sna) | VOA Shona | 6207/ 887/ 1773 | 16.2 | 195,834 |
| Kiswahili (swa) | VOA Swahili | 6593/ 942/ 1883 | 12.7 | 251,678 |
| Setswana (tsn) | MAFAND-MT (Adelani et al., 2022b) | 3489/ 499/ 996 | 8.8 | 141,069 |
| Akan/Twi (twi) | MAFAND-MT (Adelani et al., 2022b) | 4240/ 605/ 1211 | 6.3 | 155,985 |
| Wolof (wol) | MAFAND-MT (Adelani et al., 2022b) | 4593/ 656/ 1312 | 7.4 | 181,048 |
| isiXhosa (xho) | Isolezwe Newspaper | 5718/ 817/ 1633 | 15.1 | 127,222 |
| Yorùbá (yor) | Voice of Nigeria and Asejere | 6877/ 983/ 1964 | 11.4 | 244,144 |
| isiZulu (zul) | Isolezwe Newspaper | 5848/ 836/ 1670 | 11.0 | 128,658 |

Table 8.2: **Languages and Data Splits for MasakhaNER 2.0 Corpus**. Language, news source, and data split in number of sentences

## 8.4 MASAKHANER 2.0 CORPUS

### 8.4.1 *Data source and collection*

We annotate news articles from local sources. The choice of the news domain is based on the availability of data for many African languages and the variety of named entities types (e.g., person names and locations) as illustrated by popular datasets such as CoNLL-03 (Tjong

Kim Sang and De Meulder, 2003).[4] Table 8.2 shows the sources and sizes of the data we use for annotation. Overall, we collected between 4.8K–11K sentences per language from either a monolingual or a translation corpus.

MONOLINGUAL CORPUS   We collect a large monolingual corpus for nine languages, mostly from local news articles except for chiShona and Kiswahili texts, which were crawled from Voice of America (VOA) websites.[5] As Yorùbá text was missing diacritics, we asked native speakers to manually add diacritics before annotation. During data collection, we ensured that the articles are from a variety of topics e.g. politics, sports, culture, technology, society, and education. In total, we collected between 8K–11K sentences per language.

TRANSLATION CORPUS   For the remaining languages for which we were unable to obtain sufficient amounts of monolingual data, we use a translation corpus, MAFAND-MT (Adelani et al., 2022b), which consists of French and English news articles translated into 11 languages. We note that translationese may lead to undesired properties, e.g., unnaturalness. However, we did not observe serious issues during the annotation. The number of sentences is constrained by the size of the MAFAND-MT corpus, which is between 4,800–8,000.

### 8.4.2  *NER Annotation Methodology*

We annotated the collected monolingual texts with the ELISA annotation tool (Lin et al., 2018) with four entity types: Personal name (`PER`), Location (`LOC`), Organization (`ORG`), and date and time (`DATE`), similar to MasakhaNER 1.0 (Adelani et al., 2021b). We made use of the MUC-6 annotation guide.[6] The annotation was carried out by three native speakers per language recruited from AI/NLP communities in Africa. To ensure high-quality annotation, we recruited a language coordinator to supervise annotation in each language. We organized two online workshops to train language coordinators on the NER annotation. As part of the training, each coordinator annotated 100 English sentences which were verified. Each coordinator then trained three annotators in their team using both English and African language texts with the support of the workshop organizers. All annotators and language coordinators received appropriate remuneration.[7]

At the end of annotation, language coordinators worked with their team to resolve disagreements using the adjudication function of ELISA, which ensures a high inter-annotator agreement score.

---

4 We also considered using Wikipedia as our data source, did not due to quality issues (Alabi et al., 2020).
5 www.voashona.com/ and www.voaswahili.com/
6 https://cs.nyu.edu/~grishman/muc6.html
7 $10 per hour, annotating about 200 sentences per hour.

| Lang. | Fleiss' Kappa | QC flags fixed? | Lang. | Fleiss' Kappa | QC flags fixed? |
|---|---|---|---|---|---|
| bam | 0.980 | ✗ | pcm | 0.966 | ✗ |
| bbj | 1.000 | ✓ | nya | 0.988 | ✓ |
| ewe | 0.991 | ✓ | sna | 0.957 | ✓ |
| fon | 0.941 | ✗ | swa | 0.974 | ✓ |
| hau | 0.950 | ✗ | tsn | 0.962 | ✗ |
| ibo | 0.965 | ✗ | twi | 0.932 | ✗ |
| kin | 0.943 | ✗ | wol | 0.979 | ✓ |
| lug | 0.950 | ✓ | xho | 0.945 | ✓ |
| luo | 0.907 | ✗ | yor | 0.950 | ✓ |
| mos | 0.927 | ✗ | zul | 0.953 | ✓ |

Table 8.3: Inter-annotator agreement for our datasets calculated using Fleiss' kappa $\kappa$ at the entity level before adjudication. QC flags (✓) are the languages that fixed the annotations for all **Q**uality **C**ontrol flagged tokens.

### 8.4.3 *Quality Control*

Annotations were automatically adjudicated when there was agreement, but were flagged for further review when annotators disagreed on mention spans or types. The process for reviewing and fixing quality control issues was voluntary and so not all languages were further reviewed (see Table 8.3).

We automatically identified positions in the annotation that were more likely to be annotation errors and flagged them for further review and correction. The automatic process flags tokens that are commonly annotated as a named entity but were not marked as a named entity in a specific position. For example, *Boston* may appear commonly as a named entity and infrequently not as a named entity, so when it is seen as not marked it was flagged. Similarly, we flagged tokens that had near-zero entropy with regard to a certain entity type, for example a token almost always annotated as ORG but very rarely annotated as PER. We also flagged potential sentence boundary errors by identifying sentences with few tokens or sentences which end in a token that appears to be an abbreviation or acronym. As shown in Table 8.3, before further adjudication and correction there was already relatively high inter-annotator agreement measured by Fleiss' Kappa at the mention level.

After quality control, we divided the annotation into training, development, and test splits consisting of 70%, 10%, and 20% of the data respectively.

| PLM | # Lang. | Languages in MasakhaNER 2.0 |
|---|---|---|
| mBERT-cased (110M) | 104 | **swa**, **yor** |
| XLM-R-base/large (270M / 550M) | 100 | **hau**, **swa**, **xho** |
| mDeBERTaV3 (276M) | 100 | **hau**, **swa**, **xho** |
| RemBERT (575M) | 110 | **hau**, **ibo**, **nya**, **sna**, **swa**, **xho**, **yor**, **zul** |
| AfriBERTa (126M) | 11 | **hau**, **ibo**, **kin**, **pcm**, **swa**, **yor** |
| AfroXLMR-base/large (270M/550M) | 20 | **hau**, **ibo**, **kin**, **nya**, **pcm**, **sna**, **swa**, **xho**, **yor**, **zul** |

Table 8.4: Language coverage and size for PLMs.

### 8.4.4 *Other NER Corpus*

We also make use of other publicly available NER datasets for the transfer learning experiments. Table 8.5 provides the NER corpus found online that we make use for determining the best transfer languages

| Language | Data Source | # Train | # dev | # test |
|---|---|---|---|---|
| Amharic (amh) | MasakhaNER 1.0 (Adelani et al., 2021b) | 1,750 | 250 | 500 |
| Arabic (ara) | ANERcorp (Benajiba, Rosso, and BenedíRuiz, 2007; Obeid et al., 2020) | 3,472 | 500 | 924 |
| Danish (dan) | DANE (Hvingelby et al., 2020) | 4,383 | 564 | 565 |
| German (deu) | CoNLL03 (Tjong Kim Sang and De Meulder, 2003) | 12,152 | 2,867 | 3,005 |
| English (eng) | CoNLL03 (Tjong Kim Sang and De Meulder, 2003) | 14,041 | 3,250 | 3,453 |
| Spanish (spa) | CoNLL02 (Tjong Kim Sang, 2002) | 8,322 | 1,914 | 1,516 |
| Farsi (fas) | PersoNER (Poostchi et al., 2016) | 4,121 | 1,000 | 2,560 |
| Finnish (fin) | FINER (Ruokolainen et al., 2019) | 13,497 | 986 | 3,512 |
| French (fra) | Europeana (Neudecker, 2016) | 9,546 | 2,045 | 2,047 |
| Hungarian (hun) | Hungarian MTI (Szarvas et al., 2006) | 4,532 | 648 | 1,294 |
| Indonesia (ind) | (Khairunnisa, Imankulova, and Komachi, 2020) | 6,707 | 1,437 | 1,438 |
| Italian (ita) | I-CAB EVALITA 2007 & 2009 (Magnini et al., 2008) | 11,227 | 4,136 | 2,068 |
| Korean (kor) | KLUE (Park et al., 2021) | 20,008 | 1,000 | 5,000 |
| Latvian (lav) | (Gruzitis et al., 2018) | 7,997 | 1,713 | 1,715 |
| Nepali (nep) | (Singh, Padia, and Joshi, 2019) | 2,301 | 328 | 659 |
| Dutch (nld) | CoNLL02 (Tjong Kim Sang, 2002) | 15,806 | 2,895 | 5,195 |
| Norwegian (nor) | (Johansen, 2019) | 15,696 | 2,410 | 1,939 |
| Portuguese (por) | Second HAREM (Freitas et al., 2010) & Paramopama (Junior et al., 2015) | 11,258 | 2,412 | 2,414 |
| Romanian (ron) | RONEC (Dumitrescu and Avram, 2020) | 5,886 | 1,000 | 2,453 |
| Swedish (swe) | "swedish_ner_corpus" on HuggingFace Datasets (Lhoest et al., 2021) | 9,000 | 1,330 | 2,000 |
| Ukrainian (ukr) | "benjamin/ner-uk" on HuggingFace Datasets (Lhoest et al., 2021) | 10,833 | 1,307 | 668 |
| Chinese (zho) | "msra_ner" on HuggingFace Datasets (Lhoest et al., 2021) | 45,057 | 3,442 | 1,721 |

Table 8.5: **Languages and Data Splits for Other NER Datasets**.

## 8.5 BASELINE EXPERIMENTS

### 8.5.1 *Baseline Models*

As baselines, we fine-tune several multilingual PLMs including multilingual BERT (mBERT) (Devlin et al., 2019), XLM-R base & large; Conneau et al., 2020, mDeBERTaV3 (He, Gao, and Chen, 2021a), AfriB-ERTa (Ogueji, Zhu, and Lin, 2021), RemBERT (Chung et al., 2021a), and AfroXLM-R base & large; Alabi et al., 2022. We fine-tune the PLMs on each language's training data and evaluate performance on the test set using HuggingFace Transformers (Wolf et al., 2020).

MASSIVELY MULTILINGUAL PLMS    Table 8.4 shows the language coverage and size of different massively multilingual PLMs trained on 100–110 languages. mBERT was pre-trained using masked language modeling (MLM) and next-sentence prediction on 104 languages including swa and yor. RemBERT was trained with a similar objective but makes use of a larger output embedding size during pre-training and covers more African languages. XLM-R was trained only with MLM on 100 languages and on a larger pre-training corpus. mDeBERTaV3 makes use of ELECTRA-style (Clark et al., 2020) pre-training, i.e., a replaced token detection (RTD) objective instead of MLM.

AFRICA-CENTRIC MULTILINGUAL PLMS    We also obtained NER models by fine-tuning two PLMs that are pre-trained on African languages. AfriBERTa (Ogueji, Zhu, and Lin, 2021) was pre-trained on less than 1GB of text covering 11 African languages including six of our focus languages, and has shown impressive performance on NER and sentiment classification for languages in its pre-training data (Adelani et al., 2021b; Muhammad et al., 2022). AfroXLM-R (Alabi et al., 2022) is a language-adapted (Pfeiffer et al., 2020b) version of XLM-R that was fine-tuned on 17 African languages and three high-resource languages widely spoken in Africa ("eng", "fra", and "ara").

BASELINE MODEL HYPER-PARAMETERS    For training NER models, we *fine-tune* PLM, we make use of a maximum sequence length of 200, batch size of 16, gradient accumulation of 2, learning rate of 5e-5, and number of epochs 50. The experiments of the large PLMs were performed on using Nvidia V100 GPU. For AfriBERTa and mBERT, we make use of Nvidia GeForce RTX-2080Ti.

8.5.2 *Baseline Results*

Table 8.6 shows the results of training NER models on each language using the eight multilingual and Africa-centric PLMs. All PLMs provided good performance in general. However, we observed worse results for mBERT and AfriBERTa especially for languages they were not pre-trained on. For instance, both models performed between 6–12 F1 worse for bbj, wol or zul compared to XLM-R-base. We hypothesize that the performance drop is largely due to the small number of African languages covered by mBERT as well as AfriBERTa's comparatively small model capacity. XLM-R-base gave much better performance ($> 1.0$ F1) on average compared to mBERT and AfriBERTa. We found the larger variants of mBERT and XLM-R, i.e., RemBERT and XLM-R-large to give much better performance ($> 2.0$ F1) than the smaller models. Their larger capacity facilitates positive transfer, yielding better performance for unseen languages. Surprisingly, mDeBERTaV3 provided slightly better results than XLM-R-large and RemBERT despite its smaller size, demonstrating the benefits of the RTD pre-training (Clark et al., 2020).

| Language | PLM pre-trained on 100+ world languages | | | | | PLM pre-trained on African languages | | |
|---|---|---|---|---|---|---|---|---|
| | mBERT cased | XLM-R base | XLM-R large | RemBERT | mDeBERTaV3 base | AfriBERTa large | AfroXLMR base | AfroXLMR large |
| bam | 78.9 | 78.7 | 79.4 | 80.1 | 80.2 | 78.6 | 79.6 | **82.2** |
| bbj | 60.6 | 72.3 | **75.2** | 74.2 | 73.5 | 71.0 | 73.3 | 74.8 |
| ewe | 86.9 | 88.5 | 89.1 | 89.2 | 89.8 | 86.9 | 89.2 | **90.3** |
| fon | 79.9 | 81.9 | 81.6 | 82.2 | 81.8 | 79.9 | 82.3 | **82.7** |
| hau | 85.2 | 83.8 | 86.3 | 84.7 | 85.4 | 85.2 | 86.6 | **87.4** |
| ibo | 87.3 | 87.8 | 87.2 | 86.4 | 88.8 | 87.3 | 88.5 | **89.6** |
| kin | 83.2 | 82.5 | 84.3 | 85.2 | 86.4 | 83.2 | 86.1 | **87.5** |
| lug | 85.5 | 86.7 | 88.1 | 87.1 | 88.7 | 85.5 | 88.1 | **89.6** |
| luo | 80.3 | 79.3 | 80.8 | 80.4 | 80.3 | 78.4 | 80.8 | **82.2** |
| mos | 71.4 | 72.7 | 74.9 | 72.7 | **76.4** | 71.4 | 74.4 | **76.4** |
| nya | 88.6 | 89.9 | 90.5 | 91.4 | 92.0 | 88.6 | 91.9 | **92.4** |
| pcm | 87.1 | 88.5 | 89.2 | 89.5 | **90.1** | 87.1 | 89.3 | 89.7 |
| sna | 92.4 | 93.6 | 94.2 | 94.8 | 95.5 | 92.4 | 95.7 | **96.2** |
| swa | 92.1 | 92.2 | 92.6 | 92.0 | 92.5 | 92.1 | 92.3 | **92.7** |
| tsn | 86.4 | 86.1 | 85.9 | 87.0 | 86.5 | 83.2 | 87.7 | **89.4** |
| twi | 75.7 | 78.7 | 79.8 | 78.5 | 79.4 | 75.7 | 78.9 | **81.1** |
| wol | 79.9 | 82.3 | 82.0 | 83.6 | 83.6 | 79.9 | 84.9 | **86.8** |
| xho | 85.0 | 87.0 | 88.1 | 88.3 | 88.1 | 85.0 | 88.6 | **89.9** |
| yor | 87.7 | 85.8 | 86.6 | 87.2 | 86.7 | 87.7 | 88.3 | **89.3** |
| zul | 81.7 | 84.6 | 86.7 | 85.5 | 88.3 | 81.7 | 88.4 | **90.6** |
| AVG | 82.8$_{\pm 0.2}$ | 84.1$_{\pm 0.1}$ | 85.1$_{\pm 0.5}$ | 85.0$_{\pm 0.2}$ | 85.7$_{\pm 0.2}$ | 83.0$_{\pm 0.2}$ | 85.7$_{\pm 0.1}$ | **87.0**$_{\pm 0.2}$ |

Table 8.6: **NER Baselines on MasakhaNER 2.0**. We compare several multilingual PLMs including the ones trained on African languages. Average is over 5 runs.

The best PLM is AfroXLM-R-large, which outperforms mDeBERTaV3, RemBERT and AfriBERTa by $+1.3$ F1, $+2.0$ F1 and $+4.0$ F1 respectively. Even the performance of its smaller variant, AfroXLM-R-base is comparable to mDeBERTaV3. Overall, our baseline results highlight that large PLMs, PLM with improved pre-training objectives, and PLMs pre-trained on the target African languages are able to achieve reasonable baseline performance. Combining these criteria provides improved performance such as AfroXLM-R-large, a large PLM trained on several African languages.

### 8.5.3 *Entity-level Analysis of MasakhaNER 2.0*

#### 8.5.3.1 *Error Analysis with ExplainaBoard*

Furthermore, using ExplainaBoard (Liu et al., 2021), we analysed the best three baseline NER models: AfroXLM-R-large, mDeBERTaV3, and XLM-R-large. We discovered that 2-token entities were easier to predict accurately than lengthier entities (4 or more words). Moreover, the result shows that all of the models have difficulty predicting zero-frequency entities effectively (entities with no occurrences in the training set). Interestingly, AfroXLMR-large is significantly better than other models for zero-frequency entities, suggesting that training PLMs on African languages promotes generalization to unseen entities. Finally, we observed that the three models perform better when predicting PER and LOC entities compared to ORG and DATE entities by up to (+5%). Table 8.7 and Table 8.8 provides more details on the error analysis.

| Language | XLM-R-large | | | | | mDeBERTaV3-base | | | | | AfroXLMR-large | | | | |
|---|---|---|---|---|---|---|---|---|---|---|---|---|---|---|---|
| | all | o-freq | Δ o-freq | long | Δ long | all | o-freq | Δ o-freq | long | Δ long | all | o-freq | Δ o-freq | long | Δ long |
| bam | 79.4 | 62.3 | -17.1 | 74.7 | -4.7 | 81.3 | 66.3 | -15.0 | 78.6 | -2.7 | 82.1 | 67.2 | -14.9 | 81.1 | -1.0 |
| bbj | 74.8 | 66.1 | -8.7 | 87.4 | 12.6 | 75.0 | 65.8 | -9.2 | 63.9 | -11.1 | 76.5 | 65.8 | -10.7 | 80.0 | 3.5 |
| ewe | 89.5 | 75.6 | -13.9 | 70.6 | -18.9 | 90.0 | 76.9 | -13.1 | 70 | -20.0 | 91.0 | 79.7 | -11.3 | 74.2 | -16.8 |
| fon | 81.5 | 71.2 | -10.3 | 69.6 | -11.9 | 83.3 | 74.5 | -8.8 | 68.1 | -15.2 | 82.8 | 73.6 | -9.2 | 68.7 | -14.1 |
| hau | 87.4 | 83.8 | -3.6 | 77.6 | -9.8 | 84.8 | 80.0 | -4.8 | 72.2 | -12.6 | 87.8 | 84.6 | -3.2 | 78.1 | -9.7 |
| ibo | 87.0 | 77.4 | -9.6 | 75.6 | -11.4 | 89.7 | 82.6 | -7.1 | 71.8 | -17.9 | 89.1 | 80.9 | -8.2 | 64.0 | -25.1 |
| kin | 84.1 | 74.9 | -9.2 | 75.3 | -8.8 | 86.2 | 79.0 | -7.2 | 75.3 | -10.9 | 87.8 | 81.7 | -6.1 | 77.1 | -10.7 |
| lug | 87.3 | 75.3 | -12.0 | 74.1 | -13.2 | 88.7 | 77.4 | -11.3 | 78.6 | -10.1 | 89.4 | 79.7 | -9.7 | 74.7 | -14.7 |
| mos | 77.1 | 69.5 | -7.6 | 55.8 | -21.3 | 78.0 | 71.2 | -6.8 | 58.9 | -19.1 | 77.5 | 70.2 | -7.3 | 60.1 | -17.4 |
| nya | 89.7 | 82.0 | -7.7 | 81.6 | -8.1 | 91.9 | 86.5 | -5.4 | 86.7 | -5.2 | 92.2 | 87.3 | -4.9 | 87.1 | -5.1 |
| pcm | 89.8 | 84.5 | -5.3 | 76.8 | -13.0 | 90.2 | 84.9 | -5.3 | 79.7 | -10.5 | 90.4 | 86.1 | -4.3 | 79.1 | -11.3 |
| sna | 94.9 | 89.9 | -5.0 | 93.3 | -1.6 | 95.3 | 91.4 | -3.9 | 92.4 | -2.9 | 96.3 | 93.9 | -2.4 | 93.9 | -2.4 |
| swa | 92.8 | 84.1 | -8.7 | 73.0 | -19.8 | 92.4 | 82.8 | -9.6 | 65.1 | -27.3 | 92.3 | 83.0 | -9.3 | 65.9 | -26.4 |
| tsn | 86.4 | 74.9 | -11.5 | 34.5 | -51.9 | 87.0 | 75.8 | -11.2 | 45.7 | -41.3 | 89.8 | 80.9 | -8.9 | 42.9 | -46.9 |
| twi | 77.9 | 65.5 | -12.4 | 52.2 | -25.7 | 80.4 | 70.9 | -9.5 | 62.3 | -18.1 | 81.4 | 72.3 | -9.1 | 63.2 | -18.2 |
| wol | 83.3 | 65.9 | -17.4 | 59.1 | -24.2 | 83.3 | 67.2 | -16.1 | 58.6 | -24.7 | 86.2 | 72.0 | -14.2 | 62.2 | -24.0 |
| xho | 88.0 | 83.2 | -4.8 | 76.7 | -11.3 | 88.0 | 83.8 | -4.2 | 76.2 | -11.8 | 90.1 | 86.5 | -3.6 | 78.5 | -11.6 |
| yor | 86.4 | 78.2 | -8.2 | 67.0 | -19.4 | 86.8 | 79.2 | -7.6 | 74.4 | -12.4 | 90.2 | 85.0 | -5.2 | 74.0 | -16.2 |
| zul | 86.4 | 83.2 | -3.2 | 69.5 | -16.9 | 89.4 | 86.1 | -3.3 | 68.8 | -20.6 | 90.1 | 87.5 | -2.6 | 67.1 | -23.0 |
| avg | 85.5 | 76.2 | -9.3 | 70.8 | -14.7 | 86.4 | 78.0 | -8.4 | 70.9 | -15.5 | 87.5 | 79.9 | -7.6 | 72.2 | -15.3 |

Table 8.7: F1 score for two varieties of hard-to-identify entities: zero-frequency entities that do not appear in the training corpus, and longer entities of four or more words.

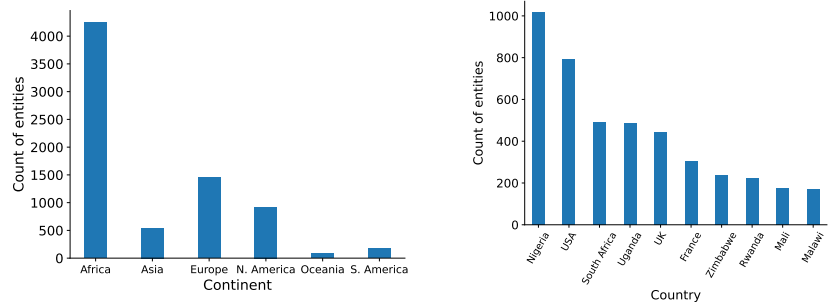

(a) Number of entities per continent  (b) Top-10 countries

Figure 8.1: **Number of entities per continent and the top-10 countries with the largest number of entities**

#### 8.5.3.2 *Dataset Geography of Entities*

Next, we analyse the geographical representativeness of the entities in our dataset, specifically, we measure the count of entities based on the countries they originate from. Following the approach of Faisal, Wang, and Anastasopoulos (2022), we first performed entity linking of named entities present in our dataset to Wikidata IDs using mGenre (De Cao et al., 2022), followed by mapping Wikidata IDs to countries.

| Language | XLM-R-large | | | | mDeBERTaV3-base | | | | AfroXLMR-large | | | |
|---|---|---|---|---|---|---|---|---|---|---|---|---|
| | DATE | LOC | ORG | PER | DATE | LOC | ORG | PER | DATE | LOC | ORG | PER |
| bam | 90.3 | 83.2 | 80.7 | 87.1 | 90.1 | 86.4 | 79.2 | 88.4 | 92.6 | 87.7 | 82.4 | 86.1 |
| bbj | 87.6 | 82.9 | 79.4 | 83.6 | 79.9 | 86.4 | 72.5 | 87.2 | 85.7 | 87.0 | 75.2 | 84.7 |
| ewe | 91.8 | 96.8 | 85.5 | 95.9 | 91.8 | 96.4 | 88.6 | 97.1 | 92.0 | 97.8 | 85.6 | 98.6 |
| fon | 85.4 | 89.2 | 86.9 | 94.6 | 86.8 | 93.3 | 89.3 | 94.3 | 85.9 | 91.9 | 86.4 | 94.6 |
| hau | 86.8 | 90.0 | 92.5 | 98.0 | 86.4 | 89.2 | 89.1 | 98.0 | 87.4 | 91 | 92.2 | 98.2 |
| ibo | 84.5 | 91.6 | 83.5 | 97.7 | 85.4 | 95.6 | 82.5 | 99.1 | 87.2 | 96.5 | 73.4 | 98.8 |
| kin | 88.4 | 92.7 | 84.0 | 94.8 | 87.4 | 95.0 | 87.8 | 97.7 | 88.1 | 95.6 | 89.1 | 99.1 |
| lug | 78.2 | 93.1 | 94.2 | 95.8 | 80.2 | 95.1 | 94.3 | 96.0 | 81.7 | 93.1 | 95.1 | 97.3 |
| mos | 80.3 | 92.7 | 74.4 | 93.1 | 81.6 | 92.1 | 78.9 | 88.3 | 83.2 | 93.7 | 75.4 | 88.9 |
| pcm | 96.6 | 91.1 | 89.7 | 96.9 | 96.1 | 93.1 | 90.9 | 97.3 | 95.6 | 92.4 | 90.9 | 97.1 |
| nya | 89.1 | 94.1 | 94.2 | 94.4 | 89.6 | 96.7 | 96.0 | 94.9 | 89.1 | 96.2 | 94.8 | 95.6 |
| sna | 95.6 | 95.6 | 96.1 | 98.1 | 96.0 | 95.1 | 96.5 | 98.7 | 96.6 | 95.4 | 97.4 | 99.3 |
| swa | 92.2 | 97.0 | 95.2 | 98.8 | 91.5 | 96.9 | 94.6 | 98.8 | 91.5 | 97.4 | 93.7 | 98.2 |
| tsn | 88.1 | 88.3 | 89.1 | 97.1 | 87.8 | 90.0 | 89.0 | 97.6 | 90.5 | 94.8 | 92.2 | 98.6 |
| twi | 66.7 | 89.3 | 79.4 | 96.1 | 76.5 | 90.4 | 82.9 | 97.5 | 75.7 | 91.4 | 85.1 | 97.7 |
| wol | 80.6 | 84.9 | 87.0 | 95.9 | 80.8 | 88.2 | 88.4 | 95.0 | 82.6 | 91.9 | 88.0 | 97.0 |
| xho | 90.7 | 91.6 | 93.1 | 96.9 | 89.7 | 92.0 | 93.4 | 98.1 | 91.1 | 93.5 | 95.0 | 98.3 |
| yor | 89.6 | 94.0 | 90.3 | 93.6 | 89.6 | 92.1 | 91.4 | 94.6 | 91.3 | 95.8 | 92.5 | 96.4 |
| zul | 85.0 | 90.1 | 87.8 | 97.1 | 92.2 | 95.5 | 88.1 | 97.1 | 90.8 | 96.2 | 91.8 | 97.2 |
| avg | 86.7 | 91.0 | 87.5 | 95.0 | 87.3 | 92.6 | 88.1 | 95.6 | 88.4 | 93.7 | 88.2 | 95.9 |

Table 8.8: F1 score for the different entity types.

Figure 8.1 shows the result of number of entities per continent and the top-10 countries with the largest representation of entities. Over 50% of the entities are from Africa, followed by Europe. This shows that the entities of MasakhaNER 2.0 properly represent the African continent. Seven out of the top-10 countries are from Africa, but also includes USA, United Kingdom and France.

### 8.5.4 *Transfer Between African NER Datasets*

African languages have a diverse set of linguistic characteristics. To demonstrate this heterogeneity, we perform a transfer learning experiment where we compare the performance of multilingual NER models jointly trained on the languages of MasakhaNER 1.0 or MasakhaNER 2.0 and perform zero-shot evaluation on both test sets. We consider three experimental settings:

(a) Train on all languages in MasakhaNER 1.0 using MasakhaNER 1.0 training data.

| | Lang. in M1.0 | Eval. on MasakhaNER 2.0 test set | | | Eval. on MasakhaNER 1.0 test set | | |
|---|---|---|---|---|---|---|---|
| | | (a) (M1.0,M1.0) | (b) (M1.0,M2.0) | (c) (M2.0,M2.0) | (a) (M1.0,M1.0) | (b) (M1.0,M2.0) | (c) (M2.0,M2.0) |
| bam | ✗ | 52.2 | 50.9 | **82.3** | - | - | - |
| bbj | ✗ | 48.4 | 49.8 | **75.5** | - | - | - |
| ewe | ✗ | 78.3 | 76.2 | **89.5** | - | - | - |
| fon | ✗ | 52.9 | 57.1 | **83.2** | - | - | - |
| hau | ✓ | 76.9 | **88.7** | 87.7 | 92.1 | 80.8 | 80.4 |
| ibo | ✓ | 86.0 | 90.1 | **92.3** | 89.2 | 84.6 | 84.3 |
| kin | ✓ | 77.6 | 87.6 | **87.2** | 79.1 | 77.7 | 77.0 |
| lug | ✓ | 83.2 | **90.0** | 89.1 | 86.0 | 79.0 | 79.8 |
| luo | ✓ | 68.6 | **82.7** | 81.8 | 86.0 | 79.0 | 79.8 |
| mos | ✗ | 55.0 | 49.6 | **75.3** | - | - | - |
| nya | ✗ | 82.1 | 80.4 | **92.2** | - | - | - |
| pcm | ✓ | 86.7 | **90.2** | 89.9 | 91.2 | 88.0 | 87.9 |
| sna | ✗ | 49.6 | 42.5 | **95.9** | - | - | - |
| swa | ✓ | 89.4 | **93.1** | 93.1 | 89.5 | 86.3 | 86.5 |
| tsn | ✗ | 80.0 | 79.4 | **89.5** | - | - | - |
| twi | ✗ | 56.6 | 57.3 | **78.8** | - | - | - |
| wol | ✓ | 73.6 | **87.0** | 86.4 | 70.8 | 71.6 | 72.1 |
| xho | ✗ | 56.9 | 47.4 | **89.7** | - | - | - |
| yor | ✓ | 69.4 | **89.7** | 89.1 | 85.0 | 85.0 | 84.8 |
| zul | ✗ | 69.9 | 64.3 | **90.7** | - | - | - |
| AVG | | 69.7$_{\pm 0.6}$ | 72.7$_{\pm 0.6}$ | 87.0$_{\pm 1.2}$ | 84.8$_{\pm 0.3}$ | 80.0$_{\pm 0.3}$ | 80.1$_{\pm 0.8}$ |

Table 8.9: **Multilingual evaluation on African NER datasets**. We compare the performance of AfroXLM-R-large trained on languages of MasakhaNER 2.0 (M2.0 for short) and MasakhaNER (M1.0 for short) and evaluated both on the same and on the other dataset. The first column indicate the languages used for training (the 10 languages from MasakhaNER (M1.0 for short) or the 20 languages from MasakhaNER 2.0 (M2.0 for short)). The second column indicates the training data. Average is over 5 runs.

(b) Train on the languages in MasakhaNER 1.0 (excl. "amh") using the MasakhaNER 2.0 training data.

(c) Train on all languages in MasakhaNER 2.0 using MasakhaNER 2.0 training data.

Table 8.9 shows the result of the three settings. When evaluating on the MasakhaNER 2.0 test set in setting (a), the performance is mostly high (> 65 F1) for languages in MasakhaNER 1.0. Most of the languages that are not in MasakhaNER 1.0 have worse zero-shot performance, typically between $48 - 60$ F1 except for ewe, nya, tsn, and zul with over 69 F1. Making use of a larger dataset, i.e., setting (b) from MasakhaNER 2.0 only provides a small improvement (+3 F1). The evaluation on setting (c) shows a large gap of about 15 F1 and 17 F1 compared to settings (b) and (a) on the MasakhaNER 2.0 test set respectively, especially for Southern Bantu languages like sna and xho. On the MasakhaNER 1.0 test set, training on the in-distribution MasakhaNER 1.0 languages and training set achieves the best performance. However, the performance gap compared to training on the MasakhaNER 2.0 data is much smaller. Overall, these results demonstrate the need to create large benchmark datasets (like

MasakhaNER 2.0) covering diverse languages with different linguistic characteristics, particularly for the African continent.

## 8.6 CROSS-LINGUAL TRANSFER

The success of cross-lingual transfer either in zero or few-shot settings depends on several factors including an appropriate selection of the best source language. Several attempts at cross-lingual transfer make use of English as the source language due to its availability of training data. However, English is unrepresentative of African languages and transfer performance is often lower for distant languages (Adelani et al., 2021b).

### 8.6.1 *Choosing Transfer Languages for NER*

Here, we follow the approach of Lin et al. (2019), LangRank, that uses source-target transfer evaluation scores and data-dependent features such as dataset size and entity overlap, and six different linguistic distance measures based on lang2vec (Littell et al., 2017) such as geographic distance ($d_{geo}$), genetic distance ($d_{gen}$), inventory distance ($d_{inv}$), syntactic distance ($d_{syn}$), phonological distance ($d_{pho}$), and featural distance ($d_{fea}$). We provide definitions of the features in Section 8.6.2. LangRank is trained using these features to determine the best transfer language in a leave-one-out setting where for each target language, we train on all other languages except the target language. We compute transfer F1 scores from a set of $N$ transfer (source) languages and evaluate on $N$ target languages, yielding $N \times N$ transfer scores.

CHOICE OF TRANSFER LANGUAGES  We selected 22 human-annotated NER datasets of diverse languages by searching the web and HuggingFace Dataset Hub (Lhoest et al., 2021). We required each dataset to contain at least the PER, ORG, and LOC types, and we limit our analysis to these types. We also added our MasakhaNER 2.0 dataset with 20 languages. In total, the datasets cover 42 languages (21 African). Each language is associated with a single dataset. Section 8.4.4 provides details about the languages, datasets, and data splits. To compute zero-shot transfer scores, we fine-tune mDeBERTaV3 on the NER dataset of a source language and perform zero-shot transfer to the target languages. We choose mDeBERTaV3 because it supports 100 languages and has the best performance among the PLMs trained on a similar number of languages.

### 8.6.2 *LangRank Feature Descriptions*

The following definitions are listed here, originally from Lin et al. (2019).

GEOGRAPHIC DISTANCE ($d_{geo}$) based on the orthodromic distance between language locations obtained from Glottolog (Hammarström, Forkel, and Haspelmath, 2018).

GENETIC DISTANCE ($d_{gen}$) based on the genealogical distance of Glottolog language tree.

INVENTORY DISTANCE ($d_{inv}$) based on the cosine distance between phonological feature vectors obtained from the PHOIBLE (a collection of 7 phonological databases) database (Moran, McCloy, and Wright, 2014).

SYNTACTIC DISTANCE ($d_{syn}$) based on cosine distance between feature vectors obtained from syntactic structures derived from WALS database (Dryer and Haspelmath, 2013).

PHONOLOGICAL DISTANCE ($d_{pho}$) based on the cosine distance between phonological feature vectors obtained from WALS and Ethnologue databases (Lewis, 2009)

FEATURAL DISTANCE ($d_{fea}$) based on the cosine distance between feature vectors combining all 5 features mentioned above.

TRANSFER LANGUAGE DATASET SIZE ($s_{tf}$) The size of the transfer language's dataset

TARGET LANGUAGE DATASET SIZE ($s_{tg}$) The size of the target language's dataset.

TRANSFER OVER TARGET SIZE RATIO ($sr$) The size of the transfer language's dataset divided by the size of the target language's dataset.

ENTITY OVERLAP ($eo$) The number of unique words that overlap between the source and target languages' training datasets.

### 8.6.3 *Single-source Transfer Results*

Figure 8.2 shows the zero-shot evaluation of training on 42 NER datasets and evaluation on the test sets of the 20 MasakhaNER 2.0 languages. On average, we find the transfer from non-African languages to be slightly worse (51.7 F1) than transfer from African languages (57.3 F1). The worst transfer result is using `bbj` as source language (41.0 F1) while the best is using `sna` (64 F1), followed by `yor` (63 F1).

We identify German (`deu`) and Finnish (`fin`) as the top-2 transfer languages among the non-African languages. In most cases, languages that are geographically and syntactically close tend to benefit most from each other. For example, `sna`, `xho`, and `zul` have very good transfer among themselves due to both syntactic and geographical closeness. Similarly, for Nigerian languages (`hau`, `ibo`, `pcm`, `yor`) and

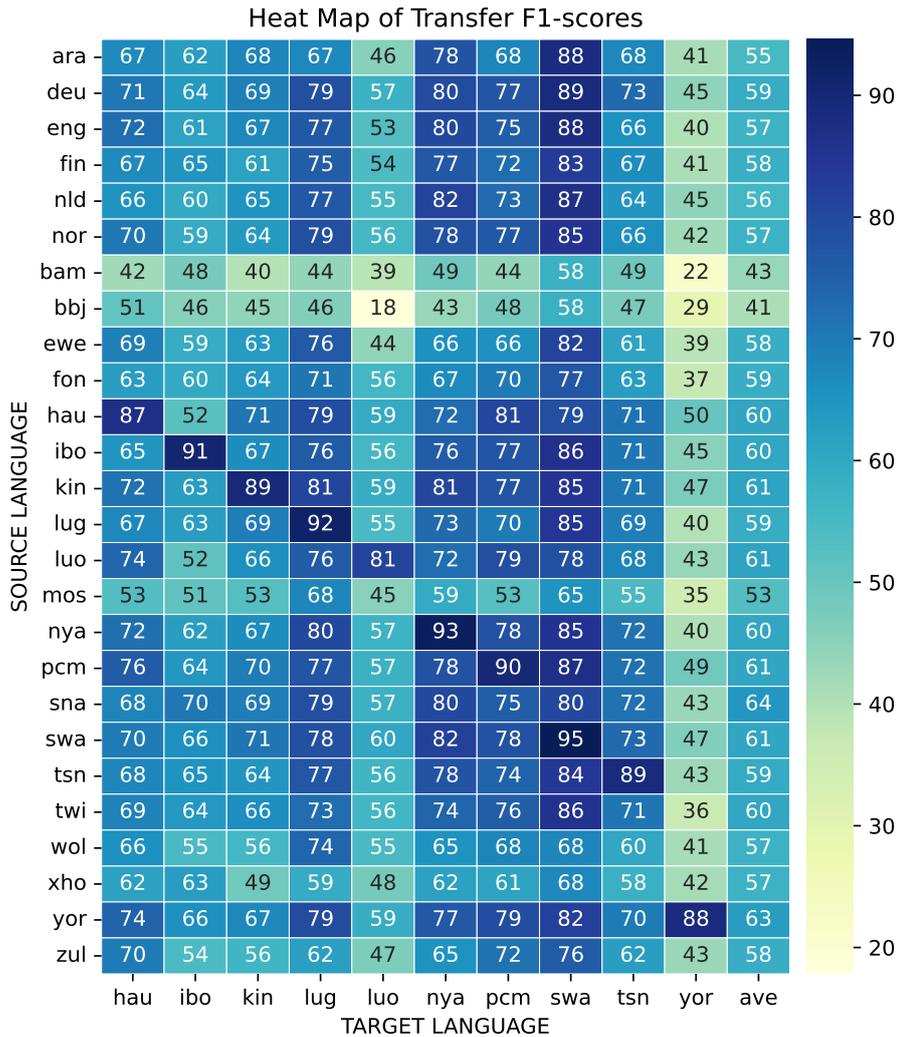

Figure 8.2: **Zero-shot Transfer** from several source languages to African languages for 10 languages in MasakhaNER 2.0 and the average (ave) over all 20 languages. Figure 8.5 shows results for each of the 20 languages.

East African languages (`kin`, `lug`, `luo`, `swa`), geographical proximity plays an important role. While most African languages prefer transfer from another African language, there are few exceptions like `swa` preferring transfer from `deu` or `ara`. The latter can be explained by the presence of Arabic loanwords in Kiswahili (Versteegh, 2001). Similarly, `nya` and `tsn` also prefer `deu`. Figure 8.6 provides results for transfer to non-African languages. We find that for non-African languages, English appears to be the best transfer on average which is not the case for African languages. The reason for this is because many of the non-African languages we evaluated on are from the Indo-European, similar to English.

### 8.6.4 *LangRank and Co-training Results*

| Target Lang. | Top-2 Transf. Lang | Top-2 LangRank Model | Top-3 features selected by LangRank model Lang 1; Lang 2 |
|---|---|---|---|
| **amh** | zho, ara | pcm, ewe | $(s_{tf}, sr, s_{tg}); (s_{tf}, s_{tg}, sr)$ |
| **bam** | twi, fon | wol, fon | $(d_{geo}, d_{inv}, sr); (d_{geo}, sr, d_{pho})$ |
| **bbj** | fon, ewe | twi, ewe | $(s_{tf}, d_{syn}, d_{geo}); (s_{tf}, d_{geo}, sr)$ |
| **ewe** | swa, twi | pcm, swa | $(d_{geo}, s_{tf}, sr); (eo, d_{geo}, s_{tf})$ |
| **fon** | mos, bbj | yor, ewe | $(d_{geo}, d_{syn}, sr); (s_{tf}, d_{geo}, d_{gen})$ |
| **hau** | pcm, yor | yor, swa | $(d_{geo}, sr, eo); (eo, sr, s_{tf})$ |
| **ibo** | sna, yor | pcm, kin | $(eo, d_{geo}, s_{tf}); (d_{geo}, sr, eo)$ |
| **kin** | hau, swa | sna, yor | $(eo, d_{geo}, s_{tf}); (eo, s_{tf}, sr)$ |
| **lug** | kin, nya | luo, zul | $(d_{geo}, sr, eo); (d_{syn}, d_{geo}, sr)$ |
| **luo** | swa, hau | lug, sna | $(d_{geo}, sr, eo); (d_{geo}, eo, sr)$ |
| **mos** | fon, ewe | yor, fon | $(d_{geo}, d_{inv}, sr); (d_{geo}, s_{tf}, sr)$ |
| **nya** | swa, nld | zul, sna | $(eo, d_{geo}, sr); (d_{geo}, eo, d_{syn})$ |
| **pcm** | hau, yor | eng, yor | $(eo, d_{gen}, d_{syn}); (eo, d_{geo}, sr)$ |
| **sna** | zul, xho | swa, zul | $(eo, sr, s_{tf}); (d_{geo}, sr, eo)$ |
| **swa** | deu, ara | ita, nld | $(sr, d_{inv}, eo); (eo, s_{tf}, sr)$ |
| **tsn** | deu, swa | swa, nya | $(eo, d_{inv}, s_{tf}); (d_{inv}, d_{geo}, d_{gen})$ |
| **twi** | swa, nya | swa, ewe | $(eo, s_{tf}, d_{geo}); (d_{geo}, s_{tf}, sr)$ |
| **wol** | fon, mos | fon, yor | $(d_{geo}, sr, s_{tf}); (sr, d_{geo}, d_{syn})$ |
| **xho** | zul, sna | zul, pcm | $(eo, d_{geo}, d_{gen}); (eo, s_{tf}, d_{inv})$ |
| **yor** | hau, pcm | fon, pcm | $(d_{geo}, d_{inv}, d_{syn}); (eo, d_{geo}, d_{inv})$ |
| **zul** | xho, sna | xho, sna | $(eo, d_{gen}, d_{geo}); (d_{syn}, sr, d_{geo})$ |

Table 8.10: **Best Transfer Languages for languages in MasakhaNER 2.0**. The ranking model features are based on the definitions in (Lin et al., 2019) like: geographic distance ($d_{geo}$), genetic distance ($d_{gen}$), inventory distance ($d_{inv}$), syntactic distance ($d_{syn}$), phonological distance ($d_{pho}$), transfer language dataset size ($s_{tf}$), transfer over target size ratio ($sr$), and entity overlap ($eo$).

We also investigate the benefit of training on the second-best language in addition to the languages selected by LangRank. We jointly train on the combined data of the top-2 transfer languages or the top-2 languages predicted by LangRank and evaluate their zero-shot performance on the target language. Table 8.10 shows the result for the top-2 transfer languages using the best from 42 × 42 transfer F1-scores and LangRank model predictions. LangRank predicted the right language as one of the top-2 best transfer language in 13 target languages. The target languages with incorrect predictions are fon, ibo, kin, lug, luo, nya, and swa. The transfer languages predicted as alternative are often in the top-5 transfer languages or are less than (−5 F1) worse than the best transfer language. For example, the best transfer language for lug is kin (81 F1) but LangRank predicted luo (76 F1). Table 8.11 gives the top-2 transfer languages for non-African languages.

| Target Lang. | Top-2 Transf. Lang | Top-2 LangRank Model | Top-3 features selected by LangRank model Lang 1; Lang 2 |
|---|---|---|---|
| **ara** | eng, deu | fas, pcm | $(eo, d_{inv}, d_{syn}); (d_{syn}, sr, d_{inv})$ |
| **dan** | nor, fin | swe, nor | $(eo, d_{gen}, d_{geo}); (eo, d_{geo}, d_{syn})$ |
| **deu** | nld, eng | dan, nld | $(d_{geo}, eo, s_{tf}, d_{syn}); (eo, d_{syn}, d_{geo})$ |
| **eng** | pcm, swe | nld, pcm | $(eo, d_{geo}, d_{syn}); (eo, d_{gen} d_{pho})$ |
| **fas** | hau, pcm | ara, eng | $(d_{syn}, d_{inv}, eo); (d_{syn}, d_{geo}, s_{tf})$ |
| **fin** | dan, eng | deu, eng | $(eo, s_{tf}, d_{geo}); (d_{syn}, d_{geo}, eo)$ |
| **fra** | swe, swa | nld, deu | $(eo, d_{syn}, d_{geo}); (d_{geo}, eo, sr)$ |
| **hun** | ukr, eng | deu, ron | $(d_{geo}, d_{syn}, eo); (d_{geo}, eo, d_{syn})$ |
| **ind** | lug, luo | zho, nld | $(s_{tg}, s_{tf}, sr); (d_{syn}, s_{tf}, eo)$ |
| **ita** | deu, spa | nld, eng | $(d_{syn}, eo, d_{geo}); (eo, d_{syn}, d_{geo})$ |
| **kor** | zho, ind | ara, nep | $(sr, s_{tf}, d_{syn}); (d_{inv}, d_{syn}, s_{tf})$ |
| **lav** | fin, dan | eng, nld | $(s_{tf}, d_{syn}, sr); (s_{tf}, d_{syn}, d_{geo})$ |
| **nep** | pcm, swa | kor, zho | $(d_{syn}, s_{tf}, d_{pho}); (s_{tf}, sr, d_{geo})$ |
| **nld** | eng, deu | eng, nor | $(eo, d_{geo}, d_{syn}); (eo, d_{geo}, s_{tf})$ |
| **nor** | dan, deu | dan, eng | $(eo, d_{geo}, s_{tf}); (eo, d_{geo}, sr)$ |
| **por** | es, nld | spa, eng | $(eo, d_{syn}, d_{gen}); (eo, d_{syn}, d_{geo})$ |
| **ron** | lav, eng | eng, ita | $(eo, d_{syn}, d_{geo}); (eo, d_{geo}, d_{syn})$ |
| **spa** | eng, por | por, lav | $(eo, d_{geo}, d_{syn}); (d_{syn}, eo, d_{geo})$ |
| **swe** | dan, nor | nor, nld | $(eo, d_{syn}, d_{geo}); (d_{syn}, d_{geo}, eo)$ |
| **ukr** | nor, eng | deu, eng | $(d_{geo}, d_{syn}, sr); (d_{syn}, d_{geo}, s_{tf})$ |
| **zho** | lav, amh | pcm, deu | $(d_{syn}, s_{tf}, s_{geo}); (d_{syn}, s_{tf}, d_{pho})$ |

Table 8.11: **Best Transfer Languages Non-African languages**. The ranking model features are based on the definitions in (Lin et al., 2019) like: geographic distance ($d_{geo}$), genetic distance ($d_{gen}$), inventory distance ($d_{inv}$), syntactic distance ($d_{syn}$), phonological distance ($d_{pho}$), transfer language dataset size ($s_{tf}$), transfer over target size ratio ($sr$), and entity overlap ($eo$).

FEATURES THAT ARE IMPORTANT FOR TRANSFER    The most important features for the selection of best language by LangRank are geographic distance ($d_{geo}$) and entity overlap ($eo$). The $d_{geo}$ is influential because named entities (e.g. name of a politician or a city) are often similar from languages spoken in the same country (e.g Nigeria with 4 languages in MasakhaNER 2.0) or region (e.g. East African languages). Similarly, we find entity overlap to have a positive Spearman correlation ($R = 0.6$) to transfer F1-score. Figure 8.3 shows the word overlap between different languages, and how they correlates with the transfer performance (F1-scores). $d_{geo}$ occurred as part of the top-3 features for 15 best transfer language and 16 second best languages. Similarly, for $eo$, it appeared 11–13 times for the top-2 transfer languages. Interestingly, dataset size was not among the most important features, highlighting the need for typologically diverse training data.

BEST TRANSFER LANGUAGE OUTPERFORMS ENGLISH    In Table 8.12, we compare the zero-shot transfer performance of the top-2 transfer

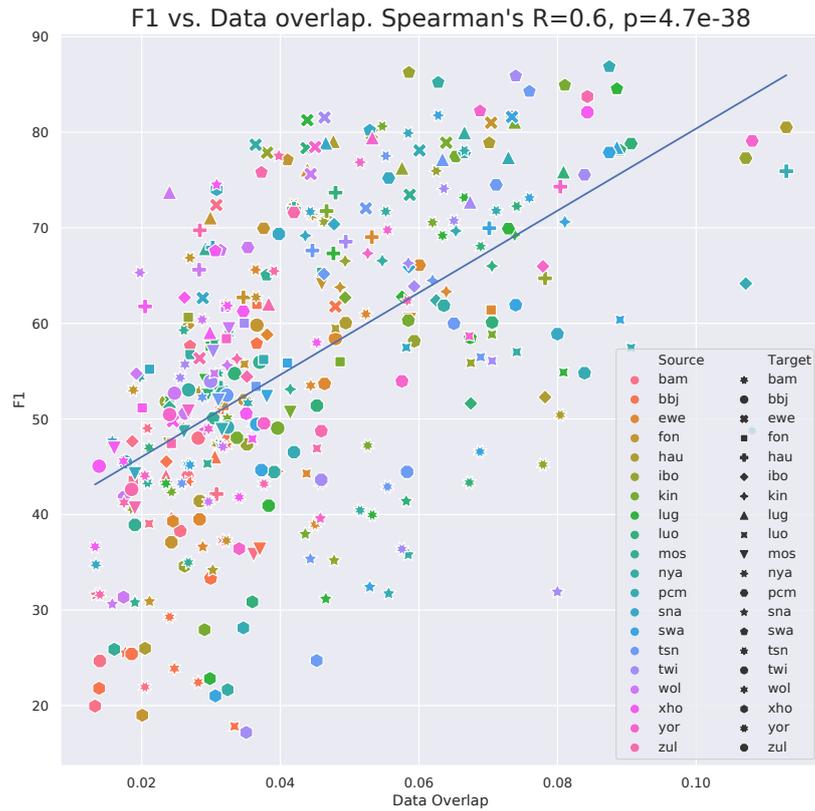

Figure 8.3: The correlation between the data overlap and F1 transfer performance. For source language $X$ and target language $Y$, denote the set of unique named entities (PER, ORG, LOC, DATE) by $T_X$ and $T_Y$ respectively. The overlap here was calculated as $\frac{|T_X \cap T_Y|}{|T_X|+|T_Y|}$, as in Lin et al. (2019).

| Target Lang. | Top-2 Transf. Lang | Top-2 LangRank Model | Target Lang. F1 | Top-1 LangRank Lang. F1 | Top-2 LangRank Lang. F1 | Top-2 Transf. Lang. F1 | Best Transf. F1 | Second Best Transf. F1 | eng Tranf. F1 |
|---|---|---|---|---|---|---|---|---|---|
| **amh** | zho, ara | pcm, ewe | 75.0 | 42.5 | 41.5 | **65.8** | 61.0 | 55.9 | 40.6 |
| **bam** | twi, fon | wol, fon | 80.4 | 47.1 | 52.8 | **55.1** | 54.3 | 53.0 | 38.4 |
| **bbj** | fon, ewe | twi, ewe | 72.9 | 53.9 | 58.8 | **60.1** | 59.8 | 58.4 | 45.8 |
| **ewe** | swa, twi | pcm, swa | 91.7 | 78.1 | 81.1 | **83.9** | 81.6 | 81.5 | 76.4 |
| **fon** | mos, bbj | yor, ewe | 84.9 | 58.4 | 64.9 | **69.9** | 65.4 | 62.0 | 50.6 |
| **hau** | pcm, yor | yor, swa | 86.9 | 74.3 | 74.8 | **77.4** | 75.9 | 74.3 | 72.4 |
| **ibo** | sna, yor | pcm, kin | 91.0 | 64.2 | 63.9 | **77.1** | 70.4 | 66.0 | 61.4 |
| **kin** | hau, swa | sna, yor | 89.5 | 69.2 | 71.8 | **74.0** | 71.1 | 70.6 | 67.4 |
| **lug** | kin, nya | luo, zul | 91.5 | 75.9 | 78.1 | **82.1** | 81.1 | 80.0 | 76.5 |
| **luo** | swa, hau | lug, sna | 81.2 | 54.9 | 61.6 | **61.1** | 60.4 | 59.5 | 53.4 |
| **mos** | fon, ewe | yor, fon | 78.9 | 50.8 | 62.5 | **65.6** | 64.2 | 60.4 | 45.4 |
| **nya** | swa, nld | zul, sna | 93.5 | 65.5 | 81.5 | **81.8** | 81.8 | 81.7 | 80.1 |
| **pcm** | hau, yor | eng, yor | 89.9 | 75.5 | 79.9 | **81.8** | 80.5 | 79.1 | 75.5 |
| **sna** | zul, xho | swa, zul | 96.0 | 32.4 | 80.0 | **80.0** | 77.5 | 74.5 | 37.1 |
| **swa** | deu, ara | ita, nld | 94.6 | 84.5 | 86.0 | **89.6** | 88.7 | 88.1 | 87.9 |
| **tsn** | deu, swa | swa, nya | 88.7 | 73.1 | 73.4 | **74.0** | 73.3 | 73.1 | 65.8 |
| **twi** | swa, nya | swa, ewe | 82.0 | 61.9 | 57.2 | **64.3** | 61.0 | 61.9 | 49.5 |
| **wol** | fon, mos | fon, yor | 85.2 | 62.0 | 59.4 | **63.0** | 62.0 | 58.9 | 44.8 |
| **xho** | zul, sna | zul, pcm | 90.8 | 83.7 | 83.0 | **84.3** | 83.7 | 74.0 | 24.5 |
| **yor** | hau, pcm | fon, pcm | 88.3 | 37.3 | 43.2 | **50.3** | 50.3 | 48.8 | 40.4 |
| **zul** | xho, sna | xho, sna | 88.6 | 82.1 | 85.5 | **85.5** | 82.1 | 69.4 | 44.7 |
| AVG | – |  | 86.7 | 63.2 | 68.6 | 72.7 | 70.8 | 68.2 | 56.1 |

Table 8.12: **Transfer Results for African languages** The best zero-shot result is **bolded**. The languages highlighted in gray have very good transfer performance (> 70%) using the best transfer language.

languages to using eng as the transfer language. They significantly outperform the eng average of 56.9 by +14 and +12 F1 for the first and second-best source language respectively when evaluating on African languages. Table 8.13 provides the zero-shot evaluation for non-African languages.

CO-TRAINING OF TOP-2 TRANSFER LANGUAGES IMPROVES PERFORMANCE Table 8.12 shows the result of co-training the top-2 transfer languages. For African languages, we find that co-training further improves zero-shot performance over the best transfer by around +3 F1. It is most significant for fon, ibo, kin and twi with 3–7 F1 improvement. Co-training the top-2 transfer languages predicted by LangRank is better than using the second-best transfer language but often performs worse than the best transfer language.

### 8.6.5 Sample Efficiency Results

Figure 8.4 shows the performance when the model is trained on a small number of target language samples compared to when the best transfer language is used prior to fine-tuning on the same number of target language samples. We show the results for four languages (which reflect common patterns across all languages) and an average

| Target Lang. | Top-2 Transf. Lang | Top-2 LangRank Model | Target Lang. F1 | Best Transf. F1 | Second Best Transf. F1 | eng Tranf. F1 | LangRank First Lang F1 | LangRank Second Lang F1 |
|---|---|---|---|---|---|---|---|---|
| ara | eng, deu | fas, hau | 82.8 | **71.5** | 69.9 | **71.5** | 55.7 | 57.9 |
| dan | nor, fin | swe, nor | 87.1 | **86.3** | 85.6 | 83.1 | 82.8 | 86.3 |
| deu | nld, eng | dan, nld | 86.5 | **79.3** | 78.8 | 78.8 | 79.3 | 79.3 |
| eng | pcm, swe | nld, pcm | 93.5 | 81.3 | 79.7 | **93.5** | 76.0 | 81.3 |
| fas | hau, pcm | ara, eng | 84.8 | **64.8** | 63.4 | 59.3 | 57.9 | 59.2 |
| fin | dan, eng | deu, eng | 93.4 | **83.7** | 83.6 | 83.6 | 80.8 | 83.6 |
| fra | swe, swa | nld, deu | 75.5 | **66.3** | 65.4 | 60.6 | 63.3 | 64.9 |
| hun | ukr, eng | deu, ron | 98.0 | **70.7** | 68.4 | 68.4 | 63.6 | 43.8 |
| ind | lug, luo | zho, nld | 93.7 | **85.9** | 85.2 | 83.9 | 78.6 | 84.1 |
| ita | deu, spa | nld, eng | 86.7 | **79.1** | 78.2 | 77.0 | 77.1 | 77.1 |
| kor | zho, deu | ara, nep | 85.7 | **31.1** | 21.5 | 12.7 | 21.3 | 11.9 |
| lav | fin, dan | eng, nld | 89.7 | **80.4** | 80.1 | 73.5 | 73.5 | 69.5 |
| nep | pcm, swa | kor, zho | 89.5 | **79.0** | 77.7 | 73.4 | 68.2 | 68.5 |
| nld | eng, deu | eng, nor | 93.4 | **85.4** | 83.7 | 85.4 | **85.4** | 79.9 |
| nor | dan, deu | dan, eng | 92.5 | **89.8** | 87.8 | 87.3 | 89.8 | 87.2 |
| por | es, nld | spa, eng | 75.0 | **77.8** | 73.5 | 72.0 | 77.8 | 72.0 |
| ron | lav, eng | eng, ita | 89.6 | **59.6** | 59.5 | 59.5 | 59.5 | 57.8 |
| spa | eng, por | por, lav | 89.6 | **83.9** | 83.6 | 83.9 | 83.6 | 77.3 |
| swe | dan, nor | dan, nld | 90.3 | **89.4** | 89.1 | 88.1 | 89.3 | 85.2 |
| ukr | nor, eng | deu, eng | 92.6 | **87.2** | 85.6 | 85.6 | 81.5 | 85.6 |
| zho | lav, amh | pcm, deu | 91.4 | **60.2** | 58.3 | 54.7 | 54.7 | 48.9 |
| AVG | – | | 88.6 | **75.8** | 74.2 | 73.2 | 71.4 | 69.6 |

Table 8.13: **Transfer Results for Non-African languages.** The best zero-shot result is **bolded**. The languages highlighted in gray have very good transfer performance (> 70%) using the best transfer language.

(ave) over the 20 languages. As seen in the figure, models achieve less than 50 F1 when we train on 100 sentences and over 75 F1 when training on 500 sentences. In practice, annotating 100 sentences takes about 30 minutes while annotating 500 sentences takes around 2h 30 minutes; therefore, slightly more annotation effort can yield a substantial quality improvement. We also find that using the best transfer language in zero-shot settings gives a performance very close to annotating 500 samples in most cases, showing the importance of transfer language selection. By additionally fine-tuning the model on 100 or 500 target language samples, we can further improve the NER performance. Figure 8.7 provides the sample efficiency results for individual languages.

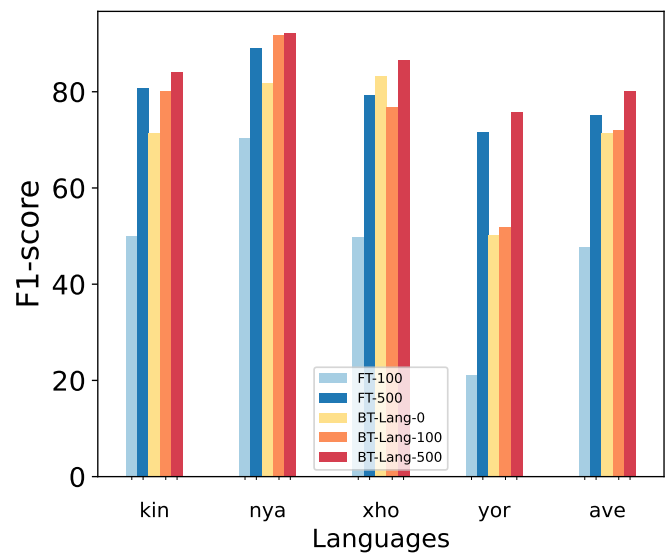

Figure 8.4: **Sample Efficiency Results** for 100 and 500 samples in the target language, model fine-tuned from a PLM (e.g. FT-100 – trained on 100 samples from the target language) or fine-tuned from the best transfer language NER model (e.g BT-Lang-0 – trained on 0 samples from the target language or zero-shot)

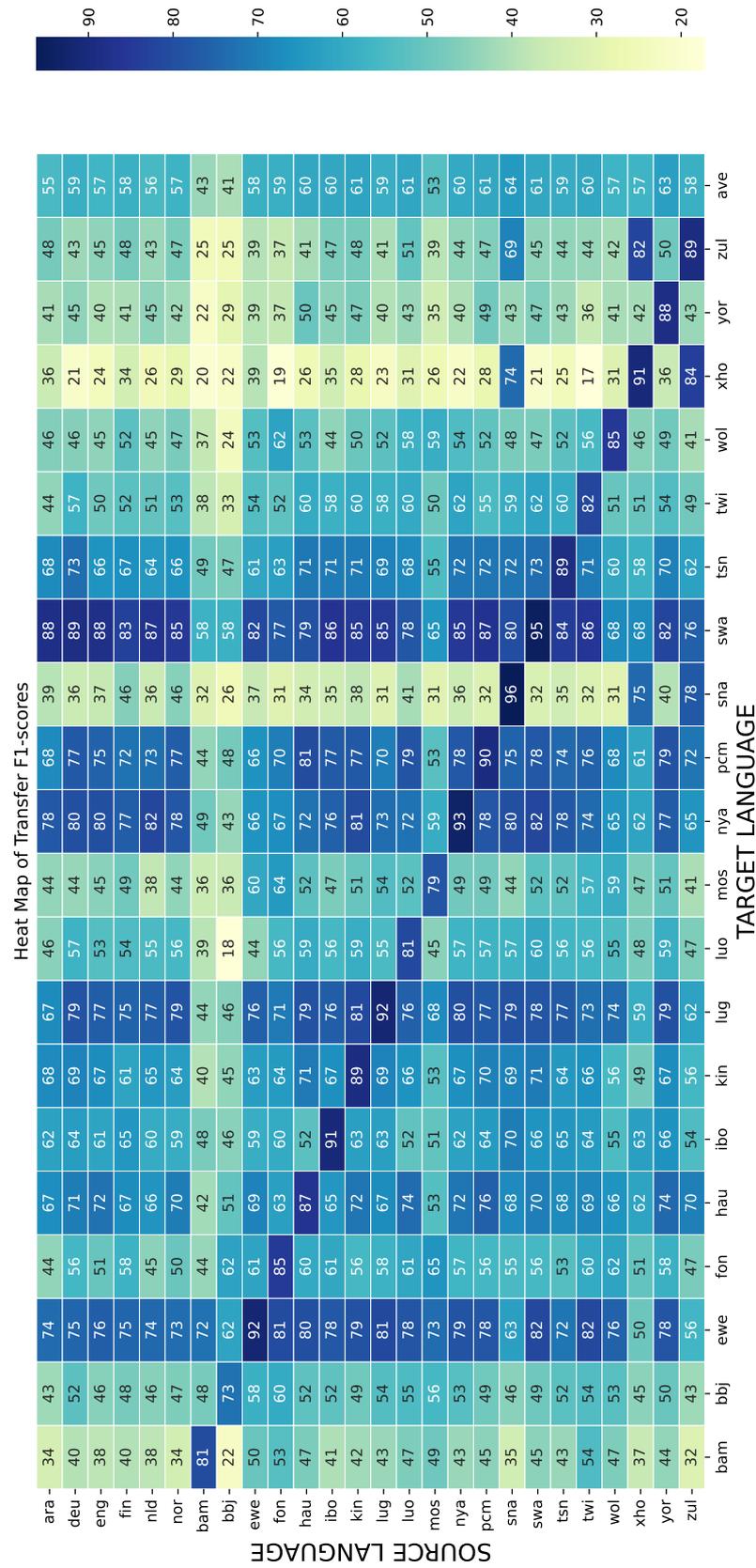

Figure 8.5: **Zero-shot Transfer** from several source languages to African languages in MasakhaNER 2.0. The languages highlighted in gray have very good transfer performance (> 70%) using the best transfer language.

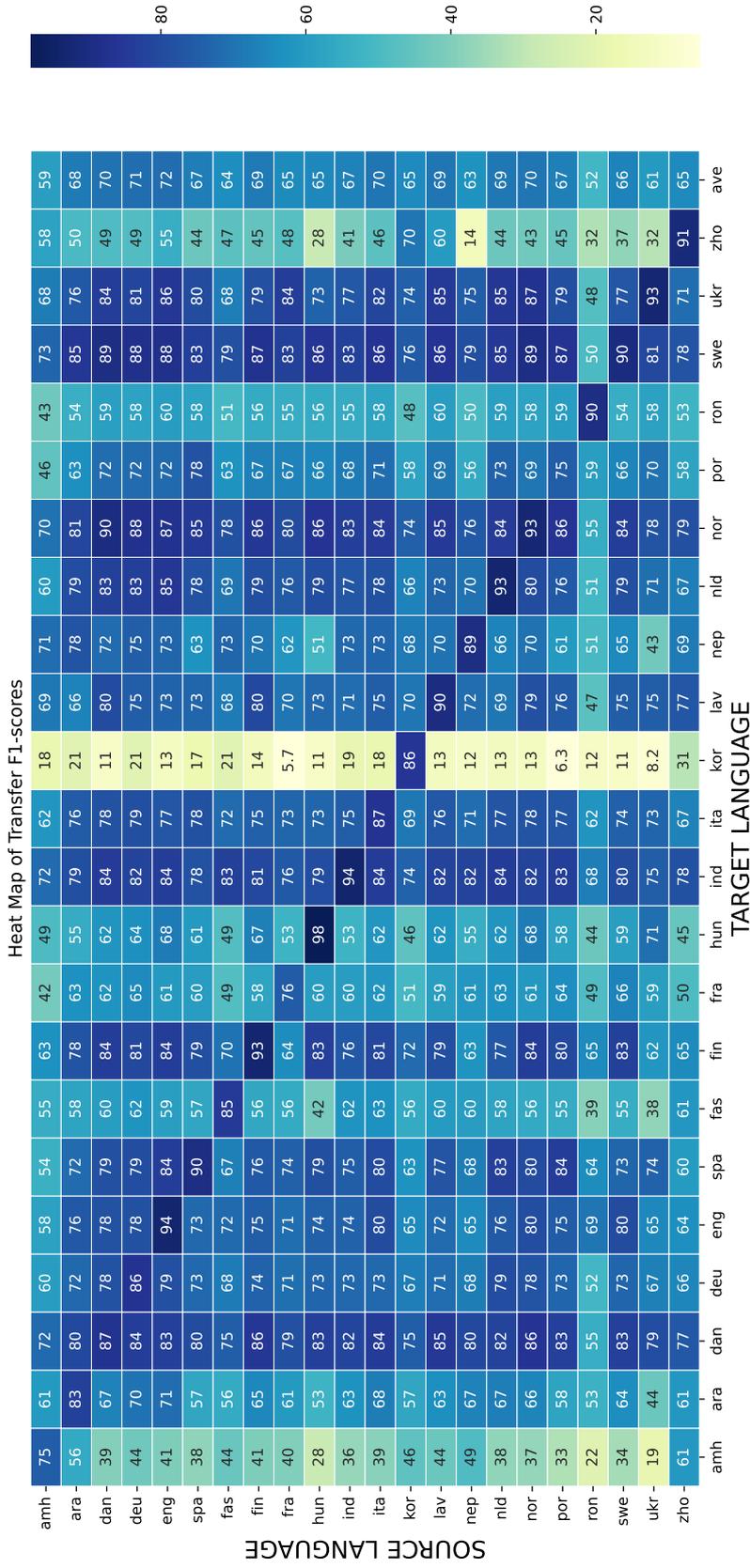

Figure 8.6: **Zero-shot Transfer** from several source languages to other languages not in MasakhaNER 2.0

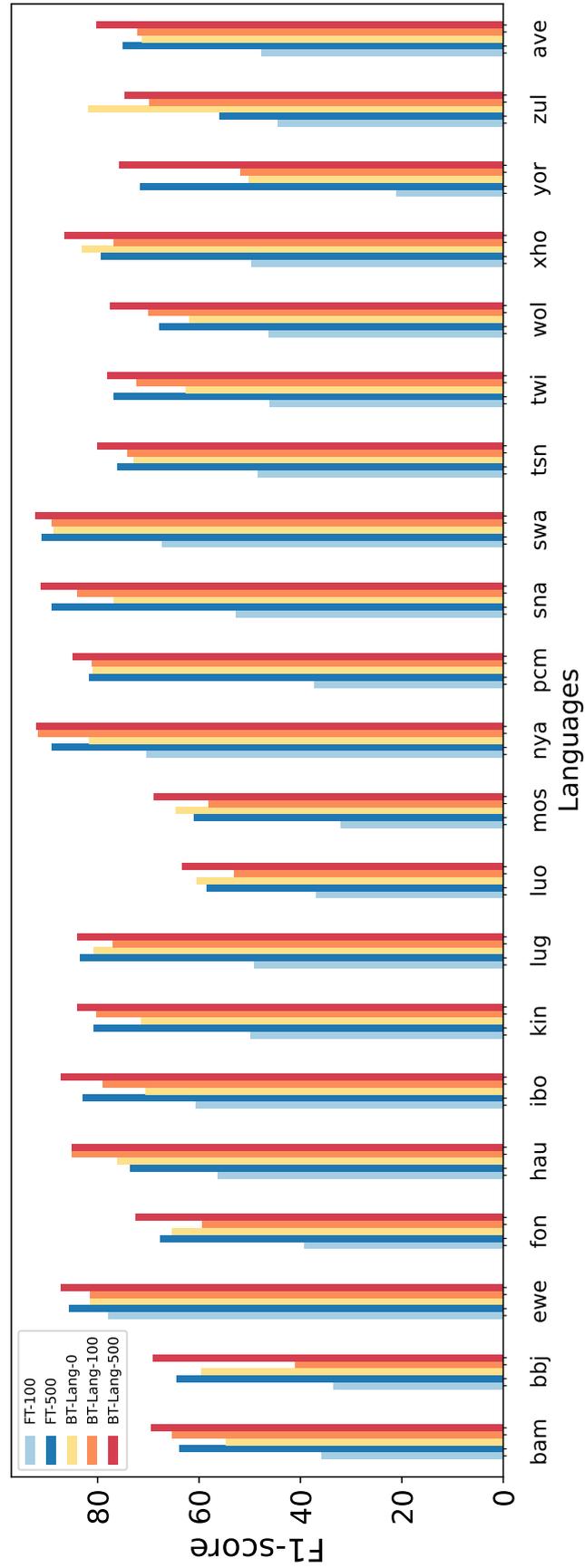

Figure 8.7: **Sample Efficiency Results** for 100 and 500 samples in the target language, model fine-tuned on a PLM (e.g. FT-100 – trained on 100 samples from the target language) or fine-tuned on the best transfer language NER model (e.g BT-Lang-0 – trained on 0 samples from the target language or zero-shot)

## 8.7 CONCLUSION

In this paper, we present the creation of MasakhaNER 2.0, the largest NER dataset for 20 diverse African languages and provide strong baseline results on the corpus by fine-tuning multilingual PLMs on in-language NER and multilingual datasets. Additionally, we analyze cross-lingual transfer in an Africa-centric setting, showing the importance of choosing the best transfer language in both zero-shot and few-shot scenarios. Using English as the default transfer language can have detrimental effects, and choosing a more appropriate language substantially improves fine-tuned NER models. By analyzing data-dependent, geographical, and typological features for transfer in NER, we conclude that geographical distance and entity overlap contribute most effectively to transfer performance.

### LIMITATIONS

NEWS DOMAIN DATA   As the data we annotated belonged to the news domain, models trained from this data may not generalize well to other domains. In particular, the models may not perform well on more casual text that may use different vocabulary, discuss different entities, and contain more orthographic variation.

GENERALIZABILTY OF TRANSFER LEARNING FINDINGS   As we only experimented with one task (NER), our findings regarding effective approaches to transfer learning for African languages and PLMs may may not generalize to other tasks (e.g. machine translation, part of speech tagging); other features of language similarity may be more important for other tasks.

EXPLAINING TRANSFER LEARNING FINDINGS   We found that the LangRank model could not predict the top transfer languages with 100% accuracy. This suggests that there are other, unknown factors that could affect transfer performance, which we did not explore. For example, there is still work to be done to understand the sociolinguistic connections and language contact conditions that may correlate with effective transfer.

Part IV

MACHINE TRANSLATION FOR AFRICAN LANGUAGES

# 9

## MULTI-DOMAIN MACHINE TRANSLATION

This Chapter[1] studies two challenges of building machine translation systems for low-resource languages: (1) lack of standardized evaluation datasets, and (2) generalizing to multiple domains outside the religious domain with large parallel corpus. We address the challenges by creating MENYO-20k, the first multi-domain parallel corpus with a special focus on clean orthography for Yorùbá–English with standardized train-test splits for benchmarking. We focus on Yorùbá, the third most spoken language (with over 40 million native speakers) in Africa without standardized evaluation set. We provide several neural MT benchmarks and compare them to the performance of popular pre-trained (massively multilingual) MT models both for the heterogeneous test set and its subdomains. Our models outperform massively multilingual models such as Google (+8.7 BLEU) and Meta M2M (+9.1 BLEU) when translating to Yorùbá, setting a high quality benchmark for future research. On further investigation with human evaluation, native speakers rated the translations into Yorùbá by our models and Google Translate to be of similar, and of high *adequacy*, but the latter is rated lower in *fluency* since it generally ignores diacritics, making it harder for the native speaker to read. The decision by Google Translate to ignore diacritics in Yorùbá translations hurt their BLEU evaluation.

### 9.1 INTRODUCTION

Neural machine translation (NMT) achieves high quality performance when large amounts of parallel sentences are available (Barrault et al., 2020). Large and freely-available parallel corpora do exist for a small number of high-resource pairs and domains. However, for low-resource languages such as Yorùbá (*yo*), one can only find few thousands of parallel sentences online[2]. In the best-case scenario, i.e. some amount of parallel data exists, one can use the Bible —the Bible is the most available resource for low-resource languages (Resnik, Olsen, and Diab, 1999)— and JW300 (Agić and Vulić, 2019). Notice that both corpora belong to the religious domain and they do not generalize well to popular domains such as news and daily conversations.

In this paper, we address this problem for the Yorùbá–English (*yo–en*) language pair by creating a multi-domain parallel dataset, MENYO-20k, which we make publicly available[3] with CC BY-NC 4.0 licence. It is a heterogeneous dataset that comprises texts obtained from news articles, TED talks, movie and radio transcripts, science and technology texts, and other short articles curated from the web and translated by professional translators. Based on the resulting train-development-test split, we provide a benchmark for the *yo–en* translation task for future research on this language pair. This allows us to properly evaluate the generalization of MT models trained on JW300 and the Bible on new domains. We further explore transfer learning approaches that can make use of a few thousand sentence pairs for domain adaptation. Finally, we analyze the effect of Yorùbá diacritics on the translation quality of pre-trained MT models, discussing in details how this affects the understanding of the translated text especially in the *en–yo* direction. We show the benefit of automatic diacritic restoration in addressing the problem of noisy diacritics.

---

1 This is based on Adelani et al. (2021c) with David Adelani, Dana Ruiter and Jesujoba Alabi contributing equally as first authors
2 http://opus.nlpl.eu
3 https://github.com/uds-lsv/menyo-20k_MT

## 9.2 THE YORÙBÁ LANGUAGE

The Yorùbá language is the third most spoken language in Africa, and it is native to south-western Nigeria and the Republic of Benin. It is one of the national languages in Nigeria, Benin and Togo, and spoken across the West African regions. The language belongs to the Niger-Congo family, and it is spoken by over 40 million native speakers (Eberhard, Simons, and Fennig, 2021).

Yorùbá has 25 letters without the Latin characters c, q, v, x and z, and with additional characters ẹ, gb, ṣ , ọ. Yorùbá is a tonal language with three tones: low, middle and high. These tones are represented by the grave (e.g. "à "), optional macron (e.g. "ā") and acute (e.g. "á") accents respectively. These tones are applied on vowels and syllabic nasals, but the mid tone is usually ignored in writings. The tone information and underdots are important for the correct pronunciation of words. Often, articles written online, including news articles such as BBC[4] ignore diacritics. Ignoring diacritics makes it difficult to identify or pronounce words except when they are embedded in context. For example, *èdè* (language), *edé* (crayfish), *ẹdẹ* (a town in Nigeria), *ẹ̀dẹ* (trap) and *ẹ̀dẹ̀* (balcony) will be mapped to *ede* without diacritics.

Machine translation might be able to learn to disambiguate the meaning of words and generate correct English even with un-diacriticized Yorùbá. However, one cannot generate correct Yorùbá if the training data is un-diacriticized. One of the purposes of our work is to build a corpus with correct and complete diacritization in several domains.

## 9.3 MENYO-20K

The dataset collection was motivated by the inability of machine translation models trained on JW300 to generalize to new domains (∀

---

4 https://www.bbc.com/yoruba

9.3 MENYO-20K 143

| Data name | Source | No. Sent. |
|---|---|---|
| **source language: en-yo** | | |
| JW News | jw.org/yo/iroyin | 3,508 |
| VON News | von.gov.ng | 3,048 |
| GV News | globalvoices.org | 2,932 |
| Yorùbá Proverbs | @yoruba_proverbs | 2,700 |
| Movie Transcript | "Unsane" on YouTube | 774 |
| UDHR | ohchr.org | 100 |
| ICT localization | from Yorùbá translators | 941 |
| Short texts | from Yorùbá translators | 687 |
| **source language: en** | | |
| TED talks | ted.com/talks | 2,945 |
| Out of His Mind | from the book author | 2,014 |
| Radio Broadcast | from Bond FM Radio | 258 |
| CC License | Creative Commons | 193 |
| Total | | 20,100 |

Table 9.1: Data collection for MENYO-20k

et al., 2020). Although ∀ et al. (2020) evaluated this for Yorùbá with surprisingly high BLEU scores, the evaluation was done on very few examples from the COVID-19 and TED Talks domains with 39 and 80 sentences respectively. Inspired by the FLoRes dataset for Nepali and Sinhala (Guzmán et al., 2019), we create a high quality test set for Yorùbá-English with few thousands of sentences in different domains to check the quality of industry MT models, pre-trained MT models, and MT models based on popular corpora such as JW300 and the Bible.

### 9.3.1 Dataset Collection for MENYO-20k

Table 9.1 summarizes the texts collected, their source, the original language of the texts and the number of sentences from each source. We collected both parallel corpora freely available on the web (e.g JW News) and monolingual corpora we are interested in translating (e.g. the TED talks) to build the MENYO-20k corpus. The JW News is different from the JW300 since they contain only news reports, and we manually verified that they are not in JW300. Some few sentences were donated by professional translators such as "short texts" in Table 9.1. Our curation followed two steps: (1) translation of monolingual texts crawled from the web by professional translators; (2) verification of translation, orthography and diacritics for parallel texts obtained online and translated. Texts obtained from the web that were judged by native speakers being high quality were verified once, the others were verified twice. The verification of translation and diacritics was done by professional translators and volunteers who are native speakers.

We provide more specific description of the data sources below.

JEHOVAH WITNESS NEWS   We collected only parallel *"newsroom"* (or *"Ìròyìn"* in Yorùbá) articles from `JW.org` website to gather texts that are not in the religious domain. As shown in Table 9.1, we collected 3,508 sentences from their website, and we manually confirmed that the sentences are not in JW300. The content of the news mostly reports persecutions of Jehovah witness members around the world, and may sometimes contain Bible verses to encourage believers.

VOICE OF NIGERIAN NEWS   We extracted parallel texts from the VON website, a Nigerian Government news website that supports seven languages with wide audience in the country (Arabic, English, Fulfulde, French, Hausa, Igbo, and Yorùbá). Despite the large availability of texts, the quality of Yorùbá texts is very poor, one can see several issues with orthography and diacritics. We asked translators and other native speakers to verify and correct each sentence.

GLOBAL VOICES NEWS   We obtained parallel sentences from the Global Voices website[5] contributed by journalists, writers and volunteers. The website supports over 50 languages, with contents mostly translated from English, French, Portuguese or Spanish.

TED TALKS TRANSCRIPTS   We selected 28 English TED talks transcripts mostly covering issues around Africa like health, gender equality, corruption, wildlife, and social media e.g "How young Africans found a voice on Twitter" (see the Table 9.2 for the selected TED talk titles). The articles were translated by a professional translator and verified by another one.

PROVERBS   Yorùbá has many proverbs and culturally referred to words of wisdom that are often referenced by elderly people. We obtained 2,700 sentences of parallel *yo–en* texts from Twitter.[6]

BOOK   With permission from the author (Bayo Adebowale) of the "Out of His Mind" book, originally published in English, we translated the entire book to Yorùbá and verified the diacritics.

SOFTWARE LOCALIZATION TEXTS (DIGITAL)   We obtained translations of some software documentations such as Kolibri[7] from past

---

[5] https://globalvoices.org
[6] Also available in https://github.com/Niger-Volta-LTI/yoruba-text
[7] https://learningequality.org/kolibri

|  | Title | Topic |
|---|---|---|
| 1 | Reducing corruption takes a specific kind of investment | Politics |
| 2 | How young Africans found a voice on Twitter | Technology |
| 3 | Mothers helping mothers fight HIV | Health |
| 4 | How women are revolutionizing Rwanda | Gender-equality |
| 5 | How community-led conservation can save wildlife | Wildlife |
| 6 | How cancer cells communicate - and how we can slow them down | Health |
| 7 | You may be accidentally investing in cigarette companies | Health |
| 8 | How deepfakes undermine truth and threaten democracy | Politics |
| 9 | What tech companies know about your kids | Technology |
| 10 | Facebook's role in Brexit - and the threat to democracy | Politics |
| 11 | How we can make energy more affordable for low-income families | Energy |
| 12 | Can we stop climate change by removing CO2 from the air? | Climate |
| 13 | A comprehensive, neighborhood-based response to COVID-19 | Health |
| 14 | Why civilians suffer more once a war is over | Human Rights |
| 15 | Lessons from the 1918 flu | Health |
| 16 | Refugees have the right to be protected | Human Rights |
| 17 | The beautiful future of solar power | Energy |
| 18 | How bees can keep the peace between elephants and humans | Wildlife |
| 19 | Will automation take away all our jobs? | Technology |
| 20 | A celebration of natural hair | Beauty |
| 21 | Your fingerprints reveal more than you think | Technology |
| 22 | Our immigration conversation is broken - here's how to have a better one | Politics |
| 23 | What I learned about freedom after escaping North Korea | Politics |
| 24 | Medical tech designed to meet Africa's needs | Health |
| 25 | What's missing from the American immigrant narrative | Education |
| 26 | A hospital tour in Nigeria | Health |
| 27 | How fake news does real harm | Politics |
| 28 | How we can stop Africa's scientific brain drain | Education |

Table 9.2: TED talks titles.

projects of professional translators. These texts include highly technical terms.

MOVIE TRANSCRIPTS    We obtained the translation of a Nigerian movie "Unsane" on YouTube from the past project of a professional translator. The language of the movie is Yorùbá and English, with transcription also provided in English.

OTHER SHORT TEXTS    Other short texts like UDHR, Creative Commons License, radio transcripts, and texts were obtained from professional translators and online sources. Table 9.1 summarizes the number of sentences obtained from each source.

Table 9.3 summarizes the figures for the MENYO-20k dataset with 20,100 parallel sentences split into 10,070 training sentences, 3,397 development sentences, and 6,633 test sentences. The test split contains 6 domains, 3 of them have more than 1000 sentences and can be used as domain test sets by themselves.

|  | **Number of Sentences** | | |
| --- | --- | --- | --- |
| **Domain** | **Train. Set** | **Dev. Set** | **Test Set** |
| *MENYO-20k* | | | |
| **News** | 4,995 | 1,391 | 3,102 |
| **TED Talks** | 507 | 438 | 2,000 |
| **Book** | - | 1,006 | 1,008 |
| **IT** | 356 | 312 | 273 |
| **Yorùbá Proverbs** | 2,200 | 250 | 250 |
| **Others** | 2,012 | 250 | 250 |
| *Standard (religious) corpora* | | | |
| **Bible** | 30,760 | – | – |
| **JW300** | 459,871 | – | – |
| *TOTAL* | 500,701 | 3,397 | 6,633 |

Table 9.3: MENYO-20k domains and training, development and test splits (top); figures for standard corpora used in this work (bottom).

9.3.2 *Other Corpora for Yorùbá and English*

PARALLEL CORPORA  For our experiments, we use two widely available parallel corpora from the religion domain, Bible and JW300 (Table 9.3, bottom). The parallel version of the Bible is not available, so we align the verses from the New International Version (NIV) for English and the Bible Society of Nigeria version (BSN) for Yorùbá. After aligning the verses, we obtain 30,760 parallel sentences. Also, we download the JW300 parallel corpus which is available for a large variety of low-resource language pairs. It has parallel corpora from English to 343 languages containing religion-related texts. From the JW300 corpus, we get 459,871 sentence pairs already tokenized with *Polyglot*[8] (Al-Rfou, 2015).

MONOLINGUAL CORPORA  We make use of additional monolingual data to train the semi-supervised MT model using back-translation. The Yorùbá monolingual texts are from the Yorùbá embedding corpus (Alabi et al., 2020), one additional book ("Ojowu") with permission from the author, JW300-*yo*, and Bible-*yo*. We only use Yorùbá texts that are properly diacritized. In order to keep the topics in the Yorùbá and English monolingual corpora close, we choose two Nigerian news websites (The Punch Newspaper[9] and Voice of Nigeria [10]) for the English monolingual corpus. The news scraped covered categories such as politics, business, sports and entertainment. Overall, we gather 475,763 monolingual sentences from the website.

---

8 https://github.com/aboSamoor/polyglot
9 https://punchng.com
10 https://von.gov.ng

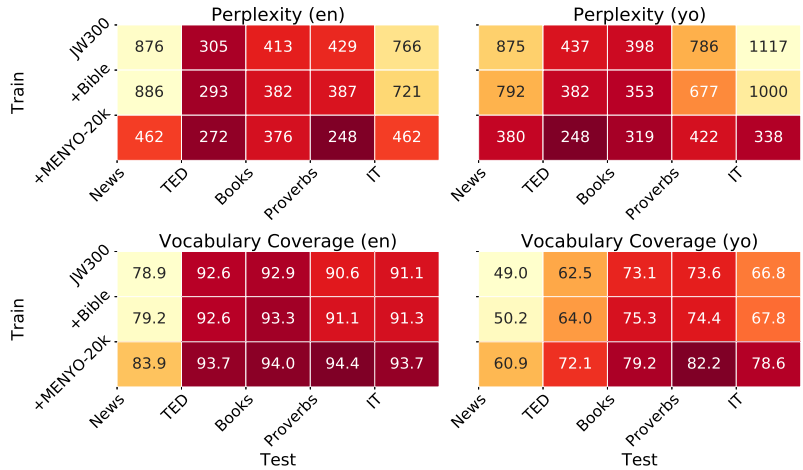

Figure 9.1: **Top:** Perplexities of KenLM 5-gram language model learned on different training corpora and tested on subsets of MENYO-20k for English (left) and Yorùbá (right) respectively. **Bottom:** Vocabulary coverage (%) of different subsets of the MENYO-20k test set per training sets for English (left) and Yorùbá (right).

### 9.3.3 *Dataset Domain Analysis*

MENYO-20k is, on purpose, highly heterogeneous. In this section we analyze the differences and how its (sub)domains depart from the characteristics of the commonly used Yorùbá–English corpora for MT.

Characterizing the domain of a dataset is a difficult task. Some metrics previously used need either large corpora or a characteristic vocabulary of the domain (Beyer, Kauermann, and Schütze, 2020; España-Bonet, Barrón-Cedeño, and Màrquez, 2020). Here, we do not have these resources and we report the overlapping vocabulary between training and test sets and the perplexity observed in the test sets when a language model (LM) is trained on the MT training corpora.

In order to estimate the perplexities, we train a language model of order 5 with KenLM (Heafield, 2011) on each of the 3 training data subsets: JW300 (named C2 for short in tables), JW300+Bible (C3), JW300+Bible+MENYO-20k (C4). Following NMT standard processing pipelines (see Section 9.4.2), we perform byte-pair encoding (BPE) (Sennrich, Haddow, and Birch, 2016a) on the corpora to avoid a large number of out-of-vocabulary tokens which, for small corpora, could alter the LM probabilities. For each of the resulting language models, we evaluate their average **perplexity** on the different domains of the test set to evaluate *compositional* domain differences (Figure 9.1, top). As expected, the average perplexity drops when adding more training data. Due to the limited domain of both JW300 and Bible, a literary style close to the Books domain, the decrease in perplexity is small when adding additional Bible data to JW300, namely −8% (*en*) and −11% (*yo*). Interestingly, both JW300 and Bible also seem to be close to the TED domain (1st and 2nd lowest perplexities for *en* and *yo* respectively), which may be due to discourse/monologue content in both training corpora. Adding the domain-diverse MENYO-20k corpus largely decreases the perplexity across all domains with a major decrease of $-66\%$ on IT (*yo*) and smallest decrease of $-1\%$ on Books (*en*). The perplexity scores correlate negatively with the resulting BLEU scores in Table 9.5, with a Pearson's $r$ ($r$) of $-0.367$ (*en*) and $-0.461$ (*yo*), underlining that compositional domain differences between training and test subsets is the main factor of differences in translation quality.

Further, to evaluate *lexical* domain differences, we calculate the **vocabulary coverage** (tokenized, not byte-pair encoded[11]) of the different domains of the test set by each of the training subsets (Figure 9.1, bottom). The vocabulary coverage increases to a large extend when MENYO-20k is added. However, while vocabulary coverage and average perplexities have a strong (negative) correlation, $r = -0.756$ (*en*) and $r = -0.689$ (*yo*), a high perplexity does not necessarily mean low vocabulary coverage. E.g., the vocabulary coverage of the IT domain by JW300 is high (91% for *en*) despite leading to high perplexities (765 for *en*). In general, vocabulary coverage of the test sets is less indicative of the resulting translation performance than perplexity, showing only a weak correlation between vocabulary coverage and BLEU, with $r = 0.150$ and $r = 0.281$ for *en* and *yo* respectively.

## 9.4 NEURAL MACHINE TRANSLATION FOR YORÙBÁ–ENGLISH

### 9.4.1 *Systems*

SUPERVISED NMT   We use the transformer-base architecture proposed by Vaswani et al. (2017) as implemented in Fairseq[12] (Ott et al., 2019). We set the drop-out at 0.3 and batch size at 10,240 tokens. For optimization, we use *adam* (Kingma and Ba, 2015) with $\beta_1 = 0.9$ and $\beta_2 = 0.98$ and a learning rate of 0.0005. The learning rate has a warmup update of 4000, using label smoothed cross-entropy loss function with label-smoothing value of 0.1.

SEMI-SUPERVISION VIA ITERATIVE BACK-TRANSLATION   We use the best performing supervised system to translate the monolingual corpora described in Section 9.3 yielding to 476*k* back-translations. This data is used together with the original corpus to train a new system. The process is repeated until convergence.

---

[11] We do not use byte-pair encoded data here, since, due to the nature of BPE, the vocabulary overlap would be close to 1 between all training and test sets.
[12] https://github.com/pytorch/fairseq

FINE-TUNING MT5   We examine a transfer learning approach by fine-tuning a massively multilingual model mT5 (Xue et al., 2021). mT5 had been pre-trained on 6.3T tokens originating from Common Crawl in 101 languages (including Yorùbá). The approach has already shown competitive results on other languages (Tang et al., 2020). In our experiments, we use mT5-base, a model with 580M parameters. We transferred all the parameters of the model including the sub-word vocabulary.

PUBLICLY AVAILABLE NMT MODELS   We further evaluate the performance of three multilingual NMT systems: OPUS-MT (Tiedemann and Thottingal, 2020), Google Multilingual NMT (GMNMT) (Arivazhagan et al., 2019) and Meta's M2M-100 (Fan et al., 2021) with 1.2B parameters. All the three pre-trained models are trained on over 100 languages. While GMNMT and M2M-100 are a single multilingual model, OPUS-MT models are for each translation direction, e.g *yo–en*. We generate the translations of the test set using the *Google Translate* interface,[13] and OPUS-MT using *Easy-NMT*.[14] For M2M-100, we make use of *Fairseq* to translate the test set.

### 9.4.2 *Experimental Settings*

DATA AND PREPROCESSING   For the MT experiments, we use the training part of our MENYO-20k corpus and two other parallel corpora, Bible and JW300 (Section 9.3). For tuning the hyper-parameters, we use the development split of the multi-domain data which has $3,397$ sentence pairs and for testing the test split with $6,633$ parallel sentences. To ensure that all the parallel corpora are in the same format, we convert the Yorùbá texts in the JW300 dataset to Unicode Normalization Form Composition (NFC), the format of the Yorùbá texts in the Bible and multi-domain dataset. Our preprocessing pipeline includes punctuation normalization, tokenization, and truecasing. For punctuation normalization and truecasing, we use the *Moses* toolkit (Koehn et al., 2007) while for tokenization, we use *Polyglot*, since it is the tokenizer used in JW300. We apply joint BPE, with a vocabulary threshold of 20 and 40k merge operations.

EVALUATION METRICS   To evaluate the models, we use tokenized BLEU (Papineni et al., 2002) score implemented in *multi-bleu.perl* and confidence intervals ($p = 95\%$) in the scoring package[15]. Since diacritics are applied on individual characters, we also use chrF, a character $n$-gram F1-score (Popović, 2015), for *en–yo* translations.

---

[13] https://translate.google.com/
[14] https://github.com/UKPLab/EasyNMT
[15] https://github.com/lvapeab/confidence_intervals

AUTOMATIC DIACRITIZATION    In order to automatically diacritize Google MNMT and M2M-100 outputs for comparison, we train an automatic diacritization system using the supervised NMT setup. We use the Yorùbá side of MENYO-20k and JW300, which use consistent diacritization. We split the resulting corpus into train (458*k* sentences), test (517 sentences) and development (500 sentences) portions. We apply a small BPE of 2k merge operations to the data. We apply noise on the diacritics by *i*) randomly removing a diacritic with probability $p = 0.3$ and *ii*) randomly replacing a diacritic with $p = 0.3$. The corrupted version of the corpus is used as the source data, and the NMT model is trained to reconstruct the original diacritics. On the test set, where the corrupted source has a BLEU (precision) of 19.0 (29.8), reconstructing the diacritics using our system lead to a BLEU (precision) of 87.0 (97.1), thus a major increase of +68.0 (+67.3) respectively.

9.4.3   *Automatic Evaluation*

INTERNAL COMPARISON    We train four basic NMT engines on different subsets of the training data: Bible (C1), JW300 (C2), JW300+Bible (C3) and JW300+Bible+MENYO-20k (C4). Further, we analyse the effect of fine-tuning for in-domain translation. For this, we fine-tune the converged model trained on JW300+Bible on MENYO-20k (C3+Transfer) and, similarly, we fine-tune the converged model trained on JW300+Bible+MENYO-20k on MENYO-20k (C4+Transfer). This yields six NMT models in total for *en–yo* and *yo–en* each. Their translation performance is evaluated on the complete MENYO-20k test set (Table 9.4, top) and later we analyze in-domain translation in Table 9.5.

As expected, the BLEU scores obtained after training on Bible only (C1) are low, with BLEU 2.2 and 1.4 for *en–yo* and *yo–en* respectively, which is due to its small amount of training data. Training on the larger JW300 corpus (C2) leads to higher scores of BLEU 7.5 (*en–yo*) and 9.6 (*yo–en*), while combining it with Bible (C3) only leads to a small increase of BLEU +0.6 and +1.2 for *en–yo* and *yo–en* respectively. When further adding MENYO-20k (C4) to the training data, the translation quality increases by +2.8 (*en–yo*) and +3.2 (*yo–en*). When, instead of adding MENYO-20k to the training pool, it is used to fine-tune the converged JW300+Bible model, (C3+Transfer) the increase in BLEU over JW300+Bible is even larger for *en–yo* (BLEU +4.2), which results in an overall top-scoring model with BLEU 12.3. For *yo–en* fine-tuning is slightly less effective (BLEU 13.2) than simply adding MENYO-20k to the training data (BLEU 14.0). As seen in Section 9.3.3, perplexities and vocabulary coverage in English are not as distant among training/test sets as in Yorùbá, so the fine-tuning step resulted less efficient.

When we use the MENYO-20k dataset to fine-tune the converged JW300+Bible+ MENYO-20k model (C4+Transfer) we observe an in-crease in BLEU over JW300+Bible for both translation directions: +4.3 for *en–yo* and +3.8 for *yo–en*. This is the best performing system and the one we use for back-translation. Table 9.4 also shows the performance of the semi-supervised system (C4+Transfer+BT). After two iterations of BT, we obtain an improvement of +3.6 BLEU points on *yo–en*. There is, however, no improvement in the *en–yo* direction probably because a significant portion of our monolingual data is based on JW300. Finally, fine-tuning mT5 with MENYO-20k does not improve over fine-tuning only the JW300+Bible system on *en–yo*, but it does for *yo–en*. Again, multilingual systems are stronger when used for English, and we need the contribution of back-translation to outperform the generic mT5.

| Model | *en–yo* | | *en–yo$^p$* | | *yo–en* | *yo–en$^u$* |
|---|---|---|---|---|---|---|
| | chrF | BLEU | chrF | BLEU | BLEU | BLEU |
| *Internal Comparison* | | | | | | |
| **C1: Bible** | 16.9 | 2.2±0.1 | – | – | 1.4±0.1 | 1.6±0.1 |
| **C2: JW300** | 29.1 | 7.5±0.2 | – | – | 9.6±0.3 | 9.3±0.3 |
| **C3: JW300+Bible** | 29.8 | 8.1±0.2 | – | – | 10.8±0.3 | 10.5±0.3 |
| +Transfer | 33.8 | 12.3±0.3 | – | – | 13.2±0.3 | 13.9±0.3 |
| **C4: JW300+Bible+MENYO-20k** | 32.5 | 10.9±0.3 | – | – | 14.0±0.3 | 14.0±0.3 |
| +Transfer | 34.3 | **12.4**±0.3 | – | – | 14.6±0.3 | – |
| + BT | 34.6 | 12.0±0.3 | – | – | <u>18.2</u>±0.4 | – |
| **mT5: mT5-base+Transfer** | 32.9 | 11.5±0.3 | – | – | 16.3±0.4 | 16.3±0.4 |
| *External Comparison* | | | | | | |
| **OPUS-MT** | – | – | – | – | 5.9±0.2 | – |
| **Google GMNMT** | 18.5 | 3.7±0.2 | 34.4 | 10.6±0.3 | **22.4**±0.5 | – |
| **Meta M2M-100** | 15.8 | 3.3±0.2 | 25.7 | 6.8±0.3 | 4.6±0.3 | – |

Table 9.4: Tokenized BLEU with confidence intervals ($p = 95\%$) and chrF scores over the full test for NMT models trained on different subsets of the training data C$_i$ (top) and performance of external systems (bottom). For Yorùbá, we analyse the effect of diacritization: *en–yo$^p$* applies an in-house diacritizer on the translations obtained from **p**re-trained models and *yo–en$^u$* reports results using **u**ndiacritized Yorùbá texts as source sentences for training (see text). Top-scoring results per block are underlined and globally boldfaced.

EXTERNAL COMPARISON   We evaluate the performance of the open source multilingual engines introduced in the previous section on the full test set (Table 9.4, bottom). **OPUS-MT**, while having no model available for *en–yo*, achieves a BLEU of 5.9 for *yo–en*. Thus, despite being trained on JW300 and other available *yo–en* corpora on OPUS, it is largely outperformed by our NMT model trained on JW300 only (BLEU +3.7). This may be caused by some of the noisy corpora in-cluded in OPUS (like CCaligned), which can depreciate the translation quality.

Meta's **M2M-100**, is also largely outperformed even by our simple JW300 baseline by 5 BLEU points in both translation directions. A manual examination of the *en–yo* LASER extractions used to train M2M-100 shows that these are very noisy similar to the findings of Kreutzer et al. (2022), which explains the poor translation performance.

Google, on the other hand, obtains impressive results with **GMNMT** for the *yo–en* direction, with BLEU 22.4. The opposite direction *en–yo*, however, shows a significantly lower performance (BLEU 3.7), being outperformed even by our simple JW300 baseline (BLEU +3.8). The difference in performance for English can be attributed to the highly multilingual but English-centric nature of the Google MNMT model. As already noticed by Arivazhagan et al. (2019), low-resourced language pairs benefit from multilinguality when translated into English, but improvements are minor when translating into the non-English language. For the other translation direction, *en–yo*, we notice that lots of diacritics are lost in Google translations, damaging the BLEU scores. Whether this drop in BLEU scores really affects understanding or not is analyzed via a human evaluation (Section 9.4.4).

DIACRITIZATION   Diacritics are important for Yorùbá embeddings (Alabi et al., 2020). However, they are often ignored in popular multilingual models (e.g. multilingual BERT (Devlin et al., 2019)), and not consistently available in training corpora and even test sets. In order to investigate whether the diacritics in Yorùbá MT can help to disambiguate translation choices, we additionally train *yo–en$^u$* equivalent models on **u**ndiacritized JW300, JW300+Bible and JW300+Bible+MENYO-20k (Table 9.4, indicated as *yo–en$^u$* in comparison to the ones with diacritics *yo–en*). Since one cannot generate correct Yorùbá text when training without diacritics, *en–yo$^u$* systems are not trained. Alternatively, we restore diacritics using our in-house diacritizer in the output of open source models that produce undiacritized text.

Results for *yo–en* are not conclusive. Diacritization is useful when only out-of-domain data is used in training (JW300, JW300+Bible[16] for testing on MENYO-20k). In this case, the domain of the training data is very different from the domain of the test set, and disambiguation is needed not to bias all the lexicon towards the religious domain. When we include in-domain data (JW300+Bible+MENYO-20k), both models perform equally well, with BLEU 14.0 for both diacritized and undiacritized versions. Diacritization is not needed when we fine-tune the model with data that shares the domain with the test set (JW300+Bible+Transfer), BLEU is 13.2 for the diacritized version vs. BLEU 13.9 for the undiacritized one.

---

16 We do not consider Bible alone. Due to its small data size, the BLEU scores are less indicative.

|  | *en–yo* | | | | | *yo–en* | | | | |
| --- | --- | --- | --- | --- | --- | --- | --- | --- | --- | --- |
|  | Prov. | News | TED | Book | IT | Prov. | News | TED | Book | IT |
| C1 | 0.8 | 1.7 | 3.1 | 3.4 | 1.5 | 1.1 | 0.9 | 2.1 | 2.4 | 0.9 |
| C2 | 2.2 | 6.4 | 9.8 | 9.8 | 4.8 | 2.6 | 8.4 | 13.1 | 9.6 | 7.0 |
| C3 | 3.5 | 6.7 | 10.7 | 11.3 | 4.9 | 4.8 | 9.5 | 14.4 | 10.9 | 7.8 |
| +Transfer | 9.0 | 10.2 | **16.1** | 15.0 | 11.8 | 8.6 | 12.5 | 16.8 | 10.8 | 9.7 |
| C4 | 7.0 | 10.0 | 12.3 | 11.5 | 10.5 | 8.7 | 13.5 | 16.7 | 11.6 | 12.4 |
| +Transfer | **10.3** | 10.9 | 15.1 | 13.2 | 13.6 | **9.3** | 14.0 | 17.8 | 11.9 | 13.7 |
| +BT | 7.5 | 11.4 | 12.9 | 14.5 | 9.7 | 7.9 | **18.6** | **20.6** | **13.3** | **16.4** |
| mT5+Transfer | 3.8 | **11.2** | 13.1 | 11.8 | 7.9 | 6.0 | 16.4 | 18.9 | 13.1 | 15.1 |

Table 9.5: Tokenized BLEU over different domains of the test set for NMT models trained on different subsets of the training data, with top-scoring results per domain in bold.

In practice, this means that, when training data is far from the desired domain, investing work for a clean diacritized Yorùbá source input can help improve the translation performance. When more data is present, the diacritization becomes less important, since context is enough for disambiguation.

When Yorùbá is the target language, diacritization is always needed. An example is the low automatic scores GMNMT (BLEU 3.7, chrF 18.5) and M2M-100 (BLEU 3.3, chrF 15.8) reach for *en–yo* translation. Table 9.4-bottom (indicated as *en–yo^p*) show the improvements after automatically restoring the diacritics, namely $BLEU + 6.9$ points, chrF +15.9 for GMNMT; and +3.5 and +9.9 for M2M-100. Even if the diacritizer is not perfect, diacritics do not seem enough to get state-of-the-art results according to automatic metrics: fine-tuning with high quality data (C4+Transfer+BT, chrF 34.6) is still better than using huge but unadapted systems.

DOMAIN DIFFERENCES   In order to analyze the domain-specific performance of the different NMT models, we evaluate each model on the different domain subsets of the test set (Table 9.5). The Proverb subset is especially difficult in both directions, as it shows the lowest translation performance across all domains, i.e. maximum BLEU of 9.04 (*en–yo*) and 8.74 (*yo–en*). This is due to the fact that proverbs often do not have literal counterparts in the target language, thus making them especially difficult to translate. The TED domain is the best performing test domain, with maximum BLEU of 16.1 (*en–yo*) and 16.8 (*yo–en*). This can be attributed to the decent base coverage of the TED domain by JW300 and Bible together (monologues) with the additional TED domain data included in the MENYO-20k training split (507 sentence pairs). Also, most BLEU results are on line with the LM perplexity results and conclusions drawn in Section 9.3.3. Due to the closeness of Bible and JW300 to the book domain, we see only small improvements of BLEU on this domain, i.e. +0.2 (*en–yo*) and

|  | *en–yo* | | | *yo–en* | | |
| Task | C4+Trf | C4+Trf+BT | GMNMT | mT5+Trf | C4+Trf+BT | GMNMT |
| --- | --- | --- | --- | --- | --- | --- |
| Adequacy | 3.12* | 3.58 | **3.69** | 3.42* | 3.41* | **4.02** |
| Fluency | **4.57*** | 4.49* | 3.74 | 4.39* | 4.18* | **4.71** |
| Diacritics acc. | **4.91*** | 4.90* | 1.74 | - | - | - |

Table 9.6: Human evaluation for *en–yo* and *yo–en* MT models (C4+Transfer (C4+Trf), C4+Trf+BT, mT5+Trf, and GMNMT) in terms of Adequacy, Fluency and Diacritics prediction accuracy. The rating that is significantly different from GMNMT is indicated by * (T-test $p < 0.05$)

+0.7 (*yo–en*), when adding MENYO-20k (C4) to the JW300+Bible (C3) training data pool. On the other hand, the IT domain benefits the most from the additional MENYO-20k data, with major gains of BLEU +5.5 (*en–yo*) and 4.6 (*yo–en*), owing to the introduction of IT domain content in the MENYO-20k training data ($\sim 1k$ sentence pairs), which is completely lacking in JW300 and Bible.

### 9.4.4  *Human Evaluation*

To have a better understanding of the quality of the translation models and the intelligibility of the translations, we compare three top performing models in *en–yo* and *yo–en*. For *en–yo*, we use **C4+Transfer**, **C4+Transfer+BT** and **GMNMT**. Although GMNMT is not the third best system according to BLEU (Table 9.4), we are interested in the study of diacritics in translation quality and intelligibility. For the *yo–en*, we choose **C4+Transfer+BT**, **mT5+Transfer** and **GMNMT** being the 3 models with the highest BLEU scores on Table 9.4.

We ask 7 native speakers of Yorùbá that are fluent in English to rate the adequacy, fluency and diacritic accuracy in a subset of test sentences. Four of them rated the *en–yo* translation direction and the others rated the opposite direction *yo–en*. We randomly select 100 sentences within the outputs of the six systems and duplicate 5 of them to check the intra-agreement consistency of our raters. Each annotator is then asked to rate 105 sentences per system on a $1-5$ Likert scale for each of the features (for English, diacritic accuracy cannot be evaluated). We calculate the agreement among raters using Krippendorff's $\alpha$. The inter-agreement per task is 0.44 (adequacy), 0.40 (fluency) and 0.87 (diacritics) for Yorùbá, and 0.71 (adequacy), 0.55 (fluency) for English language. We observe that a lot of raters often rate the fluency score for many sentences with the same values (e.g 4 or 5), which results to a lower Krippendorff's $\alpha$ for fluency. The intra-agreement for the four Yorùbá raters are 0.75, 0.91, 0.66, and 0.87, while the intra-agreement for the three English raters across all evaluation tasks are 0.92, 0.71, and 0.81.

For *yo–en*, our evaluators rated on average GMNMT to be the best in terms of adequacy (4.02 out of 5) and fluency (4.71), followed by mT5+Transfer, which shows that fine-tuning massively multilingual models also benefits low resource languages MT especially in terms of fluency (4.39). This contradicts the results of the automatic evaluation which prefers C4+Transfer+BT over mT5+Transfer.

For *en–yo*, GMNMT is still the best in terms of adequacy (3.69) followed by C4+Transfer+BT, but performs the worst in terms of fluency and diacritics prediction accuracy. So, the bad quality of the diacritics affects fluency and drastically penalises automatic metrics such as BLEU, but does not interfere with the intelligibility of the translations as shown by the good average adequacy rating. Automatic diacritic restoration for Yorùbá (Orife, 2018a; Orife et al., 2020) can therefore be very useful to improve translation quality. C4+Transfer and C4+Transfer+BT perform similarly with high scores in terms of fluency and near perfect score in diacritics prediction accuracy ($4.91 \pm 0.1$) as a result of being trained on cleaned corpora.

## 9.5 RELATED WORK

In order to make MT available for a broader range of linguistic communities, recent years have seen an effort in creating new **parallel corpora** for low-resource language pairs. Recently, Guzmán et al. (2019) provided novel supervised, semi-supervised and unsupervised benchmarks for Indo-Aryan languages {Sinhala,Nepali}–English on an evaluation set of professionally translated sentences sourced from the Sinhala, Nepali and English Wikipedias.

Novel parallel corpora focusing on **African languages** cover South African languages ({Afrikaans, isiZulu, Northern Sotho, Setswana, Xitsonga}–English) (Groenewald and Fourie, 2009) with MT benchmarks evaluated in Martinus and Abbott (2019), as well as multidomain (News, Wikipedia, Twitter, Conversational) Amharic–English (Hadgu, Beaudoin, and Aregawi, 2020) and multidomain (Government, Wikipedia, News etc.) Igbo–English (Ezeani et al., 2020). Further, the LORELEI project (Strassel and Tracey, 2016) has created parallel corpora for a variety of low-resource language pairs, including a number of Niger-Congo languages such as {isiZulu, Twi, Wolof, Yorùbá }–English. However, these are not open-access. On the contrary, Masakhane ($\forall$ et al., 2020) is an ongoing participatory project focusing on creating new freely-available parallel corpora and MT benchmark models for a large variety of African languages.

While creating parallel resources for low-resource language pairs is one approach to increase the number of linguistic communities covered by MT, this does not scale to the sheer amount of possible language combinations. Another research line focuses on **low-resource MT** from the modeling side, developing methods which allow a MT system to learn the translation task with smaller amounts of supervisory signals. This is done by exploiting the weaker supervisory signals in larger amounts of available monolingual data, e.g. by identifying additional parallel data in monolingual corpora (Artetxe and Schwenk, 2019a; Schwenk et al., 2021a,b), comparable corpora (Ruiter, España-Bonet, and Genabith, 2019; Ruiter et al., 2021), or by including auto-encoding (Currey, Miceli Barone, and Heafield, 2017) or language modeling tasks (Gulcehre et al., 2015; Ramachandran, Liu, and Le, 2017) during training. Low-resource language pairs can benefit from high-resource languages through transfer learning (Zoph et al., 2016), e.g. in a zero-shot setting (Johnson et al., 2017a), by using pre-trained language models (Lample and Conneau, 2019), or finding an optimal path of pivoting through related languages (Leng et al., 2019). By adapting the model hyperparameters to the low-resource scenario, Sennrich and Zhang (2019) were able to achieve impressive improvements over a standard NMT system.

## 9.6 CONCLUSION

We present MENYO-20k, a novel *en–yo* multi-domain parallel corpus for machine translation and domain adaptation. By defining a standardized train-development-test split of this corpus, we provide several NMT benchmarks for future research on the *en–yo* MT task. Further, we analyze the domain differences on the MENYO-20k corpus and the translation performance of NMT models trained on religion corpora, such as JW300 and Bible, across the different domains. We show that, despite consisting of only 10*k* parallel sentences, adding the MENYO-20k corpus train split to JW300 and Bible largely improves the translation performance over all domains. Further, we train a variety of supervised, semi-supervised and fine-tuned MT benchmarks on available *en–yo* corpora, creating a high quality baseline that outperforms current massively multilingual models, e.g. Google MNMT by BLEU +18.8 (*en–yo*). This shows the positive impact of using smaller amounts of high-quality data (e.g. C4+Transfer, BLEU 12.4) that takes into account language-specific characteristics, i.e. diacritics, over massive amounts of noisy data (Meta M2M-100, BLEU 3.3). Apart from having low BLEU scores, our human evaluation reveals that models trained on low-quality diacritics (Google MNMT) suffer especially in fluency, while still being intelligible to the reader. While correctly diacritized data is vital for translating *en–yo*, it only has an impact on the quality of *yo–en* translation quality when there is a domain mismatch between training and testing data.

# 10

## LEVERAGING PRE-TRAINED MODELS FOR AFRICAN NEWS TRANSLATION

In Chapter 9, we create a multi-domain benchmark evaluation on Yorùbá language. We also evaluate the performance of one pre-trained model i.e mT5, and show that their performance is better than training a Transformer model on a large parallel corpus (e.g. more than 400K sentences). However, this analysis was only performed on a language and not extended to a small corpus (e.g. 2K-5K sentences) which is the reality of many African languages especially for non-religious domains.

Chapter[1] investigates how to optimally leverage existing pre-trained models to create low-resource translation systems for 21 African languages. We focus on two questions: 1) How can pre-trained models be used for languages not included in the initial pre-training? and 2) How can the resulting translation models effectively transfer to new domains? To answer these questions, we create a new African news corpus covering 21 languages, of which eight languages are not part of any existing evaluation dataset. We demonstrate that the most effective strategy for transferring both to additional languages and to additional domains is to fine-tune large pre-trained models on small quantities of high-quality translation data.

## 10.1 INTRODUCTION

Enormous efforts have been invested in making language and translation models more multilingual while leveraging the maximal amount of data for training, most prominently large crawls of monolingual and parallel data from the web (El-Kishky et al., 2020; Schwenk et al., 2021a,b; Xue et al., 2021). The resulting models are now capable of translating between hundreds of languages, including language pairs that in isolation do not have large collections of parallel data (Fan et al., 2021; Tang et al., 2020; Xue et al., 2022). For example, M2M-100 (Goyal et al., 2022) can translate (with low accuracy) between Hausa and Yorùbá, two of the most widely spoken languages in Nigeria, even though there is barely any parallel data available for training. For languages that are not included in the set of training languages, the model would have no knowledge on how to generate translations. Does this mean there is no hope for languages that do not have large presence on the web and are therefore not included in these pre-trained models?

We investigate *how large-scale pre-trained models can be leveraged for the translation of unseen low-resource languages and domains*. We address this question by studying 21 African languages that are largely underrepresented in NLP research (Joshi et al., 2020; ∀ et al., 2020) and further have little to no training data available (§10.3). These languages provide an ideal testbed for two challenging knowledge transfer tasks: **(1)** How can pre-trained models create translations for languages unseen at training time? and **(2)** Since training data may only exist in single domain (i.e. religious texts), how can a model be trained in one domain and translate another effectively at test time?

---

1 This is based on Adelani et al. (2022b)

These questions are extremely relevant for our chosen languages because they all have millions of native speakers and a massive need for translation technologies. For example, news concerning the African continent are almost exclusively published in English, French, or Arabic, and thereby inaccessible for speakers of only native African languages. This creates a bottleneck for information transmission, which becomes even more critical in times of crises (Anastasopoulos et al., 2020; Öktem et al., 2021; Öktem, Plitt, and Tang, 2020). Furthermore, the task of translating news has historically played a central role in translation research, e.g. in shared tasks since 2008 (Callison-Burch et al., 2008) and as a test for determining human parity (Hassan et al., 2018; Läubli, Sennrich, and Volk, 2018; Toral et al., 2018). To spur the development of dedicated news translation models for Africa, we construct a benchmark of news translation for translating between 21 native African languages and English or French (§10.4).

This allows us to compare three approaches to leveraging large-scale multilingual models for the translation of previously unseen languages: **(1)** zero-shot transfer, **(2)** continual pre-training on monolingual data, and **(3)** multi-domain fine-tuning on parallel data (§10.5). We find that fine-tuning pre-trained models on a few thousand sentences of high quality bitext is remarkably effective, and can be further augmented with continual pre-training on African languages and fine-tuning on news domain data (§10.6).

Our contributions are the following:[2]

1. We create a **new African news corpus** for machine translation (following principles of participatory research ∀ et al. (2020)) covering 21 African languages.

2. We **adapt several multilingual pre-trained models** (MT5, ByT5, mBART, M2M-100) to these largely unseen languages, and evaluate their quality on news translation.

3. We quantify the **effectiveness of small in-domain translation sets** by measuring domain transfer effects and comparing fine-tuning strategies.

We find that having a targeted collection of translations is surprisingly effective, showcasing the power of local knowledge in so-called "zero-resource" scenarios (Bird, 2020). This paints a promising picture for the development of NLP technology for understudied languages: being able to customize these models for new language of interest with as little as 2k sentences and a few fine-tuning steps, MT developers and users from any language community are less dependent on choices and monetary interest of industry powerhouses from the Global North (Paullada, 2020).

10.2 related work

african mt datasets. One of the major challenges of developing MT models for African languages is lack of data. There are many attempts to automatically crawl and align sentences from the web (Schwenk et al., 2021a,b). Nevertheless, the resulting corpora for many African languages are typically small and of poor quality (Kreutzer et al., 2022). Other cleaner parallel sources are mostly from religious sources, like the Bible covering over 1600 languages (McCarthy et al., 2020) and JW300 (Agić and Vulić, 2019) from `JW.org` with over 343 languages, including over 100 African languages. Apart from the training dataset, evaluation datasets are needed to test the performance of multilingual MT models. The FLORES-101 (Goyal et al., 2022) evaluation set, sourced from Wikipedia and manually translated, covers the largest number of languages, including 20 African languages. Finally, while other evaluation datasets for translating into or from African languages have been developed (Ali, Caines, and Malavi, 2021; Azunre et al., 2021a; Emezue and Dossou, 2020; Gezmu, Nürnberger, and Bati, 2021; Nyoni and Bassett, 2021; Siminyu et al., 2021), unfortunately there are only a few African languages with evaluation datasets in the news domain (Adelani et al., 2021c; Ezeani et al., 2020; Mabuya, Abbott, and Marivate, 2021) but ours covers 11 African languages (§10.4).

---

[2] All data, models and code are publicly available on https://github.com/masakhane-io/lafand-mt under academic license.

LOW-RESOURCE MT. Interest in low-resource MT has been increasing both within the MT research community (Haddow et al., 2021), as well as in native speaker communities (Azunre et al., 2021b; Mager et al., 2021; ∀ et al., 2020). On the modeling side, many techniques have been developed: unsupervised MT (Lample et al., 2018a) leverages monolingual data, single multilingual models capable of translating between many languages (Aharoni, Johnson, and Firat, 2019; Fan et al., 2021; Firat, Cho, and Bengio, 2016; Johnson et al., 2017b), multilingual unsupervised models leverage a related language (with parallel data) to assist translating the low-resource language that might not even have any monolingual data (Ko et al., 2021). Unfortunately, unsupervised MT typically performs poorly on low-resource languages (Marchisio, Duh, and Koehn, 2020).

Transfer learning from high-resource languages has achieved more promising results: Transfer from multilingual pre-trained language models (PLM), like mBART50 (Tang et al., 2020) and MT5 (Xue et al., 2021), and large-scale multilingual MT often outperforms bilingual MT (Tran et al., 2021; Yang et al., 2021). For low-resource languages this strategy outperforms the baseline (Transformer) models (Adelani et al., 2021c; Birch et al., 2021; Lee et al., 2022). The performance can be further improved by large scale pre-training (Emezue and Dossou, 2021; Reid et al., 2021).

## 10.3 FOCUS LANGUAGES AND THEIR DATA

FOCUS LANGUAGES. We focus on 21 African languages with varying quantities of available data (Joshi et al., 2020), including moderately low-resource languages such as Swahili and Hausa, and very low-resource languages such as Ghomálá'[3] with the Bible being its largest available corpus. Table 10.1 provides an overview of the focus languages, including their language families, location and the number of speakers. The languages are from four language families: Afro-Asiatic (e.g. Hausa), Nilo-Saharan (e.g. Luo), English Creole (e.g. Nigerian-Pidgin/Naija) and Niger-Congo. Most of the languages (17 out of 21) are from the Niger-Congo family, which is the largest language family in Africa. Six of the languages are predominantly spoken in Francophone countries of Africa, while the remainder are predominantly spoken in Anglophone countries of Africa. In contrast to previous work (Gowda et al., 2021; ∀ et al., 2020), we do not focus exclusively on translation to/from English since this is not the primary language of the

---

3 Spoken by an estimated 1.1M people in Cameroon

| Target Language | Family | African Region | No. of Speakers |
|---|---|---|---|
| Amharic (amh) | Afro-Asiatic / Semitic | East | 33M |
| Bambara (bam) | NC / Manding | West | 14M |
| Ghomálá' (bbj) | NC / Grassfields | Central | 1M |
| Éwé (ewe) | NC / Kwa | West | 7M |
| Fon (fon) | NC / Volta-Niger | West | 2M |
| Hausa (hau) | Afro-Asiatic / Chadic | West | 63M |
| Igbo (ibo) | NC / Volta-Niger | West | 27M |
| Kinyarwanda (kin) | NC / Bantu | East | 12M |
| Luganda (lug) | NC / Bantu | East | 7M |
| Luo (luo) | Nilo-Saharan | East | 4M |
| Mossi (mos) | NC / Gur | West | 8M |
| Naija (pcm) | English-Creole | West | 75M |
| Chichewa (nya) | NC / Bantu | East & Central | 14M |
| chiShona (sna) | NC / Bantu | East & Central | 12M |
| Swahili (swa) | NC / Bantu | East & Central | 98M |
| Setswana (tsn) | NC / Bantu | South | 14M |
| Akan/Twi (twi) | NC / Kwa | West | 9M |
| Wolof (wol) | NC / Senegambia | West | 5M |
| Yorùbá (yor) | NC / Volta-Niger | West | 42M |
| isiXhosa (xho) | NC / Volta-Niger | West | 9M |
| isiZulu (zul) | NC / Bantu | South | 27M |

Table 10.1: **Language, Language family (NC: Niger-Congo), and number of speakers**

Francophone Africa community. All languages are spoken by at least one million speakers.

LANGUAGE CHARACTERISTICS. All languages are written in Latin script, using letters of the basic Latin alphabet with a few omissions (e.g "c", "q", "x", "z") and additions (e.g. "ɛ", "ɔ", "ŋ", "ọ", including digraphs like "gb", "kp", "gh", and sometimes more than two-character letters). 17 of the languages are tonal, and about nine make use of diacritics. Many African languages are morphologically rich. For example, all Bantu languages are agglutinative. Fon, Mossi, and Yorùbá are highly isolating. All languages follow the Subject-Verb-Object sentence structure like English and French. Section 2.2 provides more details.

EXISTING PARALLEL CORPORA. We curate publicly available parallel data for our focus languages, which consists primarily of text in the religious domain. For most African languages, the largest available parallel corpora is JW300 (Agić and Vulić, 2019), sourced from jw.org, which publishes biblical texts as well as lifestyle and opinion columns. Varying quantities of data are available for 16 of the 21 focus languages. Amharic, Éwé, Igbo, Kinyarwanda, Chichewa, chiShona, Swahili, Setswana, Twi, isiXhosa, Yorùbá, and isiZulu have over 400K parallel sentences. Hausa and Mossi have slightly more than 200K parallel sentences, while Fon and Naija have around 30K sentences. For the remaining five languages that are not in the JW300 corpus,[4] we make use of the Bible.[5] We aligned the sentences automatically by the verses (around 31k in total). Ghomálá' only has the New Testament with 8k verses. Bambara and Wolof are missing some verses and books, leading to a total size of 28K and 22K. Table 10.2 summarizes this information about the religious (REL) corpora.

| Target Language | Source Lang. | NEWS Source | Split Sizes | REL Source | Total Size |
|---|---|---|---|---|---|
| Amharic (amh) | English | Global Voices | –/ 899/ 1037 | JW300 | 667K |
| Bambara (bam) | French | Maliweb.net | 3302/ 1484/ 1600 | Bible | 28K |
| Ghomálá' (bbj) | French | Cameroun Web | 2232/ 1133/ 1430 | Bible | 8K |
| Éwé (ewe) | French | Benin Web TV | 2026/ 1414/ 1563 | JW300 | 618K |
| Fon (fon) | French | ORTB, Nation, Héraut, Matin Libre, LB Libéré, LE Précis, Visages. | 2637/ 1227/ 1579 | JW300 | 32K |
| Hausa (hau) | English | WMT2021: Khamenei.v1, Premium Times, Global Voices | 5865/ 1300/ 1500 | JW300 | 236K |
| Igbo (ibo) | English | (Ezeani et al., 2020) | 6998/ 1500/ 1500 | JW300 | 415K |
| Kinyarwanda (kin) | English | Voice of America | –/ 460/ 1006 | JW300 | 485K |
| Luganda (lug) | English | Independent Uganda | 4075/ 1500/ 1500 | Bible | 31K |
| Luo (luo) | English | Lolwe, Standard Media | 4262/ 1500/ 1500 | Bible | 31K |
| Mossi (mos) | French | Burkina24, Lefaso | 2287/ 1478/ 1574 | JW300 | 216K |
| Chichewa (nya) | English | Voice of America | –/ 483/ 1004 | JW300 | 775K |
| Naija (pcm) | English | Daily Trust Nigeria | 4790/ 1484/ 1564 | JW300 | 23K |
| Shona (sna) | English | Voice of America | –/ 556/ 1005 | JW300 | 761K |
| Swahili (swa) | English | Global Voices, OPUS | 30782/ 1791/ 1835 | JW300 | 872K |
| Setswana (tsn) | English | SABC News | 2100/ 1340/ 1500 | JW300 | 870K |
| Akan/Twi (twi) | English | StarrFM, Citi News | 3337/ 1284/ 1500 | JW300 | 601K |
| Wolof (wol) | French | Seneweb, Jotna, Yerim Post, Socialnetlink | 3360/ 1506/ 1500 | Bible | 22K |
| Yorùbá (yor) | English | (Adelani et al., 2021c) | 6644/ 1544/ 1558 | JW300 | 460K |
| isiXhosa (xho) | English | Voice of America | –/ 486/ 1002 | JW300 | 991K |
| isiZulu (zul) | English | (Mabuya, Abbott, and Marivate, 2021) | 3500/ 1239/ 998 | JW300 | 667K |

Table 10.2: **Data Details for MAFAND-MT Corpus**. Language, news source, news (NEWS), and religious domain (REL) data split. The languages highlighted in gray did not previously have news-domain data before MAFAND-MT.

## 10.4 MAFAND-MT AFRICAN NEWS CORPUS

### 10.4.1 Data Collection Process

We introduce our newly translated news corpus; MAFAND-MT—**M**asakhane **A**nglo & **F**ranco **A**frica **N**ews **D**ataset for **M**achine **T**ranslation. Table 10.2 gives the news source and data splits for 15 African languages which includes six languages (bam, bbj, ewe, fon, mos, wol) spoken predominantly in Francophone Africa and 9 languages (kin, lug, luo, nya, pcm, sna, tsn, twi, xho) spoken predominantly in Anglophone Africa. About 4 languages only have `DEV` and `TEST` sets which are about 500 and 1000 sentences respectively, the languages are: `kin`, `nya`, `sna`, and `zul`. The MAFAND-MT corpus was created in three steps:

1. **Crawling and preprocessing** of news websites from local newspapers that are publishing in English and French. Raw texts from the web were segmented into sentences. Most languages were crawled from one or two sites, except for Wolof and Fon that were crawled from four and seven news websites respectively due to local French language newspapers having very few articles. We also ensured that the articles came from a variety of topics e.g. politics, sports, culture, technology, society, religion, and education. This was carried out by native speakers of the target language with source language proficiency.

2. **Translation** of 1.5k–8k sentences by professional translators. The translation process took one to four months depending on the availability of the translators.

3. **Quality control** was provided by native speakers, who discussed and, if possible, fixed problematic translations and ran automatic checks to detect misspellings, duplicated sentences, and alignment problems.

---

4 Some languages like Luo and Luganda are covered by JW300 but are no longer available at the time of paper writing.
5 Crawled/downloaded from https://ebible.org/, except for Bambara that we obtained from https://live.bible.is/ and Ghomálá' from www.beblia.com

Following the recommendations of ∀ et al. (2020), we design the process to be *participatory*: Everyone involved in the corpus creation is a native speaker of the respective target languages and has societal knowledge about the communities that speak those languages. This is particularly important for curation and quality control to ensure that the resulting material is appropriate and relevant for stakeholders of the final MT models (Kreutzer et al., 2022; ∀ et al., 2020). Furthermore, everyone received appropriate remuneration. To enable cross-disciplinary knowledge transfer between participants in the individual steps, every language was assigned a coordinator. The coordinator conducted the initial curation in the first step, and communicated with translators and quality checkers throughout the following steps.

OTHER AVAILABLE PARALLEL CORPORA. We found six African languages with available parallel texts in the news domain: Amharic[6], Hausa[7], Igbo (Ezeani et al., 2020), Swahili[8], Yorùbá (Adelani et al., 2021c), and isiZulu (Mabuya, Abbott, and Marivate, 2021). Table 10.2 provides news source, the `TRAIN`, `DEV` and `TEST` splits. Section 10.4.2 provides details on the pre-processing of the available news corpora.

10.4.2  *Available Parallel Corpora*

We found six African languages with publicly available parallel texts in the news domain: Amharic, Hausa, Igbo, Swahili, Yorùbá, and isiZulu.

AMHARIC  We combined the Global Voices corpus on OPUS (Tiedemann, 2012) with new articles from the Global Voices website[9]. In total, we have 1,936 parallel sentences that we divide into `DEV` and `TEST` splits.

---

6  https://opus.nlpl.eu/GlobalVoices.php
7  https://www.statmt.org/wmt21/translation-task.html
8  https://sw.globalvoices.org/
9  https://am.globalvoices.org/

HAUSA    The Hausa Khamenei[10] corpus contains 5,898 sentences, we split them into `TRAIN` (3,098), `DEV` (1,300), and `TEST` split (1,500). We noticed that this dataset was created in Iran, which is not the geographical location of Hausa speakers. To diversify the texts, we decided to add 2767 *newly translated* sentences from Global Voices and Premium times news websites which covers more Nigerian and West African news – which is the location of native speakers of Hausa. In total, the training sentences increased to 5,865.

IGBO    The Igbo corpus (Ezeani et al., 2020) has 9,998 sentences, we extract 6,998 sentences for `TRAIN`, and the remaining for `DEV` and `TEST` splits.

SWAHILI    The Global Voices corpus on OPUS contains 30,782 sentences, which we use for the `TRAIN` split. We additionally crawled newer (2019–2021) publications of Swahili articles from the Global Voices website[11], this gives a total of 3,626 sentences, they were aligned and manually verified by Swahili speakers. They are split into the `DEV` and `TEST` splits.

YORÙBÁ    The MENYO-20k (Adelani et al., 2021c) corpus contains sentences from different domains (TED talks, books, software localization, proverbs, and news), from which we select the news domain sentences for the `TRAIN`, `DEV` and `TEST` splits.

ISIZULU    The Umsuka corpus (Mabuya, Abbott, and Marivate, 2021) contains 9,703 training sentences and 1,984 evaluation sentences. 4,739 training sentences were translated from English-isiZulu, and the remaining from isiZulu-English. We only keep the training sentences translated into isiZulu, and split them into 3,500 for `TRAIN` and 1,239 sentences for `DEV`. From the existing evaluation set we select only the 998 English-isiZulu translations for `TEST`. Umsuka provides two translations for each English sentence, but we use only the first.

### 10.4.3 *Monolingual News Corpus*

To adapt available multilingual pre-trained models via continued pre-training to African languages, we curated texts from the 17 highest-resourced African languages and three non-native African languages that are widely spoken on the continent (Arabic, English, and French). The selection of African languages is based on their coverage in mC4 (Xue et al., 2021), AfriBERTa corpora (Ogueji, Zhu, and Lin, 2021), and other publicly available news websites like VOA and BBC. We limited the size of the corpus extracted from mC4 to the first 30 million sentences (roughly 1GB of data) for Afrikaans, Amharic, Arabic, English, French, and Swahili. In total, we collected about 12.3 GB of data. Table 10.3 provides data source and size of the pre-training corpus. The African languages pre-trained are: Afrikaans, Amharic, Hausa, Igbo, Malagasy, Chichewa, Oromo, Naija, Kinyarwanda, Kirundi, Shona, Somali, Sesotho, Swahili, isiXhosa, Yorùbá, and isiZulu.

---

10 https://www.statmt.org/wmt21/translation-task.html

11 https://sw.globalvoices.org/

| Language | Source | Size (MB) | No. of sentences |
| --- | --- | --- | --- |
| Afrikaans (afr) | mC4 (subset) (Xue et al., 2021) | 752.2MB | 3,697,430 |
| Amharic (amh) | mC4 (subset), and VOA | 1,300MB | 2,913,801 |
| Arabic (ara) | mC4 (subset) | 1,300MB | 3,939,375 |
| English (eng) | mC4 (subset), and VOA | 2,200MB | 8,626,571 |
| French (fra) | mC4 (subset), and VOA | 960MB | 4,731,196 |
| Hausa (hau) | mC4 (all), and VOA | 594.1MB | 3,290,382 |
| Igbo (ibo) | mC4 (all), and AfriBERTa Corpus (Ogueji, Zhu, and Lin, 2021) | 287.5MB | 1,534,825 |
| Malagasy (mg) | mC4 (all) | 639.6MB | 3,304,459 |
| Chichewa (nya) | mC4 (all), Chichewa News Corpus (Siminyu et al., 2021) | 373.8MB | 2,203,040 |
| Oromo (orm) | AfriBERTa Corpus, and VOA | 67.3MB | 490,399 |
| Naija (pcm) | AfriBERTa Corpus, and VOA | 54.8MB | 166,842 |
| Rwanda-Rundi (kir/kin) | AfriBERTa Corpus, KINNEWS & KIRNEWS (Niyongabo et al., 2020), and VOA | 84MB | 303,838 |
| Shona (sna) | mC4 (all), and VOA | 545.2MB | 2,693,028 |
| Somali (som) | mC4 (all), and VOA | 1,000MB | 3,480,960 |
| Sesotho (sot) | mC4 (all) | 234MB | 1,107,565 |
| Swahili (swa) | mC4 (all) | 823.5MB | 4,220,346 |
| isiXhosa (xho) | mC4 (all), and Isolezwe Newspaper | 178.4MB | 832,954 |
| Yorùbá (yor) | mC4 (all), Alaroye News, Asejere News, Awikonko News, BBC, and VON (Adelani et al., 2021b) | 179.3MB | 897,299 |
| isiZulu (zul) | mC4 (all), and Isolezwe Newspaper | 700.7MB | 3,252,035 |

Table 10.3: Monolingual Corpora (after pre-processing – we followed AfriBERTa (Ogueji, Zhu, and Lin, 2021) approach), their sources and size (MB), and number of sentences.

## 10.5 MODELS AND METHODS

### 10.5.1 *Baseline Models*

We experiment with pre-trained multilingual models and our own bilingual MT baselines. We focus on pre-trained models that are approximately 500M parameters, both for computational feasibility and comparability across various different models.

TRANSFORMER BASELINE. We train Transformer (Vaswani et al., 2017) sequence-to-sequence models from scratch for each language pair using JoeyNMT (Kreutzer, Bastings, and Riezler, 2019). We tokenize the bitext using a joint SentencePiece[12] unigram model (Kudo, 2018), with a character coverage of 1.0 and a maximum sentence length of 4096 tokens and create a vocabulary of 10K subwords. Models are trained on the concatenation of REL and NEWS corpora for each language.

PRE-TRAINED MODELS. We consider three language models, MT5 (Xue et al., 2021), ByT5 (a token-free T5) (Xue et al., 2022), mBART50 (Tang et al., 2020), and the multilingual translation model M2M-100 (Fan et al., 2021) for our experiments. We use MT5-base and ByT5-base, and M2M-100 with 418M parameters. Table 10.4 gives the pre-trained model size, number of African languages covered, and the focus languages supported.

---

[12] https://github.com/google/sentencepiece

| Pre-trained Model (PM) | PM Size | # African Lang. | Focus languages covered |
|---|---|---|---|
| MT5/ByT5 | 580M | 13 | amh, hau, ibo, nya, sna, swa, yor, xho, zul |
| Afri[*T5] | 580M | 17 | amh, hau, ibo, kin, nya, sna, pcm, swa, xho, yor, zul |
| mBART50 | 610M | 2 | swa |
| AfriMBART | 610M | 17 | amh, hau, ibo, kin, nya, sna, pcm, swa, xho, yor, zul |
| M2M-100 | 418M | 17 | amh, hau, ibo, lug, swa, tsn, wol, xho, yor, zul |

Table 10.4: **Language coverage and size for pre-trained models**. Afri[*T5] refers to AfriMT5/ByT5.

10.5.2 *Transfer Learning Across Languages*

We describe two methods for adding new languages to existing models: continual pre-training and many-to-many multilingual translation.

CONTINUAL PRE-TRAINING. The effectiveness of PLMs is limited on extremely low-resource languages because they rarely, if ever, occur in the pre-training corpus (Liu, Winata, and Fung, 2021; Wang et al., 2020b). As shown in Table 10.4, even for MT5 and M2M-100, which cover 100 languages, less than half of the African languages under study are included. To adapt the existing PLMs to our languages corpora and domains, we apply continual pre-training (Gururangan et al., 2020; Liu, Winata, and Fung, 2021) using our collected monolingual corpus. Specifically, before fine-tuning on the parallel MT data, models are pre-trained with their original training objective and vocabulary[13] on the monolingual corpus. Pre-training parameters can be found in the appendix. We refer to the models adapted to African languages as AfriMT5, AfriByT5, and AfriMBART.

MANY-TO-MANY TRANSLATION. We fine-tuned M2M-100 for African multilingual translation to create English- and French-centric models. For the English-centric model, the M2M-100 model was fine-tuned on the news data for en–{hau, ibo, lug, luo, pcm, swa, tsn, twi, yor, zul} while the French-centric model is trained on fr–{bam, bbj, ewe, fon, mos, wol}. Languages not included in the pre-trained M2M-100 model were assigned the language code of a language included in M2M-100 but excluded from our study.

10.5.3 *Transfer Learning Across Domains*

As there is very limited MT data on the news domain, we compare different methods that combine the *large* data from the religious domain (REL) and the *small* data from the NEWS domain (NEWS) to fine-tune M2M-100:

---

13 Changing the vocabulary (Gururangan et al., 2020) to fit the languages, or adding MT-focused training objectives for word alignment (Liu, Winata, and Fung, 2021) can potentially improve the performance further, which we leave for future work.

|  | | | *fr-xx* | | | | | | | *en-xx* | | | | | | | |
| Model | bam | bbj | ewe | fon | mos | wol | hau | ibo | lug | luo | pcm | swa | tsn | twi | yor | zul | AVG |
| --- | --- | --- | --- | --- | --- | --- | --- | --- | --- | --- | --- | --- | --- | --- | --- | --- | --- |
| | | | | | | **BLEU** | | | | | | | | | | | |
| M2M-100 0-shot | – | – | – | – | – | 1.3 | 0.4 | 2.8 | – | – | – | 20.1 | 1.1 | – | 2.1 | 5.6 | – |
| MT5 | 1.5 | 0.4 | 2.2 | 1.6 | 0.1 | 0.9 | 4.8 | 18.0 | 3.0 | 3.1 | 34.1 | 25.1 | 3.4 | 1.7 | 4.8 | 11.7 | 7.3 |
| AfriMT5 | 2.1 | 0.8 | 3.7 | 2.5 | 0.1 | 1.8 | 6.9 | 19.6 | 5.2 | 4.6 | **35.0** | 26.7 | 7.0 | 2.7 | 6.2 | 13.2 | 8.6 |
| ByT5 | 9.5 | 1.8 | 5.5 | 3.8 | 0.1 | 6.0 | 9.3 | 21.8 | 12.1 | 8.4 | 30.1 | 24.4 | 14.7 | 6.0 | 7.5 | 14.0 | 10.9 |
| AfriByT5 | 11.4 | 2.2 | 5.2 | 3.7 | 0.2 | 6.4 | 10.3 | 22.7 | 13.1 | 8.9 | 30.0 | 24.7 | 17.0 | 6.1 | 7.6 | 15.3 | 11.5 |
| mBART50 | 18.6 | 2.4 | 5.3 | 6.2 | 0.8 | 9.7 | 13.9 | 21.1 | 12.0 | 10.0 | 34.1 | 25.8 | 16.8 | 7.5 | 10.0 | **21.2** | 13.5 |
| AfriMBART | 15.3 | 2.4 | 5.7 | 4.4 | 0.6 | 8.6 | 12.7 | 22.4 | 10.0 | 9.8 | 30.0 | 22.7 | 12.8 | 6.3 | 9.6 | 20.1 | 12.1 |
| M2M-100 | **22.7** | **2.9** | **6.4** | **7.1** | **1.0** | **12.4** | **16.3** | **24.7** | **14.3** | **11.5** | 33.9 | **26.7** | 24.7 | **8.8** | **12.8** | 21.0 | **15.5** |
| M2M-100-EN/FR | 18.5 | 2.2 | 6.2 | 4.3 | 0.8 | 10.6 | 7.0 | 22.4 | 8.9 | 9.5 | 34.9 | 26.4 | 19.7 | 7.0 | 5.6 | 15.6 | 12.5 |
| | | | | | | **CHRF** | | | | | | | | | | | |
| M2M-100 0-shot | – | – | – | – | – | 4.3 | 12.4 | 19.0 | – | – | – | 47.7 | 8.7 | – | 10.4 | 20.1 | |
| MT5 | 10.0 | 7.4 | 9.7 | 11.5 | 7.9 | 9.1 | 28.6 | 41.1 | 24.9 | 21.6 | 64.1 | 53.7 | 22.8 | 17.8 | 20.8 | 36.0 | 24.2 |
| AfriMT5 | 14.0 | 12.7 | 16.6 | 14.8 | 8.2 | 13.8 | 33.5 | 43.1 | 30.4 | 25.7 | **64.7** | 55.1 | 31.5 | 21.5 | 24.3 | 40.3 | 28.1 |
| ByT5 | 27.8 | 17.7 | 23.8 | 16.1 | 8.8 | 22.9 | 33.9 | 46.5 | 40.0 | 32.2 | 58.1 | 52.5 | 38.6 | 27.9 | 25.5 | 40.3 | 32.0 |
| AfriByT5 | 31.4 | 19.9 | 24.1 | 16.5 | 9.8 | 23.8 | 35.3 | 47.4 | 42.2 | 33.6 | 58.0 | 52.8 | 42.1 | 29.0 | 26.0 | 42.9 | 33.4 |
| mBART50 | 42.3 | 22.0 | 27.7 | 25.7 | 16.0 | 31.9 | 41.9 | 45.9 | 41.1 | 36.7 | 64.2 | 54.4 | 43.0 | 35.6 | 31.1 | 50.2 | 38.1 |
| AfriMBART | 40.4 | 20.1 | 26.9 | 24.1 | 15.1 | 30.9 | 41.8 | 47.4 | 38.6 | 36.7 | 54.9 | 52.7 | 40.3 | 34.2 | 31.1 | 49.3 | 36.5 |
| M2M-100 | **48.2** | **23.1** | **30.9** | **27.6** | **16.7** | **35.7** | **44.0** | **50.0** | **45.5** | **39.0** | 64.0 | **56.4** | **52.0** | **38.2** | **35.9** | **51.2** | **41.2** |
| M2M-100-EN/FR | 43.4 | 20.6 | 29.4 | 23.2 | 16.3 | 32.8 | 33.3 | 46.9 | 38.8 | 36.5 | 64.5 | 55.4 | 47.1 | 33.6 | 25.3 | 42.9 | 36.9 |

Table 10.5: **Results adding African Languages to Pre-Trained Models, en/fr-xx**. We calculate BLEU and CHRF on the news domain when training on only NEWS data from MAFAND-MT.

1. REL+NEWS: Fine-tuning on the aggregation of REL and NEWS.
2. REL→NEWS: Training on REL, followed by fine-tuning on NEWS.
3. REL+NEWS→NEWS: REL+NEWS, followed by additional fine-tuning on NEWS.

Each fine-tuning stage lasts for three epochs. We evaluate translation quality with BLEU (Papineni et al., 2002) using SacreBLEU (Post, 2018)[14] and ChrF (Popović, 2015).

### 10.5.4 *Model Hyper-parameters and Reproducibility of Results*

For the pre-trained models, we fine-tune the models using HuggingFace transformer tool (Wolf et al., 2020) with the default learning rate ($5e-5$), batch size of 10, maximum source length & maximum target length of 200, beam size of 10, and number of epochs is 3 except for models trained on only NEWS which we set to 10. We make All the experiments were performed on a single GPU (Nvidia V100).

For fine-tuning pre-trained models, especially for mBART50 that only supports two African languages, the target language is required to be specified during decoding from among those that the model has seen during pre-training, we follow past works (Cahyawijaya et al., 2021; Lee et al., 2022; Madaan, Sharma, and Singla, 2020) in selecting another closely-related language that is represented in the pre-trained model. For convenience, we make use of Swahili (sw) as the target language when an African language is not represented since Swahili is represented in all the pre-trained models. The only exception is Nigerian-Pidgin, where we make use of French (fr) since it is closely related to English. When a language is represented in the pre-trained model like M2M-100 has seen Yorùbá (yo), we make use of the correct language code.

---

[14] "intl" tokenizer, all data comes untokenized.

| | *xx-fr* | | | | | | *xx-en* | | | | | | | | | | |
|---|---|---|---|---|---|---|---|---|---|---|---|---|---|---|---|---|---|
| Model | bam | bbj | ewe | fon | mos | wol | hau | ibo | lug | luo | pcm | swa | tsn | twi | yor | zul | AVG |
| | | | | | | | **BLEU** | | | | | | | | | | |
| M2M-100 0-shot | – | – | – | – | – | 0.8 | 2.2 | 6.4 | – | – | – | 25.2 | 3.3 | – | 3.0 | 13.8 | – |
| MT5 | 2.5 | 0.9 | 1.1 | 2.4 | 0.7 | 1.3 | 8.5 | 18.9 | 12.6 | 6.4 | 42.2 | 29.5 | 9.5 | 4.6 | 12.3 | 22.4 | 11.0 |
| AfriMT5 | 6.4 | 2.0 | 2.1 | 4.2 | 1.2 | 2.9 | 11.7 | 19.5 | 15.5 | 9.7 | 44.6 | 30.6 | 16.1 | 8.4 | 13.8 | 24.0 | 13.3 |
| ByT5 | 10.0 | 2.7 | 4.1 | 4.9 | 1.5 | 7.2 | 13.4 | 21.0 | 19.8 | 12.1 | 39.4 | 27.1 | 18.6 | 9.8 | 11.5 | 22.8 | 14.1 |
| AfriByT5 | 13.8 | 4.4 | 4.5 | 5.8 | 2.2 | 9.0 | 13.8 | 20.7 | **21.1** | 12.5 | 39.5 | 27.0 | 19.7 | 10.5 | 11.9 | 24.0 | 15.1 |
| mBART50 | 6.8 | 0.3 | 1.7 | 0.8 | 0.6 | 6.3 | 12.1 | 13.2 | 14.5 | 9.1 | 44.2 | 29.0 | 2.0 | 0.5 | 8.1 | 31.1 | 11.3 |
| AfriMBART | 8.1 | 2.3 | 3.0 | 4.5 | 1.7 | 3.2 | 12.9 | 15.5 | 13.1 | 8.0 | 43.7 | 29.2 | 7.2 | 6.5 | 9.5 | 33.0 | 12.6 |
| M2M-100 | **22.1** | **5.4** | 6.9 | 8.4 | **2.8** | 10.3 | **16.9** | 19.0 | 20.0 | 13.0 | 43.8 | 29.8 | 20.0 | 10.9 | **16.0** | **37.8** | **17.7** |
| M2M-100-EN/FR | **22.1** | 5.1 | **7.4** | **9.1** | 2.1 | **10.5** | 11.4 | 20.3 | 19.8 | **14.0** | **45.2** | 30.0 | **21.4** | **11.7** | 13.4 | 9.5 | 15.8 |
| | | | | | | | **CHRF** | | | | | | | | | | |
| M2M-100 0-shot | – | – | – | – | – | 12.3 | 23.7 | 29.7 | – | – | – | 51.6 | 21.1 | – | 18.3 | 35.7 | – |
| MT5 | 19.4 | 15.1 | 17.0 | 17.9 | 10.9 | 16.2 | 31.0 | 43.5 | 36.3 | 26.1 | 66.9 | 53.7 | 32.2 | 25.2 | 31.1 | 43.9 | 30.4 |
| AfriMT5 | 27.7 | 19.6 | 21.1 | 21.4 | 13.2 | 21.6 | 35.1 | 44.9 | 40.2 | 32.2 | 68.4 | 54.5 | 39.6 | 31.2 | 33.9 | 45.9 | 34.4 |
| ByT5 | 31.2 | 21.8 | 24.8 | 20.5 | 15.4 | 26.2 | 34.4 | 46.4 | 45.4 | 34.1 | 62.0 | 50.6 | 42.4 | 32.9 | 31.4 | 42.5 | 35.1 |
| AfriByT5 | 34.8 | 25.5 | 24.9 | 22.0 | 16.2 | 29.3 | 35.0 | 46.4 | **47.1** | 35.0 | 62.1 | 50.5 | 43.4 | 33.4 | 32.0 | 43.7 | 36.3 |
| mBART50 | 26.0 | 17.1 | 20.9 | 20.2 | 17.1 | 26.6 | 32.3 | 37.9 | 39.0 | 31.0 | 68.2 | 53.5 | 20.1 | 19.4 | 26.7 | 49.0 | 31.6 |
| AfriMBART | 31.4 | 22.9 | 27.2 | 26.3 | 17.0 | 25.0 | 36.7 | 42.0 | 40.4 | 29.8 | 67.8 | 53.5 | 31.4 | 30.6 | 30.0 | 51.7 | 35.2 |
| M2M-100 | **45.9** | 26.5 | 30.9 | 27.5 | **17.7** | 33.8 | **39.4** | 46.1 | 46.4 | 36.7 | 68.6 | 54.8 | 45.2 | 35.1 | **38.1** | **55.5** | **40.5** |
| M2M-100-EN/FR | 45.6 | **26.9** | **32.2** | **28.7** | 17.0 | **34.3** | 35.1 | **46.6** | 46.0 | **37.6** | **69.0** | **55.0** | **46.3** | **36.0** | 35.2 | 31.5 | 38.9 |

Table 10.6: **Results adding African Languages to Pre-Trained Models, xx-en/fr**. We calculate BLEU and CHRF on the news domain when training on only `NEWS` data from MAFAND-MT.

To train AfriMT5 and ByT5, we start with MT5 and ByT5. We pre-train with the learning rate $1e-4$, $10,000$ warm up steps and a batch size of 2048 for one epoch. For mBART50, we pre-train with learning rate of $5e-5$ for $50,000$ steps using Fairseq (Ott et al., 2019) without modifying the mBART50 vocabulary. Table 10.11 has the names of all the models that are publicly available on HuggingFace Model Hub[15]. In total, we have 365 models from 22 x 16 bilingual models (for languages with training set), 8 models (for languages with only dev/test split), two English/French-centric models, and three adapted models to African languages (i.e AfriMT5, AfriByT5, and AfriMBART).

## 10.6 RESULTS AND DISCUSSION

We successfully adapt several multilingual pre-trained models to previously unseen African languages and quantify the effectiveness of small in-domain

---

[15] https://huggingface.co/masakhane

| | *fr-xx* | | | | | | *en-xx* | | | | | | | | | | AVG |
|---|---|---|---|---|---|---|---|---|---|---|---|---|---|---|---|---|---|
| **Model** | bam | bbj | ewe | fon | mos | wol | hau | ibo | lug | luo | pcm | swa | tsn | twi | yor | zul | |
| | | | | | | **BLEU** | | | | | | | | | | | |
| Transformer | | | | | | | | | | | | | | | | | |
| REL+NEWS | 7.3 | 0.1 | 6.2 | 2.9 | 2.1 | 3.1 | 10.7 | 22.4 | 4.6 | 3.7 | 11.7 | 26.2 | 28.1 | 8.7 | 9.7 | 16.5 | 10.2 |
| REL→NEWS | 5.1 | 0.2 | 5.4 | 2.8 | 1.7 | 2.3 | 11.7 | 22.7 | 3.9 | 3.3 | 11.9 | 26.3 | 29.7 | 8.7 | 8.4 | 20.3 | 10.3 |
| REL+NEWS→NEWS | 8.5 | 0.3 | 6.5 | 3.2 | **2.2** | 3.7 | 12.0 | 23.6 | 5.1 | 4.3 | 13.8 | 26.6 | 29.3 | 9.0 | 9.7 | 20.1 | 11.1 |
| M2M-100 | | | | | | | | | | | | | | | | | |
| REL+NEWS | 23.0 | 2.8 | 7.7 | 6.5 | 0.9 | 11.2 | 13.1 | 24.7 | 13.9 | 11.6 | **35.1** | 23.3 | 29.0 | 9.7 | 12.4 | 18.3 | 15.2 |
| REL→NEWS | 20.3 | **3.1** | 7.7 | **7.5** | 1.1 | 12.0 | 15.1 | **26.0** | 15.4 | 11.9 | 35.0 | **27.7** | **31.9** | 10.0 | 13.4 | **22.9** | 16.3 |
| REL+NEWS→NEWS | **24.7** | **3.1** | **8.9** | 7.4 | 1.1 | **12.7** | **15.5** | 25.8 | **15.7** | **12.0** | 34.2 | 27.3 | **31.9** | **10.2** | **13.9** | 22.6 | **16.7** |
| | | | | | | **CHRF** | | | | | | | | | | | |
| Transformer | | | | | | | | | | | | | | | | | |
| REL+NEWS | 25.6 | 9.6 | 30.6 | 14.5 | 17.7 | 18.9 | 36.7 | 46.7 | 30.5 | 26.4 | 37.8 | 55.3 | 55.0 | 36.7 | 30.6 | 50.0 | 32.7 |
| REL→NEWS | 18.2 | 11.2 | 27.1 | 15.4 | 18.3 | 15.9 | 37.4 | 47.2 | 28.7 | 24.4 | 38.3 | 55.5 | 56.3 | 36.6 | 28.9 | 53.0 | 32.0 |
| REL+NEWS→NEWS | 27.4 | 12.8 | 31.5 | 16.5 | 19.9 | 20.2 | 38.3 | 48.3 | 30.6 | 27.7 | 42.6 | 55.6 | 56.3 | 37.7 | 30.6 | 53.4 | 34.3 |
| M2M-100 | | | | | | | | | | | | | | | | | |
| REL+NEWS | 46.8 | 22.1 | 36.7 | 26.2 | 16.0 | 33.5 | 39.4 | 50.1 | 44.5 | 38.1 | 64.7 | 53.0 | 57.2 | 39.7 | 35.2 | 53.1 | 41.0 |
| REL→NEWS | 44.1 | 22.6 | 34.1 | 27.7 | 16.8 | 34.7 | 42.3 | 51.3 | 45.6 | 38.6 | 64.7 | 57.2 | 59.3 | 40.6 | 37.1 | 56.3 | 42.1 |
| REL+NEWS→NEWS | 49.9 | 23.5 | 37.5 | 28.5 | 16.8 | 35.8 | 42.7 | 51.3 | 46.9 | 39.4 | 64.2 | 57.0 | 59.5 | 40.8 | 37.4 | 56.3 | 43.0 |

Table 10.7: **Results adapting to Domain Shift, en/fr-xx**. We calculate BLEU and ChrF on the news domain when training on different combinations of REL and NEWS.

translation datasets. We discuss the effects of domain shift and analyze mitigation strategies.

### 10.6.1 *Adaptation to the Focus Languages*

We demonstrate that fine-tuning with a few thousand high-quality bitext is effective for adding new languages to pre-trained models. Further, continuing to pre-train to specialize models to African languages further improves performance.

ZERO-SHOT TRANSLATION. Table 10.5 and Table 10.6 gives the result of zero-shot evaluation on NEWS. We evaluate only on the M2M-100 dataset because it has been pre-trained on parallel texts with a few of our focus languages. We observe very poor performance (< 5 BLEU) on the languages except for zul (> 13 BLEU) and swa (> 20 BLEU) in both translation directions. For swa, its likely that the performance is reasonable because M2M-100 has seen more bitext during pre-training (2.4M sentences in CCAligned (El-Kishky et al., 2020)). Other African languages except for Afrikaans have less than 600K sentences in CCAligned, and are also of a lower quality (Kreutzer et al., 2022) which affect overall zero-shot performance.

PERFORMANCE AFTER FINE-TUNING. We found impressive performance after fine-tuning PLMs and M2M-100 on few thousand sentences (mostly 2K–7K sentences, except for swa with 30K sentences), including languages not seen during pre-training. For *en/fr-xx*, MT5 has a poor transfer performance with average BLEU of 7.2, despite being pre-trained on 101 languages. ByT5 outperforms MT5 by over 3 BLEU on average, even though their performances were reported to be similar in previous work (Xue et al., 2022). This indicates that ByT5 might be preferable over MT5 when translating low-resource languages. Surprisingly, mBART50 that was only pre-trained on 50 languages and 2 African languages outperformed MT5 and ByT5 which are pre-trained on 101 languages. Overall, we found M2M-100 to be the best model, most likely because it was pre-trained on a translation task. In general, BLEU scores are relatively low (< 15 BLEU for 9 out of 16 languages for en/fr-xx and 7 in xx-en/fr) even when fine-tuning M2M-100 on in-domain data, which suggests that developing more effective methods for fine-tuning might be a promising future direction. The languages with the best quality according to BLEU on the target side are pcm, swa and tsn, and pcm, zul, and swa on the source side.

|  | *xx-fr* | | | | | | *xx-en* | | | | | | | | | | |
| Model | bam | bbj | ewe | fon | mos | wol | hau | ibo | lug | luo | pcm | swa | tsn | twi | yor | zul | AVG |
| --- | --- | --- | --- | --- | --- | --- | --- | --- | --- | --- | --- | --- | --- | --- | --- | --- | --- |
| | | | | | | | **BLEU** | | | | | | | | | | |
| Transformer | | | | | | | | | | | | | | | | | |
| REL+NEWS | 4.9 | 0.6 | 6.3 | 2.2 | 3.7 | 2.2 | 11.2 | 17.4 | 5.6 | 3.1 | 19.5 | 28.0 | 23.9 | 9.8 | 12.0 | 27.3 | 11.1 |
| REL→NEWS | 4.7 | 0.8 | 6.5 | 2.4 | 3.1 | 2.5 | 11.0 | 17.4 | 6.3 | 1.8 | 19.0 | 27.9 | 24.6 | 10.1 | 11.0 | 28.5 | 11.1 |
| REL+NEWS→NEWS | 5.8 | 1.0 | 7.1 | 2.4 | **4.1** | 2.6 | 13.2 | 18.2 | 6.8 | 3.7 | 21.4 | 28.7 | 24.5 | 10.4 | 12.6 | 30.1 | 12.0 |
| M2M-100 | | | | | | | | | | | | | | | | | |
| REL+NEWS | 24.0 | 5.8 | 10.9 | 9.7 | 2.3 | 10.1 | 15.5 | 21.1 | 21.1 | 13.3 | **44.6** | 29.4 | 27.0 | 12.5 | 17.4 | 30.6 | 18.5 |
| REL→NEWS | 20.3 | 5.9 | 11.4 | 9.6 | 2.3 | 10.5 | 17.4 | **21.9** | 20.6 | 13.7 | 44.3 | **30.6** | 27.7 | **13.2** | **18.0** | 36.0 | 19.0 |
| REL+NEWS→NEWS | **25.8** | **6.3** | **11.6** | **9.9** | 2.6 | **11.5** | **18.2** | 21.5 | **22.4** | **14.3** | 44.0 | 30.5 | **27.8** | **13.2** | **18.0** | **38.1** | **19.7** |
| | | | | | | | **CHRF** | | | | | | | | | | |
| Transformer | | | | | | | | | | | | | | | | | |
| REL+NEWS | 24.7 | 12.6 | 29.4 | 16.1 | **17.6** | 19.9 | 31.7 | 43.1 | 26.9 | 23.0 | 47.8 | 53.5 | 49.8 | 34.4 | 33.4 | 49.6 | 32.1 |
| REL→NEWS | 23.0 | 12.7 | 29.8 | 16.6 | 17.2 | 18.3 | 30.6 | 42.8 | 28.7 | 20.0 | 47.3 | 53.3 | 50.8 | 34.4 | 32.2 | 50.4 | 31.8 |
| REL+NEWS→NEWS | 26.5 | 14.7 | 30.7 | 17.6 | 18.8 | 21.8 | 33.8 | 44.0 | 29.5 | 24.7 | 50.8 | 54.1 | 50.6 | 35.1 | 34.4 | 51.4 | 33.7 |
| M2M-100 | | | | | | | | | | | | | | | | | |
| REL+NEWS | 47.1 | 27.5 | 36.4 | 27.9 | 16.6 | 34.0 | 39.4 | 47.5 | 47.2 | 37.3 | **68.9** | 54.7 | 53.0 | 38.4 | 40.2 | 53.3 | 41.8 |
| REL→NEWS | 44.5 | 27.7 | 37.0 | 28.2 | 16.8 | 34.4 | 42.3 | **48.0** | 47.0 | 38.0 | 68.7 | **55.8** | 53.6 | **38.7** | 40.7 | 56.4 | 42.4 |
| REL+NEWS→NEWS | **49.0** | **28.5** | **37.2** | **28.9** | 17.2 | **35.3** | **42.7** | 47.9 | **48.5** | **38.3** | 68.6 | 55.7 | **54.0** | **38.7** | **41.0** | **57.7** | **43.1** |

Table 10.8: **Results adapting to Domain Shift, xx-en/fr**. We calculate BLEU and ChrF on the news domain when training on different combinations of REL and NEWS.

BLEU scores are higher when translating from an African language, which is expected due to the more frequent exposure to English and French on the target side during pre-training, and BLEU being penalized more for morphologically rich languages like bbj, lug, swa, tsn, and zul). The ChrF metric works better for them. For example, fine-tuning M2M-100 on NEWS and evaluating on zul has a BLEU of 21.0 in *en/fr-xx*, and BLEU of 37.8 in the *xx-en/fr* showing a large gap in performance in both directions. However, with the ChrF, we find a smaller performance gap (51.2 in *en/fr-xx* and 55.5 in the *xx-en/fr*.

CONTINUAL PRE-TRAINING.   We observe an improvement in BLEU when we utilize AfriMT5 and AfriByT5, for languages included in our continual pre-training corpus (see Table 10.3). Other languages also benefit despite not being seen during continual pre-training, possibly due to language similarity. For example, AfriByT5 on *fr-bam* improved by 1.9 BLEU over ByT5 and AfriMT5 on *en-tsn* improved by 3.6 BLEU over MT5. On average, AfriMT5 improved over MT5 by 1.3 BLEU in en/fr-xx and 2.4 BLEU in the *xx-en/fr*. The improvement for AfriByT5 was much smaller: 0.6 and 0.9 BLEU in *en/fr-xx* and *xx-en/fr* translation directions. For AfriMBART, we did not see any improvement on average, only the performance of hau (1.5 BLEU) and ibo (0.7 BLEU) improved in *en/fr-xx* direction. However, in the *xx-en/fr* direction, fon, tsn, twi, and zul improved by 2.7–6.0 BLEU.

MANY-TO-MANY MULTILINGUAL MT. Training on the combined news corpus from all languages that use French or English separately does not appear to help much. We see slight improvements for most languages only in the *xx-en/fr* direction.

10.6.2 *Adaptation to the News Domain*

To improve over the baseline performance on NEWS, we train bilingual Transformer models (as a baseline) and M2M-100 on a combination of REL and NEWS. We chose M2M-100 because it was the best performing model. Table 10.7 gives the BLEU on three settings: REL+NEWS, REL→NEWS, and REL+NEWS→NEWS. In general, the improvement depends on the size of REL corpus. For languages trained on the Bible such as bbj, bam, lug, luo, and wol, the improvement is minimal. For M2M-100, the REL+NEWS performance does not improve over NEWS despite the larger quantity of training data. This demonstrates that increasing the size in the target domain is the most helpful strategy (see Figure 10.5). Similarly, combining REL+NEWS is not very helpful for xx-en/fr. An alternative approach is REL→NEWS, which allows the model to develop a good understanding of the desired language before adapting to the news domain. We observe an increase on 1.1 BLEU over REL+NEWS in the en/fr-xx direction. However, the best strategy is REL+NEWS→NEWS, especially for xx-en/fr where it yields an improvement over NEWS and REL+NEWS by 2.0 and 1.5 BLEU, respectively.

| *bam-fr* | |
| --- | --- |
| SRC | Ani k'a fɔu ye ko cɛmancɛ fanga bɛ sigi ntuloma saba kan. |
| TGT | Et leur dire que la transition se repose sur trois piliers. |
| REL | Et qu'on leur dise que la puissance du milieu est sur trois sauterelles; |
| REL+NEWS→NEWS | Et de leur dire que la force de la transition repose sur trois piliers. |
| *lug-en* | |
| SRC | Murasaki Shikibu yawandiika ekitabo ekijjuvu akaasookera ddala mu nsi yonna. |
| TGT | Murasaki Shikibu wrote the world's first full novel. |
| REL | And Murshach Shikib writes a full scroll of the first in all the earth. |
| REL+NEWS→NEWS | Murasaki Shikibu wrote a complete book first in the world. |

Table 10.9: **Example translations** for M2M-100 fine-tuned on REL or REL+NEWS→NEWS. Terms in red are typical for biblical texts, while the terms in blue are more neutral expressions.

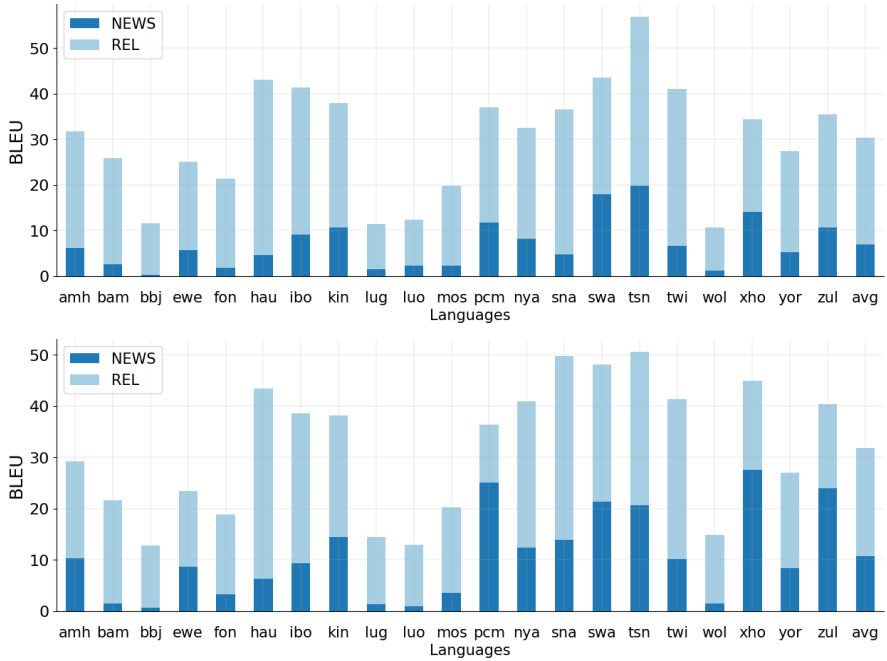

Figure 10.1: **Domain shift** of M2M-100 Transformer models trained on en/fr-xx (top) or xx-en/fr (bottom) REL domain and tested on the NEWS vs. REL domains.

10.6.3 *Analysis of Domain Shift*

IS A SMALL IN-DOMAIN SET ESSENTIAL FOR FINE-TUNING? If we train models *only* on previously available religious data, they are not capable of translating news well due to the strong *domain bias*. This is illustrated in Figure 10.1: All models perform much worse on NEWS than on the REL domain. When the quantity of religious training data is small, the loss in translation performance on the news test set is largest, c.f. bbj (8k of REL data) with a drop of -95.5% BLEU or bam (-93.5%, 28k) and luo (-93.5%, 31k).

This indicates that when the REL training data is sparse, it is insufficient to teach the M2M-100 model a more general understanding required for translating NEWS. However, when the religious training data is larger, this loss is reduced, c.f. when translating to zul (667k, -67%), swa (-69.3%, 872k), and tsn (-71%, 870k). While this is the general trend, pcm, whose religious training data is small (23k), has the lowest drop in performance (-59.3%), which may be due to the strong similarity to its source language.

HOW MANY SENTENCES IN THE TARGET DOMAIN ARE REQUIRED? Figure 10.5 shows how for three selected language pairs with a large (fr-bam), medium (eng-ibo) and relatively small (eng-swa) domain gap, the quality of target domain translations improves as we increase the size of the target domain corpus. For all three pairs, fine-tuning M2M-100 or ByT5 on 2.5*k* sentence pairs of in-domain data (NEWS) is sufficient to outperform the bilingual Transformer baselines that were additionally trained on larger amounts of out-of-domain data (REL). Surprisingly, this procedure not only works for languages included during pre-training (swa), but also for previously unseen languages (ibo, bam). M2M-100 tends to adapt to the new data more quickly than ByT5, but in all cases, models continue to learn with additional in-domain data. This shows how much more effectively a small number of in-domain translations can be used when they serve for fine-tuning multilingual pre-trained models rather than training bilingual MT models from scratch.

| Evaluation Domain | Tuned on NEWS | hau | ibo | lug | luo | swa | wol | yor | zul |
|---|---|---|---|---|---|---|---|---|---|
| *en/fr-xx* | | | | | | | | | |
| FLORES | ✗ | 2.6 | 2.8 | 0.8 | – | 20.9 | 0.6 | 1.5 | 3.3 |
| FLORES | ✓ | 13.0 | 19.9 | 7.6 | 13.7 | 27.1 | 8.2 | 13.4 | 19.2 |
| REL | ✗ | 1.2 | 1.0 | 0.0 | – | 11.0 | 0.0 | 0.4 | 1.6 |
| REL | ✓ | 8.8 | 10.3 | 3.3 | 5.4 | 14.6 | 6.7 | 10.6 | 13.0 |
| NEWS | ✗ | 0.6 | 4.1 | 2.3 | – | 21.4 | 1.2 | 2.4 | 5.6 |
| NEWS | ✓ | 20.8 | 31.6 | 22.6 | 16.4 | 31.4 | 19.9 | 25.5 | 27.6 |
| *xx-en/fr* | | | | | | | | | |
| FLORES | ✗ | 8.0 | 7.2 | 3.7 | – | 26.9 | 3.0 | 3.8 | 11.9 |
| FLORES | ✓ | 14.6 | 12.0 | 7.7 | 11.8 | 25.8 | 7.5 | 9.3 | 19.2 |
| REL | ✗ | 6.4 | 3.7 | 0.5 | – | 15.4 | 0.4 | 0.9 | 8.5 |
| REL | ✓ | 7.2 | 6.0 | 1.7 | 2.5 | 13.9 | 1.7 | 5.7 | 12.5 |
| NEWS | ✗ | 2.6 | 9.1 | 7.2 | – | 27.8 | 1.0 | 3.9 | 15.7 |
| NEWS | ✓ | 17.6 | 22.8 | 24.4 | 15.8 | 32.0 | 12.3 | 17.5 | 39.0 |

Table 10.10: **spBLEU on Wikipedia domain** (FLORES) and REL for M2M-100 before (✗) and after (✓) fine-tuning on NEWS. Baseline evaluation on NEWS is highlighted in gray.

EXAMPLES OF DOMAIN BIAS. To illustrate the challenge of overcoming domain bias, we show examples translating from bam and lug in Table 10.9. The M2M-100 model fine-tuned only on REL succeeds in roughly capturing the meaning of the sources, but using biblical terms, such as "scroll" instead of "novel". Adding our news corpus to fine-tuning resolves these issues (e.g. "book").

HOW GENERAL IS OUR NEWS CORPUS? Table 10.10 shows the zero-shot evaluation of M2M-100 fine-tuned on our small NEWS corpora on other domains: religious (REL) and Wikipedia (FLORES). We evaluated the Wikipedia domain on the FLORES *devtest* and the REL domain on either JW300 or Bible (lug, luo, wol). As a baseline, we evaluated the zero-shot performance of M2M-100 (not fine-tuned, ✗) on FLORES[16] using spBLEU (i.e. sentencepiece BLEU (Goyal et al., 2022)). We noticed very poor performance except for Swahili — as discussed in §10.6.1. After fine-tuning on our new data (✓), transfer is largely improved across the bench (up to +17 BLEU for en-ibo). The same trend holds for the religious domain. This shows that even though our data comes from the news domain, it helped the model generalize to other domains. Hence, expanding African news corpora and developing better MT models for news pays off even for other domains of interest.

---

16 except for Luo which is not supported

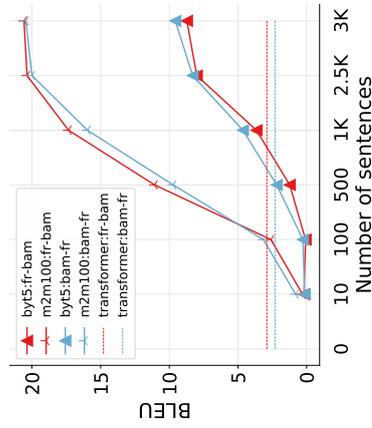

Figure 10.2: eng-swa

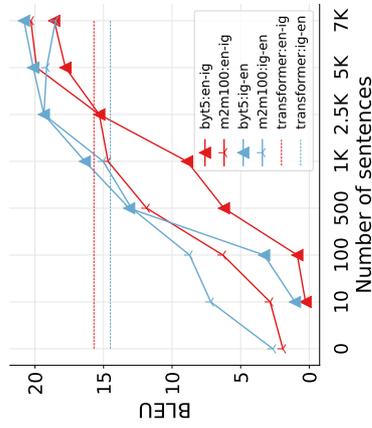

Figure 10.3: eng-ibo

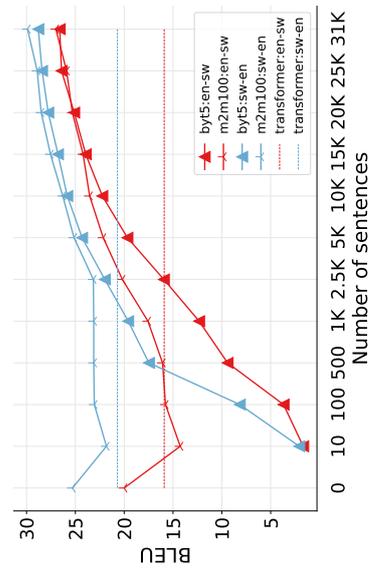

Figure 10.4: fr-bam

Figure 10.5: **Number of fine-tuning sentences** needed to exceed the performance of a bilingual Transformer model.

| Model Name | HuggingFace Model name | Remark |
| --- | --- | --- |
| AfriMT5 | `masakhane/afri-mt5-base` | mT5-base adaptation to 17 African languages, English, French and Arabic. |
| AfriByT5 | `masakhane/afri-byt5-base` | ByT5-base adaptation to 17 African languages, English, French and Arabic. |
| AfriMBART | `masakhane/afri-mbart50` | mBART50 adaptation to 17 African languages, English, French and Arabic. |
| NEWS (MT5) | `masakhane/mt5_{src}_{tgt}_news` | MT5 fine-tuned on {src}-{tgt} direction using parallel NEWS corpus. |
| NEWS (AfriMT5) | `masakhane/afrimt5_{src}_{tgt}_news` | AfriMT5 fine-tuned on {src}-{tgt} direction using parallel NEWS corpus. |
| NEWS (ByT5) | `masakhane/byt5_{src}_{tgt}_news` | ByT5 fine-tuned on {src}-{tgt} direction using parallel NEWS corpus. |
| NEWS (AfriByT5) | `masakhane/afribyt5_{src}_{tgt}_news` | AfriByT5 fine-tuned on {src}-{tgt} direction using parallel NEWS corpus. |
| NEWS (mBART50) | `masakhane/mbart50_{src}_{tgt}_news` | mBART50 fine-tuned on {src}-{tgt} direction using parallel NEWS corpus. |
| NEWS (AfriByT5) | `masakhane/afrimbart_{src}_{tgt}_news` | AfriMBART fine-tuned on {src}-{tgt} direction using parallel NEWS corpus. |
| NEWS (M2M-100) | `masakhane/m2m100_418M_{src}_{tgt}_news` | M2M-100 fine-tuned on {src}-{tgt} direction using parallel NEWS corpus. |
| NEWS (M2M-100-EN) | `masakhane/m2m100_418M-EN-NEWS` | M2M-100 fine-tuned on NEWS data that are English-centric i.e en–{hau, ibo, lug, luo, pcm, swa, tsn, twi, yor, zul} |
| NEWS (M2M-100-FR) | `masakhane/m2m100_418M-FR-NEWS` | M2M-100 fine-tuned on NEWS data that are French-centric i.e fr–{bam, bbj, ewe, fon, mos, wol}. |
| REL | `masakhane/m2m100_418M_{src}_{tgt}_rel` | M2M-100 fine-tuned on {src}-{tgt} direction using parallel REL corpus. |
| REL+NEWS | `masakhane/m2m100_418M_{src}_{tgt}_rel_news` | M2M-100 fine-tuned on {src}-{tgt} direction using parallel REL+NEWS corpus. |
| REL→NEWS | `masakhane/m2m100_418M_{src}_{tgt}_rel_ft` | M2M-100 fine-tuned on {src}-{tgt} direction using parallel REL corpus and additional fine-tuning on NEWS |
| REL+NEWS→NEWS | `masakhane/m2m100_418M_{src}_{tgt}_rel_news_ft` | M2M-100 fine-tuned on {src}-{tgt} direction using parallel REL+NEWS and additional fine-tuning on NEWS |

Table 10.11: Model names on HuggingFace Model Hub. For bilingual models, supply the correct **src** or **tgt** language. English/French make use of a 2-letter language code i.e en or fr, while all the African languages make us of 3-letter language codes e.g yor.

## 10.7 CONCLUSION

We have created MAFAND-MT, a corpus of 21 African languages to study translation systems for low-resource languages in the news domain. We investigate how to most effectively adapt large-scale pre-trained models to incorporate new languages and new domains. Our findings suggest that as little as 2k sentences are sufficient for fine-tuning, with an improved performance, paving the way for others to create new translation systems without relying on large collections of web-sourced text. This has strong implications for languages that are spoken by millions but lack presence on the web.

In the future, we hope to expand our coverage to additional under-resourced languages, and to develop even more effective fine-tuning objectives.

## 10.8 LIMITATIONS AND RISKS

Despite the promising results, our work has the following limitations:

1. **Translation quality**: Even the best model scores low BLEU on some of the reported languages (`bbj`, `mos`, `zul`), in particular when translating into them.

2. **Evaluation**: Our evaluation is focused on BLEU. We report ChrF results as well, but without a deeper human evaluation, we cannot make claims about the absolute quality of the translations. Manual inspections of translations gave us the impression that translations are surprisingly fluent and make good use of language-specific expressions when translating into English or French, but that errors in grammar and logic can be easily overlooked. Automatic reference-based metrics like BLEU and ChrF might not be able to capture the semantic relatedness to the reference sufficiently, as well potentially being tricked by word matches in incoherent phrases.

3. **Language bias**: We have shown that even when not included in pre-training, and without large out-of-domain data, significant gains in translation quality can be achieved. However, language-specific biases, in terms of resourcedness, morphology, standardization, inclusion in pre-trained models and available corpora, or relatedness to other languages, still affect the relative quality of translations, and require more efforts to be overcome.

4. **Domain limitations**: While we showed a rapid adaptation to the news domain and the auxiliary benefit of the religious domain, our study also revealed how automatically estimated translation quality drops when the test domain is narrow. Therefore, future work should aim to expand the study to multiple test domains and develop systematic methods for distilling knowledge from multiple narrow domains.

5. **Language coverage**: Africa has thousands of other languages that are not covered in our study but deserve the same attention. We hope that our work is encouraging enough to inspire native speakers of those languages not covered here to collect translations, run our code, and report their findings to the NLP research community, so that we can make joint progress in developing language technology for more people.

We believe that our translation models carry similar risks of causing harm by inaccurate and biased translations as the underlying large pre-trained models. M2M-100 is trained on large collections of texts crawled from the web, and the quality for most of the languages studied here is questionable (Kreutzer et al., 2022). Our fine-tuning successes show that some obvious biases can be overcome when the quality of the fine-tuning set is controlled (see the examples in Section 10.6.3), but we cannot guarantee that biases prevailing in the pre-training corpus or more subtle biases will not occur with other inputs. Together with a careful human evaluation, this should be the main concern for future work on the produced models. The methodology of rapid fine-tuning might also be misused to tune the models towards harmful content or purposes that harm the speakers of the languages presented here.

Part V

CONCLUSION AND FUTURE WORK

# 11

## CONCLUSION AND FUTURE WORK

This chapter summarizes the challenges of NLP for African languages, our approach to address some of the challenges, including suggestions for future directions.

## 11.1 CONCLUSION

In this dissertation, we identified several factors for the under-representation of African languages including societal factors and data-related factors. Some societal factors include lack of government support for indigenous languages, weak language policies by many African countries, the impact of colonialism, as well as geographical and language diversity of NLP researchers ∀ et al., 2020. The data-related factors are (1) lack of labelled and (2) large unlabelled data to leverage self-supervised pre-training of multilingual representation models – which serve as foundational models to build NLP models for several tasks. We focus on data-related factors. While it is difficult to address the lack of large unlabelled and labelled data issues, we show that by leveraging small human annotated data and recent advances in deep learning such as self-supervised pre-training, weakly-supervised learning, and transfer learning, we can build effective NLP models for African languages. We evaluated models developed using these techniques on several NLP tasks such as named entity recognition, machine translation, news topic classification, and sentiment classification with impressive performance. The annotated datasets and models created are available on Github[1]. Below, we summarize the main results:

### 11.1.1 *Dominance of multilingual PLMs over word embeddings*

There are two approaches that are often used for training models for natural language understanding (NLU) tasks using deep learning: (1) Traditional feature-based models, where static embeddings are extracted from word embeddings models like FastText or GloVe and used to *initialize* deep learning models like BiLSTM. (2) *fine-tuning* a PLM on a new task.

In Chapter 3, we compared the performance of feature-based models like the CNN-BiLSTM-CRF to multilingual PLM like XLM-RoBERTa on the NER task. We also investigated the difference in performance with different training data scales or sample sizes (500, 1K, 2K, and 4K sentences). Our evaluation on 21 languages showed that XLM-R is more data efficient since it gives an impressive performance with less training data (e.g. 500 sentences) compared to CNN-BiLSTM-CRF. This shows the dominance of fine-tuning a PLM for an NLU task over feature-based models, especially when there are only limited labelled data available. Interestingly, we found the performance of XLM-R with 2K sentences to be the same as CNN-BiLSTM-CRF with 4K sentences. This shows that using PLMs can help reduce the cost of annotation even up to half, providing opportunities to create labelled datasets for many low-resourced languages with a small budget.

---

[1] https://github.com/dadelani/africanlp-resources

### 11.1.2 *Developing better quality multilingual representation models with small curated monolingual data*

In Chapter 4, we evaluated the quality of pre-trained FastText word embeddings for two African languages (Twi and Yorùbá) using a word similarity task based on translated wordsim-353 word pairs. Our evaluation shows that they are of poor quality because the pre-training corpora are either small or of poor quality. The pre-trained FastText embedding for Twi was trained on Wikipedia with less than 6K tokens and does not follow the correct orthography (which we termed *noisy*). Similarly, the texts used for Yorùbá was very noisy, especially the Common Crawl corpus. We addressed this, by curating better quality texts (with less than 100MB per language), and used them to train better quality word embeddings. Our evaluation on wordSim-353 shows that the embeddings capture better word similarities which is an indication of better quality.

We extended our analysis to multilingual pre-trained models by comparing pre-trained multilingual BERT (mBERT) model to mBERT *adapted* to Yorùbá language with less than 100MB of data on the NER task. Our evaluation shows that we are able to more than double the performance of NER by training on our adapted mBERT model (60 F1) compared to the pre-trained mBERT model (27 F1).

### 11.1.3 *Weakly-supervised learning for limited labelled data scenario*

In Chapter 6, we show the effectiveness of applying distant and weak supervision to realistic low-resource scenarios by evaluating on African languages. Unlike many approaches that try to simulate low-resource scenarios (e.g. using small training data) by using high-resource languages (Hedderich and Klakow, 2018; Lange, Hedderich, and Klakow, 2019), there are other challenges in working with low-resource languages such as, the source of large entity list needed for automatic annotation, sourcing for native speakers to write rules, and low-quality multilingual representation models. By working with native speakers, we are able to surmount some of these challenges – native speakers provided rules for DATE annotation and large entity lists for automatic annotation. To alleviate some of the negative effects of the errors in automatic annotation, we integrate noise-handling methods to the NER models. Our evaluation on two different deep learning architectures (BiLSTM and mBERT) show that distant supervision can be successfully leveraged in a realistic low-resource scenario where it can more than double a classifier's performance.

### 11.1.4 *Participatory research for addressing the lack of labelled datasets*

In Chapter 7, we show the effectiveness of participatory research (∀ et al., 2020) for addressing the lack of labelled datasets for African languages. By bringing together different stakeholders like native speakers, data curators and evaluation experts from the Masakhane community, we created MasakhaNER – the first large, publicly available, high-quality dataset for named entity recognition (NER) in ten African languages. MasakhaNER dataset is from the *news* domain. Through the guidance of native speakers, we are able to select the best news source for annotation. There are many news sources, but some have issues of mixed dialects (e.g. BBC and VOA for Kinyarwanda) and non-standard orthography (e.g. BBC for Yorùbá), through participatory research, we are able to avoid these mistakes. Also, we are able to get volunteers from the Masakhane community who are enthusiastic to create the dataset for their language.

Following this participatory research approach, we expanded the NER dataset to 21 African languages (see Chapter 8), and created a machine translation evaluation dataset known as MAFAND-MT for 21 languages (see, Chapter 10). Aside from MT and NER, we also created news-topic classification dataset for five African languages known as ANTC (see, Chapter 5).

### 11.1.5 *Transfer learning for adapting to new tasks, languages and domains*

Most of the models developed in this dissertation benefited from the impressive cross-lingual transfer abilities of multilingual PLMs to new tasks, languages and domains. We highlight a few results below:

- **Transferring to new tasks** In Chapters §5, §7, §8, §9, and §10, we show the impressive performance of multilingual PLMs (Conneau et al., 2020; Devlin et al., 2019; Xue et al., 2021) to several NLP tasks like NER, sentiment classification, news topic classification, and machine translation, especially for languages they were not pre-trained on.

- **Transferring to new languages** We showed the impressive performance of adapting multilingual PLMs to a single language and multiple languages. This adaptation involves fine-tuning a PLM on unlabelled data in one or multiple languages. We call adaptation to a single language as *language-adaptive fine-tuning* (LAFT) and adaptation to multiple languages, *multilingual adaptive fine-tuning* (MAFT). In Chapter 7, we show that applying LAFT on XLM-R improves performance (on average over 10 languages) by more than (3 F1) over the XLM-R. Similarly, using MAFT, we created new multilingual PLMs for 17 African languages: AfroXLMR for NLU tasks (§5), AfriMT5, AfriByT5, and AfriMBART for text generation tasks (§10).

- **Transferring to new domains** It is not sufficient to just perform adaptation to a new language. We found that adapting to a new domain gives an additional improvement in performance. In Chapter 5, we show that we can improve the zero-shot cross-lingual performance of adapters (Pfeiffer et al., 2020a) by training language adapters on the monolingual texts of the domain of interest. Our evaluation is based on the *news* domain, and NER task. Similarly, for machine translation (§10), we found out that the best way to adapt to a new domain is to fine-tune large pre-trained models like M2M-100 (Fan et al., 2021) on a few thousand high quality translation data.

## 11.2 FUTURE WORK

In this dissertation, we lay the foundation for NLP for African languages. There are other challenges that still needs to be addressed. We highlight four major future directions below:

11.2.1 *Developing more labelled datasets for African languages*

In Chapter 3, we described the labelled datasets that are available for African languages in several tasks. Only NER, machine translation, text summarization and news topic classification cover more than 10 African languages. There is a need to expand to more tasks like part-of-speech and other universal dependencies tasks, sentiment classification, hate-speech detection, question answering, natural language inference, causal commonsense reasoning, slot-filling, intent detection, text-to-speech and speech recognition. Large-scale labelled dataset collection is still going in collaboration with Masakhane. In 2023, Masakhane members developed part-of-speech (Dione et al., 2023), news topic classification (Adelani et al., 2023), sentiment classification (Muhammad et al., 2023), and open-retrieval question answering (Ogundepo et al., 2023) labelled datasets for 10 or more African languages. However, more datasets are still needed especially for speech processing. This poses lots of challenges due to the cost of annotation. We hope to continue to leverage the participatory research approach to create small labelled data (e.g. 1,000 samples), and leverage the impressive capabilities of multilingual PLMs to achieve good performance.

11.2.2 *Developing of Africa-centric and efficient PLMs*

There are several multilingual PLMs and large language models (OpenAI, 2023; Touvron et al., 2023a,b), but they often exclude African languages. There is a need to continue developing more Africa-centric multilingual models to support more African languages. Currently, there are only few available PLMs like AfriBERTa (Ogueji, Zhu, and Lin, 2021), which covers 11 languages. The other models were developed during this PhD study—AfroXLMR, AfriMBART, AfriMT5, and AfriByT5, which cover 20 languages. The last four are the adaptation of existing multilingual PLMs. There is a need to train new models from the scratch that cover more languages (e.g. 50 African languages), preferably with different architectures like ELECTRA (Clark et al., 2020) and DeBERTa (He, Gao, and Chen, 2021b). Also, language family-specific PLMs or country-specific PLMs would be useful for geographical and linguistically related languages, like BantuBERT for Bantu languages and NaijaBERT for Nigerian languages.

Finally, the development or adaptation of autoregressive large language models and their instruction-tuned variants for African languages should be prioritized due to their recent improved capabilities especially with the launch of ChatGPT[2] model (Muennighoff et al., 2023; Scao et al., 2022; Yong et al., 2023). As we develop language models, it is important to make them more data, parameter and computationally efficient, so that they are easy to adapt to new languages with less data and easy to run by researchers from the low-resource community who do not have access to high-end GPUs (Dettmers et al., 2023; Hu et al., 2022; Liu et al., 2022; Pfeiffer et al., 2020a).

---

2 https://openai.com/blog/chatgpt

11.2.3 *Research on Speech processing*

Africa has over 2000 languages, many of which are only spoken. In some cases, many native speakers do not know how to write them. In general, research on speech processing has received less attention. There is a need to create corpora for speech technologies like automatic speech recognition, text-to-speech, speech translation, and other speech related tasks. Many of these speech tasks still require having some speech-text paired data. Recently, there are few attempts to develop speech technologies on Bible corpus (Black, 2019; Meyer et al., 2022; Ogayo, Neubig, and Black, 2022; Zanon Boito et al., 2020) which is the largest resource for many African languages. However, this biases the models to the religious domain. Other research focuses on developing self-supervised techniques that require less speech-text paired data (Babu et al., 2021; Baevski et al., 2020; Bapna et al., 2022a; Chung et al., 2021b), which could be used for a range of speech processing tasks. In general, there is a need to continue in this line of research that that require zero or few paired speech-text data that can scale to several under-resourced African languages, especially those absent during pre-training (Bapna et al., 2022b; Pratap et al., 2023).

11.2.4 *Development of basic linguistic tools*

Many African languages still lack basic linguistic tools like keyboards, tokenizers, sentence segmenters, spell-checkers, diacritizers (Orife, 2018a; Orife et al., 2020), monolingual or bilingual dictionaries, and morphological analyzers. We need to prioritize this.

Abadji, Julien, Pedro Ortiz Suarez, Laurent Romary, and Benoît Sagot (Jan. 2022). "Towards a Cleaner Document-Oriented Multilingual Crawled Corpus." In: *arXiv e-prints*, arXiv:2201.06642, arXiv:2201.06642. arXiv: 2201.06642 [cs.CL].

Abdaoui, Amine, Camille Pradel, and Grégoire Sigel (Nov. 2020). "Load What You Need: Smaller Versions of Mutililingual BERT." In: *Proceedings of SustaiNLP: Workshop on Simple and Efficient Natural Language Processing*. Online: Association for Computational Linguistics, pp. 119–123. DOI: 10.18653/v1/2020.sustainlp-1.16. URL: https://aclanthology.org/2020.sustainlp-1.16.

Abdul-Mageed, Muhammad, AbdelRahim Elmadany, and El Moatez Billah Nagoudi (Aug. 2021). "ARBERT & MARBERT: Deep Bidirectional Transformers for Arabic." In: *Proceedings of the 59th Annual Meeting of the Association for Computational Linguistics and the 11th International Joint Conference on Natural Language Processing (Volume 1: Long Papers)*. Online: Association for Computational Linguistics, pp. 7088–7105. DOI: 10.18653/v1/2021.acl-long.551. URL: https://aclanthology.org/2021.acl-long.551.

Abdulmumin, Idris and Bashir Shehu Galadanci (2019). "hauWE: Hausa Words Embedding for Natural Language Processing." In: *2019 2nd International Conference of the IEEE Nigeria Computer Chapter (NigeriaComputConf)*, pp. 1–6. DOI: 10.1109/NigeriaComputConf45974.2019.8949674.

Adebara, Ife and Muhammad Abdul-Mageed (May 2022). "Towards Afrocentric NLP for African Languages: Where We Are and Where We Can Go." In: *Proceedings of the 60th Annual Meeting of the Association for Computational Linguistics (Volume 1: Long Papers)*. Dublin, Ireland: Association for Computational Linguistics, pp. 3814–3841. DOI: 10.18653/v1/2022.acl-long.265. URL: https://aclanthology.org/2022.acl-long.265.

Adegbola, Tunde (2016). "Pattern-based Unsupervised Induction of Yoruba Morphology." In: *Proceedings of WWW 2016 Companion*, pp. 599–604.

Adegbola, Tunde and Lydia U. Odilinye (2012). "Quantifying the effect of corpus size on the quality of automatic diacritization of Yoruba texts." In: *Spoken Language Technologies for Under-Resourced Languages*.

Adelani, D., Dana Ruiter, J. Alabi, Damilola Adebonojo, Adesina Ayeni, Mofetoluwa Adeyemi, Ayodele Awokoya, and C. España-Bonet (2021a). "MENYO-20k: A Multi-domain English-Yorùbá Corpus for Machine Translation and Domain Adaptation." In: *ArXiv* abs/2103.08647.

Adelani, David Ifeoluwa, Michael A. Hedderich, D. Zhu, Esther van den Berg, and Dietrich Klakow (2020). "Distant Supervision and Noisy Label Learning for Low Resource Named Entity Recognition: A Study on Hausa and Yorùbá." In: *ArXiv* abs/2003.08370.

Adelani, David Ifeoluwa et al. (2021b). "MasakhaNER: Named Entity Recognition for African Languages." In: *Transactions of the Association for Computational Linguistics* 9, pp. 1116–1131. DOI: 10.1162/tacl_a_00416. URL: https://aclanthology.org/2021.tacl-1.66.

Adelani, David Ifeoluwa et al. (Dec. 2022a). "AfroNER: Africa-centric Transfer Learning for Named Entity Recognition." In: *Under Submission*.

Adelani, David Ifeoluwa et al. (2023). "MasakhaNEWS: News Topic Classification for African languages." In.

Adelani, David, Dana Ruiter, Jesujoba Alabi, Damilola Adebonojo, Adesina Ayeni, Mofe Adeyemi, Ayodele Esther Awokoya, and Cristina España-Bonet (Aug. 2021c). "The Effect of Domain and Diacritics in Yoruba–English Neural Machine Translation." In: *Proceedings of Machine Translation Summit XVIII: Research Track*. Virtual: Association for Machine Translation in the Americas, pp. 61–75. URL: https://aclanthology.org/2021.mtsummit-research.6.

Adelani, David et al. (July 2022b). "A Few Thousand Translations Go a Long Way! Leveraging Pre-trained Models for African News Translation." In: *Proceedings of the 2022 Conference of the North American Chapter of the Association for Computational Linguistics: Human Language Technologies*. Seattle, United States: Association for Computational Linguistics, pp. 3053–3070. DOI: 10.18653/v1/2022.naacl-main.223. URL: https://aclanthology.org/2022.naacl-main.223.

Agić, Željko and Ivan Vulić (July 2019). "JW300: A Wide-Coverage Parallel Corpus for Low-Resource Languages." In: *Proceedings of the 57th Annual Meeting of the Association for Computational Linguistics*. Florence, Italy: Association for Computational Linguistics, pp. 3204–3210. DOI: 10.18653/v1/P19-1310. URL: https://aclanthology.org/P19-1310.

Aharoni, Roee, Melvin Johnson, and Orhan Firat (June 2019). "Massively Multilingual Neural Machine Translation." In: *Proceedings of the 2019 Conference of the North American Chapter of the Association for Computational Linguistics: Human Language Technologies, Volume 1 (Long and Short Papers)*. Minneapolis, Minnesota: Association for Computational Linguistics, pp. 3874–3884. DOI: 10.18653/v1/N19-1388. URL: https://aclanthology.org/N19-1388.

Ahuja, Kabir, Shanu Kumar, Sandipan Dandapat, and Monojit Choudhury (May 2022). "Multi Task Learning For Zero Shot Performance Prediction of Multilingual Models." In: *Proceedings of the 60th Annual Meeting of the Association for Computational Linguistics (Volume 1: Long Papers)*. Dublin, Ireland: Association for Computational Linguis-

tics, pp. 5454–5467. DOI: 10.18653/v1/2022.acl-long.374. URL: https://aclanthology.org/2022.acl-long.374.

Akbik, Alan, Duncan Blythe, and Roland Vollgraf (Aug. 2018). "Contextual String Embeddings for Sequence Labeling." In: *Proceedings of the 27th International Conference on Computational Linguistics*. Santa Fe, New Mexico, USA: Association for Computational Linguistics, pp. 1638–1649. URL: https://www.aclweb.org/anthology/C18-1139.

Al-Rfou, Rami (2015). "Polyglot: A massive multilingual natural language processing pipeline." PhD thesis. Stony Brook University.

Alabi, Jesujoba Oluwadara, David Ifeoluwa Adelani, Marius Mosbach, and Dietrich Klakow (2022). "Multilingual Language Model Adaptive Fine-Tuning: A Study on African Languages." In: *ArXiv* abs/2204.06487.

Alabi, Jesujoba, Kwabena Amponsah-Kaakyire, David Adelani, and Cristina España-Bonet (May 2020). "Massive vs. Curated Embeddings for Low-Resourced Languages: the Case of Yorùbá and Twi." English. In: *Proceedings of the 12th Language Resources and Evaluation Conference*. Marseille, France: European Language Resources Association, pp. 2754–2762. ISBN: 979-10-95546-34-4. URL: https://www.aclweb.org/anthology/2020.lrec-1.335.

Alebiosu, Tajudeen Afolabi (2016). "Language Attitudes and the Issue of Dominance: The Nigerian Experience." In: *English Language, Literature Culture* 1.3, pp. 21–29. DOI: 10.11648/j.ellc.20160103.12. eprint: https://article.sciencepublishinggroup.com/pdf/10.11648.j.ellc.20160103.12. URL: https://doi.org/10.11648/j.ellc.20160103.12.

Algan, Görkem and Ilkay Ulusoy (2019). "Image Classification with Deep Learning in the Presence of Noisy Labels: A Survey." In: *CoRR* abs/1912.05170. arXiv: 1912.05170.

Ali, Felermino M. D. A., Andrew Caines, and Jaimito L. A. Malavi (2021). "Towards a parallel corpus of Portuguese and the Bantu language Emakhuwa of Mozambique." In: *ArXiv* abs/2104.05753.

Anastasopoulos, Antonios et al. (Dec. 2020). "TICO-19: the Translation Initiative for COvid-19." In: *Proceedings of the 1st Workshop on NLP for COVID-19 (Part 2) at EMNLP 2020*. Online: Association for Computational Linguistics. DOI: 10.18653/v1/2020.nlpcovid19-2.5. URL: https://aclanthology.org/2020.nlpcovid19-2.5.

Ansell, Alan, Edoardo Maria Ponti, Anna Korhonen, and Ivan Vulić (2021). *Composable Sparse Fine-Tuning for Cross-Lingual Transfer*. arXiv: 2110.07560 [cs.CL].

Ansell, Alan, Edoardo Ponti, Anna Korhonen, and Ivan Vulić (May 2022). "Composable Sparse Fine-Tuning for Cross-Lingual Transfer." In: *Proceedings of the 60th Annual Meeting of the Association for Computational Linguistics (Volume 1: Long Papers)*. Dublin, Ireland: Association for Computational Linguistics, pp. 1778–1796. DOI:

10.18653/v1/2022.acl-long.125. URL: https://aclanthology.org/2022.acl-long.125.

Antoun, Wissam, Fady Baly, and Hazem Hajj (May 2020). "AraBERT: Transformer-based Model for Arabic Language Understanding." English. In: *Proceedings of the 4th Workshop on Open-Source Arabic Corpora and Processing Tools, with a Shared Task on Offensive Language Detection*. Marseille, France: European Language Resource Association, pp. 9–15. ISBN: 979-10-95546-51-1. URL: https://aclanthology.org/2020.osact-1.2.

Aplonova, Ekaterina and Francis M. Tyers (2017). "Towards a dependency-annotated treebank for Bambara." In: *Proceedings of the 16th International Workshop on Treebanks and Linguistic Theories*. Prague, Czech Republic, pp. 138–145. URL: https://aclanthology.org/W17-7618.

Aribandi, Vamsi et al. (2022). "ExT5: Towards Extreme Multi-Task Scaling for Transfer Learning." In: *International Conference on Learning Representations*. URL: https://openreview.net/forum?id=Vzh1BFUCiIX.

Arivazhagan, Naveen et al. (2019). "Massively Multilingual Neural Machine Translation in the Wild: Findings and Challenges." In: *arXiv e-prints 1907.05019*. arXiv: 1907.05019 [cs.CL]. URL: http://arxiv.org/abs/1907.05019.

Artetxe, Mikel, Sebastian Ruder, and Dani Yogatama (July 2020). "On the Cross-lingual Transferability of Monolingual Representations." In: *Proceedings of the 58th Annual Meeting of the Association for Computational Linguistics*. Online: Association for Computational Linguistics, pp. 4623–4637. DOI: 10.18653/v1/2020.acl-main.421. URL: https://aclanthology.org/2020.acl-main.421.

Artetxe, Mikel and Holger Schwenk (July 2019a). "Margin-based Parallel Corpus Mining with Multilingual Sentence Embeddings." In: *Proceedings of the 58th Annual Meeting of the Association for Computational Linguistics*. Florence, Italy: Association for Computational Linguistics, pp. 3197–3203.

Artetxe, Mikel and Holger Schwenk (2019b). "Massively Multilingual Sentence Embeddings for Zero-Shot Cross-Lingual Transfer and Beyond." In: *Transactions of the Association for Computational Linguistics* 7, pp. 597–610. DOI: 10.1162/tacl_a_00288. URL: https://aclanthology.org/Q19-1038.

Asahiah, F. O., O. A. Odejobi, and E. R. Adagunodo (2017). "Restoring Tone-Marks in Standard Yoruba Electronic Text: Improved Model." In: *Computer Science* 18.3, pp. 301–315.

Asahiah, Franklin O (2014). "Development of a Standard Yoruba Digital Text Automatic Diacritic Restoration System." In: *Phd. Thesis, Obafemi Awolowo University, Ile-Ife, Nigeria*.

Augustinus, Liesbeth, Peter Dirix, Daniel van Niekerk, Ineke Schuurman, Vincent Vandeghinste, Frank Van Eynde, and Gerhard van Huyssteen (May 2016). "AfriBooms: An Online Treebank for Afrikaans." In: *Proceedings of the Tenth International Conference on Language Resources and Evaluation (LREC'16)*. Portorož, Slovenia: European Language Resources Association (ELRA), pp. 677–682. URL: https://aclanthology.org/L16-1107.

Azime, Israel Abebe and Nebil Mohammed (2021). "An Amharic News Text classification Dataset." In: *ArXiv* abs/2103.05639.

Azunre, Paul et al. (2021a). "English-Twi Parallel Corpus for Machine Translation." In: *ArXiv* abs/2103.15625.

Azunre, Paul et al. (2021b). "NLP for Ghanaian Languages." In: *AfricaNLP Workshop* abs/2103.15475.

Babou, Cheikh Anta and Michele Loporcaro (2016). "Noun classes and grammatical gender in Wolof." In: *Journal of African Languages and Linguistics* 37.1, pp. 1–57. DOI: doi:10.1515/jall-2016-0001. URL: https://doi.org/10.1515/jall-2016-0001.

Babu, Arun et al. (2021). "XLS-R: Self-supervised Cross-lingual Speech Representation Learning at Scale." In: *ArXiv* abs/2111.09296.

Baevski, Alexei, Sergey Edunov, Yinhan Liu, Luke Zettlemoyer, and Michael Auli (2019). "Cloze-driven Pretraining of Self-attention Networks." In: *Proceedings of the 2019 Conference on Empirical Methods in Natural Language Processing and the 9th International Joint Conference on Natural Language Processing, EMNLP-IJCNLP 2019, Hong Kong, China, November 3-7, 2019*, pp. 5359–5368. DOI: 10.18653/v1/D19-1539. URL: https://doi.org/10.18653/v1/D19-1539.

Baevski, Alexei, Yuhao Zhou, Abdelrahman Mohamed, and Michael Auli (2020). "wav2vec 2.0: A Framework for Self-Supervised Learning of Speech Representations." In: *Advances in Neural Information Processing Systems*. Ed. by H. Larochelle, M. Ranzato, R. Hadsell, M.F. Balcan, and H. Lin. Vol. 33. Curran Associates, Inc., pp. 12449–12460. URL: https://proceedings.neurips.cc/paper/2020/file/92d1e1eb1cd6f9fba3227870bb6d7f07-Paper.pdf.

Bañón, Marta et al. (July 2020). "ParaCrawl: Web-Scale Acquisition of Parallel Corpora." In: *Proceedings of the 58th Annual Meeting of the Association for Computational Linguistics*. Online: Association for Computational Linguistics, pp. 4555–4567. DOI: 10.18653/v1/2020.acl-main.417. URL: https://aclanthology.org/2020.acl-main.417.

Bapna, Ankur, Colin Cherry, Yu Zhang, Ye Jia, Melvin Johnson, Yong Cheng, Simran Khanuja, Jason Riesa, and Alexis Conneau (2022a). "mSLAM: Massively multilingual joint pre-training for speech and text." In: *ArXiv* abs/2202.01374.

Bapna, Ankur et al. (2022b). "XTREME-S: Evaluating Cross-lingual Speech Representations." In: *Proc. Interspeech 2022*.

Barrault, Loïc et al., eds. (Nov. 2020). *Proceedings of the Fifth Conference on Machine Translation*. Online: Association for Computational Linguistics. URL: https://www.aclweb.org/anthology/2020.wmt-1.

Bekker, Alan Joseph and Jacob Goldberger (2016). "Training deep neural-networks based on unreliable labels." In: *2016 IEEE International Conference on Acoustics, Speech and Signal Processing, ICASSP 2016, Shanghai, China, March 20-25, 2016*, pp. 2682–2686. DOI: 10.1109/ICASSP.2016.7472164. URL: https://doi.org/10.1109/ICASSP.2016.7472164.

Benajiba, Yassine, Paolo Rosso, and José Miguel BenedíRuiz (2007). "ANERsys: An Arabic Named Entity Recognition System Based on Maximum Entropy." In: *Computational Linguistics and Intelligent Text Processing*. Ed. by Alexander Gelbukh. Berlin, Heidelberg: Springer Berlin Heidelberg, pp. 143–153. ISBN: 978-3-540-70939-8.

Bengio, Yoshua, Réjean Ducharme, and Pascal Vincent (2000). "A Neural Probabilistic Language Model." In: *Advances in Neural Information Processing Systems*. Ed. by T. Leen, T. Dietterich, and V. Tresp. Vol. 13. MIT Press. URL: https://proceedings.neurips.cc/paper/2000/file/728f206c2a01bf572b5940d7d9a8fa4c-Paper.pdf.

Benikova, Darina, Chris Biemann, and Marc Reznicek (May 2014). "NoSta-D Named Entity Annotation for German: Guidelines and Dataset." In: *Proceedings of the Ninth International Conference on Language Resources and Evaluation (LREC'14)*. Reykjavik, Iceland: European Language Resources Association (ELRA), pp. 2524–2531. URL: http://www.lrec-conf.org/proceedings/lrec2014/pdf/276_Paper.pdf.

Beukman, Michael (2022). "Analysing the effects of transfer learning on low-resourced named entity recognition performance." In: *3rd Workshop on African Natural Language Processing*. URL: https://openreview.net/forum?id=HKWMFqfN8b5.

Beyer, Anne, Göran Kauermann, and Hinrich Schütze (May 2020). "Embedding Space Correlation as a Measure of Domain Similarity." English. In: *Proceedings of the 12th Language Resources and Evaluation Conference*. Marseille, France: European Language Resources Association, pp. 2431–2439. ISBN: 979-10-95546-34-4. URL: https://www.aclweb.org/anthology/2020.lrec-1.296.

Birch, Alexandra et al. (Aug. 2021). "Surprise Language Challenge: Developing a Neural Machine Translation System between Pashto and English in Two Months." In: *Proceedings of Machine Translation Summit XVIII: Research Track*. Virtual: Association for Machine Translation in the Americas, pp. 92–102. URL: https://aclanthology.org/2021.mtsummit-research.8.

Bird, Steven (Dec. 2020). "Decolonising Speech and Language Technology." In: *Proceedings of the 28th International Conference on Computational Linguistics*. Barcelona, Spain (Online): International Committee on Computational Linguistics, pp. 3504–3519. DOI: 10.18653/v1/2020.coling-main.313. URL: https://aclanthology.org/2020.coling-main.313.

Black, Alan W (2019). "CMU Wilderness Multilingual Speech Dataset." In: *ICASSP 2019 - 2019 IEEE International Conference on Acoustics, Speech and Signal Processing (ICASSP)*, pp. 5971–5975. DOI: 10.1109/ICASSP.2019.8683536.

Bodomo, AB and CO Marfo (2002). "The Morphophonology of Noun Classes in Dagaare and Akan." In: URL: http://hdl.handle.net/10722/109050.

Bojanowski, Piotr, Edouard Grave, Armand Joulin, and Tomas Mikolov (2017). "Enriching Word Vectors with Subword Information." In: *Transactions of the Association for Computational Linguistics* 5, pp. 135–146. DOI: 10.1162/tacl_a_00051. URL: https://aclanthology.org/Q17-1010.

Brown, Tom et al. (2020). "Language Models are Few-Shot Learners." In: *Advances in Neural Information Processing Systems*. Ed. by H. Larochelle, M. Ranzato, R. Hadsell, M.F. Balcan, and H. Lin. Vol. 33. Curran Associates, Inc., pp. 1877–1901. URL: https://proceedings.neurips.cc/paper/2020/file/1457c0d6bfcb4967418bfb8ac142f64a-Paper.pdf.

Cahyawijaya, Samuel et al. (Nov. 2021). "IndoNLG: Benchmark and Resources for Evaluating Indonesian Natural Language Generation." In: *Proceedings of the 2021 Conference on Empirical Methods in Natural Language Processing*. Online and Punta Cana, Dominican Republic: Association for Computational Linguistics, pp. 8875–8898. DOI: 10.18653/v1/2021.emnlp-main.699. URL: https://aclanthology.org/2021.emnlp-main.699.

Caines, Andrew (Oct. 2019). *The Geographic Diversity of NLP Conferences*. URL: http://www.marekrei.com/blog/geographic-diversity-of-nlp-conferences/.

Callison-Burch, Chris, Philipp Koehn, Christof Monz, Josh Schroeder, and Cameron Shaw Fordyce, eds. (June 2008). *Proceedings of the Third Workshop on Statistical Machine Translation*. Columbus, Ohio: Association for Computational Linguistics. URL: https://aclanthology.org/W08-0300.

Camacho-Collados, Jose, Mohammad Taher Pilehvar, Nigel Collier, and Roberto Navigli (Aug. 2017). "SemEval-2017 Task 2: Multilingual and Cross-lingual Semantic Word Similarity." In: *Proceedings of the 11th International Workshop on Semantic Evaluation (SemEval-2017)*. Vancouver, Canada: Association for Computational Linguistics, pp. 15–26. DOI: 10.18653/v1/S17-2002. URL: https://www.aclweb.org/anthology/S17-2002.

Cao, Yixin, Zikun Hu, Tat-Seng Chua, Zhiyuan Liu, and Heng Ji (2019). "Low-Resource Name Tagging Learned with Weakly Labeled Data." In: *Proceedings of the 2019 Conference on Empirical Methods in Natural Language Processing and the 9th International Joint Conference on Natural Language Processing, EMNLP-IJCNLP 2019, Hong Kong,*

*China, November 3-7, 2019*, pp. 261–270. DOI: 10.18653/v1/D19-1025. URL: https://doi.org/10.18653/v1/D19-1025.

Caron, Bernard, Marine Courtin, Kim Gerdes, and Sylvain Kahane (Aug. 2019). "A Surface-Syntactic UD Treebank for Naija." In: *Proceedings of the 18th International Workshop on Treebanks and Linguistic Theories (TLT, SyntaxFest 2019)*. Paris, France: Association for Computational Linguistics, pp. 13–24. DOI: 10.18653/v1/W19-7803. URL: https://aclanthology.org/W19-7803.

Chau, Ethan C. and Noah A. Smith (Nov. 2021). "Specializing Multilingual Language Models: An Empirical Study." In: *Proceedings of the 1st Workshop on Multilingual Representation Learning*. Punta Cana, Dominican Republic: Association for Computational Linguistics, pp. 51–61. DOI: 10.18653/v1/2021.mrl-1.5. URL: https://aclanthology.org/2021.mrl-1.5.

Chaudhary, Aditi, Chunting Zhou, Lori S. Levin, Graham Neubig, David R. Mortensen, and Jaime G. Carbonell (2018). "Adapting Word Embeddings to New Languages with Morphological and Phonological Subword Representations." In: *EMNLP*. Ed. by Ellen Riloff, David Chiang, Julia Hockenmaier, and Jun'ichi Tsujii. Association for Computational Linguistics, pp. 3285–3295. ISBN: 978-1-948087-84-1. URL: https://www.aclweb.org/anthology/D18-1366/.

Chen, Chenhua, Yue Zhang, and Yuze Gao (2018). "Learning How to Self-Learn: Enhancing Self-Training Using Neural Reinforcement Learning." In: *2018 International Conference on Asian Language Processing, IALP 2018, Bandung, Indonesia, November 15-17, 2018*, pp. 25–30. DOI: 10.1109/IALP.2018.8629107. URL: https://doi.org/10.1109/IALP.2018.8629107.

Chen, Xinxiong, Lei Xui, Liu Zhiyuan, Maosong Sun, and Huanbo Luan (2015). "Joint Learning of Character and Word Embeddings." In: *Proceedings of the Twenty-Fourth International Joint Conference on Artificial Intelligence (IJCAI)*. IJCAI, pp. 1236–1242.

Chi, Zewen, Li Dong, Furu Wei, Nan Yang, Saksham Singhal, Wenhui Wang, Xia Song, Xian-Ling Mao, Heyan Huang, and Ming Zhou (June 2021). "InfoXLM: An Information-Theoretic Framework for Cross-Lingual Language Model Pre-Training." In: *Proceedings of the 2021 Conference of the North American Chapter of the Association for Computational Linguistics: Human Language Technologies*. Online: Association for Computational Linguistics, pp. 3576–3588. DOI: 10.18653/v1/2021.naacl-main.280. URL: https://aclanthology.org/2021.naacl-main.280.

Chiu, Jason P.C. and Eric Nichols (2016). "Named Entity Recognition with Bidirectional LSTM-CNNs." In: *Transactions of the Association for Computational Linguistics* 4, pp. 357–370. DOI: 10.1162/tacl_a_00104. URL: https://www.aclweb.org/anthology/Q16-1026.

Chowdhery, Aakanksha et al. (2022). "PaLM: Scaling Language Modeling with Pathways." In: *ArXiv* abs/2204.02311.

Christianson, Caitlin, Jason Duncan, and Boyan A. Onyshkevych (2018). "Overview of the DARPA LORELEI Program." In: *Machine Translation* 32.1-2, pp. 3–9. DOI: 10.1007/s10590-017-9212-4. URL: https://doi.org/10.1007/s10590-017-9212-4.

Chung, Hyung Won, Thibault Fevry, Henry Tsai, Melvin Johnson, and Sebastian Ruder (2021a). "Rethinking Embedding Coupling in Pre-trained Language Models." In: *International Conference on Learning Representations*. URL: https://openreview.net/forum?id=xpFFI_NtgpW.

Chung, Yu-An, Yu Zhang, Wei Han, Chung-Cheng Chiu, James Qin, Ruoming Pang, and Yonghui Wu (2021b). "w2v-BERT: Combining Contrastive Learning and Masked Language Modeling for Self-Supervised Speech Pre-Training." In: *2021 IEEE Automatic Speech Recognition and Understanding Workshop (ASRU)*, pp. 244–250. DOI: 10.1109/ASRU51503.2021.9688253.

Clark, Kevin, Minh-Thang Luong, Quoc V. Le, and Christopher D. Manning (2020). "ELECTRA: Pre-training Text Encoders as Discriminators Rather Than Generators." In: *ICLR*. URL: https://openreview.net/pdf?id=r1xMH1BtvB.

Conneau, Alexis, Kartikay Khandelwal, Naman Goyal, Vishrav Chaudhary, Guillaume Wenzek, Francisco Guzmán, Edouard Grave, Myle Ott, Luke Zettlemoyer, and Veselin Stoyanov (July 2020). "Unsupervised Cross-lingual Representation Learning at Scale." In: *Proceedings of the 58th Annual Meeting of the Association for Computational Linguistics*. Online: Association for Computational Linguistics, pp. 8440–8451. DOI: 10.18653/v1/2020.acl-main.747. URL: https://aclanthology.org/2020.acl-main.747.

Conneau, Alexis, Ruty Rinott, Guillaume Lample, Adina Williams, Samuel R. Bowman, Holger Schwenk, and Veselin Stoyanov (2018). "XNLI: Evaluating Cross-lingual Sentence Representations." In: *Proceedings of the 2018 Conference on Empirical Methods in Natural Language Processing*. Brussels, Belgium: Association for Computational Linguistics.

Cui, Yiming, Wanxiang Che, Ting Liu, Bing Qin, Ziqing Yang, Shijin Wang, and Guoping Hu (2021). "Pre-Training With Whole Word Masking for Chinese BERT." In: *IEEE/ACM Transactions on Audio, Speech, and Language Processing* 29, pp. 3504–3514.

Currey, Anna, Antonio Valerio Miceli Barone, and Kenneth Heafield (Sept. 2017). "Copied Monolingual Data Improves Low-Resource Neural Machine Translation." In: *Proceedings of the Second Conference on Machine Translation*. Copenhagen, Denmark: Association for Computational Linguistics, pp. 148–156. DOI: 10.18653/v1/W17-4715. URL: https://aclanthology.org/W17-4715.

David, Davis (Dec. 2020). "Swahili : News Classification Dataset." Version 0.2. In: The news version contains both train and test sets.

DOI: `10.5281/zenodo.5514203`. URL: https://doi.org/10.5281/zenodo.5514203.

Davody, Ali, David Ifeoluwa Adelani, Thomas Kleinbauer, and Dietrich Klakow (2022). "TOKEN is a MASK: Few-shot Named Entity Recognition with Pre-trained Language Models." In: *ArXiv* abs/2206.07841.

De Cao, Nicola, Ledell Wu, Kashyap Popat, Mikel Artetxe, Naman Goyal, Mikhail Plekhanov, Luke Zettlemoyer, Nicola Cancedda, Sebastian Riedel, and Fabio Petroni (2022). "Multilingual Autoregressive Entity Linking." In: *Transactions of the Association for Computational Linguistics* 10, pp. 274–290. DOI: `10.1162/tacl_a_00460`. URL: https://aclanthology.org/2022.tacl-1.16.

De Pauw, Guy, Peter W Wagacha, and Dorothy Atieno Abade (2007). "Unsupervised Induction of Dholuo Word Classes using Maximum Entropy Learning." In: *Proceedings of the First International Computer Science and ICT Conference*, p. 8. DOI: `http://hdl.handle.net/11295/44250`.

Dembowski, Julia, Michael Wiegand, and Dietrich Klakow (2017). "Language Independent Named Entity Recognition using Distant Supervision." In: *Proceedings of Language and Technology Conference*.

Dettmers, Tim, Artidoro Pagnoni, Ari Holtzman, and Luke Zettlemoyer (2023). "QLoRA: Efficient Finetuning of Quantized LLMs." In: *ArXiv* abs/2305.14314. URL: https://api.semanticscholar.org/CorpusID:258841328.

Devlin, Jacob, Ming-Wei Chang, Kenton Lee, and Kristina Toutanova (June 2019). "BERT: Pre-training of Deep Bidirectional Transformers for Language Understanding." In: *Proceedings of the 2019 Conference of the North American Chapter of the Association for Computational Linguistics: Human Language Technologies, Volume 1 (Long and Short Papers)*. Minneapolis, Minnesota: Association for Computational Linguistics, pp. 4171–4186. DOI: `10.18653/v1/N19-1423`. URL: https://aclanthology.org/N19-1423.

Dimmendaal, Gerrit Jan (2016). "On stable and unstable features in Nilo-Saharan." In.

Dione, Cheikh Bamba (Aug. 2019). "Developing Universal Dependencies for Wolof." In: *Proceedings of the Third Workshop on Universal Dependencies (UDW, SyntaxFest 2019)*. Paris, France: Association for Computational Linguistics, pp. 12–23. DOI: `10.18653/v1/W19-8003`. URL: https://aclanthology.org/W19-8003.

Dione, Cheikh M. Bamba et al. (July 2023). "MasakhaPOS: Part-of-Speech Tagging for Typologically Diverse African languages." In: *Proceedings of the 61st Annual Meeting of the Association for Computational Linguistics (Volume 1: Long Papers)*. Toronto, Canada: Association for Computational Linguistics, pp. 10883–10900. DOI: `10.18653/v1/2023.acl-long.609`. URL: https://aclanthology.org/2023.acl-long.609.

Dryer, Matthew S. and Martin Haspelmath, eds. (2013). *WALS Online*. Leipzig: Max Planck Institute for Evolutionary Anthropology. URL: https://wals.info/.

Dumitrescu, Stefan Daniel and Andrei-Marius Avram (May 2020). "Introducing RONEC - the Romanian Named Entity Corpus." English. In: *Proceedings of the 12th Language Resources and Evaluation Conference*. Marseille, France: European Language Resources Association, pp. 4436–4443. ISBN: 979-10-95546-34-4. URL: https://aclanthology.org/2020.lrec-1.546.

Eberhard, David M., Gary F. Simons, and Charles D. Fennig (2021). *Ethnologue: Languages of the World. Twenty-third edition.* SIL International. URL: http://www.ethnologue.com.

Ebrahimi, Abteen et al. (May 2022). "AmericasNLI: Evaluating Zero-shot Natural Language Understanding of Pretrained Multilingual Models in Truly Low-resource Languages." In: *Proceedings of the 60th Annual Meeting of the Association for Computational Linguistics (Volume 1: Long Papers)*. Dublin, Ireland: Association for Computational Linguistics, pp. 6279–6299. DOI: 10.18653/v1/2022.acl-long.435. URL: https://aclanthology.org/2022.acl-long.435.

Eiselen, Roald (May 2016). "Government Domain Named Entity Recognition for South African Languages." In: *Proceedings of the Tenth International Conference on Language Resources and Evaluation (LREC'16)*. Portorož, Slovenia: European Language Resources Association (ELRA), pp. 3344–3348. URL: https://www.aclweb.org/anthology/L16-1533.

El-Kishky, Ahmed, Vishrav Chaudhary, Francisco Guzmán, and Philipp Koehn (Nov. 2020). "CCAligned: A Massive Collection of Cross-lingual Web-Document Pairs." In: *Proceedings of the 2020 Conference on Empirical Methods in Natural Language Processing (EMNLP 2020)*. Online: Association for Computational Linguistics, pp. 5960–5969. DOI: 10.18653/v1/2020.emnlp-main.480. URL: https://www.aclweb.org/anthology/2020.emnlp-main.480.

Emenanjo, Nolue (1978). *Elements of Modern Igbo Grammar - a descriptive approach*. Ibadan, Nigeria: Oxford University Press.

Emezue, Chris Chinenye and Bonaventure F. P. Dossou (Nov. 2021). "MMTAfrica: Multilingual Machine Translation for African Languages." In: *Proceedings of the Sixth Conference on Machine Translation*. Online: Association for Computational Linguistics, pp. 398–411. URL: https://aclanthology.org/2021.wmt-1.48.

Emezue, Chris Chinenye and Femi Pancrace Bonaventure Dossou (July 2020). "FFR v1.1: Fon-French Neural Machine Translation." In: *Proceedings of the The Fourth Widening Natural Language Processing Workshop*. Seattle, USA: Association for Computational Linguistics, pp. 83–87. DOI: 10.18653/v1/2020.winlp-1.21. URL: https://aclanthology.org/2020.winlp-1.21.

Esch, Daan van, Elnaz Sarbar, Tamar Lucassen, Jeremy O'Brien, Theresa Breiner, Manasa Prasad, Evan Crew, Chieu Nguyen, and Franccoise Beaufays (2019). "Writing Across the World's Languages: Deep Internationalization for Gboard, the Google Keyboard." In: *ArXiv* abs/1912.01218.

España-Bonet, Cristina, Alberto Barrón-Cedeño, and Lluís Màrquez (May 2020). "Tailoring and Evaluating the Wikipedia for in-Domain Comparable Corpora Extraction." In: *arXiv e-prints 2005.01177*, pp. 1–26. arXiv: 2005.01177 [cs.CL].

Ezeani, Ignatius, Paul Rayson, I. Onyenwe, C. Uchechukwu, and M. Hepple (2020). "Igbo-English Machine Translation: An Evaluation Benchmark." In: *ArXiv* abs/2004.00648.

Fagbolu, O., A. Ojoawo, K. Ajibade, and B. Alese (2015). "Digital Yoruba Corpus." In: *International Journal of Innovative Science, Engineering & Technology* 2.8, pp. 918–926.

Faisal, Fahim, Yinkai Wang, and Antonios Anastasopoulos (May 2022). "Dataset Geography: Mapping Language Data to Language Users." In: *Proceedings of the 60th Annual Meeting of the Association for Computational Linguistics (Volume 1: Long Papers)*. Dublin, Ireland: Association for Computational Linguistics, pp. 3381–3411. DOI: 10.18653/v1/2022.acl-long.239. URL: https://aclanthology.org/2022.acl-long.239.

Fan, Angela et al. (2021). "Beyond English-Centric Multilingual Machine Translation." In: *Journal of Machine Learning Research* 22.107, pp. 1–48. URL: http://jmlr.org/papers/v22/20-1307.html.

Fang, Meng and Trevor Cohn (2016). "Learning when to trust distant supervision: An application to low-resource POS tagging using cross-lingual projection." In: *Proceedings of the 20th SIGNLL Conference on Computational Natural Language Learning*. DOI: 10.18653/v1/K16-1018.

Finkelstein, I., E. Gabrilovich, Y. Mathias, E. Rivlin, Z. Solan, and G. Wolfman (2001). "Placing search in context: The concept revisited." In: *10th International Conference on World Wide Web*, pp. 406–414.

Firat, Orhan, Kyunghyun Cho, and Yoshua Bengio (June 2016). "Multi-Way, Multilingual Neural Machine Translation with a Shared Attention Mechanism." In: *Proceedings of the 2016 Conference of the North American Chapter of the Association for Computational Linguistics: Human Language Technologies*. San Diego, California: Association for Computational Linguistics, pp. 866–875. DOI: 10.18653/v1/N16-1101. URL: https://aclanthology.org/N16-1101.

Firth, J.R. (1957). "A synopsis of linguistic theory 1930-1955." In: *Studies in linguistic analysis*, pp. 1–32.

FitzGerald, Jack G. M. et al. (2022). "MASSIVE: A 1M-Example Multilingual Natural Language Understanding Dataset with 51 Typologically-Diverse Languages." In: *ArXiv* abs/2204.08582.

Fleiss, Joseph L (1971). "Measuring nominal scale agreement among many raters." In: *Psychological bulletin* 76.5, p. 378.

Frankle, Jonathan and Michael Carbin (2019). "The Lottery Ticket Hypothesis: Finding Sparse, Trainable Neural Networks." In: *International Conference on Learning Representations*. URL: https://openreview.net/forum?id=rJl-b3RcF7.

Freitas, Cláudia, Cristina Mota, Diana Santos, Hugo Gonçalo Oliveira, and Paula Carvalho (May 2010). "Second HAREM: Advancing the State of the Art of Named Entity Recognition in Portuguese." In: *Proceedings of the Seventh International Conference on Language Resources and Evaluation (LREC'10)*. Valletta, Malta: European Language Resources Association (ELRA). URL: http://www.lrec-conf.org/proceedings/lrec2010/pdf/412_Paper.pdf.

Frenay, B. and M. Verleysen (2014). "Classification in the Presence of Label Noise: A Survey." In: *IEEE Transactions on Neural Networks and Learning Systems* 25.5, pp. 845–869. DOI: 10.1109/TNNLS.2013.2292894.

Fu, Jinlan, Pengfei Liu, and Graham Neubig (Nov. 2020). "Interpretable Multi-dataset Evaluation for Named Entity Recognition." In: *Proceedings of the 2020 Conference on Empirical Methods in Natural Language Processing (EMNLP)*. Online: Association for Computational Linguistics, pp. 6058–6069. DOI: 10.18653/v1/2020.emnlp-main.489. URL: https://www.aclweb.org/anthology/2020.emnlp-main.489.

Gezmu, Andargachew Mekonnen, A. Nürnberger, and Tesfaye Bayu Bati (2021). "Extended Parallel Corpus for Amharic-English Machine Translation." In: *ArXiv* abs/2104.03543.

Gordon, Andrew, Zornitsa Kozareva, and Melissa Roemmele (June 2012). "SemEval-2012 Task 7: Choice of Plausible Alternatives: An Evaluation of Commonsense Causal Reasoning." In: *\*SEM 2012: The First Joint Conference on Lexical and Computational Semantics – Volume 1: Proceedings of the main conference and the shared task, and Volume 2: Proceedings of the Sixth International Workshop on Semantic Evaluation (SemEval 2012)*. Montréal, Canada: Association for Computational Linguistics, pp. 394–398. URL: https://aclanthology.org/S12-1052.

Government, Rwanda (2014). *Official Gazette number 41 bis of 13/10/2014*. URL: https://gazettes.africa/archive/rw/2014/rw-government-gazette-dated-2014-10-13-no-41bis.pdf.

Gowda, Thamme, Zhao Zhang, Chris Mattmann, and Jonathan May (Aug. 2021). "Many-to-English Machine Translation Tools, Data, and Pretrained Models." In: *Proceedings of the 59th Annual Meeting of the Association for Computational Linguistics and the 11th International Joint Conference on Natural Language Processing: System Demonstrations*. Online: Association for Computational Linguistics, pp. 306–316. DOI: 10.18653/v1/2021.acl-demo.37. URL: https://aclanthology.org/2021.acl-demo.37.

Goyal, Naman, Cynthia Gao, Vishrav Chaudhary, Peng-Jen Chen, Guillaume Wenzek, Da Ju, Sanjana Krishnan, Marc'Aurelio Ranzato, Francisco Guzmán, and Angela Fan (May 2022). "The Flores-101 Evaluation Benchmark for Low-Resource and Multilingual Machine Translation." In: *Transactions of the Association for Computational Linguistics* 10, pp. 522–538. ISSN: 2307-387X. DOI: 10.1162/tacl_a_00474. eprint: https://direct.mit.edu/tacl/article-pdf/doi/10.1162/tacl\_a\_00474/2020699/tacl\_a\_00474.pdf. URL: https://doi.org/10.1162/tacl\_a\_00474.

Grave, Edouard, Piotr Bojanowski, Prakhar Gupta, Armand Joulin, and Tomas Mikolov (2018). "Learning Word Vectors for 157 Languages." In: *CoRR* abs/1802.06893. URL: http://arxiv.org/abs/1802.06893.

Groenewald, Hendrik J. and Wildrich Fourie (May 2009). "Introducing the Autshumato Integrated Translation Environment." In: *Proceedings of the 13th Annual conference of the European Association for Machine Translation*. Barcelona, Spain: European Association for Machine Translation, pp. 190–196. URL: https://www.aclweb.org/anthology/2009.eamt-1.26.

Gruzitis, Normunds, Lauma Pretkalnina, Baiba Saulite, Laura Rituma, Gunta Nespore-Berzkalne, Arturs Znotins, and Peteris Paikens (May 2018). "Creation of a Balanced State-of-the-Art Multilayer Corpus for NLU." In: *Proceedings of the Eleventh International Conference on Language Resources and Evaluation (LREC 2018)*. Miyazaki, Japan: European Language Resources Association (ELRA). URL: https://aclanthology.org/L18-1714.

Gulcehre, Caglar, Orhan Firat, Kelvin Xu, Kyunghyun Cho, Loic Barrault, Huei-Chi Lin, Fethi Bougares, Holger Schwenk, and Yoshua Bengio (2015). "On using monolingual corpora in neural machine translation." In: *arXiv preprint arXiv:1503.03535*.

Gururangan, Suchin, Ana Marasović, Swabha Swayamdipta, Kyle Lo, Iz Beltagy, Doug Downey, and Noah A. Smith (July 2020). "Don't Stop Pretraining: Adapt Language Models to Domains and Tasks." In: *Proceedings of the 58th Annual Meeting of the Association for Computational Linguistics*. Online: Association for Computational Linguistics, pp. 8342–8360. DOI: 10.18653/v1/2020.acl-main.740. URL: https://www.aclweb.org/anthology/2020.acl-main.740.

Gururangan, Suchin, Swabha Swayamdipta, Omer Levy, Roy Schwartz, Samuel Bowman, and Noah A. Smith (June 2018). "Annotation Artifacts in Natural Language Inference Data." In: *Proceedings of the 2018 Conference of the North American Chapter of the Association for Computational Linguistics: Human Language Technologies, Volume 2 (Short Papers)*. New Orleans, Louisiana: Association for Computational Linguistics, pp. 107–112. DOI: 10.18653/v1/N18-2017. URL: https://aclanthology.org/N18-2017.

Guzmán, Francisco, Peng-Jen Chen, Myle Ott, Juan Pino, Guillaume Lample, Philipp Koehn, Vishrav Chaudhary, and Marc'Aurelio

Ranzato (Nov. 2019). "The FLORES Evaluation Datasets for Low-Resource Machine Translation: Nepali–English and Sinhala–English." In: *Proceedings of the 2019 Conference on Empirical Methods in Natural Language Processing and the 9th International Joint Conference on Natural Language Processing (EMNLP-IJCNLP)*. Hong Kong, China: Association for Computational Linguistics, pp. 6098–6111. DOI: [10.18653/v1/D19-1632](https://www.aclweb.org/anthology/D19-1632). URL: https://www.aclweb.org/anthology/D19-1632.

Haddow, Barry, Rachel Bawden, Antonio Valerio Miceli Barone, Jindvrich Helcl, and Alexandra Birch (2021). "Survey of Low-Resource Machine Translation." In: *ArXiv* abs/2109.00486.

Hadgu, Asmelash Teka, Adam Beaudoin, and Abel Aregawi (2020). "Evaluating Amharic Machine Translation." In: *arXiv e-prints 2003.14386*. arXiv: 2003.14386 [cs.CL].

Hammarström, Harald, Robert Forkel, and Martin Haspelmath (2018). "Glottolog 3.0." In: *Max Planck Institute for the Science of Human History*.

Harris, Z. (1954). "Distributional structure." In: *Word* 10.23, pp. 146–162.

Hasan, Tahmid, Abhik Bhattacharjee, Md. Saiful Islam, Kazi Mubasshir, Yuan-Fang Li, Yong-Bin Kang, M. Sohel Rahman, and Rifat Shahriyar (Aug. 2021). "XL-Sum: Large-Scale Multilingual Abstractive Summarization for 44 Languages." In: *Findings of the Association for Computational Linguistics: ACL-IJCNLP 2021*. Online: Association for Computational Linguistics, pp. 4693–4703. DOI: 10.18653/v1/2021.findings-acl.413. URL: https://aclanthology.org/2021.findings-acl.413.

Hassan, Hany et al. (2018). "Achieving Human Parity on Automatic Chinese to English News Translation." In: *CoRR* abs/1803.05567. URL: http://arxiv.org/abs/1803.05567.

Hassan, Samer and Rada Mihalcea (Aug. 2009). "Cross-lingual Semantic Relatedness Using Encyclopedic Knowledge." In: *Proceedings of the 2009 Conference on Empirical Methods in Natural Language Processing*. Singapore: Association for Computational Linguistics, pp. 1192–1201. URL: https://www.aclweb.org/anthology/D09-1124.

He, Pengcheng, Jianfeng Gao, and Weizhu Chen (2021b). *DeBERTaV3: Improving DeBERTa using ELECTRA-Style Pre-Training with Gradient-Disentangled Embedding Sharing*. arXiv: 2111.09543 [cs.CL].

He, Pengcheng, Jianfeng Gao, and Weizhu Chen (2021a). "DeBERTaV3: Improving DeBERTa using ELECTRA-Style Pre-Training with Gradient-Disentangled Embedding Sharing." In: *ArXiv* abs/2111.09543.

Heafield, Kenneth (July 2011). "KenLM: Faster and Smaller Language Model Queries." In: *Proceedings of the Sixth Workshop on Statistical Machine Translation*. Edinburgh, Scotland: Association for Computational Linguistics, pp. 187–197. URL: https://aclanthology.org/W11-2123.

Hedderich, Michael A., David Adelani, Dawei Zhu, Jesujoba Alabi, Udia Markus, and Dietrich Klakow (Nov. 2020). "Transfer Learning and Distant Supervision for Multilingual Transformer Models: A Study on African Languages." In: *Proceedings of the 2020 Conference on Empirical Methods in Natural Language Processing (EMNLP)*. Online: Association for Computational Linguistics, pp. 2580–2591. DOI: 10.18653/v1/2020.emnlp-main.204. URL: https://www.aclweb.org/anthology/2020.emnlp-main.204.

Hedderich, Michael A. and Dietrich Klakow (July 2018). "Training a Neural Network in a Low-Resource Setting on Automatically Annotated Noisy Data." In: *Proceedings of the Workshop on Deep Learning Approaches for Low-Resource NLP*. Association for Computational Linguistics.

Hedderich, Michael A., Lukas Lange, and Dietrich Klakow (2021). "ANEA: Distant Supervision for Low-Resource Named Entity Recognition." In: *ArXiv* abs/2102.13129.

Hochreiter, Sepp and Jürgen Schmidhuber (1997). "Long Short-Term Memory." In: *Neural Computation* 9, pp. 1735–1780.

Houlsby, Neil, Andrei Giurgiu, Stanislaw Jastrzebski, Bruna Morrone, Quentin De Laroussilhe, Andrea Gesmundo, Mona Attariyan, and Sylvain Gelly (June 2019). "Parameter-Efficient Transfer Learning for NLP." In: *Proceedings of the 36th International Conference on Machine Learning*. Ed. by Kamalika Chaudhuri and Ruslan Salakhutdinov. Vol. 97. Proceedings of Machine Learning Research. PMLR, pp. 2790–2799. URL: https://proceedings.mlr.press/v97/houlsby19a.html.

Howard, Jeremy and Sebastian Ruder (2018). "Universal Language Model Fine-tuning for Text Classification." In: *Proceedings of ACL 2018*.

Hu, Edward J, yelong shen, Phillip Wallis, Zeyuan Allen-Zhu, Yuanzhi Li, Shean Wang, Lu Wang, and Weizhu Chen (2022). "LoRA: Low-Rank Adaptation of Large Language Models." In: *International Conference on Learning Representations*. URL: https://openreview.net/forum?id=nZeVKeeFYf9.

Hu, Junjie, Sebastian Ruder, Aditya Siddhant, Graham Neubig, Orhan Firat, and Melvin Johnson (July 2020). "XTREME: A Massively Multilingual Multi-task Benchmark for Evaluating Cross-lingual Generalisation." In: *Proceedings of the 37th International Conference on Machine Learning*. Ed. by Hal Daumé III and Aarti Singh. Vol. 119. Proceedings of Machine Learning Research. PMLR, pp. 4411–4421. URL: https://proceedings.mlr.press/v119/hu20b.html.

Huang, Zhiheng, W. Xu, and Kailiang Yu (2015). "Bidirectional LSTM-CRF Models for Sequence Tagging." In: *ArXiv* abs/1508.01991.

Hvingelby, Rasmus, Amalie Brogaard Pauli, Maria Barrett, Christina Rosted, Lasse Malm Lidegaard, and Anders Søgaard (May 2020). "DaNE: A Named Entity Resource for Danish." English. In: *Proceed-

*ings of the 12th Language Resources and Evaluation Conference*. Marseille, France: European Language Resources Association, pp. 4597–4604. ISBN: 979-10-95546-34-4. URL: https://aclanthology.org/2020.lrec-1.565.

Ishola, Olájídé and Daniel Zeman (May 2020). "Yorùbá Dependency Treebank (YTB)." English. In: *Proceedings of the 12th Language Resources and Evaluation Conference*. Marseille, France: European Language Resources Association, pp. 5178–5186. ISBN: 979-10-95546-34-4. URL: https://aclanthology.org/2020.lrec-1.637.

Jehovah's Witnesses (2019). URL: https://www.jw.org/tw.

Jiang, Chao, Hsiang-Fu Yu, Cho-Jui Hsieh, and Kai-Wei Chang (June 2018). "Learning Word Embeddings for Low-Resource Languages by PU Learning." In: *Proceedings of the 2018 Conference of the North American Chapter of the Association for Computational Linguistics: Human Language Technologies, Volume 1 (Long Papers)*. New Orleans, Louisiana: Association for Computational Linguistics, pp. 1024–1034. DOI: 10.18653/v1/N18-1093. URL: https://www.aclweb.org/anthology/N18-1093.

Jiao, Xiaoqi, Yichun Yin, Lifeng Shang, Xin Jiang, Xiao Chen, Linlin Li, Fang Wang, and Qun Liu (Nov. 2020). "TinyBERT: Distilling BERT for Natural Language Understanding." In: *Findings of the Association for Computational Linguistics: EMNLP 2020*. Online: Association for Computational Linguistics, pp. 4163–4174. DOI: 10.18653/v1/2020.findings-emnlp.372. URL: https://aclanthology.org/2020.findings-emnlp.372.

Johansen, Bjarte (2019). "Named-Entity Recognition for Norwegian." In: *Proceedings of the 22nd Nordic Conference on Computational Linguistics, NoDaLiDa*.

Johnson, Melvin et al. (Oct. 2017a). "Google's Multilingual Neural Machine Translation System: Enabling Zero-Shot Translation." In: *Transactions of the Association for Computational Linguist* 5, pp. 339–351.

Johnson, Melvin et al. (2017b). "Google's Multilingual Neural Machine Translation System: Enabling Zero-Shot Translation." In: *Transactions of the Association for Computational Linguistics* 5, pp. 339–351. DOI: 10.1162/tacl_a_00065. URL: https://www.aclweb.org/anthology/Q17-1024.

Joshi, Ishani, Purvi Koringa, and Suman Mitra (2019). "Word Embeddings in Low Resource Gujarati Language." In: *2019 International Conference on Document Analysis and Recognition Workshops (ICDARW)*. Vol. 5, pp. 110–115. DOI: 10.1109/ICDARW.2019.40090.

Joshi, Pratik, Sebastin Santy, Amar Budhiraja, Kalika Bali, and Monojit Choudhury (July 2020). "The State and Fate of Linguistic Diversity and Inclusion in the NLP World." In: *Proceedings of the 58th Annual Meeting of the Association for Computational Linguistics*. Online: Association for Computational Linguistics, pp. 6282–6293. DOI:

10.18653/v1/2020.acl-main.560. URL: https://aclanthology.org/2020.acl-main.560.

Joubarne, Colette and Diana Inkpen (2011). "Comparison of Semantic Similarity for Different Languages Using the Google n-gram Corpus and Second-Order Co-occurrence Measures." In: *Advances in Artificial Intelligence*. Ed. by Cory Butz and Pawan Lingras. Berlin, Heidelberg: Springer Berlin Heidelberg, pp. 216–221. ISBN: 978-3-642-21043-3.

Junior, C., H. Macedo, T. Bispo, F. Oliveira, N. Silva, and L. Barbosa (2015). "Paramopama: a Brazilian-Portuguese Corpus for Named Entity Recognition." In: *12th National Meeting on Artificial and Computational Intelligence (ENIAC)*.

K, Karthikeyan, Zihan Wang, Stephen Mayhew, and Dan Roth (2020). "Cross-Lingual Ability of Multilingual BERT: An Empirical Study." In: *International Conference on Learning Representations*. URL: https://openreview.net/forum?id=HJeT3yrtDr.

Kahane, Sylvain, Martine Vanhove, Rayan Ziane, and Bruno Guillaume (Dec. 2021). "A morph-based and a word-based treebank for Beja." In: *Proceedings of the 20th International Workshop on Treebanks and Linguistic Theories (TLT, SyntaxFest 2021)*. Sofia, Bulgaria: Association for Computational Linguistics, pp. 48–60. URL: https://aclanthology.org/2021.tlt-1.5.

Karamolegkou, Antonia and Sara Stymne (May 2021). "Investigation of Transfer Languages for Parsing Latin: Italic Branch vs. Hellenic Branch." In: *Proceedings of the 23rd Nordic Conference on Computational Linguistics (NoDaLiDa)*. Reykjavik, Iceland (Online): Linköping University Electronic Press, Sweden, pp. 315–320. URL: https://aclanthology.org/2021.nodalida-main.32.

Kastner, Itamar and Matthew A. Tucker (2019). "Non-concatenative Morphology." In.

Khairunnisa, Siti Oryza, Aizhan Imankulova, and Mamoru Komachi (Dec. 2020). "Towards a Standardized Dataset on Indonesian Named Entity Recognition." In: *Proceedings of the 1st Conference of the Asia-Pacific Chapter of the Association for Computational Linguistics and the 10th International Joint Conference on Natural Language Processing: Student Research Workshop*. Suzhou, China: Association for Computational Linguistics, pp. 64–71. URL: https://aclanthology.org/2020.aacl-srw.10.

Kingma, Diederik P. and Jimmy Ba (2015). "Adam: A method for stochastic optimization." In: *International Conference for Learning Representations (ICLR)*.

Ko, Wei-Jen, Ahmed El-Kishky, Adithya Renduchintala, Vishrav Chaudhary, Naman Goyal, Francisco Guzmán, Pascale Fung, Philipp Koehn, and Mona Diab (Aug. 2021). "Adapting High-resource NMT Models to Translate Low-resource Related Languages without Parallel Data." In: *Proceedings of the 59th Annual Meeting of the Association* *for Computational Linguistics and the 11th International Joint Conference on Natural Language Processing (Volume 1: Long Papers)*. Online: Association for Computational Linguistics, pp. 802–812. DOI: `10.18653/v1/2021.acl-long.66`. URL: `https://aclanthology.org/2021.acl-long.66`.

Koehn, Philipp et al. (June 2007). "Moses: Open Source Toolkit for Statistical Machine Translation." In: *Proceedings of the 45th Annual Meeting of the Association for Computational Linguistics Companion Volume Proceedings of the Demo and Poster Sessions*. Prague, Czech Republic: Association for Computational Linguistics, pp. 177–180. URL: `https://aclanthology.org/P07-2045`.

Konoshenko, Maria Yu and Dasha Shavarina (2019). "A microtypological survey of noun classes in Kwa." In: *Journal of African Languages and Linguistics* 40, pp. 114 –75.

Kreutzer, Julia, Jasmijn Bastings, and Stefan Riezler (Nov. 2019). "Joey NMT: A Minimalist NMT Toolkit for Novices." In: *Proceedings of the 2019 Conference on Empirical Methods in Natural Language Processing and the 9th International Joint Conference on Natural Language Processing (EMNLP-IJCNLP): System Demonstrations*. Hong Kong, China: Association for Computational Linguistics, pp. 109–114. DOI: `10.18653/v1/D19-3019`. URL: `https://aclanthology.org/D19-3019`.

Kreutzer, Julia et al. (2021). "Quality at a Glance: An Audit of Web-Crawled Multilingual Datasets." In: *ArXiv* abs/2103.12028.

Kreutzer, Julia et al. (2022). "Quality at a Glance: An Audit of Web-Crawled Multilingual Datasets." In: *Transactions of the Association for Computational Linguistics* 10, pp. 50–72. DOI: `10.1162/tacl_a_00447`. URL: `https://aclanthology.org/2022.tacl-1.4`.

Kudo, Taku (2018). "Subword Regularization: Improving Neural Network Translation Models with Multiple Subword Candidates." In: *Proceedings of the 56th Annual Meeting of the Association for Computational Linguistics, ACL 2018, Melbourne, Australia, July 15-20, 2018, Volume 1: Long Papers*. Ed. by Iryna Gurevych and Yusuke Miyao. Association for Computational Linguistics, pp. 66–75. DOI: `10.18653/v1/P18-1007`. URL: `https://aclanthology.org/P18-1007/`.

Lafferty, John D., Andrew McCallum, and Fernando C. N. Pereira (2001). "Conditional Random Fields: Probabilistic Models for Segmenting and Labeling Sequence Data." In: *Proceedings of the Eighteenth International Conference on Machine Learning*. ICML '01. San Francisco, CA, USA: Morgan Kaufmann Publishers Inc., pp. 282–289. ISBN: 1-55860-778-1. URL: `http://dl.acm.org/citation.cfm?id=645530.655813`.

Lakew, Surafel M., Matteo Negri, and Marco Turchi (2020). *Low Resource Neural Machine Translation: A Benchmark for Five African Languages*. arXiv: `2003.14402 [cs.CL]`.

Lample, Guillaume, Miguel Ballesteros, Sandeep Subramanian, Kazuya Kawakami, and Chris Dyer (2016). "Neural Architectures for Named Entity Recognition." In: *Proceedings of NAACL-HLT 2016*.

Lample, Guillaume and Alexis Conneau (2019). "Cross-lingual Language Model Pretraining." In: *Advances in Neural Information Processing Systems*. Ed. by H. Wallach, H. Larochelle, A. Beygelzimer, F. d'Alché-Buc, E. Fox, and R. Garnett. Vol. 32. Curran Associates, Inc., pp. 7059–7069. URL: https://proceedings.neurips.cc/paper/2019/file/c04c19c2c2474dbf5f7ac4372c5b9af1-Paper.pdf.

Lample, Guillaume, Alexis Conneau, Ludovic Denoyer, and Marc'Aurelio Ranzato (2018a). "Unsupervised Machine Translation Using Monolingual Corpora Only." In: *International Conference on Learning Representations*. URL: https://openreview.net/forum?id=rkYTTf-AZ.

Lample, Guillaume, Alexis Conneau, Marc'Aurelio Ranzato, Ludovic Denoyer, and Hervé Jégou (2018b). "Word translation without parallel data." In: *6th International Conference on Learning Representations, ICLR 2018, Vancouver, BC, Canada, April 30 - May 3, 2018, Conference Track Proceedings*. URL: https://openreview.net/forum?id=H196sainb.

Lange, Lukas, Michael A. Hedderich, and Dietrich Klakow (2019). "Feature-Dependent Confusion Matrices for Low-Resource NER Labeling with Noisy Labels." In: *Proceedings of the 2019 Conference on Empirical Methods in Natural Language Processing and the 9th International Joint Conference on Natural Language Processing (EMNLP-IJCNLP)*. Association for Computational Linguistics, pp. 3545–3550. DOI: 10.18653/v1/D19-1362.

Läubli, Samuel, Rico Sennrich, and Martin Volk (Nov. 2018). "Has Machine Translation Achieved Human Parity? A Case for Document-level Evaluation." In: *Proceedings of the 2018 Conference on Empirical Methods in Natural Language Processing*. Brussels, Belgium: Association for Computational Linguistics, pp. 4791–4796. DOI: 10.18653/v1/D18-1512. URL: https://aclanthology.org/D18-1512.

Lauscher, Anne, Vinit Ravishankar, Ivan Vulić, and Goran Glavaš (Nov. 2020). "From Zero to Hero: On the Limitations of Zero-Shot Language Transfer with Multilingual Transformers." In: *Proceedings of the 2020 Conference on Empirical Methods in Natural Language Processing (EMNLP)*. Online: Association for Computational Linguistics, pp. 4483–4499. DOI: 10.18653/v1/2020.emnlp-main.363. URL: https://www.aclweb.org/anthology/2020.emnlp-main.363.

Le, Hang, Loïc Vial, Jibril Frej, Vincent Segonne, Maximin Coavoux, Benjamin Lecouteux, Alexandre Allauzen, Benoit Crabbé, Laurent Besacier, and Didier Schwab (May 2020). "FlauBERT: Unsupervised Language Model Pre-training for French." English. In: *Proceedings of the 12th Language Resources and Evaluation Conference*. Marseille, France: European Language Resources Association, pp. 2479–2490.

ISBN: 979-10-95546-34-4. URL: https://aclanthology.org/2020.lrec-1.302.

Lee, En-Shiun, Sarubi Thillainathan, Shravan Nayak, Surangika Ranathunga, David Adelani, Ruisi Su, and Arya McCarthy (May 2022). "Pre-Trained Multilingual Sequence-to-Sequence Models: A Hope for Low-Resource Language Translation?" In: *Findings of the Association for Computational Linguistics: ACL 2022*. Dublin, Ireland: Association for Computational Linguistics, pp. 58–67. DOI: 10.18653/v1/2022.findings-acl.6. URL: https://aclanthology.org/2022.findings-acl.6.

Lee, Kuang-Huei, Xiaodong He, Lei Zhang, and Linjun Yang (2018). "CleanNet: Transfer Learning for Scalable Image Classifier Training With Label Noise." In: *2018 IEEE Conference on Computer Vision and Pattern Recognition, CVPR 2018, Salt Lake City, UT, USA, June 18-22, 2018*, pp. 5447–5456. DOI: 10.1109/CVPR.2018.00571.

Leng, Yichong, Xu Tan, Tao Qin, Xiang-Yang Li, and Tie-Yan Liu (2019). "Unsupervised Pivot Translation for Distant Languages." In: *Proceedings of the 57th Annual Meeting of the Association for Computational Linguistics*, pp. 175–183.

Leviant, Ira and Roi Reichart (2015). "Judgment Language Matters: Multilingual Vector Space Models for Judgment Language Aware Lexical Semantics." In: *CoRR* abs/1508.00106. arXiv: 1508.00106. URL: http://arxiv.org/abs/1508.00106.

Lewis, M Paul (2009). *Ethnologue: Languages of the world Sixteenth Edition*. SIL international. URL: http://www.ethnologue.com/16/.

Lewis, Mike, Yinhan Liu, Naman Goyal, Marjan Ghazvininejad, Abdelrahman Mohamed, Omer Levy, Veselin Stoyanov, and Luke Zettlemoyer (July 2020). "BART: Denoising Sequence-to-Sequence Pre-training for Natural Language Generation, Translation, and Comprehension." In: *Proceedings of the 58th Annual Meeting of the Association for Computational Linguistics*. Online: Association for Computational Linguistics, pp. 7871–7880. DOI: 10.18653/v1/2020.acl-main.703. URL: https://aclanthology.org/2020.acl-main.703.

Lhoest, Quentin et al. (Nov. 2021). "Datasets: A Community Library for Natural Language Processing." In: *Proceedings of the 2021 Conference on Empirical Methods in Natural Language Processing: System Demonstrations*. Online and Punta Cana, Dominican Republic: Association for Computational Linguistics, pp. 175–184. DOI: 10.18653/v1/2021.emnlp-demo.21. URL: https://aclanthology.org/2021.emnlp-demo.21.

Li, Wen, Limin Wang, Wei Li, Eirikur Agustsson, and Luc Van Gool (2017). "WebVision Database: Visual Learning and Understanding from Web Data." In: *CoRR* abs/1708.02862. arXiv: 1708.02862.

Liang, Chen, Yue Yu, Haoming Jiang, Siawpeng Er, Ruijia Wang, Tuo Zhao, and Chao Zhang (2020). "BOND: Bert-Assisted Open-Domain Named Entity Recognition with Distant Supervision." In: *ACM*

*SIGKDD International Conference on Knowledge Discovery and Data Mining*.

Lin, Xi Victoria et al. (2021). "Few-shot Learning with Multilingual Language Models." In: *ArXiv* abs/2112.10668.

Lin, Ying, Cash Costello, Boliang Zhang, Di Lu, Heng Ji, James Mayfield, and Paul McNamee (July 2018). "Platforms for Non-speakers Annotating Names in Any Language." In: *Proceedings of ACL 2018, System Demonstrations*. Melbourne, Australia: Association for Computational Linguistics, pp. 1–6. DOI: 10.18653/v1/P18-4001. URL: https://www.aclweb.org/anthology/P18-4001.

Lin, Yu-Hsiang et al. (July 2019). "Choosing Transfer Languages for Cross-Lingual Learning." In: *Proceedings of the 57th Annual Meeting of the Association for Computational Linguistics*. Florence, Italy: Association for Computational Linguistics, pp. 3125–3135. DOI: 10.18653/v1/P19-1301. URL: https://aclanthology.org/P19-1301.

Lison, Pierre, Jeremy Barnes, and Aliaksandr Hubin (Aug. 2021). "skweak: Weak Supervision Made Easy for NLP." In: *Proceedings of the 59th Annual Meeting of the Association for Computational Linguistics and the 11th International Joint Conference on Natural Language Processing: System Demonstrations*. Online: Association for Computational Linguistics, pp. 337–346. DOI: 10.18653/v1/2021.acl-demo.40. URL: https://aclanthology.org/2021.acl-demo.40.

Littell, Patrick, David R. Mortensen, Ke Lin, Katherine Kairis, Carlisle Turner, and Lori Levin (Apr. 2017). "URIEL and lang2vec: Representing languages as typological, geographical, and phylogenetic vectors." In: *Proceedings of the 15th Conference of the European Chapter of the Association for Computational Linguistics: Volume 2, Short Papers*. Valencia, Spain: Association for Computational Linguistics, pp. 8–14. URL: https://aclanthology.org/E17-2002.

Liu, Haokun, Derek Tam, Muqeeth Mohammed, Jay Mohta, Tenghao Huang, Mohit Bansal, and Colin Raffel (2022). "Few-Shot Parameter-Efficient Fine-Tuning is Better and Cheaper than In-Context Learning." In: *Advances in Neural Information Processing Systems*. Ed. by Alice H. Oh, Alekh Agarwal, Danielle Belgrave, and Kyunghyun Cho. URL: https://openreview.net/forum?id=rBCvMG-JsPd.

Liu, Pengfei, Jinlan Fu, Yang Xiao, Weizhe Yuan, Shuaichen Chang, Junqi Dai, Yixin Liu, Zihuiwen Ye, and Graham Neubig (Aug. 2021). "ExplainaBoard: An Explainable Leaderboard for NLP." In: *Proceedings of the 59th Annual Meeting of the Association for Computational Linguistics and the 11th International Joint Conference on Natural Language Processing: System Demonstrations*. Online: Association for Computational Linguistics, pp. 280–289. DOI: 10.18653/v1/2021.acl-demo.34. URL: https://aclanthology.org/2021.acl-demo.34.

Liu, Weijie, Peng Zhou, Zhiruo Wang, Zhe Zhao, Haotang Deng, and Qi Ju (July 2020a). "FastBERT: a Self-distilling BERT with Adaptive Inference Time." In: *Proceedings of the 58th Annual Meeting of the*

*Association for Computational Linguistics*. Online: Association for Computational Linguistics, pp. 6035–6044. DOI: 10.18653/v1/2020.acl-main.537. URL: https://aclanthology.org/2020.acl-main.537.

Liu, Yinhan, Jiatao Gu, Naman Goyal, Xian Li, Sergey Edunov, Marjan Ghazvininejad, Mike Lewis, and Luke Zettlemoyer (2020b). "Multilingual Denoising Pre-training for Neural Machine Translation." In: *Transactions of the Association for Computational Linguistics* 8, pp. 726–742. DOI: 10.1162/tacl_a_00343. URL: https://aclanthology.org/2020.tacl-1.47.

Liu, Yinhan, Myle Ott, Naman Goyal, Jingfei Du, Mandar Joshi, Danqi Chen, Omer Levy, Mike Lewis, Luke S. Zettlemoyer, and Veselin Stoyanov (2019). "RoBERTa: A Robustly Optimized BERT Pretraining Approach." In: *ArXiv* abs/1907.11692.

Liu, Zihan, Genta Indra Winata, and Pascale Fung (Aug. 2021). "Continual Mixed-Language Pre-Training for Extremely Low-Resource Neural Machine Translation." In: *Findings of the Association for Computational Linguistics: ACL-IJCNLP 2021*. Online: Association for Computational Linguistics, pp. 2706–2718. DOI: 10.18653/v1/2021.findings-acl.239. URL: https://aclanthology.org/2021.findings-acl.239.

Luo, Fuli, Wei Wang, Jiahao Liu, Yijia Liu, Bin Bi, Songfang Huang, Fei Huang, and Luo Si (Aug. 2021). "VECO: Variable and Flexible Cross-lingual Pre-training for Language Understanding and Generation." In: *Proceedings of the 59th Annual Meeting of the Association for Computational Linguistics and the 11th International Joint Conference on Natural Language Processing (Volume 1: Long Papers)*. Online: Association for Computational Linguistics, pp. 3980–3994. DOI: 10.18653/v1/2021.acl-long.308. URL: https://aclanthology.org/2021.acl-long.308.

MBS (2020). *Téereb Injiil: La Bible Wolof – Ancien Testament*. http://biblewolof.com/.

Ma, Xuezhe and Eduard Hovy (Aug. 2016). "End-to-end Sequence Labeling via Bi-directional LSTM-CNNs-CRF." In: *Proceedings of the 54th Annual Meeting of the Association for Computational Linguistics (Volume 1: Long Papers)*. Berlin, Germany: Association for Computational Linguistics, pp. 1064–1074. DOI: 10.18653/v1/P16-1101. URL: https://www.aclweb.org/anthology/P16-1101.

Mabuya, Rooweither, Jade Abbott, and Vukosi Marivate (June 2021). *Umsuka English - isiZulu Parallel Corpus*. Thank you to Facebook Research for funding the creation of this dataset. DOI: 10.5281/zenodo.5035171. URL: https://doi.org/10.5281/zenodo.5035171.

Madaan, Lovish, Soumya Sharma, and Parag Singla (Nov. 2020). "Transfer Learning for Related Languages: Submissions to the WMT20 Similar Language Translation Task." In: *Proceedings of the Fifth Conference on Machine Translation*. Online: Association for Computational

Linguistics, pp. 402–408. URL: https://aclanthology.org/2020.wmt-1.46.

Mager, Manuel et al. (June 2021). "Findings of the AmericasNLP 2021 Shared Task on Open Machine Translation for Indigenous Languages of the Americas." In: *Proceedings of the First Workshop on Natural Language Processing for Indigenous Languages of the Americas*. Online: Association for Computational Linguistics, pp. 202–217. DOI: 10.18653/v1/2021.americasnlp-1.23. URL: https://aclanthology.org/2021.americasnlp-1.23.

Magnini, Bernardo et al. (May 2008). "Evaluation of Natural Language Tools for Italian: EVALITA 2007." In: *Proceedings of the Sixth International Conference on Language Resources and Evaluation (LREC'08)*. Marrakech, Morocco: European Language Resources Association (ELRA). URL: http://www.lrec-conf.org/proceedings/lrec2008/pdf/630_paper.pdf.

Mahajan, Dhruv, Ross Girshick, Vignesh Ramanathan, Kaiming He, Manohar Paluri, Yixuan Li, Ashwin Bharambe, and Laurens van der Maaten (2018). "Exploring the Limits of Weakly Supervised Pretraining." In: *Computer Vision – ECCV 2018*. Ed. by Vittorio Ferrari, Martial Hebert, Cristian Sminchisescu, and Yair Weiss. Springer International Publishing, pp. 185–201.

Malema, Gabofetswe and Ontiretse Ishmael (Feb. 2022). "Parts of Speech Tagging: A Setswana Relative." In: *Journal of Physics: Conference Series* 2188.1, p. 012002. DOI: 10.1088/1742-6596/2188/1/012002. URL: https://doi.org/10.1088/1742-6596/2188/1/012002.

Marchisio, Kelly, Kevin Duh, and Philipp Koehn (Nov. 2020). "When Does Unsupervised Machine Translation Work?" In: *Proceedings of the Fifth Conference on Machine Translation*. Online: Association for Computational Linguistics, pp. 571–583. URL: https://aclanthology.org/2020.wmt-1.68.

Martin, Gati, Medard Edmund Mswahili, Young-Seob Jeong, and Jiyoung Woo (July 2022). "SwahBERT: Language Model of Swahili." In: *Proceedings of the 2022 Conference of the North American Chapter of the Association for Computational Linguistics: Human Language Technologies*. Seattle, United States: Association for Computational Linguistics, pp. 303–313. DOI: 10.18653/v1/2022.naacl-main.23. URL: https://aclanthology.org/2022.naacl-main.23.

Martin, Louis, Benjamin Muller, Pedro Javier Ortiz Suárez, Yoann Dupont, Laurent Romary, Éric de la Clergerie, Djamé Seddah, and Benoît Sagot (July 2020). "CamemBERT: a Tasty French Language Model." In: *Proceedings of the 58th Annual Meeting of the Association for Computational Linguistics*. Online: Association for Computational Linguistics, pp. 7203–7219. DOI: 10.18653/v1/2020.acl-main.645. URL: https://aclanthology.org/2020.acl-main.645.

Martinus, Laura and Jade Z Abbott (2019). "A focus on neural machine translation for African languages." In: *arXiv preprint arXiv:1906.05685*. URL: https://arxiv.org/abs/1906.05685.

Mayhew, Stephen, Snigdha Chaturvedi, Chen-Tse Tsai, and Dan Roth (Nov. 2019). "Named Entity Recognition with Partially Annotated Training Data." In: *Proceedings of the 23rd Conference on Computational Natural Language Learning (CoNLL)*. Hong Kong, China: Association for Computational Linguistics, pp. 645–655. DOI: 10.18653/v1/K19-1060. URL: https://www.aclweb.org/anthology/K19-1060.

Mayhew, Stephen and Dan Roth (2018). "TALEN: Tool for Annotation of Low-resource ENtities." In: *Proceedings of ACL 2018: System Demonstrations*. URL: http://aclweb.org/anthology/P18-4014.

Mayhew, Stephen, Tatiana Tsygankova, and Dan Roth (Nov. 2019). "ner and pos when nothing is capitalized." In: *Proceedings of the 2019 Conference on Empirical Methods in Natural Language Processing and the 9th International Joint Conference on Natural Language Processing (EMNLP-IJCNLP)*. Hong Kong, China: Association for Computational Linguistics, pp. 6256–6261. DOI: 10.18653/v1/D19-1650. URL: https://aclanthology.org/D19-1650.

McCann, Bryan, James Bradbury, Caiming Xiong, and Richard Socher (2017). "Learned in Translation: Contextualized Word Vectors." In: *NIPS*.

McCarthy, Arya D., Rachel Wicks, Dylan Lewis, Aaron Mueller, Winston Wu, Oliver Adams, Garrett Nicolai, Matt Post, and David Yarowsky (May 2020). "The Johns Hopkins University Bible Corpus: 1600+ Tongues for Typological Exploration." English. In: *Proceedings of the 12th Language Resources and Evaluation Conference*. Marseille, France: European Language Resources Association, pp. 2884–2892. ISBN: 979-10-95546-34-4. URL: https://aclanthology.org/2020.lrec-1.352.

McKellar, Cindy A. (2014). "An English to Xitsonga statistical machine translation system for the government domain." In.

Meyer, Josh et al. (2022). "BibleTTS: a large, high-fidelity, multilingual, and uniquely African speech corpus." In: *ArXiv* abs/2207.03546.

Mikolov, Tomas, Kai Chen, Greg Corrado, and Jeff Dean (2013a). "Efficient estimation of word representations in vector space." In: *Proceedings of the Workshop at International Conference on Learning Representations (ICLR)*, pp. 1–12.

Mikolov, Tomas, Ilya Sutskever, Kai Chen, Greg S Corrado, and Jeff Dean (2013b). "Distributed Representations of Words and Phrases and their Compositionality." In: *Advances in Neural Information Processing Systems*. Ed. by C.J. Burges, L. Bottou, M. Welling, Z. Ghahramani, and K.Q. Weinberger. Vol. 26. Curran Associates, Inc. URL: https://proceedings.neurips.cc/paper/2013/file/9aa42b31882ec039965f3c4923ce901b-Paper.pdf.

Moran, Steven, D McCloy, and R Wright (2014). *PHOIBLE Online. Max Planck Institute for Evolutionary Anthropology, Leipzig*.

Muennighoff, Niklas et al. (July 2023). "Crosslingual Generalization through Multitask Finetuning." In: *Proceedings of the 61st Annual Meeting of the Association for Computational Linguistics (Volume 1: Long Papers)*. Toronto, Canada: Association for Computational Linguistics, pp. 15991–16111. DOI: 10.18653/v1/2023.acl-long.891. URL: https://aclanthology.org/2023.acl-long.891.

Muhammad, Shamsuddeen Hassan et al. (2022). "NaijaSenti: A Nigerian Twitter Sentiment Corpus for Multilingual Sentiment Analysis." In: *ArXiv* abs/2201.08277.

Muhammad, Shamsuddeen Hassan et al. (2023). *AfriSenti: A Twitter Sentiment Analysis Benchmark for African Languages*. DOI: 10.48550/arXiv.2302.08956. URL: https://arxiv.org/abs/2302.08956.

Muller, Benjamin, Antonios Anastasopoulos, Benoît Sagot, and Djamé Seddah (June 2021). "When Being Unseen from mBERT is just the Beginning: Handling New Languages With Multilingual Language Models." In: *Proceedings of the 2021 Conference of the North American Chapter of the Association for Computational Linguistics: Human Language Technologies*. Online: Association for Computational Linguistics, pp. 448–462. DOI: 10.18653/v1/2021.naacl-main.38. URL: https://aclanthology.org/2021.naacl-main.38.

NLLB-Team et al. (2022). "No Language Left Behind: Scaling Human-Centered Machine Translation." In: *ArXiv* abs/2207.04672.

Neubig, Graham et al. (2017). "DyNet: The Dynamic Neural Network Toolkit." In: *ArXiv* abs/1701.03980.

Neudecker, Clemens (May 2016). "An Open Corpus for Named Entity Recognition in Historic Newspapers." In: *Proceedings of the Tenth International Conference on Language Resources and Evaluation (LREC'16)*. Portorož, Slovenia: European Language Resources Association (ELRA), pp. 4348–4352. URL: https://aclanthology.org/L16-1689.

Nguyen, T., C. K. Mummadi, T. P. N. Ngo, T. H. P. Nguyen, L. Beggel, and T. Brox (2020). "SELF: learning to filter noisy labels with self-ensembling." In: *International Conference on Learning Representations (ICLR)*.

Nichols, Johanna and Balthasar Bickel (2013). "Possessive Classification." In: *The World Atlas of Language Structures Online*. Ed. by Matthew S. Dryer and Martin Haspelmath. Leipzig: Max Planck Institute for Evolutionary Anthropology. URL: https://wals.info/chapter/59.

Nivre, Joakim et al. (May 2016). "Universal Dependencies v1: A Multilingual Treebank Collection." In: *Proceedings of the Tenth International Conference on Language Resources and Evaluation (LREC'16)*. Portorož, Slovenia: European Language Resources Association (ELRA), pp. 1659–1666. URL: https://aclanthology.org/L16-1262.

Niyongabo, Rubungo Andre, Qu Hong, Julia Kreutzer, and Li Huang (Dec. 2020). "KINNEWS and KIRNEWS: Benchmarking Cross-Lingual Text Classification for Kinyarwanda and Kirundi." In: *Proceedings of the 28th International Conference on Computational Linguistics*. Barcelona, Spain (Online): International Committee on Computational Linguistics, pp. 5507–5521. DOI: 10.18653/v1/2020.coling-main.480. URL: https://www.aclweb.org/anthology/2020.coling-main.480.

Nurse, Derek and Gerard Philippson, eds. (Mar. 2006). *The Bantu Languages*. Routledge Language Family Series. London, England: Routledge.

Nyoni, Evander and Bruce A. Bassett (2021). "Low-Resource Neural Machine Translation for Southern African Languages." In: *ArXiv* abs/2104.00366.

Nzeyimana, Antoine and Andre Niyongabo Rubungo (May 2022). "KinyaBERT: a Morphology-aware Kinyarwanda Language Model." In: *Proceedings of the 60th Annual Meeting of the Association for Computational Linguistics (Volume 1: Long Papers)*. Dublin, Ireland: Association for Computational Linguistics, pp. 5347–5363. DOI: 10.18653/v1/2022.acl-long.367. URL: https://aclanthology.org/2022.acl-long.367.

Obeid, Ossama, Nasser Zalmout, Salam Khalifa, Dima Taji, Mai Oudah, Bashar Alhafni, Go Inoue, Fadhl Eryani, Alexander Erdmann, and Nizar Habash (May 2020). "CAMeL Tools: An Open Source Python Toolkit for Arabic Natural Language Processing." English. In: *Proceedings of the 12th Language Resources and Evaluation Conference*. Marseille, France: European Language Resources Association, pp. 7022–7032. ISBN: 979-10-95546-34-4. URL: https://aclanthology.org/2020.lrec-1.868.

Offiong Mensah, Eyo (2012). "Grammaticalization in Nigerian Pidgin." In: *Íkala, revista de lenguaje y cultura* 17.2, pp. 167–179.

Ogayo, Perez, Graham Neubig, and Alan W. Black (2022). "Building African Voices." In: *ArXiv* abs/2207.00688.

Ogueji, Kelechi, Yuxin Zhu, and Jimmy Lin (Nov. 2021). "Small Data? No Problem! Exploring the Viability of Pretrained Multilingual Language Models for Low-resourced Languages." In: *Proceedings of the 1st Workshop on Multilingual Representation Learning*. Punta Cana, Dominican Republic: Association for Computational Linguistics, pp. 116–126. DOI: 10.18653/v1/2021.mrl-1.11. URL: https://aclanthology.org/2021.mrl-1.11.

Ogundepo, Odunayo et al. (2023). *AfriQA: Cross-lingual Open-Retrieval Question Answering for African Languages*. arXiv: 2305.06897 [cs.CL].

Ojarikre, Anthony (2013). "Perspectives and problems of codifying Nigerian pidgin English orthography." In: *Perspectives* 3.12.

Öktem, Alp, Eric DeLuca, Rodrigue Bashizi, Eric Paquin, and Grace Tang (2021). "Congolese Swahili Machine Translation for Humani-

tarian Response." In: *AfricaNLP Workshop*. arXiv: 2103.10734. URL: https://arxiv.org/abs/2103.10734.

Öktem, Alp, Mirko Plitt, and Grace Tang (2020). "Tigrinya Neural Machine Translation with Transfer Learning for Humanitarian Response." In: *AfricaNLP Workshop*. arXiv: 2003.11523. URL: https://arxiv.org/abs/2003.11523.

Onovbiona, Ijite Blessing (Dec. 2012). "Serial Verb Construction in Nigerian Pidgin." In.

Onyenwe, Ikechukwu E. and Mark Hepple (2016). "Predicting Morphologically-Complex Unknown Words in Igbo." In: *Text, Speech, and Dialogue*. Ed. by Petr Sojka, Aleš Horák, Ivan Kopeček, and Karel Pala. Cham: Springer International Publishing, pp. 206–214. ISBN: 978-3-319-45510-5.

Onyenwe, Ikechukwu E., Mark Hepple, Uchechukwu Chinedu, and Ignatius Ezeani (May 2019). "Toward an Effective Igbo Part-of-Speech Tagger." In: *ACM Trans. Asian Low-Resour. Lang. Inf. Process.* 18.4. ISSN: 2375-4699. DOI: 10.1145/3314942. URL: https://doi.org/10.1145/3314942.

OpenAI (2023). "GPT-4 Technical Report." In: *ArXiv* abs/2303.08774. URL: https://api.semanticscholar.org/CorpusID:257532815.

Orife, Iroro Fred Ọ̀nọ̀mẹ̀ (2018a). "Sequence-to-Sequence Learning for Automatic Yorùbá Diacritic Restoration." In: *Proceedings of the Interspeech*, pp. 27–35.

Orife, Iroro (2018b). "Attentive Sequence-to-Sequence Learning for Diacritic Restoration of YorùBá Language Text." In: *Proc. Interspeech 2018*, pp. 2848–2852.

Orife, Iroro, D. Adelani, Timi E. Fasubaa, Victor Williamson, Wuraola Fisayo Oyewusi, Olamilekan Wahab, and Kọ́lá Túbọ̀sún (2020). "Improving Yorùbá Diacritic Restoration." In: *ArXiv* abs/2003.10564.

Ortiz Suárez, Pedro Javier, Benoît Sagot, and Laurent Romary (2019). "Asynchronous pipelines for processing huge corpora on medium to low resource infrastructures." en. In: ed. by Piotr Bański, Adrien Barbaresi, Hanno Biber, Evelyn Breiteneder, Simon Clematide, Marc Kupietz, Harald Lüngen, and Caroline Iliadi. Proceedings of the Workshop on Challenges in the Management of Large Corpora (CMLC-7) 2019. Cardiff, 22nd July 2019. Mannheim: Leibniz-Institut für Deutsche Sprache, pp. 9 –16. DOI: 10.14618/ids-pub-9021. URL: http://nbn-resolving.de/urn:nbn:de:bsz:mh39-90215.

Osam, E. Kweku (2003). "An Introduction to the Verbal and Multi-Verbal System of Akan." In: *Proceedings of the workshop on Multi-Verb Constructions*. URL: https://web.archive.org/web/20140407085659/http://www.ling.hf.ntnu.no/tross/osam.pdf.

Ott, Myle, Sergey Edunov, Alexei Baevski, Angela Fan, Sam Gross, Nathan Ng, David Grangier, and Michael Auli (June 2019). "fairseq: A Fast, Extensible Toolkit for Sequence Modeling." In: *Proceedings of the 2019 Conference of the North American Chapter of the Associa-* *tion for Computational Linguistics (Demonstrations)*. Minneapolis, Minnesota: Association for Computational Linguistics, pp. 48–53. DOI: 10.18653/v1/N19-4009. URL: https://aclanthology.org/N19-4009.

Ouyang, Xuan, Shuohuan Wang, Chao Pang, Yu Sun, Hao Tian, Hua Wu, and Haifeng Wang (Nov. 2021). "ERNIE-M: Enhanced Multilingual Representation by Aligning Cross-lingual Semantics with Monolingual Corpora." In: *Proceedings of the 2021 Conference on Empirical Methods in Natural Language Processing*. Online and Punta Cana, Dominican Republic: Association for Computational Linguistics, pp. 27–38. DOI: 10.18653/v1/2021.emnlp-main.3. URL: https://aclanthology.org/2021.emnlp-main.3.

Palen-Michel, Chester, June Kim, and Constantine Lignos (June 2022). "Multilingual Open Text Release 1: Public Domain News in 44 Languages." In: *Proceedings of the Language Resources and Evaluation Conference*. Marseille, France: European Language Resources Association, pp. 2080–2089. URL: https://aclanthology.org/2022.lrec-1.224.

Pan, Xiaoman, Boliang Zhang, Jonathan May, Joel Nothman, Kevin Knight, and Heng Ji (July 2017). "Cross-lingual Name Tagging and Linking for 282 Languages." In: *Proceedings of the 55th Annual Meeting of the Association for Computational Linguistics (Volume 1: Long Papers)*. Vancouver, Canada: Association for Computational Linguistics, pp. 1946–1958. DOI: 10.18653/v1/P17-1178. URL: https://www.aclweb.org/anthology/P17-1178.

Papineni, Kishore, Salim Roukos, Todd Ward, and Wei-Jing Zhu (July 2002). "Bleu: a Method for Automatic Evaluation of Machine Translation." In: *Proceedings of the 40th Annual Meeting of the Association for Computational Linguistics*. Philadelphia, Pennsylvania, USA: Association for Computational Linguistics, pp. 311–318. DOI: 10.3115/1073083.1073135. URL: https://aclanthology.org/P02-1040.

Park, Sungjoon et al. (2021). *KLUE: Korean Language Understanding Evaluation*. arXiv: 2105.09680 [cs.CL].

Paul, Debjit, Mittul Singh, Michael A. Hedderich, and Dietrich Klakow (June 2019). "Handling Noisy Labels for Robustly Learning from Self-Training Data for Low-Resource Sequence Labeling." In: *Proceedings of the 2019 Conference of the North American Chapter of the Association for Computational Linguistics: Student Research Workshop*. Minneapolis, Minnesota: Association for Computational Linguistics, pp. 29–34. DOI: 10.18653/v1/N19-3005.

Paullada, Amandalynne (2020). "How Does Machine Translation Shift Power?" In: *Resistance in AI Workshop*. URL: https://drive.google.com/file/d/1wO5UOxTThrcCiU-gEJm_KBShxL_YXEXx/view.

Payne, Doris L., Sara Pacchiarotti, and Mokaya Bosire, eds. (2017). *Diversity in African languages*. Contemporary African Linguistics 1. Berlin: Language Science Press.

Pennington, Jeffrey, Richard Socher, and Christopher Manning (Oct. 2014). "GloVe: Global Vectors for Word Representation." In: *Proceedings of the 2014 Conference on Empirical Methods in Natural Language Processing (EMNLP)*. Doha, Qatar: Association for Computational Linguistics, pp. 1532–1543. DOI: 10.3115/v1/D14-1162. URL: https://www.aclweb.org/anthology/D14-1162.

Peters, Matthew, Mark Neumann, Mohit Iyyer, Matt Gardner, Christopher Clark, Kenton Lee, and Luke Zettlemoyer (June 2018). "Deep Contextualized Word Representations." In: *Proceedings of the 2018 Conference of the North American Chapter of the Association for Computational Linguistics: Human Language Technologies, Volume 1 (Long Papers)*. New Orleans, Louisiana: Association for Computational Linguistics, pp. 2227–2237. DOI: 10.18653/v1/N18-1202. URL: https://www.aclweb.org/anthology/N18-1202.

Pfeiffer, Jonas, Ivan Vuli, Iryna Gurevych, and Sebastian Ruder (2020a). "MAD-X: An Adapter-based Framework for Multi-task Cross-lingual Transfer." In: *Proceedings of EMNLP 2020*.

Pfeiffer, Jonas, Ivan Vulić, Iryna Gurevych, and Sebastian Ruder (Nov. 2020b). "MAD-X: An Adapter-Based Framework for Multi-Task Cross-Lingual Transfer." In: *Proceedings of the 2020 Conference on Empirical Methods in Natural Language Processing (EMNLP)*. Online: Association for Computational Linguistics, pp. 7654–7673. DOI: 10.18653/v1/2020.emnlp-main.617. URL: https://aclanthology.org/2020.emnlp-main.617.

Pfeiffer, Jonas, Ivan Vulić, Iryna Gurevych, and Sebastian Ruder (Nov. 2021). "UNKs Everywhere: Adapting Multilingual Language Models to New Scripts." In: *Proceedings of the 2021 Conference on Empirical Methods in Natural Language Processing*. Online and Punta Cana, Dominican Republic: Association for Computational Linguistics, pp. 10186–10203. DOI: 10.18653/v1/2021.emnlp-main.800. URL: https://aclanthology.org/2021.emnlp-main.800.

Phillipson, Robert and Tove Skutnabb-Kangas (2010). "Language rights in postcolonial Africa." In: *Linguistic Human Rights: Overcoming Linguistic Discrimination*. Ed. by Tove Skutnabb-Kangas and Robert Phillipson. De Gruyter Mouton, pp. 335–346. DOI: doi:10.1515/9783110866391.335. URL: https://doi.org/10.1515/9783110866391.335.

Pires, Telmo, Eva Schlinger, and Dan Garrette (July 2019). "How Multilingual is Multilingual BERT?" In: *Proceedings of the 57th Annual Meeting of the Association for Computational Linguistics*. Florence, Italy: Association for Computational Linguistics, pp. 4996–5001. DOI: 10.18653/v1/P19-1493. URL: https://aclanthology.org/P19-1493.

Plank, Barbara, Anders Søgaard, and Yoav Goldberg (Aug. 2016). "Multilingual Part-of-Speech Tagging with Bidirectional Long Short-Term Memory Models and Auxiliary Loss." In: *Proceedings of the 54th Annual Meeting of the Association for Computational Linguistics (Volume*

*2: Short Papers)*. Berlin, Germany: Association for Computational Linguistics, pp. 412–418. DOI: 10.18653/v1/P16-2067. URL: https://www.aclweb.org/anthology/P16-2067.

Ponti, Edoardo Maria, Goran Glavaš, Olga Majewska, Qianchu Liu, Ivan Vulić, and Anna Korhonen (Nov. 2020). "XCOPA: A Multilingual Dataset for Causal Commonsense Reasoning." In: *Proceedings of the 2020 Conference on Empirical Methods in Natural Language Processing (EMNLP)*. Online: Association for Computational Linguistics, pp. 2362–2376. DOI: 10.18653/v1/2020.emnlp-main.185. URL: https://aclanthology.org/2020.emnlp-main.185.

Poostchi, Hanieh, Ehsan Zare Borzeshi, Mohammad Abdous, and Massimo Piccardi (Dec. 2016). "PersoNER: Persian Named-Entity Recognition." In: *Proceedings of COLING 2016, the 26th International Conference on Computational Linguistics: Technical Papers*. Osaka, Japan: The COLING 2016 Organizing Committee, pp. 3381–3389. URL: https://aclanthology.org/C16-1319.

Popović, Maja (Sept. 2015). "chrF: character n-gram F-score for automatic MT evaluation." In: *Proceedings of the Tenth Workshop on Statistical Machine Translation*. Lisbon, Portugal: Association for Computational Linguistics, pp. 392–395. DOI: 10.18653/v1/W15-3049. URL: https://aclanthology.org/W15-3049.

Post, Matt (Oct. 2018). "A Call for Clarity in Reporting BLEU Scores." In: *Proceedings of the Third Conference on Machine Translation: Research Papers*. Brussels, Belgium: Association for Computational Linguistics, pp. 186–191. DOI: 10.18653/v1/W18-6319. URL: https://aclanthology.org/W18-6319.

Poth, Clifton, Jonas Pfeiffer, Andreas Rücklé, and Iryna Gurevych (Nov. 2021). "What to Pre-Train on? Efficient Intermediate Task Selection." In: *Proceedings of the 2021 Conference on Empirical Methods in Natural Language Processing*. Online and Punta Cana, Dominican Republic: Association for Computational Linguistics, pp. 10585–10605. DOI: 10.18653/v1/2021.emnlp-main.827. URL: https://aclanthology.org/2021.emnlp-main.827.

Pratap, Vineel et al. (2023). "Scaling Speech Technology to 1, 000+ Languages." In: *ArXiv* abs/2305.13516. URL: https://api.semanticscholar.org/CorpusID:258841617.

Pruksachatkun, Yada, Jason Phang, Haokun Liu, Phu Mon Htut, Xiaoyi Zhang, Richard Yuanzhe Pang, Clara Vania, Katharina Kann, and Samuel R. Bowman (July 2020). "Intermediate-Task Transfer Learning with Pretrained Language Models: When and Why Does It Work?" In: *Proceedings of the 58th Annual Meeting of the Association for Computational Linguistics*. Online: Association for Computational Linguistics, pp. 5231–5247. DOI: 10.18653/v1/2020.acl-main.467. URL: https://aclanthology.org/2020.acl-main.467.

Radford, Alec and Karthik Narasimhan (2018). "Improving Language Understanding by Generative Pre-Training." In.

Radford, Alec, Jeff Wu, Rewon Child, David Luan, Dario Amodei, and Ilya Sutskever (2019). "Language Models are Unsupervised Multitask Learners." In.

Raffel, Colin, Noam Shazeer, Adam Roberts, Katherine Lee, Sharan Narang, Michael Matena, Yanqi Zhou, Wei Li, and Peter J. Liu (2020). "Exploring the Limits of Transfer Learning with a Unified Text-to-Text Transformer." In: *Journal of Machine Learning Research* 21.140, pp. 1–67. URL: http://jmlr.org/papers/v21/20-074.html.

Rahimi, Afshin, Yuan Li, and Trevor Cohn (July 2019). "Massively Multilingual Transfer for NER." In: *Proceedings of the 57th Annual Meeting of the Association for Computational Linguistics*. Florence, Italy: Association for Computational Linguistics, pp. 151–164. URL: https://www.aclweb.org/anthology/P19-1015.

Ralethe, Sello (May 2020). "Adaptation of Deep Bidirectional Transformers for Afrikaans Language." English. In: *Proceedings of the 12th Language Resources and Evaluation Conference*. Marseille, France: European Language Resources Association, pp. 2475–2478. ISBN: 979-10-95546-34-4. URL: https://aclanthology.org/2020.lrec-1.301.

Ramachandran, Prajit, Peter Liu, and Quoc Le (Sept. 2017). "Unsupervised Pretraining for Sequence to Sequence Learning." In: *Proceedings of the 2017 Conference on Empirical Methods in Natural Language Processing*. Copenhagen, Denmark: Association for Computational Linguistics, pp. 383–391. DOI: 10.18653/v1/D17-1039. URL: https://aclanthology.org/D17-1039.

Ratinov, Lev and Dan Roth (June 2009). "Design Challenges and Misconceptions in Named Entity Recognition." In: *Proceedings of the Thirteenth Conference on Computational Natural Language Learning (CoNLL-2009)*. Boulder, Colorado: Association for Computational Linguistics, pp. 147–155. URL: https://www.aclweb.org/anthology/W09-1119.

Ratner, Alexander J, Christopher M De Sa, Sen Wu, Daniel Selsam, and Christopher Ré (2016). "Data Programming: Creating Large Training Sets, Quickly." In: *Advances in Neural Information Processing Systems 29*. Ed. by D. D. Lee, M. Sugiyama, U. V. Luxburg, I. Guyon, and R. Garnett. Curran Associates, Inc., pp. 3567–3575.

Ratner, Alexander, Stephen H. Bach, Henry Ehrenberg, Jason Fries, Sen Wu, and Christopher Ré (July 2019). "Snorkel: rapid training data creation with weak supervision." In: *The VLDB Journal*. ISSN: 0949-877X. DOI: 10.1007/s00778-019-00552-1. URL: https://doi.org/10.1007/s00778-019-00552-1.

Rehbein, Ines and Josef Ruppenhofer (July 2017). "Detecting annotation noise in automatically labelled data." In: *Proceedings of the 55th Annual Meeting of the Association for Computational Linguistics (Volume 1: Long Papers)*. Vancouver, Canada: Association for Computational Linguistics, pp. 1160–1170. DOI: 10.18653/v1/P17-1107. URL: https://www.aclweb.org/anthology/P17-1107.

Reid, Machel, Junjie Hu, Graham Neubig, and Yutaka Matsuo (Nov. 2021). "AfroMT: Pretraining Strategies and Reproducible Benchmarks for Translation of 8 African Languages." In: *Proceedings of the 2021 Conference on Empirical Methods in Natural Language Processing*. Online and Punta Cana, Dominican Republic: Association for Computational Linguistics, pp. 1306–1320. DOI: 10.18653/v1/2021.emnlp-main.99. URL: https://aclanthology.org/2021.emnlp-main.99.

Reimers, Nils and Iryna Gurevych (Nov. 2019). "Sentence-BERT: Sentence Embeddings using Siamese BERT-Networks." In: *Proceedings of the 2019 Conference on Empirical Methods in Natural Language Processing*. Association for Computational Linguistics. URL: https://arxiv.org/abs/1908.10084.

Resnik, Philip, Mari Broman Olsen, and Mona T. Diab (1999). "The Bible as a Parallel Corpus: Annotating the 'Book of 2000 Tongues'." In: *Computers and the Humanities* 33, pp. 129–153.

Rijhwani, Shruti, Shuyan Zhou, Graham Neubig, and Jaime Carbonell (July 2020). "Soft Gazetteers for Low-Resource Named Entity Recognition." In: *Proceedings of the 58th Annual Meeting of the Association for Computational Linguistics*. Online: Association for Computational Linguistics, pp. 8118–8123. DOI: 10.18653/v1/2020.acl-main.722. URL: https://www.aclweb.org/anthology/2020.acl-main.722.

Rosenthal, Sara, Noura Farra, and Preslav Nakov (2017). "SemEval-2017 task 4: Sentiment analysis in Twitter." In: *Proceedings of the 11th international workshop on semantic evaluation (SemEval-2017)*, pp. 502–518.

Ruder, Sebastian, Matthew E. Peters, Swabha Swayamdipta, and Thomas Wolf (June 2019). "Transfer Learning in Natural Language Processing." In: *Proceedings of the 2019 Conference of the North American Chapter of the Association for Computational Linguistics: Tutorials*. Minneapolis, Minnesota: Association for Computational Linguistics, pp. 15–18. DOI: 10.18653/v1/N19-5004. URL: https://aclanthology.org/N19-5004.

Ruder, Sebastian et al. (Nov. 2021). "XTREME-R: Towards More Challenging and Nuanced Multilingual Evaluation." In: *Proceedings of the 2021 Conference on Empirical Methods in Natural Language Processing*. Online and Punta Cana, Dominican Republic: Association for Computational Linguistics, pp. 10215–10245. DOI: 10.18653/v1/2021.emnlp-main.802. URL: https://aclanthology.org/2021.emnlp-main.802.

Ruiter, Dana, Cristina España-Bonet, and Josef van Genabith (July 2019). "Self-Supervised Neural Machine Translation." In: *Proceedings of the 57th Annual Meeting of the Association for Computational Linguistics*. Florence, Italy: Association for Computational Linguistics, pp. 1828–1834. DOI: 10.18653/v1/P19-1178. URL: https://aclanthology.org/P19-1178.

Ruiter, Dana, Dietrich Klakow, Josef van Genabith, and Cristina España-Bonet (2021). "Integrating Unsupervised Data Generation into Self-Supervised Neural Machine Translation for Low-Resource Languages." In: *Proceedings of Machine Translation Summit (Research Track)*. virtual: European Association for Machine Translation.

Ruokolainen, Teemu, Pekka Kauppinen, Miikka Silfverberg, and Krister Lindén (2019). "A finnish news corpus for named entity recognition." In: *Language Resources and Evaluation*, pp. 1–26.

Saleva, Jonne and Constantine Lignos (2021). "Mining Wikidata for Name Resources for African Languages." In: *ArXiv* abs/2104.00558.

Sang, Erik F and Fien De Meulder (2003). "Introduction to the CoNLL-2003 shared task: Language-independent named entity recognition." In: *Proceedings of CoNLL 2003*.

Sangal, Rajeev, Dipti Misra Sharma, and Anil Kumar Singh (2008). "Proceedings of the IJCNLP-08 Workshop on Named Entity Recognition for South and South East Asian Languages." In: URL: https://www.aclweb.org/anthology/I08-5000.

Sanh, Victor, Lysandre Debut, Julien Chaumond, and Thomas Wolf (2019). "DistilBERT, a distilled version of BERT: smaller, faster, cheaper and lighter." In: *ArXiv* abs/1910.01108.

Scao, Teven Le et al. (2022). "BLOOM: A 176B-Parameter Open-Access Multilingual Language Model." In: *ArXiv* abs/2211.05100. URL: https://api.semanticscholar.org/CorpusID:253420279.

Scheible, Raphael, Fabian Thomczyk, Patric Tippmann, Victor Jaravine, and Martin Boeker (2020). "GottBERT: a pure German Language Model." In: *ArXiv* abs/2012.02110.

Schwenk, Holger, Vishrav Chaudhary, Shuo Sun, Hongyu Gong, and Francisco Guzmán (Apr. 2021a). "WikiMatrix: Mining 135M Parallel Sentences in 1620 Language Pairs from Wikipedia." In: *Proceedings of the 16th Conference of the European Chapter of the Association for Computational Linguistics: Main Volume*. Online: Association for Computational Linguistics, pp. 1351–1361. DOI: 10.18653/v1/2021.eacl-main.115. URL: https://aclanthology.org/2021.eacl-main.115.

Schwenk, Holger, Guillaume Wenzek, Sergey Edunov, Edouard Grave, Armand Joulin, and Angela Fan (Aug. 2021b). "CCMatrix: Mining Billions of High-Quality Parallel Sentences on the Web." In: *Proceedings of the 59th Annual Meeting of the Association for Computational Linguistics and the 11th International Joint Conference on Natural Language Processing (Volume 1: Long Papers)*. Online: Association for Computational Linguistics, pp. 6490–6500. DOI: 10.18653/v1/2021.acl-long.507. URL: https://aclanthology.org/2021.acl-long.507.

Sennrich, Rico, Barry Haddow, and Alexandra Birch (Aug. 2016a). "Improving Neural Machine Translation Models with Monolingual Data." In: *Proceedings of the 54th Annual Meeting of the Association for Computational Linguistics (Volume 1: Long Papers)*. Berlin, Germany: Association for Computational Linguistics, pp. 86–96.

Sennrich, Rico, Barry Haddow, and Alexandra Birch (Aug. 2016b). "Neural Machine Translation of Rare Words with Subword Units." In: *Proceedings of the 54th Annual Meeting of the Association for Computational Linguistics (Volume 1: Long Papers)*. Berlin, Germany: Association for Computational Linguistics, pp. 1715–1725. DOI: 10.18653/v1/P16-1162. URL: https://www.aclweb.org/anthology/P16-1162.

Sennrich, Rico and Biao Zhang (July 2019). "Revisiting Low-Resource Neural Machine Translation: A Case Study." In: *Proceedings of the 57th Annual Meeting of the Association for Computational Linguistics*. Florence, Italy: Association for Computational Linguistics, pp. 211–221.

Seyoum, Binyam Ephrem, Yusuke Miyao, and Baye Yimam Mekonnen (May 2018). "Universal Dependencies for Amharic." In: *Proceedings of the Eleventh International Conference on Language Resources and Evaluation (LREC 2018)*. Miyazaki, Japan: European Language Resources Association (ELRA). URL: https://aclanthology.org/L18-1350.

Shaalan, K. (2014). "A Survey of Arabic Named Entity Recognition and Classification." In: *Computational Linguistics* 40, pp. 469–510.

Shliazhko, Oleh, Alena Fenogenova, Maria Tikhonova, Vladislav Mikhailov, Anastasia Kozlova, and Tatiana Shavrina (2022). "mGPT: Few-Shot Learners Go Multilingual." In: *ArXiv* abs/2204.07580.

Shode, Iyanuoluwa, David Ifeoluwa Adelani, and Anna Feldman (2022). "YOSM: A NEW YORUBA SENTIMENT CORPUS FOR MOVIE REVIEWS." In: *3rd Workshop on African Natural Language Processing*. URL: https://openreview.net/forum?id=rRzx5qzVIb9.

Shode, Iyanuoluwa, David Ifeoluwa Adelani, JIng Peng, and Anna Feldman (July 2023). "NollySenti: Leveraging Transfer Learning and Machine Translation for Nigerian Movie Sentiment Classification." In: *Proceedings of the 61st Annual Meeting of the Association for Computational Linguistics (Volume 2: Short Papers)*. Toronto, Canada: Association for Computational Linguistics, pp. 986–998. DOI: 10.18653/v1/2023.acl-short.85. URL: https://aclanthology.org/2023.acl-short.85.

Siminyu, Kathleen et al. (2021). "AI4D - African Language Program." In: *ArXiv* abs/2104.02516.

Singh, O. M., A. Padia, and A. Joshi (Dec. 2019). "Named Entity Recognition for Nepali Language." In: *2019 IEEE 5th International Conference on Collaboration and Internet Computing (CIC)*, pp. 184–190. DOI: 10.1109/CIC48465.2019.00031.

Strassel, Stephanie and Jennifer Tracey (May 2016). "LORELEI Language Packs: Data, Tools, and Resources for Technology Development in Low Resource Languages." In: *Proceedings of the Tenth International Conference on Language Resources and Evaluation (LREC'16)*. Portorož, Slovenia: European Language Resources Association (ELRA), pp. 3273–3280. URL: https://www.aclweb.org/anthology/L16-1521.

Strötgen, Jannik, Michael Gertz, Graeme Hirst, and Ruihong Huang (2018). "Domain-Sensitive Temporal Tagging." In: *Computational Linguistics* 44.2.

Sun, Zhiqing, Hongkun Yu, Xiaodan Song, Renjie Liu, Yiming Yang, and Denny Zhou (July 2020). "MobileBERT: a Compact Task-Agnostic BERT for Resource-Limited Devices." In: *Proceedings of the 58th Annual Meeting of the Association for Computational Linguistics*. Online: Association for Computational Linguistics, pp. 2158–2170. DOI: 10.18653/v1/2020.acl-main.195. URL: https://aclanthology.org/2020.acl-main.195.

Szarvas, György, Richárd Farkas, László Felföldi, András Kocsor, and János Csirik (May 2006). "A highly accurate Named Entity corpus for Hungarian." In: *Proceedings of the Fifth International Conference on Language Resources and Evaluation (LREC'06)*. Genoa, Italy: European Language Resources Association (ELRA). URL: http://www.lrec-conf.org/proceedings/lrec2006/pdf/365_pdf.pdf.

Tang, Y., C. Tran, X. Li, P. Chen, Naman Goyal, Vishrav Chaudhary, Jiatao Gu, and Angela Fan (2020). "Multilingual Translation with Extensible Multilingual Pretraining and Finetuning." In: *ArXiv* abs/2008.00401.

Tapo, Allahsera Auguste, Bakary Coulibaly, Sébastien Diarra, Christopher Homan, Julia Kreutzer, Sarah Luger, Arthur Nagashima, Marcos Zampieri, and Michael Leventhal (Dec. 2020). "Neural Machine Translation for Extremely Low-Resource African Languages: A Case Study on Bambara." In: *Proceedings of the 3rd Workshop on Technologies for MT of Low Resource Languages*. Suzhou, China: Association for Computational Linguistics, pp. 23–32. URL: https://aclanthology.org/2020.loresmt-1.3.

Tay, Yi, Vinh Q. Tran, Sebastian Ruder, Jai Gupta, Hyung Won Chung, Dara Bahri, Zhen Qin, Simon Baumgartner, Cong Yu, and Donald Metzler (2022). "Charformer: Fast Character Transformers via Gradient-based Subword Tokenization." In: *International Conference on Learning Representations*. URL: https://openreview.net/forum?id=JtBRnrlOEFN.

The Editors of Encyclopedia Britannica (Jan. 2007). "Malagasy languages." In: *Encyclopedia Britannica*.

Tiedemann, Jörg (May 2012). "Parallel Data, Tools and Interfaces in OPUS." In: *Proceedings of the Eighth International Conference on Language Resources and Evaluation (LREC'12)*. Istanbul, Turkey: European Language Resources Association (ELRA), pp. 2214–2218. URL: http://www.lrec-conf.org/proceedings/lrec2012/pdf/463_Paper.pdf.

Tiedemann, Jörg and Santhosh Thottingal (Nov. 2020). "OPUS-MT – Building open translation services for the World." In: *Proceedings of the 22nd Annual Conference of the European Association for Machine Translation*. Lisboa, Portugal: European Association for Ma-chine Translation, pp. 479–480. URL: https://www.aclweb.org/anthology/2020.eamt-1.61.

Tjong Kim Sang, Erik F. (2002). "Introduction to the CoNLL-2002 Shared Task: Language-Independent Named Entity Recognition." In: *COLING-02: The 6th Conference on Natural Language Learning 2002 (CoNLL-2002)*. URL: https://aclanthology.org/W02-2024.

Tjong Kim Sang, Erik F. and Fien De Meulder (2003). "Introduction to the CoNLL-2003 Shared Task: Language-Independent Named Entity Recognition." In: *Proceedings of the Seventh Conference on Natural Language Learning at HLT-NAACL 2003*, pp. 142–147. URL: https://www.aclweb.org/anthology/W03-0419.

Toral, Antonio, Sheila Castilho, Ke Hu, and Andy Way (Oct. 2018). "Attaining the Unattainable? Reassessing Claims of Human Parity in Neural Machine Translation." In: *Proceedings of the Third Conference on Machine Translation: Research Papers*. Brussels, Belgium: Association for Computational Linguistics, pp. 113–123. DOI: 10.18653/v1/W18-6312. URL: https://aclanthology.org/W18-6312.

Touvron, Hugo et al. (2023a). "LLaMA: Open and Efficient Foundation Language Models." In: *ArXiv* abs/2302.13971. URL: https://api.semanticscholar.org/CorpusID:257219404.

Touvron, Hugo et al. (2023b). "Llama 2: Open Foundation and Fine-Tuned Chat Models." In: *ArXiv* abs/2307.09288. URL: https://api.semanticscholar.org/CorpusID:259950998.

Traill, Anthony (Nov. 2015). "click languages." In: *Encyclopedia Britannica*.

Tran, Chau, Shruti Bhosale, James Cross, Philipp Koehn, Sergey Edunov, and Angela Fan (2021). "Facebook AI WMT21 News Translation Task Submission." In: *arXiv preprint arXiv:2108.03265*.

Turian, Joseph, Lev-Arie Ratinov, and Yoshua Bengio (July 2010). "Word Representations: A Simple and General Method for Semi-Supervised Learning." In: *Proceedings of the 48th Annual Meeting of the Association for Computational Linguistics*. Uppsala, Sweden: Association for Computational Linguistics, pp. 384–394. URL: https://aclanthology.org/P10-1040.

Varab, Daniel and Natalie Schluter (Nov. 2021). "MassiveSumm: a very large-scale, very multilingual, news summarisation dataset." In: *Proceedings of the 2021 Conference on Empirical Methods in Natural Language Processing*. Online and Punta Cana, Dominican Republic: Association for Computational Linguistics, pp. 10150–10161. DOI: 10.18653/v1/2021.emnlp-main.797. URL: https://aclanthology.org/2021.emnlp-main.797.

Vaswani, Ashish, Noam Shazeer, Niki Parmar, Jakob Uszkoreit, Llion Jones, Aidan N Gomez, Ł ukasz Kaiser, and Illia Polosukhin (2017). "Attention is All you Need." In: *Advances in Neural Information Processing Systems 30*. Ed. by I. Guyon, U. V. Luxburg, S. Bengio, H. Wallach, R. Fergus, S. Vishwanathan, and R. Garnett. Curran

Associates, Inc., pp. 5998–6008. URL: http://papers.nips.cc/paper/7181-attention-is-all-you-need.pdf.

Veit, Andreas, Neil Alldrin, Gal Chechik, Ivan Krasin, Abhinav Gupta, and Serge J. Belongie (2017). "Learning From Noisy Large-Scale Datasets With Minimal Supervision." In: *Proceedings of the IEEE Conference on Computer Vision and Pattern Recognition (CVPR)*. DOI: 10.1109/CVPR.2015.7298885.

Versteegh, Kees (Dec. 2001). "Linguistic Contacts Between Arabic and Other Languages." In: *Arabica* 48, pp. 470–508. DOI: 10.1163/157005801323163825.

Voigt, Rainer M. (1987). "The two prefix-conjugations in East Cushitic, East Semitic, and Chadic." In: *Bulletin of the School of Oriental and African Studies*.

Vries, Wietse de, Martijn Wieling, and Malvina Nissim (May 2022). "Make the Best of Cross-lingual Transfer: Evidence from POS Tagging with over 100 Languages." In: *Proceedings of the 60th Annual Meeting of the Association for Computational Linguistics (Volume 1: Long Papers)*. Dublin, Ireland: Association for Computational Linguistics, pp. 7676–7685. DOI: 10.18653/v1/2022.acl-long.529. URL: https://aclanthology.org/2022.acl-long.529.

Wang, Alex, Amanpreet Singh, Julian Michael, Felix Hill, Omer Levy, and Samuel Bowman (Nov. 2018). "GLUE: A Multi-Task Benchmark and Analysis Platform for Natural Language Understanding." In: *Proceedings of the 2018 EMNLP Workshop BlackboxNLP: Analyzing and Interpreting Neural Networks for NLP*. Brussels, Belgium: Association for Computational Linguistics, pp. 353–355. DOI: 10.18653/v1/W18-5446. URL: https://aclanthology.org/W18-5446.

Wang, Hao, Bing Liu, Chaozhuo Li, Yan Yang, and Tianrui Li (2019). "Learning with Noisy Labels for Sentence-level Sentiment Classification." In: *Proceedings of the 2019 Conference on Empirical Methods in Natural Language Processing and the 9th International Joint Conference on Natural Language Processing, EMNLP-IJCNLP 2019*, pp. 6285–6291. DOI: 10.18653/v1/D19-1655.

Wang, Wenhui, Hangbo Bao, Shaohan Huang, Li Dong, and Furu Wei (Aug. 2021). "MiniLMv2: Multi-Head Self-Attention Relation Distillation for Compressing Pretrained Transformers." In: *Findings of the Association for Computational Linguistics: ACL-IJCNLP 2021*. Online: Association for Computational Linguistics, pp. 2140–2151. DOI: 10.18653/v1/2021.findings-acl.188. URL: https://aclanthology.org/2021.findings-acl.188.

Wang, Wenhui, Furu Wei, Li Dong, Hangbo Bao, Nan Yang, and Ming Zhou (2020a). "MiniLM: Deep Self-Attention Distillation for Task-Agnostic Compression of Pre-Trained Transformers." In: *Advances in Neural Information Processing Systems*. Ed. by H. Larochelle, M. Ranzato, R. Hadsell, M. F. Balcan, and H. Lin. Vol. 33. Curran Associates, Inc., pp. 5776–5788. URL: https://proceedings.neurips.

cc/paper/2020/file/3f5ee243547dee91fbd053c1c4a845aa-Paper.pdf.

Wang, Zihan, Karthikeyan K, Stephen Mayhew, and Dan Roth (Nov. 2020b). "Extending Multilingual BERT to Low-Resource Languages." In: *Findings of the Association for Computational Linguistics: EMNLP 2020*. Online: Association for Computational Linguistics, pp. 2649–2656. DOI: 10.18653/v1/2020.findings-emnlp.240. URL: https://aclanthology.org/2020.findings-emnlp.240.

Wilie, Bryan et al. (Dec. 2020). "IndoNLU: Benchmark and Resources for Evaluating Indonesian Natural Language Understanding." In: *Proceedings of the 1st Conference of the Asia-Pacific Chapter of the Association for Computational Linguistics and the 10th International Joint Conference on Natural Language Processing*. Suzhou, China: Association for Computational Linguistics, pp. 843–857. URL: https://aclanthology.org/2020.aacl-main.85.

Wolf, Thomas et al. (2019). "HuggingFace's Transformers: State-of-the-art Natural Language Processing." In: *ArXiv* abs/1910.03771.

Wolf, Thomas et al. (Oct. 2020). "Transformers: State-of-the-Art Natural Language Processing." In: *Proceedings of the 2020 Conference on Empirical Methods in Natural Language Processing: System Demonstrations*. Online: Association for Computational Linguistics, pp. 38–45. DOI: 10.18653/v1/2020.emnlp-demos.6. URL: https://aclanthology.org/2020.emnlp-demos.6.

Wu, Shijie and Mark Dredze (Nov. 2019). "Beto, Bentz, Becas: The Surprising Cross-Lingual Effectiveness of BERT." In: *Proceedings of the 2019 Conference on Empirical Methods in Natural Language Processing and the 9th International Joint Conference on Natural Language Processing (EMNLP-IJCNLP)*. Hong Kong, China: Association for Computational Linguistics, pp. 833–844. DOI: 10.18653/v1/D19-1077. URL: https://aclanthology.org/D19-1077.

Xia, Mengzhou, Antonios Anastasopoulos, Ruochen Xu, Yiming Yang, and Graham Neubig (July 2020). "Predicting Performance for Natural Language Processing Tasks." In: *Proceedings of the 58th Annual Meeting of the Association for Computational Linguistics*. Online: Association for Computational Linguistics, pp. 8625–8646. DOI: 10.18653/v1/2020.acl-main.764. URL: https://aclanthology.org/2020.acl-main.764.

Xiao, Tong, Tian Xia, Yi Yang, Chang Huang, and Xiaogang Wang (2015). "Learning from massive noisy labeled data for image classification." In: *Proceedings of the IEEE conference on computer vision and pattern recognition*, pp. 2691–2699. DOI: 10.1109/CVPR.2015.7298885.

Xu, Liang et al. (Dec. 2020). "CLUE: A Chinese Language Understanding Evaluation Benchmark." In: *Proceedings of the 28th International Conference on Computational Linguistics*. Barcelona, Spain (Online): International Committee on Computational Linguistics,

pp. 4762–4772. DOI: 10.18653/v1/2020.coling-main.419. URL: https://aclanthology.org/2020.coling-main.419.

Xue, Linting, Aditya Barua, Noah Constant, Rami Al-Rfou, Sharan Narang, Mihir Kale, Adam Roberts, and Colin Raffel (2022). "ByT5: Towards a Token-Free Future with Pre-trained Byte-to-Byte Models." In: *Transactions of the Association for Computational Linguistics* 10, pp. 291–306. DOI: 10.1162/tacl_a_00461. URL: https://aclanthology.org/2022.tacl-1.17.

Xue, Linting, Noah Constant, Adam Roberts, Mihir Kale, Rami Al-Rfou, Aditya Siddhant, Aditya Barua, and Colin Raffel (June 2021). "mT5: A Massively Multilingual Pre-trained Text-to-Text Transformer." In: *Proceedings of the 2021 Conference of the North American Chapter of the Association for Computational Linguistics: Human Language Technologies*. Online: Association for Computational Linguistics, pp. 483–498. DOI: 10.18653/v1/2021.naacl-main.41. URL: https://aclanthology.org/2021.naacl-main.41.

Yadav, Vikas and Steven Bethard (Aug. 2018). "A Survey on Recent Advances in Named Entity Recognition from Deep Learning models." In: *Proceedings of the 27th International Conference on Computational Linguistics*. Santa Fe, New Mexico, USA: Association for Computational Linguistics, pp. 2145–2158. URL: https://www.aclweb.org/anthology/C18-1182.

Yamada, Ikuya, Akari Asai, Hiroyuki Shindo, Hideaki Takeda, and Yuji Matsumoto (Nov. 2020). "LUKE: Deep Contextualized Entity Representations with Entity-aware Self-attention." In: *Proceedings of the 2020 Conference on Empirical Methods in Natural Language Processing (EMNLP)*. Online: Association for Computational Linguistics, pp. 6442–6454. DOI: 10.18653/v1/2020.emnlp-main.523. URL: https://www.aclweb.org/anthology/2020.emnlp-main.523.

Yang, Jian et al. (2021). "Multilingual Machine Translation Systems from Microsoft for WMT21 Shared Task." In: *ArXiv* abs/2111.02086.

Yang, Yaosheng, Wenliang Chen, Zhenghua Li, Zhengqiu He, and Min Zhang (2018). "Distantly Supervised NER with Partial Annotation Learning and Reinforcement Learning." In: *Proceedings of COLING 2018*. URL: https://www.aclweb.org/anthology/C18-1183.

Yimam, Seid Muhie, Hizkiel Mitiku Alemayehu, Abinew Ayele, and Chris Biemann (Dec. 2020). "Exploring Amharic Sentiment Analysis from Social Media Texts: Building Annotation Tools and Classification Models." In: *Proceedings of the 28th International Conference on Computational Linguistics*. Barcelona, Spain (Online): International Committee on Computational Linguistics, pp. 1048–1060. DOI: 10.18653/v1/2020.coling-main.91. URL: https://aclanthology.org/2020.coling-main.91.

Yimam, Seid Muhie, Abinew Ali Ayele, Gopalakrishnan Venkatesh, Ibrahim Gashaw, and Chris Biemann (2021). "Introducing Various Semantic Models for Amharic: Experimentation and Evaluation with Multiple Tasks and Datasets." In: *Future Internet* 13.11. ISSN: 1999-5903. DOI: 10.3390/fi13110275. URL: https://www.mdpi.com/1999-5903/13/11/275.

Yohannes, Hailemariam Mehari and Toshiyuki Amagasa (2022). "Named-entity recognition for a low-resource language using pre-trained language model." In: *Proceedings of the 37th ACM/SIGAPP Symposium on Applied Computing*.

Yong, Zheng Xin et al. (July 2023). "BLOOM+1: Adding Language Support to BLOOM for Zero-Shot Prompting." In: *Proceedings of the 61st Annual Meeting of the Association for Computational Linguistics (Volume 1: Long Papers)*. Toronto, Canada: Association for Computational Linguistics, pp. 11682–11703. DOI: 10.18653/v1/2023.acl-long.653. URL: https://aclanthology.org/2023.acl-long.653.

Zanon Boito, Marcely, William Havard, Mahault Garnerin, Éric Le Ferrand, and Laurent Besacier (May 2020). "MaSS: A Large and Clean Multilingual Corpus of Sentence-aligned Spoken Utterances Extracted from the Bible." English. In: *Proceedings of the 12th Language Resources and Evaluation Conference*. Marseille, France: European Language Resources Association, pp. 6486–6493. ISBN: 979-10-95546-34-4. URL: https://aclanthology.org/2020.lrec-1.799.

Zhang, Boliang, Ying Lin, Xiaoman Pan, Di Lu, Jonathan May, Kevin Knight, and Heng Ji (2018). "ELISA-EDL: A Cross-lingual Entity Extraction, Linking and Localization System." In: *Proceedings of NAACL-HLT 2018: Demonstrations*. DOI: 10.18653/v1/N18-5009. URL: http://aclweb.org/anthology/N18-5009.

Zhang, Xiang, Junbo Jake Zhao, and Yann LeCun (2015). "Character-level Convolutional Networks for Text Classification." In: *NIPS*.

Zoph, Barret, Deniz Yuret, Jonathan May, and Kevin Knight (Nov. 2016). "Transfer Learning for Low-Resource Neural Machine Translation." In: *Proceedings of the 2016 Conference on Empirical Methods in Natural Language Processing*. Austin, Texas: Association for Computational Linguistics, pp. 1568–1575. DOI: 10.18653/v1/D16-1163. URL: https://aclanthology.org/D16-1163.

∀ et al. (2020). "Participatory Research for Low-resourced Machine Translation: A Case Study in African Languages." In: *Findings of the Association for Computational Linguistics: EMNLP 2020*. Online. URL: https://www.aclweb.org/anthology/2020.findings-emnlp.195.